\documentclass{article}
\usepackage[utf8]{inputenc}
\usepackage{main}
\usepackage{microtype}
\usepackage{graphicx}
\usepackage{times}
\usepackage{amsmath}
\usepackage{amssymb}
\usepackage{enumitem}
\usepackage{booktabs}
\usepackage{xcolor}  

\usepackage{longtable}
\usepackage{array}
\usepackage{natbib}
\definecolor{mydarkblue}{rgb}{0,0.08,0.45}
\usepackage[colorlinks=true,linkcolor=mydarkblue,citecolor=mydarkblue,filecolor=mydarkblue,urlcolor=mydarkblue]{hyperref}
\usepackage{fancyhdr}
\usepackage[table]{xcolor}
\usepackage[most]{tcolorbox} 
\usepackage{listings}
\usepackage{wrapfig}
\usepackage{mdframed}
\usepackage{listings}
\usepackage{xcolor}
\usepackage{enumitem}
\usepackage{fontawesome5}
\usepackage{multirow}
\usepackage{makecell}    
\usepackage{xcolor}
\usepackage{pifont}
\definecolor{gred}{RGB}{250, 210, 207}
\definecolor{coolblue1}{rgb}{0.91, 0.94, 0.98}
\definecolor{coolblue2}{rgb}{0.76, 0.85, 0.94}
\definecolor{coolblue3}{rgb}{0.54, 0.72, 0.87}
\definecolor{coolblue4}{rgb}{1, 1, 1}

\usepackage{xspace}
\usepackage{cleveref}
\usepackage{multirow}

\definecolor{reviewboxbg}{RGB}{224,240,244}   
\definecolor{commentboxbg}{RGB}{238,238,238}  
\definecolor{mutedred}{RGB}{180, 70, 60}      
\definecolor{mutedgreen}{RGB}{60, 120, 75}    

\newcommand{\cmark}{\textcolor{green!70!black}{\ding{51}}}
\newcommand{\xmark}{\textcolor{red}{\ding{55}}}
\newcommand{\pmark}{\textcolor{orange!90!black}{$\triangle$}}

\newenvironment{itemize*}%
 {\leftmargini=10pt\begin{itemize}%
  \setlength{\itemsep}{0pt}%
  \setlength{\parskip}{0pt}%
  }%
 {\end{itemize}}
\newenvironment{enumerate*}%
 {\begin{enumerate}%
  \setlength{\itemsep}{0pt}%
  \setlength{\parskip}{0pt}}%
 {\end{enumerate}}

\mdfdefinestyle{reviewbox}{
    linecolor=blue!50!black,
    linewidth=0.8pt,
    roundcorner=3pt,
    innertopmargin=8pt,
    innerbottommargin=8pt,
    innerleftmargin=12pt,
    innerrightmargin=12pt,
    skipabove=8pt,
    skipbelow=8pt,
    frametitlebackgroundcolor=blue!50!black,
    frametitlefont=\normalfont\small\color{white},
    frametitlerule=false,
    frametitleaboveskip=4pt,
    frametitlebelowskip=4pt,
    frametitlealignment=\raggedright,
}

\newcommand{\commentbar}[1]{%
    \par\noindent
    \hspace{1em}%
    \begin{minipage}{\dimexpr\linewidth-1em\relax}%
        \textcolor{blue!50!black}{\rule[-0.3em]{1pt}{1em}}\hspace{6pt}%
        \parbox[t]{\dimexpr\linewidth-10pt\relax}{%
            \textcolor{blue!50!black}{\textit{\textbf{Comment:}}} #1%
        }%
    \end{minipage}%
    \par
}

\usepackage{titletoc}

\newcommand\DoToC{%
  \startcontents
  \noindent\begin{minipage}{\textwidth}
    \printcontents{}{1}{\noindent \textbf{\Large{Table of Contents in Appendix}}\vskip1pt\vskip3pt}%
  \end{minipage}%
  \vskip1pt\vskip3pt
}

\lstdefinestyle{reviewcode}{
    basicstyle=\footnotesize\ttfamily,
    backgroundcolor=\color{gray!8},
    frame=single,
    rulecolor=\color{gray!40},
    framesep=6pt,
    xleftmargin=20pt,
    xrightmargin=20pt,
    language=Python,
    keywordstyle=\color{blue!60!black}\bfseries,
    commentstyle=\color{gray!60}\itshape,
    stringstyle=\color{teal!70!black},
    showstringspaces=false,
    breaklines=true,
    breakatwhitespace=true,
    aboveskip=8pt,
    belowskip=8pt,
}

\mdfdefinestyle{categorybox}{
    linecolor=blue!50!black,
    linewidth=0.8pt,
    roundcorner=3pt,
    innertopmargin=8pt,
    innerbottommargin=8pt,
    innerleftmargin=12pt,
    innerrightmargin=12pt,
    skipabove=10pt,
    skipbelow=6pt,
    frametitlebackgroundcolor=blue!50!black,
    frametitlefont=\normalfont\small\color{white},
    frametitlerule=false,
    frametitleaboveskip=4pt,
    frametitlebelowskip=4pt,
    nobreak=true,
    frametitlealignment=\raggedright,
}

\mdfdefinestyle{categoryboxR}{
    linewidth=1pt,
    linecolor=blue!50!black,
    backgroundcolor=blue!7,
    fontcolor=black,
    leftmargin=0pt, rightmargin=0pt,
    innertopmargin=8pt, innerbottommargin=8pt,
    innerleftmargin=10pt, innerrightmargin=10pt,
    frametitlefont=\normalfont\bfseries,
    frametitlebackgroundcolor=blue!15,
    frametitlerule=true,
    frametitlerulecolor=blue!50!black,
    frametitlerulewidth=0.5pt
}

\mdfdefinestyle{expertcommentbox}{
    linecolor=black,
    linewidth=0.6pt,
    backgroundcolor=white,
    roundcorner=3pt,
    innertopmargin=6pt,
    innerbottommargin=6pt,
    innerleftmargin=8pt,
    innerrightmargin=8pt,
    skipabove=3pt,
    skipbelow=12pt,
}

\mdfdefinestyle{categoryboxW}{
    linecolor=mutedred,
    linewidth=0.8pt,
    roundcorner=3pt,
    innertopmargin=8pt,
    innerbottommargin=8pt,
    innerleftmargin=12pt,
    innerrightmargin=12pt,
    skipabove=10pt,
    skipbelow=6pt,
    frametitlebackgroundcolor=mutedred,
    frametitlefont=\normalfont\small\color{white},
    frametitlerule=false,
    frametitleaboveskip=4pt,
    frametitlebelowskip=4pt,
    frametitlealignment=\raggedright,
}

\mdfdefinestyle{categoryboxS}{
    linecolor=mutedgreen,
    linewidth=0.8pt,
    roundcorner=3pt,
    innertopmargin=8pt,
    innerbottommargin=8pt,
    innerleftmargin=12pt,
    innerrightmargin=12pt,
    skipabove=10pt,
    skipbelow=6pt,
    frametitlebackgroundcolor=mutedgreen,
    frametitlefont=\normalfont\small\color{white},
    frametitlerule=false,
    frametitleaboveskip=4pt,
    frametitlebelowskip=4pt,
    frametitlealignment=\raggedright,
}

\begin{document}

\title{\textbf{On the limits and opportunities of AI reviewers: Reviewing the reviews of Nature-family papers with 45 expert scientists}}

\author{
\textbf{Seungone Kim}$^{1}$ \quad
\textbf{Dongkeun Yoon}$^{2}$ \quad
\textbf{Kiril Gashteovski}$^{3,4}$ \quad
\textbf{Juyoung Suk}$^{2}$ \quad
\textbf{Jinheon Baek}$^{2}$ \\
\textbf{Pranjal Aggarwal}$^{1}$ \quad
\textbf{Ian Wu}$^{1}$ \quad
\textbf{Viktor Zaverkin}$^{5}$ \quad
\textbf{Spase Petkoski}$^{4,6}$ \quad
\textbf{Daniel R. Schrider}$^{7}$ \\
\textbf{Ilija Dukovski}$^{4,8}$ \quad
\textbf{Francesco Santini}$^{9,10}$ \quad
\textbf{Biljana Mitreska}$^{11}$ \quad
\textbf{Yong Jeong}$^{2}$ \quad
\textbf{Kyeongha Kwon}$^{2}$ \\
\textbf{Young Min Sim}$^{2}$ \quad
\textbf{Dragana Manasova}$^{12}$ \quad
\textbf{Arthur Porto}$^{13}$ \quad
\textbf{Biljana Mojsoska}$^{14}$ \quad
\textbf{Makoto Takamoto}$^{3}$ \\
\textbf{Marko Shuntov}$^{15}$ \quad
\textbf{Ruoqi Liu}$^{16}$ \quad
\textbf{Hyunjoo Jenny Lee}$^{2}$ \quad
\textbf{Niyazi Ulas Dinç}$^{17}$ \quad
\textbf{Yehhyun Jo}$^{18}$ \\
\textbf{Sunkyu Han}$^{2}$ \quad
\textbf{Chungwoo Lee}$^{2}$ \quad
\textbf{Huishan Li}$^{2}$ \quad
\textbf{Esther H. R. Tsai}$^{19}$ \quad
\textbf{Ergun Simsek}$^{20}$ \\
\textbf{Khushboo Shafi}$^{2}$ \quad
\textbf{Yeonseung Chung}$^{2}$ \quad
\textbf{Jihye Park}$^{21}$ \quad
\textbf{Aleksandar Shulevski}$^{4,22}$ \\
\textbf{Henrik Christiansen}$^{3}$ \quad
\textbf{Yoosang Son}$^{2}$ \quad
\textbf{Elly Knight}$^{23}$ \quad
\textbf{Amanda Montoya}$^{24}$ \\
\textbf{Jeongyoun Ahn}$^{2}$ \quad
\textbf{Christian Langkammer}$^{25}$ \quad
\textbf{Heera Moon}$^{2}$ \quad
\textbf{Changwon Yoon}$^{2}$ \\
\textbf{Nikola Stikov}$^{4,26,27}$ \quad
\textbf{Mooseok Jang}$^{2}$ \quad
\textbf{Edward Choi}$^{2}$ \quad
\textbf{Junhan Kim}$^{2}$ \quad
\textbf{Yeon Sik Jung}$^{2}$ \\
\textbf{Woo Youn Kim}$^{2}$ \quad
\textbf{Jae Kyoung Kim}$^{2}$ \quad
\textbf{Ishraq Md Anjum}$^{20}$ \quad
\textbf{Hyun Uk Kim}$^{2}$ \quad
\textbf{Drew Bridges}$^{1}$ \\
\textbf{Carolin Lawrence}$^{3}$ \quad
\textbf{Xiang Yue}$^{1}$ \quad
\textbf{Alice Oh}$^{2}$ \quad
\textbf{Akari Asai}$^{1}$ \quad
\textbf{Sean Welleck}$^{1}$ \quad
\textbf{Graham Neubig}$^{1}$ \\
\textsuperscript{1}Carnegie Mellon University \quad
\textsuperscript{2}KAIST \quad
\textsuperscript{3}NEC Laboratories Europe \\
\textsuperscript{4}Ss. Cyril and Methodius University in Skopje \quad
\textsuperscript{5}INM - Leibniz Institute for New Materials;\\ Saarland University German Research Center for Artificial Intelligence (DFKI) \\
\textsuperscript{6}Aix Marseille University, INSERM \quad
\textsuperscript{7}University of North Carolina at Chapel Hill \quad
\textsuperscript{8}Boston University \\
\textsuperscript{9}University of Basel \quad
\textsuperscript{10}University Hospital of Basel \quad
\textsuperscript{11}University of Manchester \\
\textsuperscript{12}Massachusetts Institute of Technology \quad
\textsuperscript{13}Florida Museum of Natural History, University of Florida \\
\textsuperscript{14}Roskilde University \quad
\textsuperscript{15}University of Copenhagen \quad
\textsuperscript{16}Stanford University \\
\textsuperscript{17}École Polytechnique Fédérale de Lausanne \quad
\textsuperscript{18}Institute for Basic Science (IBS) \\
\textsuperscript{19}Brookhaven National Laboratory \quad
\textsuperscript{20}University of Maryland Baltimore County \\
\textsuperscript{21}Lawrence Berkeley National Laboratory \quad
\textsuperscript{22}The Netherlands Institute for Radio Astronomy \\
\textsuperscript{23}University of Alberta \quad
\textsuperscript{24}The University of Texas MD Anderson Cancer Center \\
\textsuperscript{25}Medical University of Graz \quad
\textsuperscript{26}Polytechnique Montréal \quad
\textsuperscript{27}Montreal Heart Institute \\
\texttt{seungone@cmu.edu \quad swelleck@andrew.cmu.edu \quad gneubig@cs.cmu.edu} \\
    \href{https://github.com/prometheus-eval/cmu-paper-reviewer}{\includegraphics[height=0.4cm]{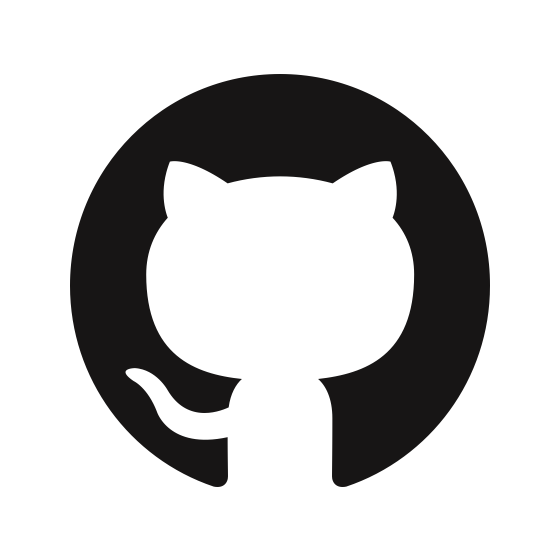} \textbf{Code}} ~
    \href{https://huggingface.co/datasets/prometheus-eval/peerreview-bench}{\includegraphics[height=0.4cm]{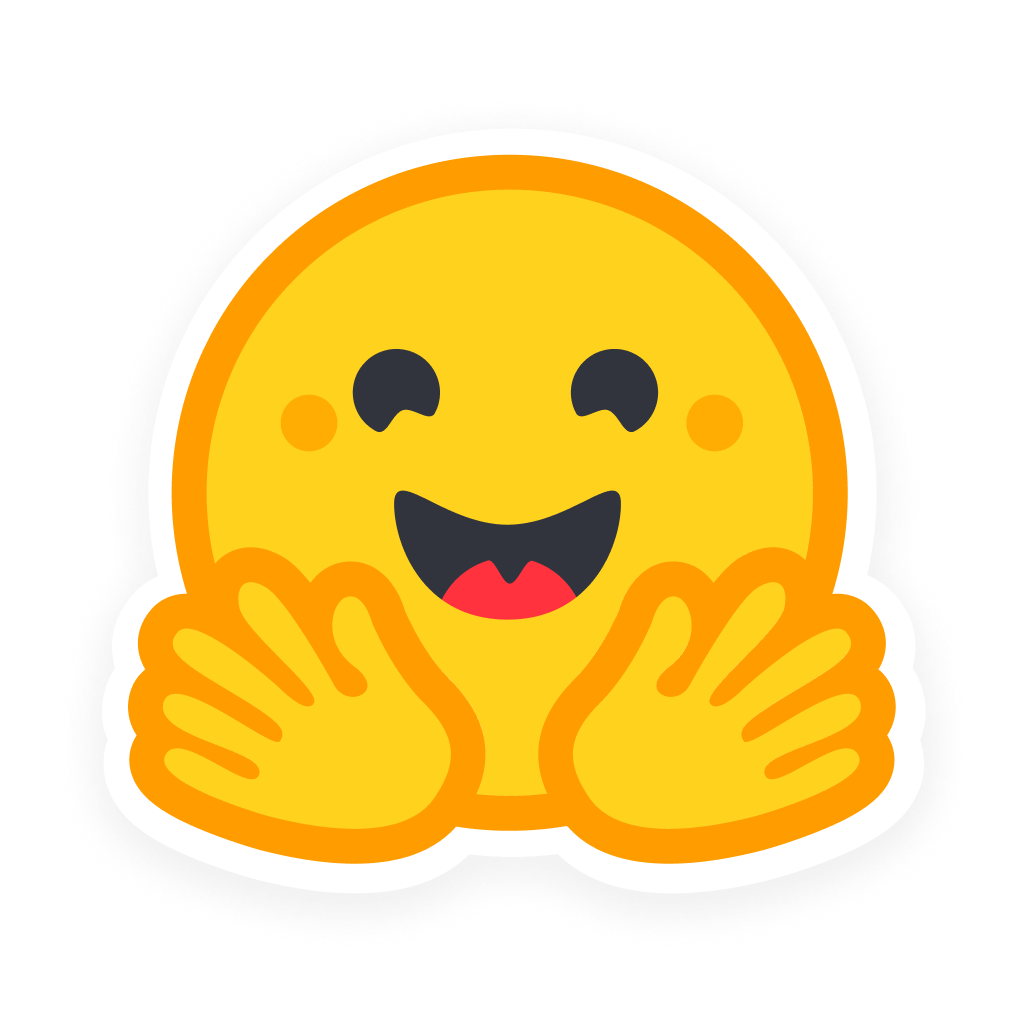} \textbf{PeerReview Bench Dataset}} ~
    \href{https://prometheus-eval.github.io/cmu-paper-reviewer/}{\includegraphics[height=0.4cm]{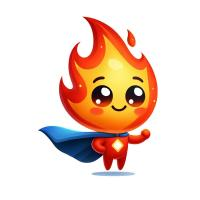} \textbf{CMU Paper Reviewer}}
}

\maketitle
\thispagestyle{fancy}
\fancyhead{}
\lhead{\includegraphics[height=0.5cm]{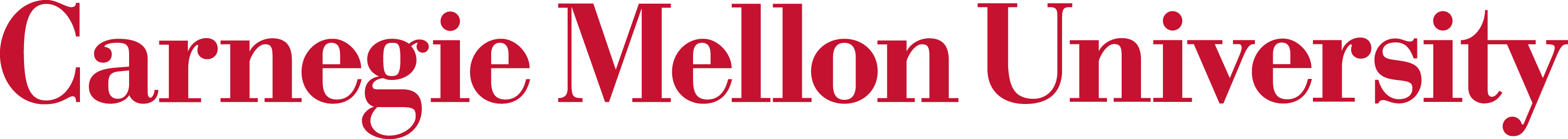}}
\rhead{%
  \raisebox{-0.1cm}{\includegraphics[height=0.8cm]{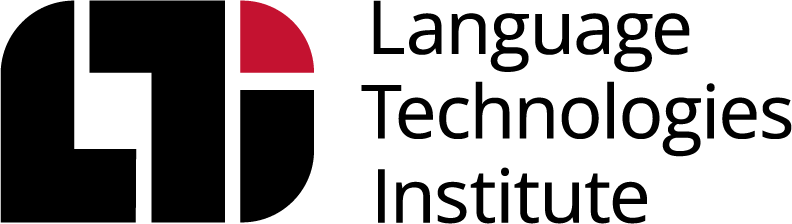}}%
}
\renewcommand{\headrulewidth}{0pt}
\setlength{\headheight}{12pt}
\addtolength{\topmargin}{00pt}
\setlength{\headsep}{3mm}

\vspace{-1.0em}

\begin{abstract}
With the advancement of AI capabilities, AI reviewers are beginning to be deployed in scientific peer review, yet their capability and credibility remain in question: many scientists simply view them as probabilistic systems without the expertise to evaluate research, while other researchers are more optimistic about their readiness without concrete evidence. Understanding what AI reviewers do well, where they fall short, and what challenges remain is essential. However, existing evaluations of AI reviewers have focused on whether their verdicts match human verdicts (\textit{e.g.}, score alignment, acceptance prediction), which is insufficient to characterize their capabilities and limits. In this paper, we close this gap through a large-scale expert annotation study, in which 45 domain scientists in Physical, Biological, and Health Sciences spent 469 hours rating 2{,}960 individual criticisms (each targeting one specific aspect of a paper) from human-written and AI-generated reviews of 82 \textit{Nature}-family papers on correctness, significance, and sufficiency of evidence. On a composite of all three dimensions, a reviewing agent powered by GPT-5.2 scores above each paper's top-rated human reviewer (60.0\% vs.\ 48.2\%, $p = 0.009$), while all three AI reviewers (including Gemini 3.0 Pro and Claude Opus 4.5) exceed the lowest-rated human across every dimension. AI reviewers' accurate criticisms are also more often rated significant and well-evidenced, and surface a distinct 26\% of issues no human raises. However, AI reviewers overlap far more than humans do (21\% vs.\ 3\% for cross-reviewer pairs), and exhibit 16 recurring weaknesses humans do not share, such as limited subfield knowledge, lack of long context management over multiple files, and overly critical stance on minor issues. Overall, our results position current AI reviewers as complements to, not substitutes for, human reviewers.
\end{abstract}
\section{Introduction}

Peer review has long served as the cornerstone of the scholarly publication system, ensuring the credibility, rigor, and cumulative advancement of scientific knowledge~\citep{gannon2001essential,kelly2014peer,siler2015measuring}. The expert scrutiny it provides catches errors before they enter the literature, surfaces methodological concerns that improve the published work, and ultimately calibrates which findings the scientific community treats as reliable. This system, however, is under unprecedented scaling pressure. The volume of scientific output is rising at a historic rate, accelerated further by the recent maturation of generative AI as a research aid~\citep{wang2023scientific,lu2026towards}, while the pool of qualified human reviewers is not expanding at a comparable pace. In major AI conferences such as \textit{NeurIPS} and \textit{ICLR}, submissions have grown so rapidly that many researchers report declining review quality~\citep{chen2025position}. In major science journals including \textit{Nature} and \textit{Science}, the median time from submission to publication has extended to 100 to 160 days~\citep{powell2016does}, delaying the feedback authors need to refine their manuscripts. LLM-agent powered reviewers, which we refer to as \textit{AI reviewers}~\citep{liu2023reviewergpt,kuznetsov2024can,bauchner2024use}, are one response now being trialed at scale, including \textit{AAAI}-26's deployment on all 22{,}977 main-track submissions~\citep{biswas2026ai} and \textit{NEJM AI}'s ``Fast Track'' process~\citep{manrai2025accelerating}. Their throughput is not bounded by reviewer availability, and they can perform tasks human reviewers often forgo under time constraints, such as literature cross-referencing and code inspection~\citep{wei2025ai}.\\

What such deployments and the existing literature do and do not tell us about AI reviewers hinges on the level at which AI reviews have been evaluated to date. This evaluation has happened chiefly at the level of \textit{aggregate outputs} (\textit{i.e.}, ``Do AI reviewers produce similar overall scores, accept-or-reject recommendations, or holistic ratings as humans?'')~\citep{saad2024exploring,zhu2025deepreview,idahl2025openreviewer,zhang2026from,lu2026towards}. Such verdict-level agreement is a fragile benchmark in principle: the NeurIPS 2014 and 2021 consistency experiments, in which roughly 10\% of submissions were independently reviewed by two committees, found that approximately half of the papers accepted by one committee were rejected by the other (49.5\% in 2014, 50.6\% in 2021)~\citep{cortes2021inconsistency,beygelzimer2023has}, indicating substantial randomness in the human verdict itself. More importantly, verdict-level agreement says nothing about the substance of the individual criticisms authors actually receive: whether they are factually correct, raise issues that matter, and are backed by credible evidence. The reports of inflated scores and generic feedback in indiscriminate AI use for reviewing~\citep{liang2024monitoring,russo2025ai} describe exactly the kind of failure verdict-level evaluation cannot see, since two reviews can arrive at the same recommendation while differing entirely in which problems they identify and how well they support them. Distinguishing whether AI reviewers offer genuine technical scrutiny or polished but superficial commentary, and whether their issues overlap with or extend beyond those humans find, requires evaluation at the criticism level.\\

We address this with a large-scale expert annotation study in which forty-five domain scientists, spanning Physical, Biological, and Health Sciences, collectively spent 469 hours scoring 2{,}960 \textit{review items} (atomic criticisms each targeting one aspect of a paper) from the human and AI-generated reviews of 82 \textit{Nature}-family papers, judging each on correctness, significance, and evidence sufficiency, with free-form qualitative feedback. Three findings emerge, which together establish that current AI reviewers could complement, but should not replace, human reviewers. First, on the composite of all three quality criteria, GPT-5.2 outperforms the top-rated human reviewer on each paper (60.0\% vs.\ 48.2\%, $p = 0.009$), and Claude Opus 4.5 and Gemini 3.0 Pro are statistically indistinguishable from the top-rated human. Specifically, AI reviewers raise more incorrect items than the top-rated human, but their correct items are more often significant and well-evidenced (\autoref{sec:performance}). Second, AI reviewers raise issues at coverage comparable to that of another human reviewer, while additionally surfacing a distinctive set of issues no human raises: a single AI reviewer recovers 27.1\% of a human reviewer's items (versus 25.8\% recovered by another human), and roughly one quarter of AI items have no similar human counterpart. However, AI reviewers overlap more substantially with each other (21.0\% for AI-AI pairs versus 3.1\% for human-human pairs), which indicates that introducing a panel of AI reviewers would likely harm diversity of perspective (\autoref{sec:overlap}). Third, AI reviewers exhibit characteristic weaknesses humans do not share: we identify 16 recurring failure modes from qualitative feedback, three of which account for most incorrect items, namely limited grasp of \textit{subfield-specific methodological conventions}, \textit{losing track of content} across long papers and supplementary materials, and an \textit{overly critical stance} that inflates minor issues (\autoref{sec:strengths_weaknesses}).\\

Based on these findings, we release two resources for the AI and scientific communities. \textsc{PeerReview Bench} is a benchmark that automatically applies our expert evaluation criteria, supporting continued tracking of AI reviewer quality without repeating the costly expert annotation as language models advance; even GPT-5.4, DeepSeek-V4-Pro, and Claude-Opus-4.7 achieve only 41.4\%, 48.5\%, and 50.5\% F1, respectively, leaving substantial headroom for improvement (\autoref{subsec:peerreview_bench}). \textsc{CMU Paper Reviewer} is an open-source AI reviewer service built on the script we used in our expert annotation study, providing authors with pre-submission feedback on their manuscripts; its review items are more often correct, significant, and well-evidenced than those from existing platforms (95.5\% vs.\ 59.8\% and 57.6\% for the Stanford Agentic Reviewer and OpenAIReview, respectively) (\autoref{subsec:cmu_paper_reviewer}). We hope these findings and resources contribute to a more constructive, evidence-based discussion of AI reviewer deployment.

\begin{figure}[t]
    \centering    \includegraphics[width=1\linewidth]{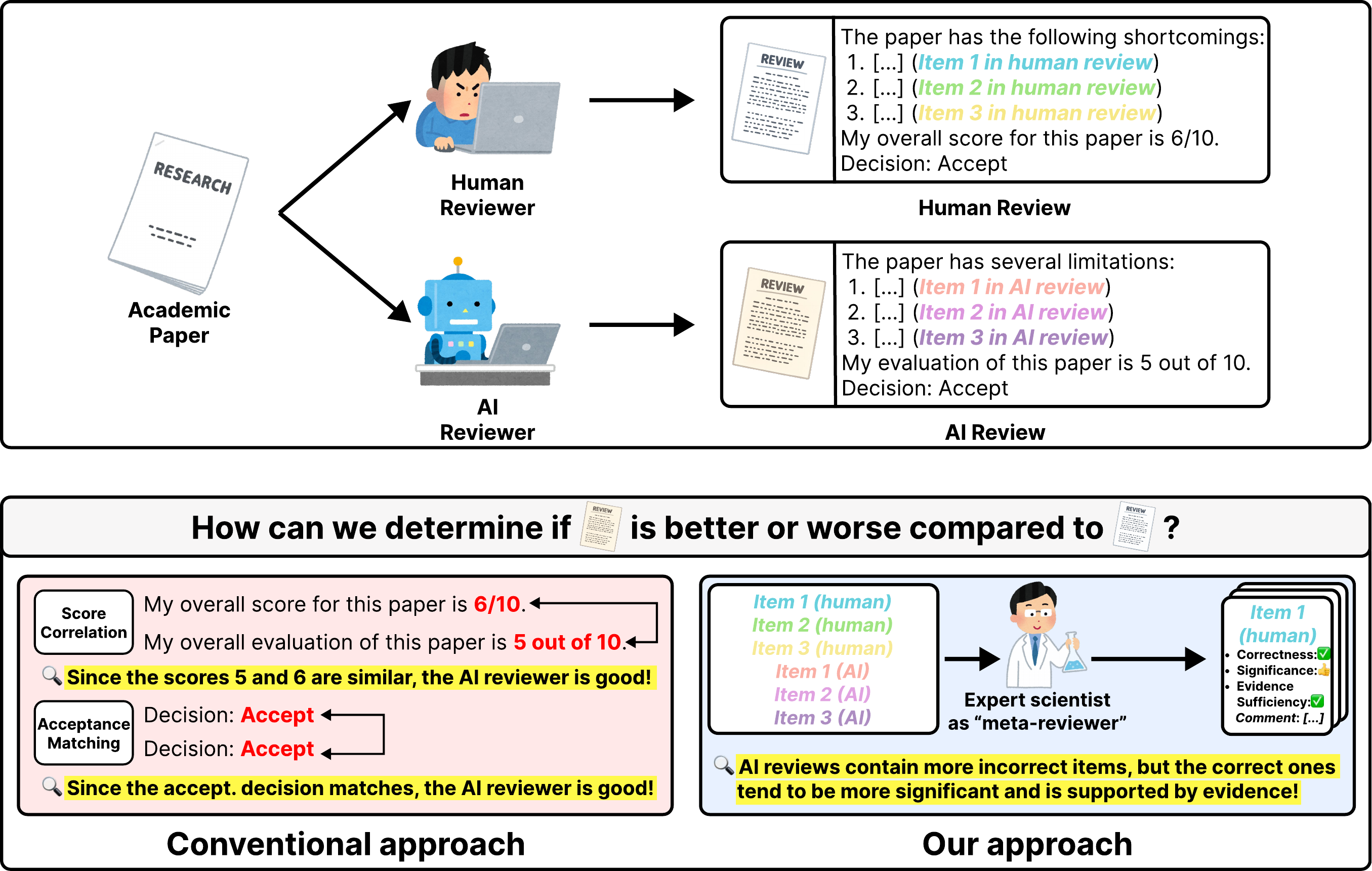}
    
    \caption{\textbf{Illustration of the motivation behind our expert annotation study.} Given a human-written review and an AI-generated review based on the same academic paper, prior works used shallow heuristics such as score correlation and acceptance matching to determine the quality of the AI-generated review. However, producing similar scores or matching accept/reject recommendations doesn't indicate that the AI-generated reviews contain useful feedback nor does it explain whether the criticisms raised by human and AI reviewer are similar. Hence, in our study, we recruit 45 scientists and judge every review item raised by human and AI reviewers for comparison.}
    \label{fig:motivation}
\vspace{-5pt}
\end{figure}

\section{Preliminaries: Expert annotation study design and experimental setup}\label{sec:preliminaries}

\paragraph{Motivation} As shown in \autoref{fig:motivation}, prior evaluations of AI reviewers have predominantly compared AI- and human-produced reviews at the aggregate output level (e.g., correlating overall scores or matching accept/reject verdicts). These aggregate views conceal what matters most in practice: at the level of each individual criticism, are AI-raised criticisms correct, do they address significant aspects of the paper, and are they supported by sufficient evidence? This is crucial because two reviews can produce similar overall scores while raising entirely different sets of criticisms, and two AI reviewers can look comparably competent in aggregate while one inflates minor concerns and another misses methodological flaws that any domain expert would catch. We therefore design an evaluation that (i) operates at the level of individual criticisms rather than aggregate scores; (ii) decomposes review quality into separable dimensions, since AI and human reviewers can differ in opposite directions across them; and (iii) spans scientific disciplines beyond AI itself, since a vast majority of prior AI-reviewer research has been conducted by AI researchers evaluating AI reviewers on AI papers (typically using OpenReview data from venues such as ICLR), and AI reviewer behavior on papers from the physical, biological, and health sciences remains largely uncharacterized.

\subsection{Methodology for reviewing a review}\label{subsec:rubric}
 
\paragraph{Definition of a review item} We define a \textbf{review item} as a single atomic criticism directed at one specific aspect of the paper. This is our unit of analysis throughout this work, in contrast to prior evaluations of AI reviewers that compare reviews at the aggregate level (e.g., overall score, accept/reject verdict). Specifically, a single peer review typically contains multiple distinct criticisms, and although separating a free-text review into atomic criticisms is non-trivial in general, human reviewers themselves conventionally use bullet points, explicit enumeration markers (e.g., ``First,'' ``Second,''), or paragraph transitions to demarcate the points they want authors to address in revision. We rely on these conventional markers to manually decompose each human peer review into review items. \autoref{fig:review-item-example1} shows two example review items extracted from the same human peer review of a physical-sciences paper.

\begin{figure}[!t]
\centering

\begin{mdframed}[
    style=reviewbox,
    frametitle={\textbf{An example of review item \# 1 from a human review}},
]
\footnotesize

\textit{``In Fig. 2, the appropriate comparison could include Carr-Parrinello MD since that is common to use for minor distortion of the geometry. At a minimum, mentioning CPMD in the main text would be less misleading to the reader.''}

\end{mdframed}


\begin{mdframed}[
    style=reviewbox,
    frametitle={\textbf{An example of review item \# 2 from a human review}},
]
\footnotesize

\textit{``Would the author advise using CASSCF, RASSCF, or better CASPT2 to properly include both static and dynamic correlations? That could represent an important application for the method, especially in the framework of dynamics. CAS-based dynamics is highly employed and reliable, being able to accelerate it by reducing the number of electronic structure calls using the approach proposed would be an important achievement.''}

\end{mdframed}

\caption{Two example review items written by the same human reviewer of a paper in the Physical Sciences. Each \textbf{review item} corresponds to a single atomic criticism directed at one aspect of the paper, making up the entire review.}
\vspace{-3mm}
\label{fig:review-item-example1}
\end{figure}

\paragraph{Evaluation criteria for evaluating a review item} During our expert annotation study, we ask domain scientists with subfield-matched expertise to rate every review item in their assigned paper along three dimensions:
\begin{itemize}[leftmargin=*]
    \item \textbf{``Correctness'' of the critique} (binary): whether the main point of the criticism is correct (i.e., the issue it raises actually exists in the paper rather than being a misreading of the manuscript) and is clearly stated.
    \item \textbf{``Significance'' of the critique} (ordinal, three-level): conditional on the criticism being correct, whether it addresses a significant aspect of the paper. The three levels are \emph{Significant} (an insightful concern that, if addressed, would meaningfully improve the paper), \emph{Marginally Significant} (e.g., typos or stylistic issues), and \emph{Not Significant} (a minor item that would be better removed from the review).
    \item \textbf{``Sufficiency of evidence'' of the critique} (binary): conditional on the criticism being correct and at least marginally significant, whether the evidence accompanying the criticism (e.g., quotes from the paper's main text, supplementary materials, or external references) is sufficient to support the main point.
\end{itemize}

The \textbf{cascading structure} reflects the logical dependency among the three dimensions: significance is only meaningful for criticisms that are correct, and evidence sufficiency is only meaningful for criticisms that are correct and at least marginally significant. We chose a three-dimensional design rather than a single overall rating so that we can identify the specific aspects in which AI reviewers are better or worse than human reviewers.

\subsection{Scope of papers for the expert annotation study}\label{subsec:dataset}

\begin{wraptable}{r}{0.56\textwidth}
\vspace{-10pt}
\centering
\fontsize{9}{11}\selectfont
\setlength{\tabcolsep}{4pt}
\begin{tabular}{lrrrr}
\toprule
\textbf{Component} & \textbf{Mean} & \textbf{Median} & \textbf{Min} & \textbf{Max} \\
\midrule
\multicolumn{5}{l}{\textit{Manuscript content}} \\
\# Words in main manuscript & 8,820 & 8,500 & 3,787 & 22,830 \\
\# Supplementary files per paper & 5.5 & 2 & 0 & 129 \\
\# Figures per paper & 4.0 & 4 & 0 & 19 \\
\# Code files per paper & 296.7 & 18 & 0 & 8,285 \\
\midrule
\multicolumn{5}{l}{\textit{Peer review}} \\
\# Human reviews per paper & 2.9 & 3 & 1 & 5 \\
\# Review items per human review & 8.8 & 7 & 1 & 72 \\
\# Review items per AI review & 4.2 & 4 & 2 & 5 \\
\bottomrule
\end{tabular}
\caption{Statistics of the 82-papers in the expert annotation study.}
\vspace{-5pt}
\label{tab:dataset-stats}
\end{wraptable}
 
\paragraph{Paper selection criteria} For our expert annotation study, we chose 82 papers from \textit{Nature} and its sister journals, spanning the physical, biological, and health sciences. Papers are included only if they meet three criteria: (1) a publicly released set of official human peer reviews under \textit{Nature}'s transparent peer review policy, so that we have human reviews for comparison; (2) a publicly available pre-review version of the manuscript on Research Square\footnote{\href{https://www.researchsquare.com/}{https://www.researchsquare.com/}}, so that AI reviewers evaluate the same manuscript that the human reviewers did; and (3) a subfield match with one of our recruited domain scientists, so that every review item can be annotated by an expert with relevant methodological knowledge. These three constraints jointly narrow the candidate pool substantially: public peer review is not the default at most venues, pre-review versions are rarely available after publication, and subfield-specific expert recruitment further restricts viable subjects.

\newsavebox{\reviewcodebox}
\begin{lrbox}{\reviewcodebox}
\begin{minipage}{0.85\linewidth}
\begin{lstlisting}[style=reviewcode]
if hermitian is True:
    # Solve the generalized eigenvalue problem for Hermitian Hamiltonian
    vals, vecs = eigh(H, S)
\end{lstlisting}
\end{minipage}
\end{lrbox}

\begin{figure}[!t]
\centering
\begin{mdframed}[
    style=reviewbox,
    frametitle={\textbf{An example of review item \#1 from an AI review}},
]
\footnotesize

\textcolor{blue!60!black}{\textbf{Claim: ``The generalized eigenproblem is not made numerically robust''}}

\vspace{2pt}
\colorbox{blue!5}{\begin{minipage}{0.97\linewidth}
\vspace{4pt}
\begin{itemize}[leftmargin=1.2em, itemsep=2pt, topsep=0pt]
\item \textbf{Main point of criticism:} The paper's implementation of the core generalized eigenvalue solve is not numerically hardened against overlap-matrix ill conditioning, so the reported active-learning MD workflow can fail unpredictably once training states become nearly linearly dependent.
\item \textbf{Evaluation criteria:} Data and methodology; Validity
\end{itemize}
\vspace{2pt}
\end{minipage}}

\vspace{8pt}
\textcolor{blue!60!black}{\textbf{Evidence:}}

\vspace{2pt}

\colorbox{blue!5}{\begin{minipage}{0.97\linewidth}
\vspace{4pt}
\textcolor{blue!50!black}{\textit{\textbf{Quote 1}} (from main text):}
\textit{``This is achieved in closed form as the diagonalization of a generalized eigenvalue problem in the basis of the training states.''}
\\
\commentbar{This is the numerical kernel of the method, so its stability determines whether the algorithm can actually be used in the long active-learning trajectories described in the paper. Because the training basis is enlarged sequentially, the overlap matrix can become nearly singular as similar geometries are added. The manuscript does not describe any canonical orthogonalization, eigenvalue truncation, or other conditioning step for this solve. That omission is a material reproducibility risk, not a minor implementation detail. A paper that claims black-box molecular dynamics should show how this step remains stable as the basis grows.}

\vspace{2pt}
\end{minipage}}

\vspace{4pt}

\colorbox{blue!5}{\begin{minipage}{0.97\linewidth}
\vspace{4pt}
\textcolor{blue!50!black}{\textit{\textbf{Quote 2}} (from submitted source code):}

\vspace{4pt}
\usebox{\reviewcodebox}
\vspace{4pt}

\commentbar{The implementation delegates the solve directly to \texttt{scipy.linalg.eigh} with the overlap matrix passed in unchanged. There is no diagnostic for the condition number of \texttt{S}, no thresholding of small eigenvalues, and no fallback path if the basis is nearly linearly dependent. A minimal reproduction of this code on a singular overlap matrix raises an exception, which shows the failure mode is real rather than hypothetical. Because the MD loop depends on repeating this operation many times, the current implementation is too brittle for the robustness claims made in the manuscript. The code needs a conditioning strategy before the method can be treated as reliably black-box.}

\vspace{2pt}
\end{minipage}}

\vspace{4pt}

\colorbox{blue!5}{\begin{minipage}{0.97\linewidth}
\vspace{4pt}
\textcolor{blue!50!black}{\textit{\textbf{Quote 3}} (from external reference, SciPy documentation):}
\textit{``If eigenvalue computation does not converge, an error occurred, or b matrix is not definite positive. Note that if input matrices are not symmetric or Hermitian, no error will be reported but results will be wrong.''}
\\
\commentbar{This warning from SciPy documents the failure mode that the current code is exposed to when the overlap matrix becomes singular or nearly singular. The paper does not report any safeguards to ensure that every reported run stays safely away from that regime. A direct reproduction of the current solver on a singular overlap matrix triggers exactly this kind of \texttt{LinAlgError}, showing that the instability is not merely theoretical. To support its robustness claims, the manuscript should either regularize the overlap matrix or report the conditioning of \texttt{S} for the presented trajectories. Without that, the method's reliability remains unproven.}

\vspace{2pt}
\end{minipage}}

\end{mdframed}
\caption{An example of a review item produced by an AI reviewer for the same paper as \autoref{fig:review-item-example1}. A review item consists of a \textbf{Claim} (main point of criticism and evaluation criteria) and \textbf{Evidence} (a set of supporting quotes with interpretive comments). Evidence quotes may be drawn from the paper's main text, supplementary materials, the source code, or external references. We instruct AI reviewers to produce at most five review items per paper.}
\label{fig:review-item-example2}
\vspace{-10pt}
\end{figure}

\paragraph{Dataset statistics} The 82 papers are drawn from \textit{Nature Communications} (73 papers), \textit{Nature} (2), \textit{Nature Computational Science} (2), \textit{Nature Ecology \& Evolution} (2), \textit{Nature Methods} (1), \textit{Nature Physics} (1), and \textit{Nature Microbiology} (1)\footnote{All manuscripts and peer review files in our dataset are released by their publishers under a CC BY 4.0 license.}, published between 10 January 2020 and 27 October 2025. Following the \textit{Nature Communications} subject taxonomy\footnote{\href{https://www.nature.com/ncomms/browse-subjects}{https://www.nature.com/ncomms/browse-subjects}}, the 82 papers span 27 subject categories: 38 in Physical Sciences, 30 in Biological Sciences, and 14 in Health Sciences. Beyond the main manuscript text, most submissions include additional components that AI reviewers can also access: 83\% have supplementary materials, 76\% have separately submitted figures, and 74\% have submitted source code. Per-paper content and review statistics are summarized in \autoref{tab:dataset-stats}.

\subsection{Reviewers: official human reviewers and frontier-LLM agents}\label{subsec:reviewers}
 
\paragraph{Human reviewers} For each paper, we use the first-round official human peer reviews released by \textit{Nature}-family journals, retaining the first three reviewers when more than three are present. Each review is decomposed into review items per \autoref{subsec:rubric}. Further details are in \autoref{appendix:peer_review_processing}.
 
\paragraph{AI reviewers} We use three frontier language models as AI reviewers: GPT-5.2, Claude Opus 4.5, and Gemini 3.0 Pro. Each model is deployed as an agent through the OpenHands software-agent-sdk \citep{wang2026openhandssoftwareagentsdk}, with filesystem access to the paper's source files (main text, supplementary materials, figures, and submitted code) and a small set of tools (shell terminal, file editor, task tracker, and a web-search tool with the paper's publisher domains blocked to prevent retrieval of the published version or peer review report). Each agent receives a prompt asking it to produce up to five review items per paper according to the six \textit{Nature} peer-review evaluation criteria, with each item structured as a \textbf{main claim} (the central point of criticism, with its associated criterion) followed by \textbf{supporting evidence} (a set of quotes from the paper's main text, supplementary materials, submitted source code, or external references, each accompanied by an interpretive comment). \autoref{fig:review-item-example2} shows an example AI-produced review item for the same paper as \autoref{fig:review-item-example1}. We generated one review per (paper, model) pair. Further details are in \autoref{appendix:ai_reviewer_details}.

\subsection{Meta-reviewers: domain scientist annotators}\label{subsec:annotators}
 
\paragraph{Meta-reviewer recruitment} Our annotator pool comprises 45 domain scientists from 25 institutions: 23 faculty members, 7 research scientists at industrial labs, national laboratories, or research institutes, 6 postdoctoral researchers, and 9 Ph.D. students. They produced 109 meta-reviews across the 82 papers (averaging 2.42 papers per scientist), totaling 469 hours of expert annotation. Further details are in \autoref{appendix:recruitment}.

\begin{wraptable}{r}{0.52\textwidth}
\centering
\fontsize{7}{9}\selectfont
\setlength{\tabcolsep}{3pt}
\begin{tabular}{lrccc}
\toprule
\textbf{Dimension} & $N$ & \textbf{\%Agree.} & \textbf{Cohen's $\kappa$} & \textbf{Gwet's AC1} \\
\midrule
Correctness (binary) & 908 & 85.8\% & 0.28\;{\scriptsize(fair)} & 0.82\;{\scriptsize(alm. perf.)} \\
Significance (3-way ord.) & 743 & 59.9\% & 0.31\;{\scriptsize(fair)} & 0.44\;{\scriptsize(moderate)} \\
Evidence (binary) & 583 & 88.0\% & 0.12\;{\scriptsize(slight)} & 0.86\;{\scriptsize(alm. perf.)} \\
\bottomrule
\end{tabular}
\caption{Inter-annotator agreement on 908 doubly-annotated review items from 27 doubly-annotated papers. $N$ decreases across dimensions due to the cascading annotation structure. Interpretation labels follow \citet{landis1977measurement}.}
\label{tab:irr}
\vspace{-3mm}
\end{wraptable}
 
\paragraph{Inter-annotator agreement} To measure inter-annotator agreement, 27 of the 82 papers were independently annotated by a second domain scientist, yielding 908 doubly-annotated review items. Because the marginal class distributions of our annotations are highly skewed, we report Gwet's AC1 alongside raw percent agreement and Cohen's $\kappa$ in \autoref{tab:irr}. Agreement is almost perfect for the two binary dimensions (correctness and evidence sufficiency) and moderate for the three-level significance scale; the substantial gap between $\kappa$ and AC1 on the two skewed binary dimensions illustrates the well-known kappa paradox under skewed marginals, motivating our choice to report AC1 as the primary chance-corrected measure. The full IRR analysis is in \autoref{appendix:irr_details}.

\paragraph{Gathering qualitative feedback} Beyond the structured judgments above, each meta-reviewer provides free-form responses at both the review-item and paper level. For each AI reviewer, they note any criticism the AI raised that other reviewers (AI or human) missed, and may annotate individual items with optional comments. We analyze these qualitative responses in \autoref{sec:strengths_weaknesses} to characterize systematic strengths and weaknesses of AI reviewers.

\paragraph{Paper-level overall survey} After completing the item-level annotations for a paper, the meta-reviewer also completes a paper-level overall survey: (i) selecting the \textbf{Top-Rated Human Reviewer} and \textbf{Lowest-Rated Human Reviewer} from the paper's official human reviewers based on overall review quality, (ii) indicating which AI reviewers match or exceed each of these two human references on overall quality, and (iii) optionally noting any items that a given reviewer raised which other reviewers (human or AI) missed. These paper-level judgments provide the top-rated and lowest-rated human baselines against which AI reviewers are compared in \autoref{sec:performance} and \autoref{sec:overlap}.
 
\paragraph{Primary unit of analysis in expert annotation study} Throughout \autoref{sec:performance} and \autoref{sec:overlap}, the unit of analysis is the paper ($n = 82$) rather than the review item ($N = 2{,}960$). For each paper we pool item-level annotations: items from the 27 doubly-annotated papers contribute two annotation rows each, while items from the remaining 55 contribute one row. This respects within-paper correlation among items rated by the same domain scientist. Reviewer-specific aggregation choices (means, paired tests, bootstrap CIs) are further described inline in each results section.


\section{In which aspects are AI reviewers better or worse than human reviewers?}\label{sec:performance}
 
\paragraph{Motivaiton \& experimental setting} To compare the three AI reviewers against the Top-Rated and Lowest-Rated Human Reviewer baselines (\autoref{subsec:annotators}), we examine each of the three rubric dimensions from \autoref{subsec:rubric} and an integrative indicator, \textit{fully positive}: a review item is fully positive iff it is rated Correct, Significant at the highest level on the 0--2 ordinal scale, \emph{and} Sufficient on evidence. The unit of analysis is the paper ($n = 82$); inferential comparisons use paired $t$-tests (with Cohen's $d$) for binary metrics and the Wilcoxon signed-rank test (with rank-biserial correlation $r$) for the ordinal significance score, all with 95\% bootstrap confidence intervals (10{,}000 paper-level resamples, percentile method). Item counts differ across reviewers because AI reviews are capped at five items per paper while human reviews are not, a stricter bar for AI; paper-level aggregation removes this asymmetry inferentially, and the item-level rates and Generalized Linear Mixed Model (GLMM) analysis in \autoref{appendix:extended_analysis_results} reach the same conclusions.
 
\begin{table}[t]
\centering
\fontsize{7.5}{10}\selectfont
\setlength{\tabcolsep}{3pt}
\begin{tabular}{lccccc}
\toprule
& \textbf{Correctness (\%)} & \textbf{Significance (0--2)} & \textbf{Evidence (\%)} & & \\
\textbf{Reviewer} & {\scriptsize mean [95\% CI]} & {\scriptsize mean [95\% CI]} & {\scriptsize mean [95\% CI]} & $n_{\text{items}}$ & $n_{\text{papers}}$ \\
\midrule
Top-Rated Human & 92.3{\scriptsize\,[89.2, 95.0]} & 1.39{\scriptsize\,[1.30, 1.49]} & 92.2{\scriptsize\,[88.5, 95.3]} & 1{,}139 & 82 \\
Lowest-Rated Human & 79.1{\scriptsize\,[73.0, 84.6]} & 1.30{\scriptsize\,[1.18, 1.42]} & 89.7{\scriptsize\,[84.7, 94.0]} & 833 & 82 \\
\midrule
GPT-5.2 & 86.2{\scriptsize\,[80.7, 91.2]} & 1.61{\scriptsize\,[1.50, 1.70]} & 97.1{\scriptsize\,[93.7, 99.5]} & 442 & 81 \\
\rowcolor{gray!6}
\multicolumn{1}{r}{{\scriptsize\textcolor{gray!70}{vs Top-Rated Human}}} & {\scriptsize $d{=}{-}0.225$,\; $p{=}.046^{*}$} & {\scriptsize $r{=}{+}0.491$,\; $p{<}.001^{*}$} & {\scriptsize $d{=}{+}0.233$,\; $p{=}.041^{*}$} & & \\
\rowcolor{gray!6}
\multicolumn{1}{r}{{\scriptsize\textcolor{gray!70}{vs Lowest-Rated Human}}} & {\scriptsize $d{=}{+}0.220$,\; $p{=}.051$} & {\scriptsize $r{=}{+}0.562$,\; $p{<}.001^{*}$} & {\scriptsize $d{=}{+}0.234$,\; $p{=}.048^{*}$} & & \\[3pt]
 
Claude Opus 4.5 & 83.7{\scriptsize\,[78.2, 88.6]} & 1.53{\scriptsize\,[1.43, 1.63]} & 96.5{\scriptsize\,[93.1, 99.1]} & 475 & 81 \\
\rowcolor{gray!6}
\multicolumn{1}{r}{{\scriptsize\textcolor{gray!70}{vs Top-Rated Human}}} & {\scriptsize $d{=}{-}0.337$,\; $p{=}.003^{*}$} & {\scriptsize $r{=}{+}0.301$,\; $p{=}.028^{*}$} & {\scriptsize $d{=}{+}0.243$,\; $p{=}.035^{*}$} & & \\
\rowcolor{gray!6}
\multicolumn{1}{r}{{\scriptsize\textcolor{gray!70}{vs Lowest-Rated Human}}} & {\scriptsize $d{=}{+}0.134$,\; $p{=}.230$} & {\scriptsize $r{=}{+}0.434$,\; $p{=}.003^{*}$} & {\scriptsize $d{=}{+}0.274$,\; $p{=}.022^{*}$} & & \\[3pt]
 
Gemini 3.0 Pro & 81.9{\scriptsize\,[76.5, 87.1]} & 1.56{\scriptsize\,[1.46, 1.65]} & 89.5{\scriptsize\,[84.0, 94.2]} & 460 & 82 \\
\rowcolor{gray!6}
\multicolumn{1}{r}{{\scriptsize\textcolor{gray!70}{vs Top-Rated Human}}} & {\scriptsize $d{=}{-}0.424$,\; $p{<}.001^{*}$} & {\scriptsize $r{=}{+}0.423$,\; $p{=}.003^{*}$} & {\scriptsize $d{=}{-}0.099$,\; $p{=}.380$} & & \\
\rowcolor{gray!6}
\multicolumn{1}{r}{{\scriptsize\textcolor{gray!70}{vs Lowest-Rated Human}}} & {\scriptsize $d{=}{+}0.088$,\; $p{=}.428$} & {\scriptsize $r{=}{+}0.460$,\; $p{=}.001^{*}$} & {\scriptsize $d{=}{-}0.012$,\; $p{=}.918$} & & \\
\bottomrule
\end{tabular}
\caption{\textbf{AI reviewers identify more significant issues but with lower factual correctness than the top-rated human reviewers.} Paper-level means with 95\% bootstrap CIs. Shaded rows: pairwise effect sizes and $p$-values versus each human reference (positive = AI higher; $d$ = Cohen's $d$ for binary metrics, $r$ = rank-biserial for ordinal significance). $^{*}p < 0.05$. See \autoref{appendix:extended_analysis_results} for item-level rates and a GLMM robustness analysis.
}
\label{tab:main-results}
\vspace{-10pt}
\end{table}
 
\begin{table}[t]
\centering
\fontsize{7}{9}\selectfont
\setlength{\tabcolsep}{4pt}
\begin{tabular}{l cc cc cc}
\toprule
& \multicolumn{2}{c}{\textbf{Per-group ``fully positive'' rate}} & & & & \\
\cmidrule(lr){2-3}
\textbf{Reviewer} & {\scriptsize Paper-level mean [95\% CI]} & {\scriptsize Item-level rate [95\% CI]} & \textbf{Fully Positive} & \textbf{Total Items} & \textbf{Papers} & \\
\midrule
Top-Rated Human    & 48.2{\scriptsize\,[42.2, 54.3]} & 47.8{\scriptsize\,[44.9, 50.7]} & 544 & 1{,}139 & 82 & \\
Lowest-Rated Human & 36.2{\scriptsize\,[30.0, 42.4]} & 29.3{\scriptsize\,[26.3, 32.5]} & 244 & 833     & 82 & \\
GPT-5.2            & 60.0{\scriptsize\,[52.3, 67.4]} & 58.6{\scriptsize\,[54.0, 63.1]} & 259 & 442     & 81 & \\
Claude Opus 4.5    & 53.1{\scriptsize\,[45.6, 60.7]} & 54.7{\scriptsize\,[50.2, 59.2]} & 260 & 475     & 81 & \\
Gemini 3.0 Pro     & 50.2{\scriptsize\,[42.7, 57.7]} & 51.3{\scriptsize\,[46.7, 55.8]} & 236 & 460     & 82 & \\
\midrule
\multicolumn{7}{@{}l}{\textbf{Pairwise paired differences} {\scriptsize (paper-level, row $-$ column; paired $t$-test)}} \\
\midrule
\textbf{Comparison} & \textbf{Diff} & \textbf{95\% CI (diff)} & \textbf{Cohen's $d$} & \textbf{$p$-value} & $n_{\text{paired}}$ & \\
\midrule
Top-Rated H. vs Lowest-Rated H.   & $+12.1\%$ & {\scriptsize $[+5.3, +18.9]$}    & $+0.39$ & $0.0007^{***}$ & 82 & \\
Top-Rated H. vs GPT-5.2           & $-11.6\%$ & {\scriptsize $[-20.3, -3.0]$}    & $-0.30$ & $0.0091^{**}$  & 81 & \\
Top-Rated H. vs Claude Opus 4.5   & $-4.9\%$  & {\scriptsize $[-13.0, +3.2]$}    & $-0.13$ & $0.2300$       & 81 & \\
Top-Rated H. vs Gemini 3.0 Pro    & $-2.0\%$  & {\scriptsize $[-10.6, +6.7]$}    & $-0.05$ & $0.6521$       & 82 & \\
\midrule
Lowest-Rated H. vs GPT-5.2        & $-23.6\%$ & {\scriptsize $[-32.3, -14.8]$}  & $-0.60$ & $<0.0001^{***}$ & 81 & \\
Lowest-Rated H. vs Claude Opus 4.5 & $-16.5\%$ & {\scriptsize $[-25.3, -7.6]$}   & $-0.41$ & $0.0004^{***}$  & 81 & \\
Lowest-Rated H. vs Gemini 3.0 Pro & $-14.1\%$ & {\scriptsize $[-23.4, -4.7]$}   & $-0.33$ & $0.0038^{**}$   & 82 & \\
\midrule
GPT-5.2 vs Claude Opus 4.5        & $+8.0\%$  & {\scriptsize $[+0.8, +15.3]$}    & $+0.25$ & $0.0294^{*}$   & 80 & \\
GPT-5.2 vs Gemini 3.0 Pro         & $+10.4\%$ & {\scriptsize $[+4.1, +16.8]$}    & $+0.36$ & $0.0016^{**}$  & 81 & \\
Claude Opus 4.5 vs Gemini 3.0 Pro & $+2.2\%$  & {\scriptsize $[-2.8, +7.3]$}     & $+0.10$ & $0.3784$       & 81 & \\
\bottomrule
\end{tabular}
\caption{\textbf{GPT-5.2 exceeds the top-rated human on aggregate review-item quality, and all three AI reviewers exceed the lowest-rated human.}
\textbf{Top:} per-group fully-positive rates aggregated paper-level (each paper weighted equally; bootstrap 95\% CI) and item-level (papers weighted by item count; Wilson 95\% CI).
\textbf{Bottom:} all 10 paired comparisons on per-paper rates, with paired difference (row $-$ column), 95\% CI, Cohen's $d$, and $p$-value; positive = first-named reviewer higher.
$^{*}p < 0.05$, $^{**}p < 0.01$, $^{***}p < 0.001$.}
\vspace{-3mm}
\label{tab:fully-positive}
\end{table}
 
\paragraph{AI reviewers are less correct but raise more significant issues than the top-rated human} \autoref{tab:main-results} reports paper-level means on each dimension together with pairwise effect sizes and $p$-values against each human baseline. On \textit{correctness}, all three AI reviewers fall below the Top-Rated Human (92.3\%) by 6 to 10 percentage points: GPT-5.2 reaches 86.2\% ($d = -0.23$, $p = .046$), Claude Opus 4.5 reaches 83.7\% ($d = -0.34$, $p = .003$), and Gemini 3.0 Pro reaches 81.9\% ($d = -0.42$, $p < .001$). On \textit{significance}, however, the direction reverses: among correct items, all three AI reviewers raise more significant criticisms than the Top-Rated Human (mean significance score 1.39 on the 0 to 2 scale), with rank-biserial correlations of $r = +0.49$ for GPT-5.2, $r = +0.30$ for Claude Opus 4.5, and $r = +0.42$ for Gemini 3.0 Pro (all $p \leq .028$). On \textit{sufficiency of evidence}, GPT-5.2 and Claude Opus 4.5 score slightly higher than the Top-Rated Human ($d = +0.23$ and $+0.24$ respectively, both $p < .05$), while Gemini 3.0 Pro is statistically indistinguishable ($d = -0.10$, $p = .380$). Compared to the Lowest-Rated Human, all three AI reviewers match or exceed on every dimension, with significance showing the largest gains ($r = +0.43$ to $+0.56$). Together, these results reveal a \textit{tradeoff between correctness and significance}: AI reviewers raise more significant issues with comparable or better evidence, but with lower correctness than the top-rated human reviewer.
 
\paragraph{On aggregate review-item quality, all three AI reviewers exceed the lowest-rated human, and GPT-5.2 exceeds the top-rated human} The dimension-level results above do not directly tell us what fraction of each reviewer's items would be fully useful to an author: a criticism that is correct but insignificant, or significant but insufficiently evidenced, provides little actionable feedback. We therefore examine the \textit{fully positive} rate, defined above. As shown in \autoref{tab:fully-positive}, the paper-level mean fully-positive rate is 48.2\% for the Top-Rated Human, 36.2\% for the Lowest-Rated Human, 60.0\% for GPT-5.2, 53.1\% for Claude Opus 4.5, and 50.2\% for Gemini 3.0 Pro. GPT-5.2 exceeds the Top-Rated Human by 11.6 percentage points ($d = -0.30$, $p = .009$; paired difference Top-Rated $-$ GPT-5.2), while Claude Opus 4.5 and Gemini 3.0 Pro are statistically indistinguishable from the Top-Rated Human ($p = .23$ and $p = .65$ respectively). All three AI reviewers substantially exceed the Lowest-Rated Human (differences of $+14.1$ to $+23.6$ percentage points, all $p \leq .004$). Among the AI reviewers, GPT-5.2 produces the highest-quality items on average, significantly above both Claude Opus 4.5 ($d = +0.25$, $p = .029$) and Gemini 3.0 Pro ($d = +0.36$, $p = .002$), which in turn are statistically indistinguishable from each other.
 
\begin{wraptable}{r}{0.52\textwidth}
\centering
\fontsize{9}{11}\selectfont
\setlength{\tabcolsep}{4pt}
\begin{tabular}{lcc}
\toprule
\textbf{Reviewer} & \textbf{vs Top-Rated} & \textbf{vs Lowest-Rated} \\
                  & {\scriptsize Win-rate [95\% CI]} & {\scriptsize Win-rate [95\% CI]} \\
\midrule
GPT-5.2         & 48.6\% {\scriptsize\,[38.7, 58.5]} & 73.4\% {\scriptsize\,[64.2, 82.4]} \\
Claude Opus 4.5  & 32.1\% {\scriptsize\,[22.5, 42.0]} & 68.8\% {\scriptsize\,[58.9, 78.4]} \\
Gemini 3.0 Pro & 30.3\% {\scriptsize\,[21.6, 39.4]} & 59.6\% {\scriptsize\,[50.4, 69.2]} \\
\bottomrule
\end{tabular}
\caption{\textbf{Expert scientists judge GPT-5.2 reviews to match or exceed the top-rated human reviewer's review on nearly half of all papers.} Fraction of papers where each AI reviewer matches or exceeds the human reference ($n = 109$ observations across 82 papers, 95\% cluster-bootstrap CIs).}
\label{tab:winrate}
\vspace{-8pt}
\end{wraptable}
 
\paragraph{Expert-judged win rates against human reviewers corroborate the aggregate picture} Beyond the item-by-item comparison, each domain scientist also provided a paper-level judgment of which AI reviewers matched or exceeded the overall quality of each human reference (\autoref{subsec:annotators}). As shown in \autoref{tab:winrate}, these holistic assessments closely track the aggregate quality results. GPT-5.2 is judged to match or exceed the Top-Rated Human on 48.6\% of papers and the Lowest-Rated Human on 73.4\%. Claude Opus 4.5 and Gemini 3.0 Pro trail GPT-5.2 substantially on the top-rated comparison (30.3\% and 32.1\%), but both clear the lowest-rated bar on a majority of papers (59.6\% and 68.8\%). This ordering is consistent with the fully-positive-rate ordering in \autoref{tab:fully-positive}, suggesting that expert holistic judgments are well-captured by the aggregate of per-item ratings.
 
\paragraph{Takeaway: AI reviewers raise more significant items but with lower correctness; only GPT-5.2 exceeds the top-rated human on aggregate} Taken together, the three analyses (dimension-level in \autoref{tab:main-results}, aggregate item quality in \autoref{tab:fully-positive}, and expert-judged paper-level matching in \autoref{tab:winrate}) agree on a consistent picture. Current frontier AI reviewers, when equipped with tool-based access to the paper's full source, its code, and external literature, produce review items that surface significant issues at a higher rate than even the best human reviewer of a paper, and is well-evidenced. They do so, however, at a cost in factual correctness: each AI reviewer raises a non-trivial fraction of items that experts judge to be incorrect or not clearly stated. The net effect as a whole is model-dependent: GPT-5.2's higher significance and evidence rates more than offset its correctness gap, so that on aggregate its reviews are judged to match or exceed the top human reviewer; for Claude Opus 4.5 and Gemini 3.0 Pro, the correctness gap is larger relative to their significance gains, placing them between the top-rated and lowest-rated human baselines. A GLMM analysis with paper-level random intercepts reaches the same conclusions (\autoref{appendix:extended_analysis_results}).

\section{To what extent do AI reviews overlap with human reviews?}\label{sec:overlap}
 
\paragraph{Motivation} Whereas \autoref{sec:performance} compared AI and human reviewers in terms of the quality of each individual review item, this section asks how the criticisms raised by different reviewers relate to one another, with neither positioned as the ground truth. The motivation is that multi-reviewer peer review draws much of its value from the \textit{diversity of perspectives} that different reviewers bring to a manuscript~\citep{page2008difference}: assigning a slate of reviewers with diverse expertise both improves coverage of the manuscript and reduces redundancy across reviews~\citep{goyal2024causal}. Whether AI reviewers contribute to or erode this diversity therefore depends not on their individual quality but on whether the targets they criticize overlap with those criticized by human reviewers.
 
\paragraph{Defining when two review items overlap} Each review item can be decomposed into three components. The \textbf{target} is the specific part of the paper being pointed at: a section, figure, equation, code function, supplementary file, or specific claim. The \textbf{criticism} is what the reviewer says about the target: what is wrong with it, what is missing, or what needs improvement. The \textbf{evidence} is what supports the criticism: quoted text from the manuscript, code blocks, or external references. To make the distinction concrete, if two reviewers both flag Figure~2 of a paper but one says the error bars are missing while the other says the color scheme is inaccessible, they share the same target (Figure~2) but raise different criticisms; if both say the error bars are missing while citing different sentences in the methods section as evidence, they share the same target and the same criticism but use different evidence.

Comparing two review items along these three components yields four mutually exclusive categories: (1) \textit{different target}; (2) \textit{same target, different criticism}; (3) \textit{same target, same criticism, different evidence}; and (4) \textit{same target, same criticism, same evidence} (near-paraphrases). We classify a pair as \textit{similar} if it shares the same target and the same criticism (categories 3 and 4), and \textit{not similar} otherwise. The taxonomy was derived from how domain scientists themselves treated overlap in the free-form paper-level survey responses; see \autoref{appendix:similarity_judge_calibration} for more details.
 
\paragraph{Automated similarity judging} The 65{,}704 cross-reviewer pairs are far too many to label manually. We use an LLM-based similarity judge (GPT-5.4) calibrated against a 164-pair set of manually annotated pairs, on which it achieves 92.7\% binary accuracy (similar vs.\ not similar) and 83.5\% accuracy on the full 4-way classification, with sensitivity 87.1\% and specificity 96.8\%. To convert the judge's apparent prevalence into an estimate of the true prevalence, we apply the Rogan-Gladen prevalence correction~\citep{rogan1978estimating}. All percentages reported in this section are Rogan-Gladen-corrected with 95\% cluster-bootstrap CIs (10{,}000 paper-level resamples) that propagate uncertainty in both the judge's error rates and cross-paper variance. Full details on the calibration set, the comparison of candidate judges, the confusion matrix, and the correction procedure are in \autoref{appendix:similarity_details}.

\begin{table*}[t]
\begin{minipage}[t]{0.45\textwidth}
\centering
\fontsize{6.5}{8.5}\selectfont
\begin{tabular}{l c c}
\toprule
\textbf{Metric} & \textbf{Value} & \textbf{95\% CI} \\
\midrule
Human items covered by 1 AI reviewer & 26.9\% & [21.3, 32.6] \\
Human items covered by 3 AI reviewers & 46.3\% & [39.4, 53.9] \\
Fully-positive human items covered by 1 AI & 36.3\% & [29.6, 43.6] \\
Fully-positive human items covered by 3 AI & 59.2\% & [50.2, 69.1] \\
\midrule
AI items covered by $\geq$1 human reviewer & 74.0\% & [65.5, 84.1] \\
\midrule
Uncovered AI items (no human match) & 26.0\% & [15.9, 34.5] \\
Uncovered AI items that are fully positive & 48.1\% & [39.6, 56.6] \\
Uncovered AI items that are correct & 81.8\% & [75.3, 87.9] \\
\bottomrule
\end{tabular}
\captionof{table}{\textbf{Coverage of human concerns by AI reviewers.} ``Covered'' = at least one cross-reviewer item pair is classified as similar (same target and same criticism). Per-reviewer-pair averaging at paper level; Rogan-Gladen-corrected.}
\label{tab:coverage-summary}
\end{minipage}%
\hfill
\begin{minipage}[t]{0.526\textwidth}
\centering
\fontsize{7.5}{9.5}\selectfont
\setlength{\tabcolsep}{3pt}
\begin{tabular}{l c c c c}
\toprule
\textbf{Metric} & \textbf{Uncovered} & \textbf{Matched} & $\chi^2$ & $p$-value \\
\midrule
Correctness & 81.8\% (79) & 84.4\% (81) & 1.45 & $0.229$ \\
Significance $\geq 1$$^{\dagger}$ & 87.4\% (75) & 91.8\% (80) & 2.81 & $0.094$ \\
Significance $= 2$$^{\dagger}$ & 57.6\% (75) & 63.5\% (80) & 14.98 & ${<}\,0.001^{***}$ \\
Evidence sufficient$^{\ddagger}$ & 93.5\% (71) & 93.2\% (80) & 0.00 & $0.971$ \\
\midrule
Fully positive (composite) & 48.1\% (79) & 54.8\% (81) & 13.71 & ${<}\,0.001^{***}$ \\
\bottomrule
\end{tabular}
\captionof{table}{\textbf{Uncovered AI items are equally correct and well-evidenced but less often rated as highly significant.} ``Uncovered''/``Matched'' = AI items with/without a similar human counterpart. $N$ (number of papers) in parentheses. $^{\dagger}$Among correct items (cascading filter). $^{\ddagger}$Among correct and at least marginally significant items. $^{***}p < 0.001$ ($\chi^2$ tests).}
\label{tab:uncovered-components}
\end{minipage}
\vspace{-3mm}
\end{table*}

\paragraph{Review items raised by AI reviewers but not by human reviewers are correct and well-evidenced} A natural first question is whether AI reviewers add anything that human reviewers do not, and whether the additions are valuable contributions or spurious artifacts. \autoref{tab:coverage-summary} shows that 74.0\% of AI-raised items have a similar counterpart in at least one human's review (we refer to these as \emph{matched}), leaving 26.0\% with no similar human counterpart (\emph{uncovered}). A component-level comparison of uncovered versus matched AI items (\autoref{tab:uncovered-components}) shows that uncovered items are not hallucinations and are not under-evidenced: their correctness rate (81.8\%) and evidence-sufficiency rate (93.5\%) are both statistically indistinguishable from those of matched AI items (84.4\% correct, $p = 0.23$; 93.2\% evidence-sufficient, $p = 0.97$). What separates the two groups is the fraction rated as \emph{highly} significant: 57.6\% for uncovered versus 63.5\% for matched ($p < 0.001$), with the same gap appearing across all three AI reviewer models. This does not contradict the \autoref{sec:performance} finding that AI reviewers as a whole raise more significant items than human reviewers: AI items that overlap with human concerns tend to be the items humans also flagged, while uncovered AI items, by definition, correspond to observations humans did not raise. We interpret that AI reviewers reliably surface the major concerns that human reviewers would also raise, and \emph{additionally} surface a non-trivial volume of valid, well-supported observations that humans miss, which tend to be relatively lower-priority items.

\begin{figure}[t]
    \centering
    \includegraphics[width=1\linewidth]{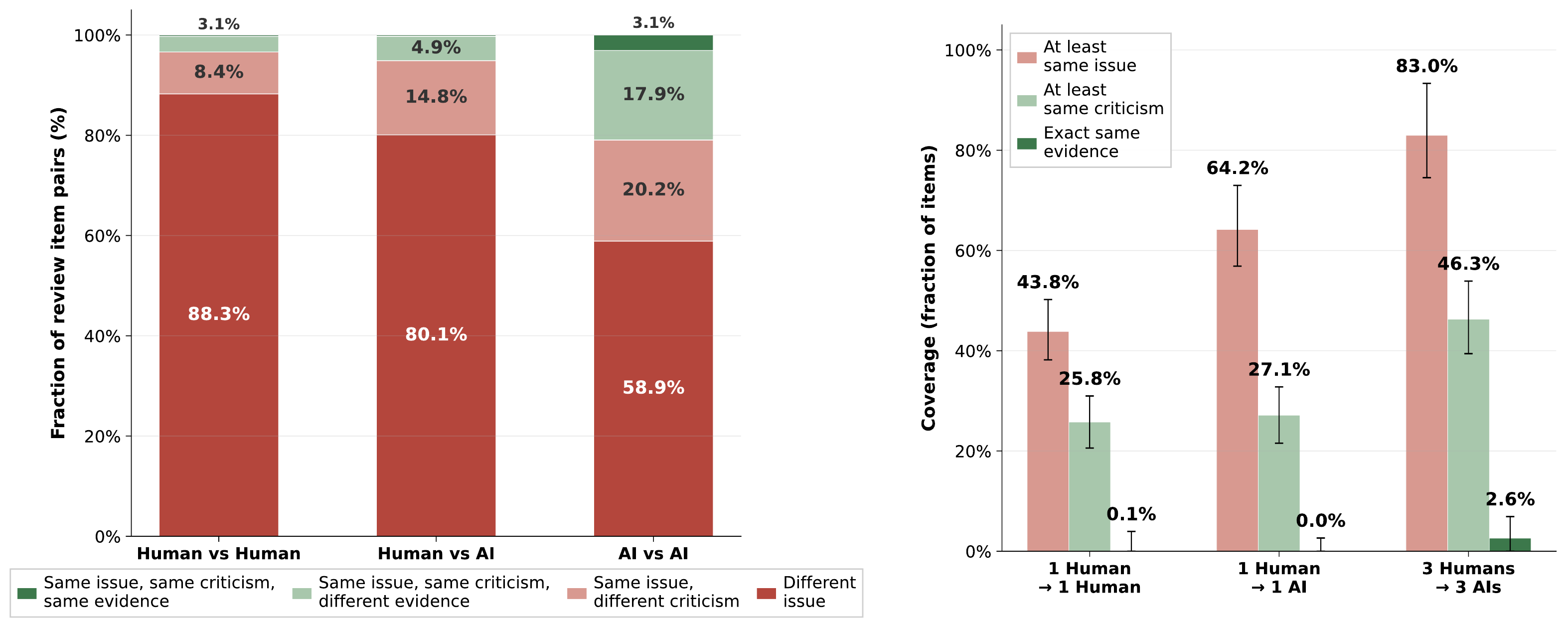}
    \caption{\textbf{AI reviewers overlap with each other much more than humans do, while AI panels match most human targets but only about half of the specific criticisms.}
    \textbf{(Left)} Distribution of cross-reviewer item pairs across the four similarity categories, for Human-Human, Human-AI, and AI-AI pair types.
    \textbf{(Right)} Fraction of one reviewer's items covered by another at three progressively stricter similarity thresholds: at least same target, at least same criticism, exact same evidence. The three configurations compare another human (reference), one AI reviewer, and the union of three AI reviewers. All percentages are Rogan-Gladen-corrected; error bars show 95\% cluster-bootstrap CIs. The full breakdowns are further explained in \autoref{appendix:similarity_breakdown}.}
    \label{fig:overlap}
    \vspace{-3mm}
\end{figure}

\paragraph{A panel of AI reviewers identifies most of the same targets as three humans, but delivers noticeably different feedback} The right panel of \autoref{fig:overlap} quantifies how much of the human review panel is surfaced by the AI panel. Replacing a single human reviewer with a single AI reviewer raises the same-target coverage of the remaining human's concerns from 43.8\% (human-vs-human) to 64.2\% (human-vs-AI), while the same-criticism coverage rises only marginally from 25.8\% to 27.1\%. Expanding to the full three-AI panel, the union of AI-raised items covers 83.0\% of the human-raised items at the same-target level but only 46.3\% at the same-criticism level. For authors, this means that if their three human reviewers were replaced with three AI reviewers, they would see feedback about most of the same parts of their paper but only about half of the specific criticisms with the same substance. This gap between target-level and criticism-level coverage adds a distinct perspective to the dimension-level comparison in \autoref{sec:performance}: AI reviewers and human reviewers converge on which parts of a paper warrant review but diverge in how they characterize what they find, meaning that an AI panel is not a drop-in replacement for a human panel.
 
\paragraph{Replacing one human reviewer with one AI reviewer minimally erodes panel diversity} \autoref{fig:overlap} (left) shows that although different AI reviewers overlap with each other (A--A, same target and same criticism: 20.9\% [16.2, 25.4]) roughly six times more than two human reviewers do (3.4\%), this high AI--AI overlap does not carry over to AI--human pairs. The rate at which an AI reviewer overlaps with a human reviewer (H--A: 5.1\% [0.3, 9.0]) is only slightly higher than the human--human baseline, and the confidence intervals of the two estimates overlap substantially. The coverage view in \autoref{fig:overlap} (right) gives the same ranking at the item level: the fraction of one human reviewer's items with a same-criticism counterpart rises only marginally when the second reviewer is an AI rather than another human (27.1\% for H--A versus 25.8\% for H--H). In other words, substituting a single AI reviewer for a human reviewer leaves the pairwise diversity of the panel nearly intact.
 
\paragraph{Human reviewers themselves surface largely disjoint sets of criticisms} In \autoref{fig:overlap} (left), the H--H baseline (3.4\% same-target-same-criticism) is itself a substantive finding worth reading carefully. Between two different human reviewers of the same paper, the remaining 96.6\% of pairs either raise different criticisms about the same target (8.4\%) or address entirely different targets (88.3\%); the coverage view shows that only 25.8\% of one human reviewer's items have a same-criticism counterpart in another human reviewer's review of the paper, and 43.8\% have any same-target counterpart at all. Human peer review in our dataset is therefore not a redundant exercise: each human reviewer brings a different set of concerns, providing empirical support for the \emph{diversity of perspectives} argument for peer review~\citep{page2008difference} and grounding the comparisons in the three preceding paragraphs.

\paragraph{Takeaway: AI reviewers can augment but not replace a human panel} Overall, AI reviewers behave less like substitutes for humans and more like an additional perspective that partially overlaps with the human one. Items that AI reviewers raise but humans do not are largely valid: 81.8\% are correct and 93.5\% are well-evidenced, with only a modest drop in the highest-significance rate compared with matched AI items. Substituting a single AI reviewer for a single human reviewer leaves pairwise panel diversity nearly intact, with H--A overlap (5.1\%) comparable to the 3.4\% H--H baseline. A panel of three AI reviewers, however, identifies most of the same parts of a paper as a human panel (83.0\% target-level coverage) but converges with humans on only about half of the specific criticisms (46.3\%); AI reviewers overlap with each other (20.9\% same-target-same-criticism) far more than humans do. These results argue for AI reviewers as augmentation to a human panel rather than as a drop-in replacement for it.

\begin{figure}[t]
    \centering
    \includegraphics[width=1\linewidth]{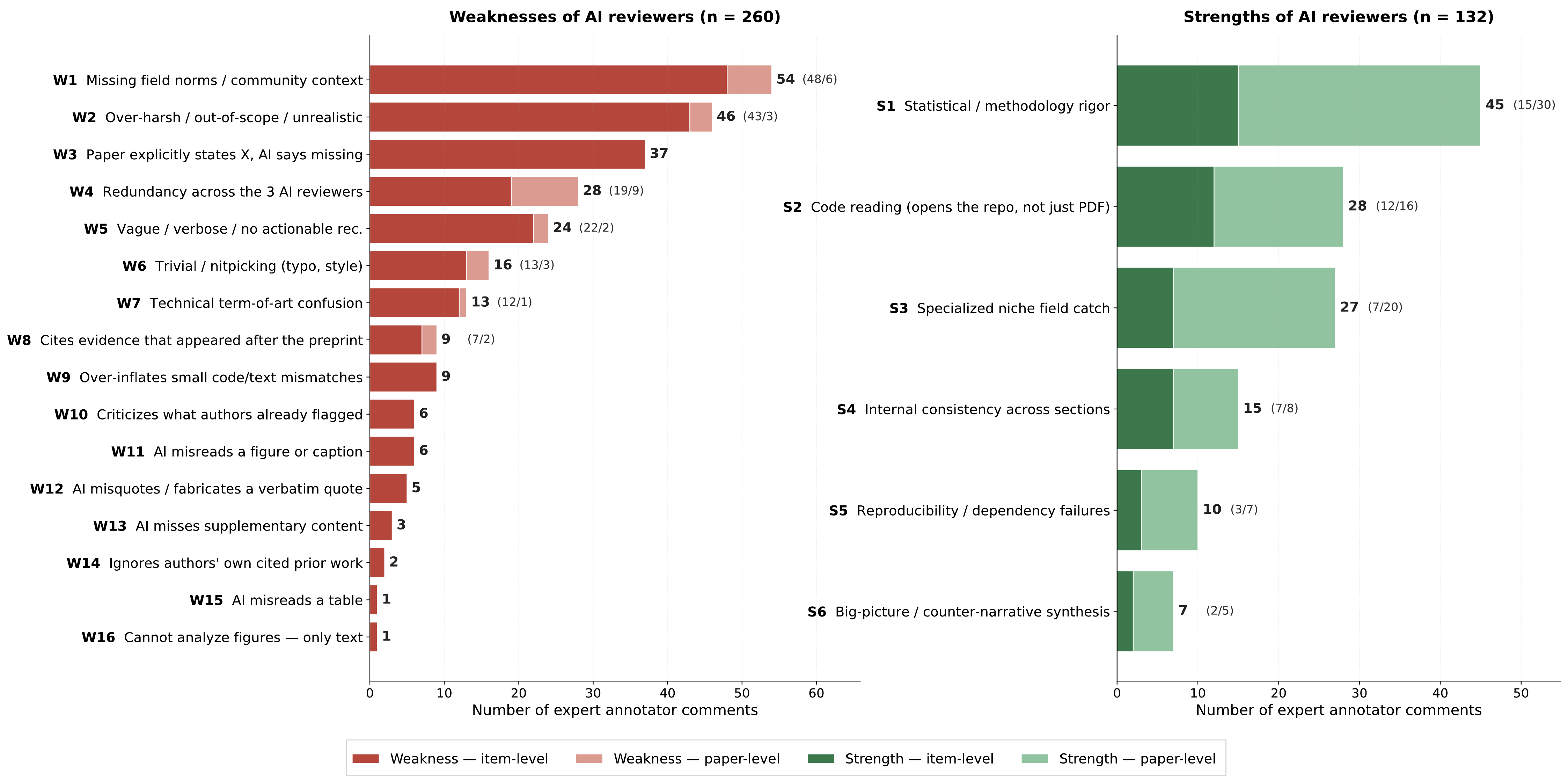}
    \caption{\textbf{Strengths and weaknesses of AI reviewers identified by domain experts.} Distribution of 442 free-form comments on AI reviews across 16 weakness categories (left, $n = 260$) and 6 strength categories (right, $n = 132$). Dark bars are item-level comments; light bars are paper-level comments. Categories are sorted by total count.}
    \label{fig:strength_weakness}
\vspace{-3mm}
\end{figure}

\section{What are the concrete strengths and weaknesses of AI reviewers?}\label{sec:strengths_weaknesses}

\paragraph{Motivation \& Setup} \autoref{sec:performance} and \autoref{sec:overlap} characterize how AI reviewers compare to human reviewers, but they do not describe the specific behaviors that drive those differences. For researchers studying AI review systems, editors considering AI-assisted review, and authors receiving AI-assisted feedback alike, a more pointed question is which behaviors of current AI reviewers most undermine their usefulness, and which behaviors genuinely outperform human reviewers and should be preserved or amplified. To address this, we examined the free-form comments that domain scientists provided alongside their item-level ratings (see \autoref{fig:annotation_sheet}), producing 767 item-level and 250 paper-level comments across the 82 papers. Of these, 392 comments pointed out a concrete strength or weakness of AI reviewers relative to human reviewers; we manually categorized them into 16 weakness ($n = 260$) and 6 strength ($n = 132$) categories (\autoref{fig:strength_weakness}). Below we discuss the most frequent categories and illustrate the three highest-signal patterns (one for the dominant weakness, one for the most concerning weakness, and one for the most differentiating strength) with one representative quote-comment pair each. Note that further explanation of the remaining categories with additional examples, and the full set of raw comments are all presented in \autoref{appendix:sw_all_comments}.

\subsection{Failure cases: Limitations of AI reviewers}\label{qualitative_analysis_weaknesses}
 
\paragraph{Overview} The five most frequently cited weaknesses account for roughly three quarters of all weakness comments ($189/260$), and they describe AI reviews as contextually uncalibrated rather than factually empty. The dominant theme is severity miscalibration against field-specific norms (\textbf{W1} missing community / field norms, $n = 54$; \textbf{W2} over-harsh or out-of-scope demands, $n = 46$): the AI reviewer's technical content is usually correct against a discipline-neutral standard for reproducibility, sample size, or out-of-distribution validation, but the severity is wrong because the reviewer lacks the accumulated knowledge of what the field treats as normal. The second theme is reading failure (\textbf{W3} paper explicitly states $X$, AI says missing, $n = 37$): the AI asserts that something is absent from the manuscript when in fact the paper provides it, whether in a different section, the supplementary material, or elsewhere in the same passage. The third theme is panel- and review-level form (\textbf{W4} redundancy across the three AI reviewers, $n = 28$; \textbf{W5} vague or verbose reviews without actionable recommendations, $n = 24$): the three AI reviewers tend to converge on overlapping criticisms (see also \autoref{sec:overlap}), and individual AI reviews are often long and dense without ending in a concrete revision request. We illustrate the dominant W1 pattern and the most directly damning W3 pattern with one example each; additional examples for these categories and the full set of examples for W2, W4, W5, and W8 (citing post-preprint evidence, $n = 9$) are in \autoref{appendix:sw_all_comments}.
 
\begin{mdframed}[style=categoryboxW, frametitle={\textbf{AI Reviewer Weakness \#1: Missing community / field norms ($n = 54$)}}]
\footnotesize
The AI reviewer flags something as a methodological gap that is in fact accepted or expected practice within the subfield of the paper. The content of the critique is typically accurate against a discipline-neutral standard (for reproducibility, sample size, out-of-distribution validation, and so on), but the severity is miscalibrated because the reviewer lacks the accumulated knowledge of what the field treats as normal.
\end{mdframed}
 
\begin{mdframed}[
    style=reviewbox,
    frametitle={\textbf{Example \#1 for W1 \textemdash{} Particle physics paper, reviewer: GPT-5.2}},
]
\footnotesize
 
\noindent\textcolor{blue!60!black}{\textbf{Claim: ``Analysis choices are not documented enough for independent reproduction''}}
 
\vspace{2pt}\par\noindent\colorbox{blue!5}{\begin{minipage}{0.97\linewidth}
\vspace{4pt}
\begin{itemize}[leftmargin=1.2em, itemsep=2pt, topsep=0pt]
\item \textbf{Main point of criticism:} The manuscript's analysis choices (simulation-fixed fit shapes, external calibration inputs, and pseudoexperiment-based uncertainties) are not documented in enough detail, and are not accompanied by reusable released artifacts, to enable independent reproduction or reanalysis of the key result.
\item \textbf{Evaluation criteria:} Data and methodology; Reproducibility
\end{itemize}
\vspace{2pt}
\end{minipage}}
 
\vspace{8pt}\par\noindent\textcolor{blue!60!black}{\textbf{Evidence:}}
 
\vspace{2pt}\par\noindent\colorbox{blue!5}{\begin{minipage}{0.97\linewidth}
\vspace{4pt}
\textcolor{blue!50!black}{\textit{\textbf{Quote 1}} (from main text):}
\textit{``The signal shape is modelled as a combination of a Gaussian and two Crystal Ball functions, all sharing the same peak position. The parameters of the signal function are determined from simulated events and are fixed in the fit to data, except for the Gaussian width, the average width of the Crystal Ball functions, and the peak position, which accounts for imperfections in the simulation.''}
\\
\commentbar{The extracted yields, and therefore the measured asymmetry, depend on simulation-derived shape parameters that are fixed rather than floated in the data fit. For an external reader, reproducing the measurement requires access not only to the data but also to the exact simulation configuration, the derived fixed parameter values, and the fitting implementation. The manuscript does not tabulate these fixed parameters or provide a machine-readable specification of the full likelihood model, which would make independent reproduction infeasible even in principle. Providing the parameter values and a compact statistical model description (or a public likelihood artifact) would significantly strengthen reproducibility.}
 
\vspace{2pt}
\end{minipage}}
 
\end{mdframed}
 
\begin{mdframed}[style=expertcommentbox]
\footnotesize\textbf{Expert comment.}~It is true that these information are not documented in the paper, however these are kept internally at CERN and the authors are not expected to share any data or simulation when publishing the paper for the full reproduction of the results. This is done in exceptional cases where relevant work has been done to provide data for example to the theory community. This comment  for this particular paper is irrelevant.
\end{mdframed}
\vspace{10pt}
 
The AI reviewer's request is reasonable by generic open-science standards: tabulated fit parameters, a machine-readable likelihood, and enough metadata for an external reanalysis. None of that is factually incorrect, and in some subfields it would be a legitimate revision request. But in papers produced at CERN (LHCb experiment as the current paper example), the relevant calibration and fit-configuration artifacts are maintained and used internally by the collaboration rather than published alongside the paper, and a specialist reviewer knows this by default. The critique is therefore not wrong in content so much as miscalibrated in severity: what the AI reads as a reproducibility gap is the community's normal publication boundary.
 
\begin{mdframed}[style=categoryboxW, frametitle={\textbf{AI Reviewer Weakness \#3: Paper explicitly states X, AI says missing ($n = 37$)}}]
\footnotesize
The AI reviewer asserts that a piece of information is absent from the manuscript when in fact the paper provides it, whether in a different section, the supplementary material, or elsewhere in the same passage. The critique itself would often be valid if the information were truly missing, but the factual premise is wrong. This happens because of limited long context management: Through out writing a review, an AI reviewer must read multiple files (context from the main text, supplementary materials, code, information retrieved from the web) and when it exceeds its context length limit, it summarizes the context, resulting in losing information it should hold to process information and write a coherent review.
\end{mdframed}
 
\begin{mdframed}[
    style=reviewbox,
    frametitle={\textbf{Example for W3 \textemdash{} Future air-quality projections paper, reviewer: Claude Opus 4.5}},
]
\footnotesize
 
\noindent\textcolor{blue!60!black}{\textbf{Claim: ``Known CMIP6 PM2.5 bias is not corrected for''}}
 
\vspace{2pt}\par\noindent\colorbox{blue!5}{\begin{minipage}{0.97\linewidth}
\vspace{4pt}
\begin{itemize}[leftmargin=1.2em, itemsep=2pt, topsep=0pt]
\item \textbf{Main point of criticism:} The paper does not adequately discuss or correct for the known systematic underestimation of PM\textsubscript{2.5} concentrations by CMIP6 models, particularly in highly polluted regions like China, which could bias the DAPP projections.
\item \textbf{Evaluation criteria:} Data and methodology; Validity
\end{itemize}
\vspace{2pt}
\end{minipage}}
 
\vspace{8pt}\par\noindent\textcolor{blue!60!black}{\textbf{Evidence:}}
 
\vspace{2pt}\par\noindent\colorbox{blue!5}{\begin{minipage}{0.97\linewidth}
\vspace{4pt}
\textcolor{blue!50!black}{\textit{\textbf{Quote 1}} (from main text):}
\textit{``We calculated future PM\textsubscript{2.5} concentration in China based on empirical formulas and future pollutant concentration data.''}
\\
\commentbar{The paper mentions using CMIP6 pollutant concentration data but does not discuss bias correction. Multiple studies have documented that CMIP6 models systematically underestimate surface PM\textsubscript{2.5} concentrations, particularly in regions with high pollution levels like East Asia, which could lead to underestimation of future DAPP if not addressed.}
 
\vspace{2pt}
\end{minipage}}
 
\end{mdframed}
 
\begin{mdframed}[style=expertcommentbox]
\footnotesize\textbf{Expert comment.}~This is factually incorrect. The paper explicitly addresses this issue. Lines 489--496 describe a calibration procedure (Equation 5) that adjusts modeled PM\textsubscript{2.5} concentrations against observed baseline period data (2012--2017).
\end{mdframed}
\vspace{10pt}
 
The AI reviewer is not missing a subheading; it is missing an entire calibration procedure (Equation 5, with its own methodological description spanning seven lines) that is already in the manuscript and addresses exactly the bias the AI is raising. The expert's reply cites the line numbers because the content is unambiguously present. What makes this critique fall under W3 rather than a more fundamental validity concern is that the underlying methodological point is valid in general (CMIP6 does underestimate surface PM\textsubscript{2.5} in East Asia), but the accusation that the paper ignores it is factually wrong. A human reviewer who reads the full Methods section will see the bias correction. This limitation should be addressed by future works working on long-context memory of LLM agents.
 
\subsection{Successful cases: Strengths of AI reviewers}\label{qualitative_analysis_strengths}
 
\paragraph{Overview} The four most frequently cited strengths account for roughly $87\%$ of all strength comments ($115/132$), and they describe a consistent pattern: AI reviewers are more diligent than human reviewers at routine but labor-intensive kinds of scrutiny, particularly scrutiny that requires reading code, checking statistical assumptions, or cross-referencing specialized literature. Statistical and methodological rigor (\textbf{S1}, $n = 45$) is the largest single category and applies across all three reviewer models: AI reviewers catch independence violations, missing validation splits, and inadequate uncertainty reporting that human reviewers often pass over. Inspecting the submitted source code (\textbf{S2}, $n = 28$) is a capability specific to our agentic framework: the AI reviewer opens the repository and uses what it finds to support or refute manuscript claims, surfacing implementation bugs and data-leakage issues that the manuscript text alone does not reveal. Domain-specific technical depth (\textbf{S3}, $n = 27$) appears when the AI recognizes that a particular word or method choice in a specific subfield carries technical commitments the paper has not backed up. Internal consistency across sections (\textbf{S4}, $n = 15$) is a related but distinct strength: catching mismatches between abstract, methods, and results that careful reading would surface. We illustrate S2 with one representative example because it is the most differentiating capability from a human-only review; additional examples are in \autoref{appendix:sw_all_comments}.
 
\begin{mdframed}[style=categoryboxS, frametitle={\textbf{AI Reviewer Strength \#2: Inspecting the submitted source code ($n = 28$)}}]
\footnotesize
The AI reviewer opens the submitted source code, reads it, and uses what it finds to support or refute specific claims in the manuscript. In several cases this surfaces concrete implementation bugs, data leakage, or mismatches between described and implemented methodology that the manuscript text alone does not reveal.
\end{mdframed}
\vspace{-2mm} 
 
\newsavebox{\sIIexIcode}
\begin{lrbox}{\sIIexIcode}
\begin{minipage}{0.85\linewidth}
\begin{lstlisting}[style=reviewcode]
def logdet_matmul(xs, w=None):
    # Combines determinants in log-domain.
    # xs: FermiNet orbitals in each determinant.
    #   Either of length 1 with shape
    #     (ndet, nelectron, nelectron)  # full_det=True
    #   or length 2 with shapes
    #     (ndet, nalpha, nalpha), (ndet, nbeta, nbeta)
    #   (full_det=False, determinants are factorised
    #    into block-diagonals for each spin channel).
\end{lstlisting}
\end{minipage}
\end{lrbox}
 
\begin{mdframed}[
    style=reviewbox,
    frametitle={\textbf{Example for S2 \textemdash{} Neural-network wavefunctions paper, reviewer: GPT-5.2}},
]
\footnotesize
 
\noindent\textcolor{blue!60!black}{\textbf{Claim: ``The wavefunction ansatz is inconsistent between the main text and the algorithm''}}
 
\vspace{2pt}\par\noindent\colorbox{blue!5}{\begin{minipage}{0.97\linewidth}
\vspace{2pt}
\begin{itemize}[leftmargin=1.2em, itemsep=2pt, topsep=0pt]
\item \textbf{Main point of criticism:} The manuscript's formal description of the solid wavefunction ansatz is internally inconsistent between the main-text expression and the provided algorithm, making it difficult to verify correctness (spin structure and determinant factorization) and undermining reproducibility of the central methodological contribution.
\item \textbf{Evaluation criteria:} Data and methodology; Reproducibility
\end{itemize}
\vspace{2pt}
\end{minipage}}
 
\vspace{4pt}\par\noindent\textcolor{blue!60!black}{\textbf{Evidence:}}
 
 
\par\noindent\colorbox{blue!5}{\begin{minipage}{0.97\linewidth}
\vspace{2pt}
\textcolor{blue!50!black}{\textit{\textbf{Quote 1}} (from main text):}
\textit{``$\Psi(\mathbf{r}) = \mathrm{Det}_1[e^{i\mathbf{k}\cdot\mathbf{r}_1}u_{mol}^{\uparrow}(d)]\mathrm{Det}_1[e^{i\mathbf{k}\cdot\mathbf{r}_1}u_{mol}^{\downarrow}(d)]$.''}
\\
\commentbar{This equation indicates a product of two determinants associated with spin-up and spin-down channels. However, the notation is ambiguous (both are labeled Det\textsubscript{1}), and it does not clearly specify whether the determinants are block-diagonalized by spin or full determinants over all electrons. In periodic systems with complex-valued orbitals, these details matter for the correctness of antisymmetry constraints and the implementation of phase factors.}
 
\vspace{2pt}
\end{minipage}}
 
\vspace{2pt}
 
\par\noindent\colorbox{blue!5}{\begin{minipage}{0.97\linewidth}
\vspace{2pt}
\textcolor{blue!50!black}{\textit{\textbf{Quote 2}} (from submitted source code):}
 
\vspace{2pt}
\usebox{\sIIexIcode}
\vspace{2pt}
 
\commentbar{The released implementation explicitly distinguishes two cases: a single full determinant over all electrons (\texttt{full\_det=True}) or a product of spin-channel determinants when factorized (\texttt{full\_det=False}). This illustrates that the determinant factorization and spin handling are subtle and explicitly parameterized in code. When contrasted with the manuscript's Algorithm~1 line that multiplies two identical determinants, it reinforces that the paper's written description does not clearly convey which mode was used. Because determinant structure can change expressivity, optimization behavior, and numerical stability, readers cannot reproduce or compare results without a clear specification in the main text.}
 
\end{minipage}}
 
\end{mdframed}
 
\begin{mdframed}[style=expertcommentbox]
\footnotesize\textbf{Expert comment.}~I find it impressive that this reviewer actually looks at the code to understand whether some parts of the paper are not properly discussed. I am not doing it on a regular basis as it would consume a substantial amount of time.
\end{mdframed}
\vspace{10pt}
 
This is the clearest expert endorsement of S2 behavior in our dataset: the annotator, a domain specialist, explicitly acknowledges that cross-checking the main-text mathematics against the provided code is something human reviewers typically do not do because it is too time-consuming. The AI reviewer caught a real ambiguity: the main text describes two spin-separated determinants, but the algorithm and code use notation consistent with two identical determinants, leaving readers unable to tell which formulation was actually executed. For a paper whose central contribution is the wavefunction ansatz, this is a genuine reproducibility issue. The main point is that the reader cannot tell, and the AI reviewer is the only reviewer in the entire panel who checked this problem.
 
\paragraph{Takeaway: Current frontier AI reviewers operating in an agentic framework provide genuine value on the rigor- and code-heavy aspects of peer review, while systematically failing on the field-context aspects.} The dominant weakness pattern is not factual error but severity miscalibration against community norms; the dominant strength pattern is willingness to read code, check statistical assumptions, and cross-reference specialized literature at a level of detail human reviewers typically do not. The most productive engineering targets for AI reviewer systems are therefore (i) calibrating severity against field-specific norms, (ii) handling long-context memory, and (iii) reducing redundancy across multiple AI reviewers, while preserving the tool-enabled scrutiny from AI reviewers.

\section{Tools for improving and using AI reviewers}\label{sec:discussion}
 
\subsection{Can we mimic the fineness of expert validation and automatically evaluate AI reviewers?}\label{subsec:peerreview_bench}

\paragraph{Motivation} The analyses in the preceding sections rest on a 469-hour expert annotation study, which is costly to repeat for every new AI reviewer model. We therefore investigate whether AI agents can serve as meta-reviewers, performing the same three-axis judgment of correctness, significance, and evidence sufficiency that domain scientists did in our study. On a held-out calibration set of 908 dual-annotated review items, the three frontier models we test all land at or near human inter-annotator agreement (Claude-Opus-4.7: 87.9\% / 56.7\% / 85.6\% vs human-human 85.8\% / 59.9\% / 88.0\%), with further analysis of AI meta-reviewers in \autoref{appendix:peerreview_bench}.

\paragraph{\textsc{PeerReview Bench}} Building on this validation, we release \textsc{PeerReview Bench}, a 78-paper benchmark that scores an AI reviewer along two metrics. \emph{Precision} is the fraction of the AI reviewer's items that the meta-reviewer judges fully positive, paralleling the analysis in \autoref{sec:performance}; this captures how often AI-raised items meet the three-axis bar. \emph{Recall} is the fraction of \emph{fully positive} human review items (those rated correct, significant, and evidence-sufficient by both annotators) that the AI reviewer also raises, paralleling the analysis in \autoref{sec:overlap}; this captures how much of the human-validated rubric the AI covers. \autoref{tab:peerreview-bench} reports the results across twelve backbone models, and two findings stand out. First, a substantial gap from the human ceiling remains on both metrics: even the top model (Claude-Opus-4.5) reaches only F1 = 50.89 (precision 75.49\%, recall 38.39\%). Second, different model families specialize in different halves of the benchmark: GPT-5.4 attains the highest precision (93.81\%) but only 26.55\% recall, while Gemini-3.0-Pro-Preview shows the opposite skew (53.35\% precision, 37.65\% recall); the Claude-Opus models balance the two best. Further details of the evaluations are in \autoref{appendix:peerreview_bench}.
 
\begin{table}[t]
\centering
\fontsize{8}{10}\selectfont
\setlength{\tabcolsep}{6pt}
\begin{tabular}{l cccc}
\toprule
\multicolumn{1}{c}{\multirow{2}{*}{\textbf{AI Reviewer backbone LLM}}} & \multicolumn{4}{c}{\multirow{1}{*}{\textbf{PeerReview Bench}}}\\
\cmidrule(lr){2-5} & \textbf{Precision} & \textbf{Recall}  & \textbf{F1 score} & \textbf{\# Review items} \\
\midrule
\quad Claude-Opus-4.7        & 71.47 & \textbf{39.00} & \underline{50.46} & 4.73\\
\quad Claude-Opus-4.5        & 75.49 & \underline{38.39} & \textbf{50.89} & 4.38\\
\quad Claude-Sonnet-4.6      & 82.03 & 29.74 & 43.65 & 4.42\\
\quad GPT-5.4                & \textbf{93.81} & 26.55 & 41.38 & 3.60\\
\quad GPT-5.2                & \underline{88.92} & 32.28 & 47.37 & 4.12\\
\quad GPT-5.4-mini           & 86.59 & 22.55 & 35.78 & 3.04\\
\quad Gemini-3.1-Pro-Preview & 59.55 & 13.92 & 22.57 & 3.31\\
\quad Gemini-3.0-Pro-Preview & 53.35 & 37.65 & 44.14 & 4.23\\
\quad Gemini-3-Flash         & 60.46 & 31.99 & 41.84 & 4.86\\
\quad DeepSeek-V4-Pro        & 76.75 & 35.47 & 48.52 & 4.59\\
\quad Kimi-K2.6              & 74.38 & 28.76 & 41.48 & 4.15\\
\quad Qwen3.6-Plus           & 56.67 & 36.27 & 44.23 & 4.62\\
\bottomrule
\end{tabular}
\caption{\textbf{\textsc{PeerReview Bench} results.} Recall is the fraction of correct, significant, and well-evidence human review items matched by review items from the AI reviewer; precision is the fraction of generated review items that are correct, significant, and well-evidence; F1 is the harmonic mean. All metrics are computed at the paper level and averaged across 78 papers. Best scores are \textbf{bolded}, second best are \underline{underlined}.}
\label{tab:peerreview-bench}
\vspace{-3mm}
\end{table}

\begin{table}[t]
\centering
\fontsize{8}{10}\selectfont
\setlength{\tabcolsep}{6pt}
\begin{tabular}{l cc | cccc}
\toprule
\multicolumn{1}{c}{\multirow{2}{*}{\textbf{AI Reviewer platforms}}}
& \multicolumn{1}{c}{\multirow{2}{*}{\textbf{LLM Backbone}}}
& \multicolumn{1}{c}{\multirow{2}{*}{\textbf{\# Max review items}}}
& \multicolumn{4}{c}{\multirow{1}{*}{\textbf{PeerReview Bench}}}\\
\cmidrule(lr){4-7}
& & & Precision & Recall & F1 & \# Review items\\
\midrule
Stanford Agentic Reviewer & N/A &  \xmark & 59.84 & \textbf{45.43} & \underline{51.65} & 11.08\\
OpenAIReview      & Claude-Opus-4.7 & \xmark & 57.57 & 40.98 & 47.88 & 18.64\\
\midrule
\multicolumn{1}{l}{\multirow{3}{*}{\textsc{CMU Paper Reviewer (Ours)}}}                & Claude-Opus-4.7 & 5 & 71.47 & 39.00 & 50.46 & 4.73\\
& GPT-5.4 & 5 & \underline{93.81} & 26.55 & 41.38 & 3.60 \\
& GPT-5.4 & 15 & \textbf{95.46}&\underline{42.32}&\textbf{58.64}&7.35\\
\bottomrule
\end{tabular}
\caption{\textbf{Comparison of publicly available AI reviewer platforms on PeerReview Bench (78 papers).}
For each platform, we evaluate the review items it produces on \textbf{PeerReview Bench}, and report Precision, Recall, and F1 score.
For Stanford Agentic Reviewer, each bullet point under the ``Weaknesses'' section is treated as a single review item; for OpenAIReview, each feedback card is treated as a single review item.
}
\label{tab:platform-comparison}
\end{table}
 
\subsection{\textsc{CMU Paper Reviewer}: An open-source reviewer platform for pre-submission feedback}\label{subsec:cmu_paper_reviewer}

We release the \textsc{CMU Paper Reviewer}\footnote{\url{https://prometheus-eval.github.io/cmu-paper-reviewer/}}, an open-source platform for authors, students, and researchers who want detailed feedback on a manuscript before submission, built on the pipeline employed in our expert annotation study. 

\paragraph{Platform features and benchmark results} The platform implements concrete mitigations for three weakness patterns documented in \autoref{sec:strengths_weaknesses}. First, to address vague or non-actionable critiques, every review item is paired with a concrete patch suggestion: a proposed manuscript edit, or a runnable code patch when source is provided. Second, to address over-harsh or out-of-scope demands, severity ratings are grounded against the manuscript's own stated limitations, and an interactive debate mode lets authors challenge a critique that the AI then defends or concedes. Third, to address the tendency to cite evidence published after the preprint, citations are annotated and optionally filtered against a user-set publication date. On \textsc{PeerReview Bench}, the platform configured with GPT-5.4 at fifteen items per paper reaches F1 = 58.64, against 51.65 for the Stanford Agentic Reviewer and 47.88 for OpenAIReview. We emphasize that the \textsc{CMU Paper Reviewer} is a pre-submission feedback tool and \textbf{should not be used at conferences or journals that prohibit AI reviewers in their official review process}; organizers interested in officially incorporating it into a sanctioned workflow are welcome to contact the authors. Implementation details, the full platform-comparison table, and intended-use guidance are in \autoref{appendix:cmu_paper_reviewer}.

\section{Conclusion}

In this paper, we characterize the strengths and weaknesses of AI reviewers through an expert annotation study in which 45 domain scientists, over 469 hours, evaluate every 2,960 review item from both AI and human reviewers across 82 Nature-family papers. Surprisingly, current AI reviewers are competitive even with the top-rated reviewers in Nature's official peer review: on the composite of correctness, significance, and evidence sufficiency, they produce a significantly higher fraction of review items per paper than the top-rated human reviewer. AI reviewers also show distinctive strengths: thorough cross-reference checking, code-level inspection, and raising valid criticisms that human reviewers miss. The weaknesses are equally clear: AI reviewers produce more factually incorrect items than humans do, with the composite advantage arising from higher significance and evidence sufficiency among items that are correct. AI reviewers' items also overlap with each other far more than human reviewers' do, so an all-AI panel would substantially narrow the diversity of perspectives. Finally, AI reviewers exhibit characteristic failure patterns rarely seen in humans: a limited grasp of \textit{subfield-specific methodological conventions}, \textit{losing track of content} across long papers and supplementary materials, and an \textit{overly critical stance} that inflates minor issues.\\

Looking forward, two threads follow from this work. For AI reviewer developers, the priority is closing the correctness gap and improving criticism calibration. Concrete next steps include inducing an understanding of subfield-specific norms, embedding better long-context management into LLM agents, and calibrating to expert judgment on when criticism is warranted versus inflated. \textsc{PeerReview Bench} offers a testbed for tracking progress on these fronts across future generations of AI reviewers. For the field, AI reviewers shouldn't be evaluated against human reviews as an implicit gold standard, but on the same per-axis standards (correctness, significance, and evidence sufficiency) that domain experts apply to human reviews. The \textsc{CMU Paper Reviewer} offers an platform authors can use today for pre-submission feedback while the community works out how AI reviewers should be integrated into venue-level review workflows. Open questions also remain (\textit{e.g.}, whether the tradeoff persists as models improve, whether the patterns generalize beyond Nature-family papers, and what governance norms should accompany operational AI deployment). The experiments from this paper provides the empirical infrastructure to answer them, and the urgency only grows as AI reviewers move further into operational deployment.


\bibliographystyle{unsrtnat}
\bibliography{ref}
\DoToC
\appendix

\section{Related Work}\label{app:related-work}
 
\paragraph{Peer Review System for Science.} The peer review system, fundamental to scientific publishing, has come under increasing strain in recent years. Manuscript submission volumes continue to rise across both journals and conferences while the pool of qualified human reviewers does not expand at the same rate, leading to reviewer fatigue, longer review timelines, and difficulty recruiting reviewers willing to provide thorough evaluations~\citep{hanson2024strain, adam2025peer, chen2025position, bauchner2024use}. As a result, median submission-to-publication times have stretched to 100 to 160 days at major science journals~\citep{powell2016does}. These growing pressures have raised concerns about review quality and consistency: the NeurIPS 2014 and 2021 consistency experiments, in which roughly 10\% of submissions were independently reviewed by two committees, found that around half of the accepted papers would have been rejected if the review process had been independently rerun, and that approximately 26\% (in 2014) and 23\% (in 2021) of papers received inconsistent accept-or-reject decisions across the two committees~\citep{cortes2021inconsistency, beygelzimer2023has}, indicating that a substantial portion of decisions reflects subjective variation rather than the underlying quality of the work. Beyond consistency, recent work has documented further structural challenges including reviewer miscalibration, reviewer-author collusion, and undetected major errors in submitted manuscripts~\citep{shah2022challenges}.
 
In response, the scientific community has begun to consider whether natural language processing and large language models might assist (rather than replace) the human peer-review process, ranging from low-level support tasks such as plagiarism detection and reviewer-paper matching to higher-level tasks such as structured feedback generation and consistency checks across reviews~\citep{kuznetsov2024can}. This consideration is no longer purely theoretical. A recent global survey of more than 1{,}600 active researchers across 111 countries reported that over 50\% of reviewers have already used AI tools while peer-reviewing manuscripts, often without clear policy guidance and in some cases against the official rules of the venue they were reviewing for~\citep{naddaf2026more}. At the venue level, several conferences have begun to formally pilot AI assistance within their review workflows: NeurIPS 2024 piloted an opt-in LLM-based author checklist assistant on a voluntary subset of submissions~\citep{goldberg2024usefulness}, ICLR 2025 ran a large-scale randomized controlled trial of an LLM-based feedback agent that nudged human reviewers toward more specific and actionable comments across more than 20{,}000 reviews~\citep{thakkar2025can}, and AAAI-26 went a step further and deployed a state-of-the-art AI reviewer on every one of its 22{,}977 main-track submissions, with surveys of authors and program-committee members reporting that participants found the AI reviews useful and, on certain dimensions such as technical accuracy and research suggestions, preferred them to human reviews~\citep{biswas2026ai}. Together, these developments mark a shift from speculative discussion of AI in peer review to its concrete operational deployment at scale.
 
\paragraph{LLM-as-a-Judge.} A parallel line of work has investigated the use of large language models themselves as evaluators of model and text outputs (LLM-as-a-Judge), establishing strong correlations with human judgement on general tasks~\citep{zheng2023judging, kim2023prometheus, kim2024prometheus} while documenting systematic biases such as position, verbosity, and self-preference effects~\citep{ye2024flask,wataoka2024selfpreference,dubois2024length,kim2025biggen,ye2025justice}.
 
\paragraph{Development of AI Reviewers.} Building on LLM-as-a-Judge methodologies, an active line of research has explored applying LLM-as-a-Judge specifically to scientific peer review (\textit{i.e.}, AI Reviewers). Early exploratory studies tested whether GPT-4 could detect inserted errors and produce structured first-pass feedback~\citep{yuan2022can, liu2023reviewergpt}. Subsequent work has produced a wide range of full review-generation systems, including multi-agent discussion~\citep{d2024marg}, standardized review generation through fine-tuning~\citep{yu2024automated}, bias-aware pipelines~\citep{tyser2024ai}, deep-thinking review generation~\citep{zhu2025deepreview}, and RL-trained reviewers~\citep{zeng2025reviewrl}. A separate strand uses LLMs as a quality-control component within end-to-end research-automation systems~\citep{lu2026towards}, while another simulates the peer-review process itself with LLM agents to study its dynamics~\citep{jin2024agentreview}. Beyond review generation, several efforts target review-process improvement directly, including a large-scale ICLR 2025 randomized study showing that LLM feedback agents can make human reviewers' reviews more informative~\citep{thakkar2026large}, LLM checklist assistance for authors~\citep{goldberg2024usefulness}, and pairwise comparison through LLM agents~\citep{zhang2026from}.
 
Notably, most existing evaluations of AI reviewers operate at the review level rather than the individual-comment level: they measure score correlation, decision alignment, or text-level similarity against human reviews and treat human reviews as the gold standard~\citep{idahl2025openreviewer, bougie2025generative}. Some evaluation frameworks have begun to introduce finer-grained components: ReviewEval performs per-claim factual verification and per-insight actionability scoring~\citep{garg2025revieweval}; TreeReview evaluates individual feedback comments through embedding-based and LLM-based matching for coverage and specificity~\citep{chang2025treereview}; and FLAWS benchmarks LLMs' ability to localize specific errors in scientific manuscripts~\citep{xi2025flaws}. Other work has begun to question the gold-standard assumption itself, either by demonstrating that LLM pairwise comparison can identify high-impact papers more accurately than rating-based aggregation~\citep{zhang2026from} or by reporting that AI reviewers can match human reviewers in predicting paper outcomes~\citep{bougie2025generative}.
 
In contrast, our work performs an evaluation that, to our knowledge, no prior framework has performed: every individual review item, from both AI and human reviewers, is annotated by a domain scientist in each paper's own field along three complementary quality axes (correctness, significance, and sufficiency of evidence). Three properties of this design are absent from prior practice taken together. The annotation is bidirectional, in that AI and human review items are subject to the same scrutiny rather than human reviews being treated as the implicit reference for AI~\citep{chang2025treereview, garg2025revieweval, idahl2025openreviewer, bougie2025generative}. The annotation is performed by domain scientists rather than by LLM judges or graduate-student annotators from outside the paper's subfield~\citep{ryu2025reviewscore}, since judgments of scientific significance and evidence sufficiency are field-dependent. And the annotation is multi-axis rather than single-dimensional or holistic, since a comment can be correct but trivial, significant but unsupported, or both significant and well-supported, and these are very different failure modes. The result is per-axis directional evidence on the specific quality dimensions on which AI reviewers outperform human reviewers and on which the reverse holds.
 

\begin{table}[t]
\centering
\fontsize{8.5}{10.5}\selectfont
\setlength{\tabcolsep}{8pt}
\renewcommand{\arraystretch}{1.15}
\begin{tabular}{l c c c c}
\toprule
\textbf{Work} & \textbf{Bidirectional} & \textbf{Per-comment} & \textbf{Multi-axis} & \textbf{Domain experts} \\
\midrule
ReviewRobot~\citep{wang2020reviewrobot}             & \pmark & \cmark & \xmark & \cmark \\
Liang et al.~\citep{liang2024can}                   & \pmark & \pmark & \xmark & \pmark \\
TreeReview~\citep{chang2025treereview}              & \xmark & \cmark & \pmark & \xmark \\
OpenReviewer~\citep{idahl2025openreviewer}          & \xmark & \xmark & \pmark & \xmark \\
GAR~\citep{bougie2025generative}                    & \xmark & \xmark & \xmark & \xmark \\
ReviewerToo~\citep{sahu2025reviewertoo}             & \pmark & \xmark & \pmark & \xmark \\
ReviewEval~\citep{garg2025revieweval}               & \xmark & \pmark & \cmark & \xmark \\
REVIEWSCORE~\citep{ryu2025reviewscore}              & \xmark & \cmark & \xmark & \xmark \\
FLAWS~\citep{xi2025flaws}                           & \xmark & \cmark & \xmark & \xmark \\
\midrule
\textbf{CMU Paper Reviewer \& PeerReview Bench (ours)}                & \cmark & \cmark & \cmark & \cmark \\
\bottomrule
\end{tabular}
\caption{\textbf{Comparison of evaluation methodologies in prior peer-review work along four properties of our design.}
\textbf{Bidirectional}: both AI and human review items are annotated under the same protocol, rather than treating human reviews as an implicit reference for AI.
\textbf{Per-comment}: the unit of evaluation is the individual review item, rather than the full review.
\textbf{Multi-axis}: each item is decomposed along multiple complementary quality axes, rather than scored on a single dimension or collapsed into one aggregate verdict.
\textbf{Domain experts}: annotations are performed by domain scientists in each paper's field, rather than by LLM judges or by annotators from outside the paper's subfield.
\cmark{} indicates that the property is fully present, \xmark{} that it is fully absent, and \pmark{} that it is partially present (see Appendix~\ref{app:related-work} for per-row justification).
No prior work satisfies all four properties simultaneously.}
\label{tab:related-work-comparison}
\end{table}
 
Table~\ref{tab:related-work-comparison} systematizes the contrast described above into four properties of our evaluation design that, taken together, are absent from prior peer-review evaluation frameworks: \textbf{bidirectional} annotation of both AI and human review items, \textbf{per-comment} unit of analysis, \textbf{multi-axis} decomposition into complementary quality dimensions, and annotation by \textbf{domain experts} in each paper's field. The justifications below explain the basis for each marking.
 
\paragraph{ReviewRobot~\citep{wang2020reviewrobot}.}
ReviewRobot generates structured review comments using a knowledge-graph approach and asks domain experts to assess the comments for validity and constructiveness, also reporting that AI-generated comments were judged better than human-written ones in 20\% of pairings.
Per-comment annotation by domain experts is therefore present, and the comparison touches both AI and human comments, justifying \cmark{} for per-comment and \cmark{} for domain experts and \pmark{} for bidirectional.
However, only a single quality dimension (validity or constructiveness) is assessed, and the comparison is preference-based rather than axis-by-axis, so multi-axis is marked \xmark{}.
 
\paragraph{Liang et al.~\citep{liang2024can}.}
Liang et al. conduct a large-scale empirical study with two components.
The retrospective component measures overlap between GPT-4-generated review comments and human reviewer comments on Nature-family journals and ICLR submissions, using GPT-4 itself as the matcher.
The prospective component surveys more than 300 researchers who received AI feedback on their own manuscripts, asking them to rate the AI feedback on a small number of properties such as specificity and helpfulness.
Bidirectional and per-comment annotation are therefore partially present (\pmark, \pmark): the overlap analysis matches human and AI comments symmetrically, but the matching is performed by an LLM rather than by humans, and the resulting metric is an aggregate overlap rate rather than a per-comment quality judgment.
Domain expertise is partially present (\pmark): the prospective survey involves authors who are experts on their own paper, but their judgments are paper-level usefulness ratings rather than per-comment quality annotations.
Multiple quality axes are not used (\xmark).
 
\paragraph{TreeReview~\citep{chang2025treereview}.}
TreeReview proposes a hierarchical question-decomposition framework for full review generation and actionable feedback comments.
Its evaluation has two parts.
For full reviews, an LLM-as-Judge scores quality on dimensions such as specificity, comprehensiveness, and technical depth, supplemented by human preference ranking.
For feedback comments, the model output is matched against human comments using precision and Jaccard similarity, with the human comments treated as the reference.
Per-comment evaluation is therefore present (\cmark) and the dimensional decomposition for full reviews is partial (\pmark), but human reviews are treated as the implicit gold standard rather than annotated alongside AI reviews (\xmark{} for bidirectional), and the annotators are LLM judges or non-expert humans (\xmark{} for domain experts).
 
\paragraph{OpenReviewer~\citep{idahl2025openreviewer}.}
OpenReviewer is an open-source specialized language model fine-tuned for generating critical reviews; the system demonstration paper evaluates it primarily through review-level quality comparisons against baseline LLMs.
Evaluation is at the full-review rather than per-comment level (\xmark), uses LLM judges and non-expert humans (\xmark{} for domain experts), treats human reviews as a reference target (\xmark{} for bidirectional), and reports multiple quality dimensions only at the review level (\pmark{} for multi-axis).
 
\paragraph{GAR~\citep{bougie2025generative}.}
Generative Reviewer Agents simulate the peer-review process and are evaluated primarily on whether AI reviewers can match human reviewers in predicting paper outcomes (acceptance or rejection).
The evaluation is at the review level rather than per-comment (\xmark), is unidirectional in the sense that AI is assessed against human-derived ground truth (\xmark{} for bidirectional), uses outcome prediction rather than multi-dimensional quality (\xmark{} for multi-axis), and does not employ domain experts as annotators (\xmark).
 
\paragraph{ReviewerToo~\citep{sahu2025reviewertoo}.}
ReviewerToo evaluates an AI program-committee member on 1{,}963 ICLR 2025 submissions and reports that the AI achieves close-to-human accept-or-reject accuracy and is rated as higher quality than the human average by an LLM judge.
The paper notes domains where AI excels (such as fact-checking and literature coverage) and where it struggles (such as assessing methodological novelty).
Both AI and human reviews are assessed (\pmark{} for bidirectional), but only by an LLM judge at the review level (\xmark{} for per-comment, \xmark{} for domain experts, \pmark{} for multi-axis).
 
\paragraph{ReviewEval~\citep{garg2025revieweval}.}
ReviewEval introduces a multi-dimensional evaluation framework for AI-generated reviews along five dimensions (alignment with human reviews, factual correctness, analytical depth, actionable insights, and adherence to guidelines), with per-claim factual verification implemented as one of its components.
Multi-axis decomposition is fully present (\cmark) and per-claim factual verification yields partial per-comment evaluation (\pmark).
However, the evaluation is one-directional in that only AI-generated reviews are scored, with human reviews used as the reference (\xmark{} for bidirectional), and the entire evaluation pipeline is implemented through LLM judges; the authors explicitly acknowledge that this reliance on LLMs risks propagating biases (\xmark{} for domain experts).
 
\paragraph{REVIEWSCORE~\citep{ryu2025reviewscore}.}
REVIEWSCORE constructs an annotated dataset of 657 review points (143 questions, 92 claims, and 422 arguments) from 40 ICLR 2021 to 2023 papers, with 1{,}748 underlying premises manually annotated, in order to train and evaluate LLMs on detecting misinformed review points.
Per-comment annotation is fully present (\cmark).
However, the annotators are 15 graduate students studying AI rather than domain scientists in each paper's subfield (\xmark{} for domain experts), the annotation targets a single axis (factuality of weaknesses and unanswerability of questions, \xmark{} for multi-axis), and the work does not compare AI-generated and human-written review items under the same protocol; rather, it benchmarks LLMs against human-annotated review points (\xmark{} for bidirectional).
 
\paragraph{FLAWS~\citep{xi2025flaws}.}
FLAWS is a benchmark of 713 paper-error pairs, constructed by systematically inserting claim-invalidating errors into peer-reviewed papers, that evaluates whether LLMs can identify and localize errors.
Per-error evaluation is at the item level (\cmark), but the task is single-axis error detection (\xmark), the evaluation is fully automated against synthetically inserted ground truth rather than annotated by domain experts (\xmark), and the work targets LLM error-detection capability rather than AI-versus-human review comparison (\xmark{} for bidirectional).

\section{Extended: Expert annotation study design and experimental setup}\label{appendix:expert_annotation_study_design}

This appendix collects the procedural and statistical details that support \autoref{sec:preliminaries}. Each subsection extends a specific element of the main text, following the order in which those elements appear there.

\begin{table}[h]
\centering
\small
\begin{tabular}{llr}
\toprule
\textbf{Broad Subject Area} & \textbf{Mid-level Category} & \textbf{Papers} \\
\midrule
\multirow{8}{*}{Physical Sciences (38)}
  & Physics & 7 \\
  & Optics and photonics & 7 \\
  & Engineering & 6 \\
  & Astronomy and planetary science & 5 \\
  & Chemistry & 4 \\
  & Nanoscience and technology & 4 \\
  & Energy science and technology & 3 \\
  & Materials science & 2 \\
\midrule
\multirow{14}{*}{Biological Sciences (30)}
  & Neuroscience & 7 \\
  & Microbiology & 5 \\
  & Ecology & 3 \\
  & Computational biology and bioinformatics & 3 \\
  & Drug discovery & 2 \\
  & Genetics & 2 \\
  & Biological techniques & 1 \\
  & Biophysics & 1 \\
  & Biotechnology & 1 \\
  & Evolution & 1 \\
  & Immunology & 1 \\
  & Molecular biology & 1 \\
  & Plant sciences & 1 \\
  & Systems biology & 1 \\
\midrule
\multirow{5}{*}{Health Sciences (14)}
  & Medical research & 5 \\
  & Neurology & 4 \\
  & Risk factors & 3 \\
  & Biomarkers & 1 \\
  & Oncology & 1 \\
\midrule
\multicolumn{2}{l}{\textbf{Total}} & \textbf{82} \\
\bottomrule
\end{tabular}
\caption{Subject category breakdown of the 82-paper dataset.}
\label{tab:extended-data-subjects}
\end{table}

\subsection{Subject category breakdown}\label{appendix:subject_categories}
 
\autoref{tab:extended-data-subjects} shows the full subject category breakdown of the 82 papers employed in our expert annotation study. This categorization is based on the \textit{Nature Communications} subject taxonomy\footnote{\href{https://www.nature.com/ncomms/browse-subjects}{https://www.nature.com/ncomms/browse-subjects}}.

\subsection{Evaluation criteria for reviewing a paper}\label{appendix:criteria_details}
 
The six \textit{Nature} peer-review evaluation criteria referenced in \autoref{subsec:reviewers} follow the \textit{Nature} Portfolio peer review policy and are presented to each AI reviewer agent verbatim. They are also shown to each domain scientist as part of the annotation guidelines. They are defined as follows, ordered by priority.
 
\begin{itemize}[leftmargin=*]
    \item \textbf{Validity.} Does the manuscript have significant flaws which should prohibit its publication?
    \item \textbf{Conclusions.} Are the conclusions and data interpretation robust, valid, and reliable?
    \item \textbf{Originality and significance.} Are the results presented of immediate interest to many people in the field of study, and/or to people from several disciplines?
    \item \textbf{Data and methodology.} Is the reporting of data and methodology sufficiently detailed and transparent to enable reproducing the results?
    \item \textbf{Appropriate use of statistics and treatment of uncertainties.} Are all error bars defined in the corresponding figure legends, and are all statistical tests appropriate and the description of any error bars and probability values accurate?
    \item \textbf{Clarity and context.} Is the abstract clear and accessible? Are the abstract, introduction, and conclusions appropriate?
\end{itemize}
 
\noindent
Criteria are ordered by priority: when the AI reviewer agent selects which criticisms to include in the final review (up to five items sorted from most to least significant), earlier criteria take precedence over later ones. For instance, a criticism based on \textit{Validity} should be prioritized over one based on \textit{Clarity and context}, all else equal. Domain scientists were not instructed to apply this priority when annotating, as their role is to evaluate each individual item on its own merits rather than to rank items.

\subsection{Evaluation criteria for reviewing a review}\label{appendix:meta_review_criteria}
 
The three-dimensional cascading rubric introduced in \autoref{subsec:rubric} is operationalized through the following exact wording presented to each domain scientist in the annotation interface. While the criteria for reviewing a paper (\autoref{appendix:criteria_details}) operate at the level of a manuscript, these criteria operate at the level of an individual review item, with each subsequent dimension assessed only when the previous one is satisfied.
 
\begin{itemize}[leftmargin=*]
    \item \textbf{Correctness} (binary). ``Is the main point of the criticism correct and clearly stated?'' The annotator selects ``Correct'' if every aspect of the main point is correct and clearly stated, and ``Not Correct'' if there is any slight doubt, i.e., at least one aspect is incorrect or not clearly stated.
 
    \item \textbf{Significance} (ordinal, three-level; only if the item is marked Correct). ``Does the main point of the criticism talk about a significant aspect of the paper that is constructive to enhance the paper rather than touching a minor issue?'' The annotator selects one of three options:
    \begin{itemize}
        \item \textit{Significant}: an item that is insightful and helpful for improving the paper.
        \item \textit{Marginally Significant}: an item that is not directly helpful for improving the paper but is still worth remaining in the review, e.g., typos, stylistic issues, or suggestions to submit the paper to a different journal.
        \item \textit{Not Significant (very marginal issue)}: a very minor item that should not affect the acceptance of the paper and is better removed from the review.
    \end{itemize}
 
    \item \textbf{Sufficiency of Evidence} (binary; only if the item is marked Correct and at least Marginally Significant). ``Is the main point of the criticism well supported by evidence or concrete reasoning?'' The annotator selects ``Evidence is sufficient'' or ``Requires more evidence.''
\end{itemize}
 
\noindent
Annotators may also add a free-form comment to any review item.

\subsection{Processing official peer review files}\label{appendix:peer_review_processing}
 
We manually extract peer reviews from each paper's publicly released Peer Review File, retaining only first-round reviewer comments directed at the same pre-review manuscript that AI reviewers evaluate; editor decision letters and author rebuttals are excluded. To maintain a consistent comparison between three AI reviewers and three human reviewers, we cap the number of human reviews at three by retaining the first three reviewers (Reviewer 1, 2, and 3) for papers with four or more reviewers. Each extracted review is then manually decomposed into \textbf{review items}, where each review item corresponds to a single atomic criticism directed at one aspect of the paper: bullet-pointed reviews are segmented at each bullet, and paragraph-based reviews are segmented at explicit enumeration markers (e.g., ``First,'' ``Second,'').

\subsection{AI reviewer configuration details}\label{appendix:ai_reviewer_details}
 
This subsection provides the full configuration of the three AI reviewer agents used in the expert annotation study (GPT-5.2, Claude Opus 4.5, and Gemini 3.0 Pro), including the agent framework, per-model hyperparameters, tool specifications, web search domain restrictions, and the full reviewer prompt.
 
\paragraph{Agent framework} Each AI reviewer is implemented as an autonomous agent using the \href{https://github.com/OpenHands/software-agent-sdk}{OpenHands Software Agent SDK} \citep{wang2025openhands}. The version used to generate the reviews for the expert annotation study described in \autoref{sec:performance} is \texttt{v1.1.0}; the version used for the \textsc{PeerReview Bench} evaluation in \autoref{subsec:peerreview_bench} is \texttt{v1.5.0}. Both versions share the same core architecture; the later version incorporates minor improvements and additional tools that do not affect the reviewer prompt or the agent's behavior on the tasks reported here.
 
\paragraph{Shared agent configuration} All three agents share the same OpenHands agent settings, summarized in \autoref{tab:openhands_agent_config}. Each agent is equipped with three built-in OpenHands tools (\texttt{TerminalTool}, \texttt{FileEditorTool}, \texttt{TaskTrackerTool}) together with the Tavily web search tool described below. The agent uses the \texttt{LLMSummariz\\ingCondenser} to manage long conversation histories: when the conversation exceeds a configured threshold, earlier turns are summarized by the same underlying language model while the first three turns (system prompt, task specification, and initial tool outputs) are preserved verbatim.

\begin{table}[h]
\centering
\small
\begin{tabular}{ll}
\toprule
\textbf{Parameter} & \textbf{Value} \\
\midrule
Framework & OpenHands SDK (v1.1.0 / v1.5.0) \\
Tools (built-in) & \texttt{TerminalTool}, \texttt{FileEditorTool}, \texttt{TaskTrackerTool} \\
Tools (external) & Tavily MCP web search \\
Max iterations & 5{,}000 \\
Max review items per paper & 5 (sorted by significance) \\
Criteria preset & Nature (six evaluation criteria) \\
Condenser & \texttt{LLMSummarizingCondenser} (\texttt{max\_size}=200, \texttt{keep\_first}=3) \\
Per-API-call timeout & 600 s \\
Prompt caching & enabled \\
\bottomrule
\end{tabular}
\caption{Shared OpenHands agent configuration for all three AI reviewer agents.}
\label{tab:openhands_agent_config}
\end{table}

\paragraph{Shared LLM configuration} All three agents are configured with identical LLM-level settings, passed to the SDK's \texttt{LLM} class and forwarded to each provider via \texttt{LiteLLM}: \texttt{reasoning\_effort = "high"}, \texttt{extended\_think\\ing\_budget = 200{,}000} tokens, \texttt{temperature = 1.0}, \texttt{drop\_params = True}, and \texttt{num\_retries = 5}. The \texttt{drop\_params = True} setting causes the SDK to silently drop any parameter that is not supported by a given provider: Anthropic models receive all three reasoning-related parameters, OpenAI (GPT) models receive \texttt{reasoning\_effort} and \texttt{temperature}, and Gemini models receive \texttt{temperature} with \texttt{reasoning\_effort} mapped to \texttt{thinking\_level = "high"}. The temperature of 1.0 is required by Anthropic's extended thinking mode and applied uniformly across all models for consistency.
 
\paragraph{Per-model specifications} Context window and output length limits for each model are listed in \autoref{tab:per_model_config}. Each row reports the maximum input and output tokens used in our configuration, which equal each provider's published maximum for the given model at the time we generated the AI reviewers for the expert annotation study (Dec 2025).

\begin{table}[h]
\centering
\small
\begin{tabular}{lllrr}
\toprule
\textbf{Model} & \textbf{Provider} & \textbf{Model identifier} & \textbf{Input tokens} & \textbf{Output tokens} \\
\midrule
GPT-5.2          & Azure AI (OpenAI) & \texttt{azure\_ai/gpt-5.2}              & 1{,}050{,}000 & 128{,}000 \\
Claude Opus 4.5  & Anthropic         & \texttt{anthropic/claude-opus-4-5}      & 200{,}000     & 64{,}000  \\
Gemini 3.0 Pro   & Google            & \texttt{gemini/gemini-3.0-pro-preview}  & 1{,}048{,}576 & 65{,}536  \\
\bottomrule
\end{tabular}
\caption{Model identifier and context/output limits for each of the three AI reviewer agents used in the expert annotation study. All three models support multimodal (image) input.}
\label{tab:per_model_config}
\end{table}

\paragraph{Tool specifications} Each agent has read and write access to the paper's source files (preprint, supplementary materials, figures, and code) via the following four tools:

\begin{itemize}
    \item \textbf{TerminalTool.} A bash-based terminal for executing shell commands, running code, and inspecting the filesystem.
    \item \textbf{FileEditorTool.} A file editor that supports reading files, creating new files, and applying targeted string replacements for writing the final review and any verification code.
    \item \textbf{TaskTrackerTool.} A lightweight task manager that lets the agent record, update, and mark off a plan of subtasks during the review.
    \item \textbf{Tavily web search.} An external web search tool provided via the \href{https://github.com/tavily-ai/tavily-mcp}{Tavily MCP server}, used to retrieve relevant literature and external references during the review.
\end{itemize}

\paragraph{Domain restrictions on web search} To prevent the AI reviewer from directly retrieving the published version of a benchmark paper or its existing peer review report, we exclude search results from four domains: \texttt{nature.com}, \texttt{researchsquare.com}, \texttt{springer.com}, and \texttt{springerlink.com}. This restriction is enforced at two levels. First, at the tool level, we apply a monkey-patch to the Tavily MCP tool executor that filters out any result whose URL matches one of the blacklisted domains before the result is returned to the agent. Second, at the prompt level, the reviewer prompt explicitly instructs the agent to avoid retrieving content from these domains. Both levels are required because an LLM agent may attempt to paraphrase or reconstruct blocked content, and because tool-level filtering alone does not prevent the agent from issuing queries that could be logged.

\subsection{AI reviewer prompt}\label{appendix:reviewer_prompt}

\autoref{fig:reviewer_prompt_part1}, \autoref{fig:reviewer_prompt_part2}, and \autoref{fig:reviewer_prompt_part3} reproduce the full reviewer prompt passed to each agent at the start of every (paper, model) run. The prompt is identical across all three models and across all 82 papers; only the placeholder \texttt{[LINK TO THE PAPER]} is replaced at runtime with the absolute path to the paper's source directory, and \texttt{[MODEL NAME]} with the identifier of the model being invoked.

\begin{figure}[h]
\begin{footnotesize}
\begin{verbatim}
You are a reviewer agent assessing the quality of a research paper.
You will be given the paper's content, images, and optionally its code
and supplementary materials.
Your task is to write a review in markdown format, where your review
must contain at most five items (from most significant to least significant).
Each item represents an atomic criticism of the paper and points out a
major issue.
If the paper contains no significant issues, then you can output zero items.


### Principles guiding your review (ordered by importance)
1. Your review must be factually correct:
   Your claims will be checked by domain experts. Any incorrect or
   unsupported criticism will undermine the credibility of your review.
   When uncertain, avoid speculation.
2. Your review must consist of only significant issues:
   Only point out problems that meaningfully affect the paper's validity,
   soundness, methodology, claims, or reproducibility. Do not focus on
   minor or cosmetic issues. If you think there are less than five
   significant issues, then you should output less than five items (even
   zero items are allowed if there are no significant issues).
3. Your review must be concise and only criticize at most five major
   aspects with detailed evidence:
   Each criticism must be supported with detailed evidence. Specifically,
   mention the contextual background of what the authors attempted to do,
   and why that was not sufficient when comparing to common practices in
   the field.


### Rules for constructing each item
1. Each item consists of exactly two components: a claim and evidence.
2. The claim is the criticism itself. In the claim, you must clearly state:
   a. What you are criticizing the paper for.
   b. On which evaluation criterion or criteria the criticism is based.
   c. Which component of the paper the criticism refers to.
3. The evidence must directly support the claim. You should quote:
   a. Exact sentences from the main paper or supplementary materials.
   b. Exact code blocks or functions from the paper's code.
   c. Exact sentences from papers in the literature (hyperlinked and cited).
4. At the end of the review, include a citation list containing all
   literature references used in your evidence.
5. The review must not include an introduction, summary, or concluding
   remarks. It must contain at most five items and a citation list.
6. All output must be valid markdown.
7. You must separate each item with a blank line.
8. Try to avoid using what the paper listed in the "Limitations" or
   "Future work" section as your claim unless it is a significant issue.
9. The items should be sorted by their importance.
10. Use the format Item 1, Item 2, ..., with no fraction or denominator.
\end{verbatim}
\end{footnotesize}
\caption{Reviewer prompt (Part 1 of 3): task description, principles, and rules for constructing each review item.}
\label{fig:reviewer_prompt_part1}
\end{figure}

\clearpage

\begin{figure}[h]
\begin{footnotesize}
\begin{verbatim}
### Required structure and format of each item
Each item must be formatted exactly as follows:

## Item N: <short title summarizing the criticism>

#### Claim
* Main point of criticism: <State what you are criticizing the paper for>
* Evaluation criteria: <which evaluation criteria the criticism is based on>

#### Evidence
* Quote: <Exact sentence(s) 1 from the paper>
   * Comment: <Explanation of why this sentence is problematic>
* Quote: <Exact sentence(s) 2 from the paper>
   * Comment: <Explanation of why this sentence is problematic>
* Quote: <Exact code block 1 from the paper's code>
   * Comment: <Explanation of why this code block is problematic>
* Quote: <Exact sentence(s) from other papers [hyperlinked citation]>
   * Comment: <Explanation of how this contradicts the paper under review>

Each comment should be 5-7 sentences long (a single paragraph).
Insert two empty lines between each item to separate them.


### Required structure and format of the citation list

#### Citation List
[1] <citation 1> (hyperlinked to the retrieved literature)
[2] <citation 2> (hyperlinked to the retrieved literature)
[3] <citation 3> (hyperlinked to the retrieved literature)

There should be at least five citations in the citation list.


### Evaluation criteria (ordered by importance)
1. Validity: Does the manuscript have significant flaws which should
   prohibit its publication?
2. Conclusions: Are the conclusions and data interpretation robust,
   valid and reliable?
3. Originality and significance: Are the results presented of immediate
   interest to many people in the field of study, and/or to people from
   several disciplines?
4. Data and methodology: Is the reporting of data and methodology
   sufficiently detailed and transparent to enable reproducing the results?
5. Appropriate use of statistics and treatment of uncertainties: Are all
   error bars defined in the corresponding figure legends and are all
   statistical tests appropriate and the description of any error bars
   and probability values accurate?
6. Clarity and context: Is the abstract clear, accessible? Are abstract,
   introduction and conclusions appropriate?

Note that earlier evaluation criteria should be prioritized over later
ones when deciding the items in the review.
\end{verbatim}
\end{footnotesize}
\caption{Reviewer prompt (Part 2 of 3): required output format for each review item and the citation list, and the six Nature evaluation criteria ordered by priority.}
\label{fig:reviewer_prompt_part2}
\end{figure}

\clearpage

\begin{figure}[h]
\begin{footnotesize}
\begin{verbatim}
### TODO list for writing your review
- [ ] Read through the paper, supplementary files, and images; construct
      a potential list of items you will criticize.
- [ ] Read through the paper's code, check the functionality of each
      file, and attempt to execute the code if possible. You may
      implement additional code to validate the claims you make.
- [ ] Devise a list of search queries to find relevant literature.
- [ ] Retrieve relevant papers, read them, and update your list of
      criticisms.
- [ ] (Very Important) Iterate through your list and ensure each
      potential criticism is factually correct, significant, and
      eligible for inclusion.
- [ ] Write the review in markdown format and save it to the designated
      review file.


### Guidelines for opening the paper files
The directory to the paper you will be reviewing is [LINK TO THE PAPER].
The directory structure contains: the main paper in Markdown
(preprint.md), a JSON listing the images and their captions, an images
directory, an optional supplementary directory, and an optional code
directory.


### Guidelines for reading the paper's code
1. The code may include a README file that explains the purpose of the
   code and how to run it. Check it before trying to run the code.
2. If the code is not executable, try to resolve dependencies, download
   the necessary datasets, and run the code to validate your claims.
3. Do not try to run the code if it is non-executable or resource-prohibitive.


### Guidelines for retrieving literature
1. Do not iterate through all the papers included in the paper's
   references. Determine which papers are most relevant.
2. Be proactive and add search queries during the review process.
3. It is recommended not only to retrieve academic papers, but also
   blog posts, news articles, datasets, and code repositories.
4. Ensure you actually read what you retrieved.


### Tips
1. The paper's markdown may contain OCR errors. Do not assume the paper
   is incorrect solely because of OCR mistakes. Do not point out that
   the manuscript is incomplete due to formatting issues.
2. Image filenames are guaranteed to be figure1.png, figure2.png, etc.
   Do not point out broken or missing figure assets.
3. The code you are reviewing does not need to be perfect; focus on
   major issues such as non-reproducible experiments or mismatches
   with descriptions rather than minor issues.
4. When refining your review, ensure that all items are factually
   correct, significant, and mutually exclusive.
\end{verbatim}
\end{footnotesize}
\caption{Reviewer prompt (Part 3 of 3): task workflow, guidelines for reading the paper and its code, guidelines for retrieving literature, and additional tips.}
\label{fig:reviewer_prompt_part3}
\end{figure}

\clearpage

\subsection{Domain scientist recruitment}\label{appendix:recruitment}
 
We recruited the 45 domain scientists described in \autoref{subsec:annotators} through a three-stage process conducted between September 2025 and April 2026. We first contacted candidate scientists via cold email describing the project and the expected time commitment. Candidates who expressed interest were then sent a list of 10 papers from our dataset and asked to identify papers they felt qualified to meta-review. Once a matching paper was confirmed, the candidate received the AI reviews (generated as described in \autoref{appendix:ai_reviewer_details}) together with the extracted human review comments for that paper in the form of the annotation sheet described in \autoref{appendix:annotation_guidelines} (see \autoref{fig:annotation_sheet}), with at least one month allotted for completion. Annotation required approximately three hours per paper on average.

\subsection{Annotation guidelines}\label{appendix:annotation_guidelines}
 
\begin{figure}[t]
    \centering
    \includegraphics[width=1\linewidth]{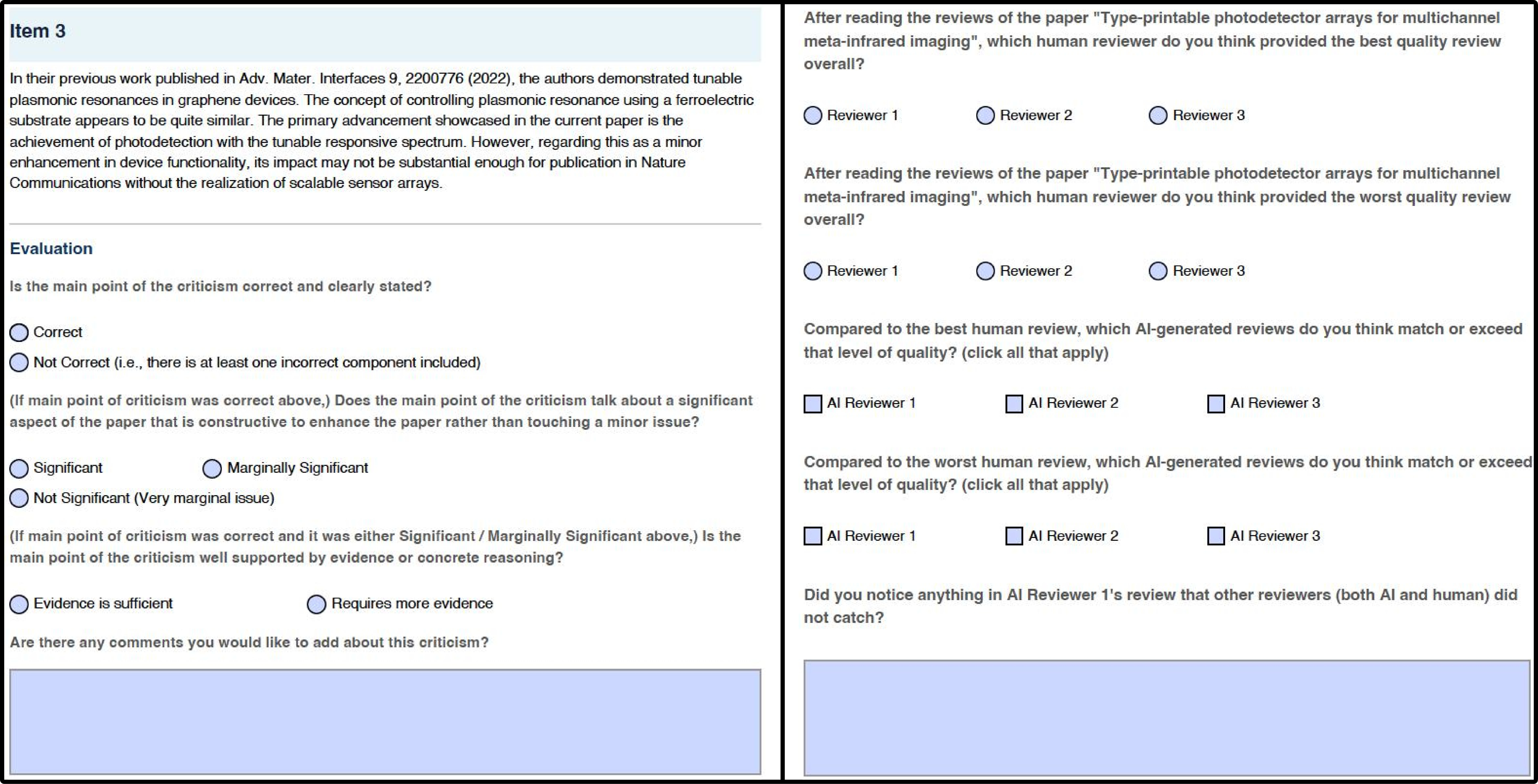}
    \caption{The annotation sheet presented to each domain scientist.
    \textbf{(Left)} \textbf{item-level annotation} for a single review item. The domain scientist judges its \textit{correctness} (binary), \textit{significance} (ordinal, three-level), and \textit{sufficiency of evidence} (binary) along a cascading structure in which each dimension is only assessed when the previous one is satisfied, and may add an optional free-form comment.
    \textbf{(Right)} \textbf{paper-level overall survey} completed after all individual items are evaluated. The domain scientist selects the top-rated and lowest-rated human reviewer, identifies the AI reviewers whose overall review quality matches or exceeds each of these two human references via multi-select, and optionally notes, for each AI reviewer, anything that AI reviewer identified that others missed.}
    \label{fig:annotation_sheet}
    \vspace{-5pt}
\end{figure}

This subsection reproduces the workflow-level annotation guidelines distributed to each domain scientist at the start of the annotation task. The item-level evaluation criteria themselves are in \autoref{appendix:meta_review_criteria}. Each annotator received these guidelines together with the custom-built PDF annotation sheet (see \autoref{fig:annotation_sheet}) containing the paper's human and AI reviews decomposed into review items. No training or calibration session was conducted before the annotation task; annotators were asked to contact the corresponding author if they had any questions about the guidelines.
 
\paragraph{Task description} Each domain scientist is asked to check the quality of criticisms included in the human reviews and AI reviews of a paper they are qualified to meta-review. The resulting meta-reviews (reviews of reviews) are used for (i) comparing the quality of human reviews and AI reviews across multiple dimensions, and (ii) constructing \textsc{PeerReview Bench}, a benchmark for evaluating future generations of AI reviewer agents (\autoref{subsec:peerreview_bench}).
 
\paragraph{Review item structure} Each point of criticism in a review is divided into atomic units called \textbf{review items}. For AI reviews, each item consists of a \textit{Main Point}, \textit{Evaluation Criteria}, and \textit{Evidence} (see \autoref{fig:review-item-example2} for an example). The concatenation of these three components is treated as a single item. For human reviews, which are written in a more free-form manner, the decomposition into items is performed manually by the authors as described in \autoref{subsec:rubric}, and each item is presented to the annotator as a single paragraph.
 
\paragraph{Item-level evaluation} For each review item, the domain scientist provides three judgments (correctness, significance, sufficiency of evidence) structured as a cascading evaluation. The exact criteria, options, and wording presented to annotators are given in \autoref{appendix:meta_review_criteria}.
 
\paragraph{Paper-level overall survey} After evaluating all individual review items, the domain scientist completes a paper-level survey comparing the reviewers, as summarized in \autoref{subsec:annotators}. The survey asks:
\begin{itemize}[leftmargin=*]
    \item ``Which human reviewer do you think provided the best quality review overall?'' (single-select among the paper's human reviewers).
    \item ``Which human reviewer do you think provided the worst quality review overall?'' (single-select).
    \item ``Compared to the best human review, which AI-generated reviews do you think match or exceed that level of quality?'' (multi-select among the three AI reviewers).
    \item ``Compared to the worst human review, which AI-generated reviews do you think match or exceed that level of quality?'' (multi-select).
    \item For each AI reviewer: ``Did you notice anything in AI Reviewer $k$'s review that other reviewers (both AI and human) did not catch?'' (free-form).
\end{itemize}
 
\noindent
Annotators may also add a free-form comment on the paper's review history as a whole. The full annotation sheet and the code used to generate it are released as part of our public code release.

\subsection{Inter-annotator agreement details}\label{appendix:irr_details}

\begin{table}[h]
\centering
\small
\setlength{\tabcolsep}{4pt}
\begin{tabular}{llrccc}
\toprule
\textbf{Dimension} & \textbf{Review type} & $N$ & \textbf{\%Agree.} & \textbf{Cohen's $\kappa$} & \textbf{Gwet's AC1} \\
\midrule
\multirow{3}{*}{Correctness (binary)}
  & Human  & 568 & 84.7\% & 0.12\;{\scriptsize(slight)}        & 0.81\;{\scriptsize(alm. perf.)} \\
  & AI     & 340 & 87.6\% & 0.47\;{\scriptsize(moderate)}       & 0.84\;{\scriptsize(alm. perf.)} \\
  & \textbf{All} & \textbf{908} & \textbf{85.8\%} & \textbf{0.28}\;{\scriptsize(fair)} & \textbf{0.82}\;{\scriptsize(alm. perf.)} \\
\midrule
\multirow{3}{*}{Significance (3-way ord.)}
  & Human  & 470 & 60.2\% & 0.34\;{\scriptsize(fair)}           & 0.43\;{\scriptsize(moderate)} \\
  & AI     & 273 & 59.3\% & 0.22\;{\scriptsize(fair)}           & 0.45\;{\scriptsize(moderate)} \\
  & \textbf{All} & \textbf{743} & \textbf{59.9\%} & \textbf{0.31}\;{\scriptsize(fair)} & \textbf{0.44}\;{\scriptsize(moderate)} \\
\midrule
\multirow{3}{*}{Evidence (binary)}
  & Human  & 353 & 83.9\% & 0.06\;{\scriptsize(slight)}         & 0.81\;{\scriptsize(alm. perf.)} \\
  & AI     & 230 & 94.3\% & 0.29\;{\scriptsize(fair)}           & 0.94\;{\scriptsize(alm. perf.)} \\
  & \textbf{All} & \textbf{583} & \textbf{88.0\%} & \textbf{0.12}\;{\scriptsize(slight)} & \textbf{0.86}\;{\scriptsize(alm. perf.)} \\
\bottomrule
\end{tabular}
\caption{Full inter-annotator agreement breakdown across 908 doubly-annotated review items from 27 doubly-annotated papers, separated by reviewer type (human vs.\ AI). $N$ decreases across dimensions due to the cascading annotation structure: significance is only annotated for correct items, and evidence sufficiency only for items that are correct and at least marginally significant. Interpretation labels follow \citet{landis1977measurement}. \textit{alm.\ perf.} = almost perfect agreement.}
\label{tab:irr_full}
\end{table}

As shown in \autoref{tab:irr_full}, comparing the human and AI subsets, Gwet's AC1 is consistently high on correctness and sufficiency of evidence, and moderate on significance, across both subsets. Raw percent agreement is similar between human and AI items on correctness (84.7\% vs.\ 87.6\%) and significance (60.2\% vs.\ 59.3\%), but higher on AI items for evidence (94.3\% vs.\ 83.9\%). The pattern across dimensions mirrors the main-text result: judgments of correctness and evidence sufficiency are largely stable across annotators, while significance is the most subjective dimension.

\section{Extended: In which aspects are AI reviewers better or worse than human reviewers?}\label{appendix:extended_analysis_results}

In this section, we provide the supplementary analyses referenced in \autoref{sec:performance}: item-level descriptive statistics across all reviewers and dimensions (\autoref{appendix:item_level_stats}), complete pairwise paired comparisons across the five reviewer groups on each primary evaluation metric (\autoref{appendix:pairwise_comparisons}), and a generalized linear mixed-effects model (GLMM) robustness analysis with paper-level random intercepts (\autoref{appendix:glmm}).
 
\subsection{Item-level descriptive statistics}\label{appendix:item_level_stats}
 
\autoref{tab:ext-itemlevel} reports item-level counts and rates for each reviewer across the three evaluation dimensions (correctness, significance, evidence sufficiency). The main text (\autoref{tab:main-results}) reports paper-level means, which is the appropriate unit of analysis for the paired inferential tests. We include item-level rates here for completeness and for comparison with the GLMM analysis in \autoref{appendix:glmm}, which directly models item-level outcomes with paper-level random intercepts. The item-level rates are close to but not identical to the paper-level means because papers contribute different numbers of items: item-level rates implicitly weight papers by their item count, while the main-text paper-level means weight each paper equally.
 
\begin{table}[h]
\centering
\fontsize{6.5}{9.5}\selectfont
\setlength{\tabcolsep}{3pt}
\begin{tabular}{l cc c cc cc c cccc cc c cc cc}
\toprule
& \multicolumn{5}{c}{\textbf{Correctness}} & & \multicolumn{6}{c}{\textbf{Significance}} & & \multicolumn{5}{c}{\textbf{Evidence}} \\
\cmidrule(lr){2-6} \cmidrule(lr){8-13} \cmidrule(lr){15-19}
& & & & \multicolumn{2}{c}{\scriptsize Breakdown} & & & & & \multicolumn{3}{c}{\scriptsize Breakdown} & & & & & \multicolumn{2}{c}{\scriptsize Breakdown} \\
\cmidrule(lr){5-6} \cmidrule(lr){11-13} \cmidrule(lr){18-19}
\textbf{Reviewer} & {\scriptsize Rate [95\% CI]} & & $n$ & {\scriptsize Corr.} & {\scriptsize Incorr.} & & {\scriptsize Mean [95\% CI]} & & $n$ & {\scriptsize Sig.} & {\scriptsize Marg.} & {\scriptsize Not} & & {\scriptsize Rate [95\% CI]} & & $n$ & {\scriptsize Suff.} & {\scriptsize Insuff.} \\
\midrule
Top-Rated Human    & 93.6{\scriptsize\,[92.0, 94.9]} & & 1{,}139 & {\scriptsize 1{,}066} & {\scriptsize 73}  & & 1.38{\scriptsize\,[1.33, 1.42]} & & 1{,}066 & {\scriptsize 570} & {\scriptsize 330} & {\scriptsize 166} & & 93.0{\scriptsize\,[91.1, 94.5]} & & 900 & {\scriptsize 837} & {\scriptsize 63} \\
Lowest-Rated Human & 86.0{\scriptsize\,[83.4, 88.1]} & & 833     & {\scriptsize 716}     & {\scriptsize 117} & & 1.06{\scriptsize\,[0.99, 1.12]} & & 716     & {\scriptsize 272} & {\scriptsize 212} & {\scriptsize 232} & & 88.2{\scriptsize\,[85.0, 90.8]} & & 484 & {\scriptsize 427} & {\scriptsize 57} \\
GPT-5.2            & 86.9{\scriptsize\,[83.4, 89.7]} & & 442     & {\scriptsize 384}     & {\scriptsize 58}  & & 1.60{\scriptsize\,[1.54, 1.66]} & & 384     & {\scriptsize 260} & {\scriptsize 94}  & {\scriptsize 30}  & & 97.5{\scriptsize\,[95.2, 98.7]} & & 354 & {\scriptsize 345} & {\scriptsize 9} \\
Claude Opus 4.5    & 84.6{\scriptsize\,[81.1, 87.6]} & & 475     & {\scriptsize 402}     & {\scriptsize 73}  & & 1.56{\scriptsize\,[1.50, 1.62]} & & 402     & {\scriptsize 261} & {\scriptsize 106} & {\scriptsize 35}  & & 97.8{\scriptsize\,[95.8, 98.9]} & & 367 & {\scriptsize 359} & {\scriptsize 8} \\
Gemini 3.0 Pro     & 82.6{\scriptsize\,[78.9, 85.8]} & & 460     & {\scriptsize 380}     & {\scriptsize 80}  & & 1.59{\scriptsize\,[1.53, 1.65]} & & 380     & {\scriptsize 249} & {\scriptsize 105} & {\scriptsize 26}  & & 91.8{\scriptsize\,[88.5, 94.2]} & & 354 & {\scriptsize 325} & {\scriptsize 29} \\
\bottomrule
\end{tabular}
\caption{\textbf{Item-level descriptive statistics across all evaluation dimensions.}
Correctness and evidence rates use Wilson score 95\% CIs; significance uses non-parametric bootstrap CIs (10{,}000 iterations).
The cascading rubric (\autoref{subsec:rubric}) explains the decreasing $n$ across metrics: significance is annotated only on items rated Correct, evidence only on items rated Correct \emph{and} at least Marginally Significant.
Paper-level statistics and paired inferential comparisons are in \autoref{tab:main-results}.}
\label{tab:ext-itemlevel}
\end{table}
 
\subsection{Complete pairwise paired comparisons}\label{appendix:pairwise_comparisons}
 
\autoref{tab:main-results} in the main text reports the six pairwise comparisons of each AI reviewer against the two human baselines. For completeness, \autoref{tab:pairwise-full} reports the full set of $\binom{5}{2} = 10$ pairwise comparisons across all five reviewer groups on each of the three primary metrics. Binary metrics (correctness, evidence sufficiency) use paired $t$-tests on per-paper rate differences with Cohen's $d$ as the effect size; the three-level ordinal significance score uses Wilcoxon signed-rank tests with rank-biserial correlation $r$ as the effect size.
 
Three patterns in \autoref{tab:pairwise-full} complement the main-text findings. First, all three AI reviewers raise more significant issues than the Lowest-Rated Human at large effect sizes ($r = -0.43$ to $-0.56$), mirroring the Top-Rated Human comparison. Second, on correctness, the gap between the Lowest-Rated Human and each AI reviewer is directionally negative but smaller in magnitude than the Top-Rated Human gap, and does not reach significance for any AI reviewer. Third, among the three AI reviewers, GPT-5.2 has a modestly higher significance rate than Claude Opus 4.5 ($r = +0.30$, $p < .05$) and a higher evidence-sufficiency rate than Gemini 3.0 Pro ($d = +0.31$, $p < .01$); Claude Opus 4.5 and Gemini 3.0 Pro are indistinguishable on significance and on correctness.
 
\begin{table}[h]
\centering
\fontsize{7.5}{9.5}\selectfont
\setlength{\tabcolsep}{4pt}
\begin{tabular}{l ccc c}
\toprule
\textbf{Comparison} & \textbf{Correctness} & \textbf{Significance} & \textbf{Evidence Sufficiency} & \textbf{$n_{\text{paired}}$} \\
                    & {\scriptsize diff (Cohen's $d$)} & {\scriptsize diff (rank-biserial $r$)} & {\scriptsize diff (Cohen's $d$)} & {\scriptsize range} \\
\midrule
Top-Rated H. vs Lowest-Rated H. & $+13.2\%^{***}$\,({\scriptsize $+0.54$}) & $+0.13$\,({\scriptsize $+0.26$})       & $+2.7\%$\,({\scriptsize $+0.11$})       & 77--82 \\
Top-Rated H. vs GPT-5.2         & $+6.1\%^{*}$\,({\scriptsize $+0.23$})   & $-0.22^{***}$\,({\scriptsize $-0.49$}) & $-5.0\%^{*}$\,({\scriptsize $-0.23$})   & 79--81 \\
Top-Rated H. vs Claude Opus 4.5 & $+8.6\%^{**}$\,({\scriptsize $+0.34$})  & $-0.13^{*}$\,({\scriptsize $-0.30$})   & $-4.8\%^{*}$\,({\scriptsize $-0.24$})   & 78--81 \\
Top-Rated H. vs Gemini 3.0 Pro  & $+10.3\%^{***}$\,({\scriptsize $+0.42$}) & $-0.17^{**}$\,({\scriptsize $-0.42$})  & $+2.4\%$\,({\scriptsize $+0.10$})       & 80--82 \\
\midrule
Lowest-Rated H. vs GPT-5.2      & $-7.4\%$\,({\scriptsize $-0.22$})       & $-0.30^{***}$\,({\scriptsize $-0.56$}) & $-5.3\%^{*}$\,({\scriptsize $-0.23$})   & 74--81 \\
Lowest-Rated H. vs Claude Opus 4.5 & $-4.8\%$\,({\scriptsize $-0.13$})    & $-0.24^{**}$\,({\scriptsize $-0.43$})  & $-7.1\%^{*}$\,({\scriptsize $-0.27$})   & 73--81 \\
Lowest-Rated H. vs Gemini 3.0 Pro  & $-2.9\%$\,({\scriptsize $-0.09$})    & $-0.27^{**}$\,({\scriptsize $-0.46$})  & $+0.3\%$\,({\scriptsize $+0.01$})       & 75--82 \\
\midrule
GPT-5.2 vs Claude Opus 4.5      & $+2.8\%$\,({\scriptsize $+0.11$})      & $+0.11^{*}$\,({\scriptsize $+0.30$})   & $+2.0\%$\,({\scriptsize $+0.13$})       & 75--80 \\
GPT-5.2 vs Gemini 3.0 Pro       & $+4.5\%$\,({\scriptsize $+0.18$})      & $+0.05$\,({\scriptsize $+0.15$})       & $+7.0\%^{**}$\,({\scriptsize $+0.31$})  & 77--81 \\
Claude Opus 4.5 vs Gemini 3.0 Pro & $+1.5\%$\,({\scriptsize $+0.06$})    & $-0.06$\,({\scriptsize $-0.22$})       & $+6.6\%^{*}$\,({\scriptsize $+0.24$})   & 78--81 \\
\bottomrule
\end{tabular}
\caption{\textbf{All pairwise paired comparisons across the five reviewer groups, on each of the three primary evaluation metrics.}
Each row reports the paired difference A $-$ B with effect size in parentheses (Cohen's $d$ for binary metrics, rank-biserial $r$ for ordinal significance); positive = first-named reviewer higher.
$n_{\text{paired}}$ range reflects cascade-eligible paired-paper counts across the three metrics (highest for correctness; lower for significance and evidence due to additional cascade conditions).
$^{*}p < 0.05$, $^{**}p < 0.01$, $^{***}p < 0.001$.
Corresponding fully-positive comparisons are in \autoref{tab:fully-positive}.
}
\label{tab:pairwise-full}
\end{table}
 
\subsection{Generalized linear mixed-effects model robustness analysis}\label{appendix:glmm}
 
As a robustness check against the paper-level paired analysis in \autoref{sec:performance}, we fit a generalized linear mixed-effects model (GLMM) that directly models item-level outcomes with a paper-level random intercept. This complements the paired analysis in two ways: (i) it explicitly models the hierarchical structure of the data (items nested within papers) rather than aggregating to the paper level, yielding greater statistical power when the within-paper variance is small; and (ii) it separates within-paper from between-paper variance via the intraclass correlation coefficient (ICC).
 
\paragraph{Model specification} For each of the three primary metrics, we fit a Bayesian binomial generalized linear mixed model (\texttt{BinomialBayesMixedGLM} with variational Bayes) of the form:
\[
\text{logit}(P(y_{ij} = 1)) = \beta_0 + \sum_k \beta_k \cdot \mathbb{1}[\text{reviewer}_{ij} = k] + u_i, \qquad u_i \sim \mathcal{N}(0, \sigma_u^2)
\]
where $y_{ij}$ is the item-level outcome for item $j$ in paper $i$, $\mathbb{1}[\text{reviewer}_{ij} = k]$ is a reviewer indicator (dummy-coded against a chosen reference category), and $u_i$ is the paper-level random intercept. For the ordinal significance score, we fit the same binomial GLMM at the ``Significant'' cut point ($y_{ij} = 1$ iff the item is rated at the highest significance level, $P(Y = 2)$), rather than a full cumulative-link mixed model, to keep the three metrics on a common log-odds scale and directly interpretable as the probability of passing the ``Significant'' bar conditional on being correct.
 
\paragraph{Interpretation} Coefficients in \autoref{tab:glmm} are on the log-odds scale; a positive coefficient means the row reviewer has higher log-odds of the outcome than the reference category. Probability-scale changes at the reference can be computed as $\sigma(\text{intercept} + \text{coef}) - \sigma(\text{intercept})$, where $\sigma$ is the logistic function. The two halves of the table refit the same model with a different reference category (Top-Rated or Lowest-Rated Human) so that all pairwise contrasts to either human baseline can be read directly off a single row.
 
\begin{table}[h]
\centering
\fontsize{7.5}{10}\selectfont
\setlength{\tabcolsep}{4pt}
\begin{tabular}{l ccc ccc}
\toprule
& \multicolumn{3}{c}{\textbf{Reference: Top-Rated Human}} & \multicolumn{3}{c}{\textbf{Reference: Lowest-Rated Human}} \\
\cmidrule(lr){2-4} \cmidrule(lr){5-7}
& \textbf{Correctness} & \textbf{Significance} & \textbf{Evidence} & \textbf{Correctness} & \textbf{Significance} & \textbf{Evidence} \\
\midrule
Intercept            & $+2.860^{***}$ & $+0.228^{***}$ & $+3.050^{***}$ & $+1.925^{***}$ & $-0.251^{***}$ & $+2.706^{***}$ \\
                     & {\scriptsize (0.057)} & {\scriptsize (0.041)} & {\scriptsize (0.086)} & {\scriptsize (0.057)} & {\scriptsize (0.041)} & {\scriptsize (0.086)} \\[3pt]
Top-Rated Human      & ---            & ---            & ---            & $+0.968^{***}$ & $+0.479^{***}$ & $+0.356^{**}$  \\
                     &                &                &                & {\scriptsize (0.126)} & {\scriptsize (0.067)} & {\scriptsize (0.137)} \\[3pt]
Lowest-Rated Human   & $-0.931^{***}$ & $-0.488^{***}$ & $-0.344^{*}$   & ---            & ---            & ---            \\
                     & {\scriptsize (0.106)} & {\scriptsize (0.083)} & {\scriptsize (0.151)} &                &                &                \\[3pt]
GPT-5.2              & $-0.639^{***}$ & $+0.684^{***}$ & $+1.173^{***}$ & $+0.295^{*}$   & $+1.163^{***}$ & $+1.508^{***}$ \\
                     & {\scriptsize (0.149)} & {\scriptsize (0.117)} & {\scriptsize (0.339)} & {\scriptsize (0.149)} & {\scriptsize (0.117)} & {\scriptsize (0.338)} \\[3pt]
Claude Opus 4.5      & $-0.846^{***}$ & $+0.506^{***}$ & $+1.314^{***}$ & $+0.090$       & $+0.985^{***}$ & $+1.648^{***}$ \\
                     & {\scriptsize (0.136)} & {\scriptsize (0.113)} & {\scriptsize (0.358)} & {\scriptsize (0.136)} & {\scriptsize (0.113)} & {\scriptsize (0.356)} \\[3pt]
Gemini 3.0 Pro       & $-1.033^{***}$ & $+0.518^{***}$ & $-0.092$       & $-0.098$       & $+0.996^{***}$ & $+0.250$       \\
                     & {\scriptsize (0.132)} & {\scriptsize (0.116)} & {\scriptsize (0.204)} & {\scriptsize (0.132)} & {\scriptsize (0.116)} & {\scriptsize (0.204)} \\
\midrule
ICC (paper)          & 0.253          & 0.272          & 0.289          & 0.253          & 0.272          & 0.290          \\
Observations         & 3{,}349        & 2{,}948        & 2{,}459        & 3{,}349        & 2{,}948        & 2{,}459        \\
Papers               & 82             & 82             & 82             & 82             & 82             & 82             \\
\bottomrule
\end{tabular}
\caption{\textbf{Mixed-effects logistic regression confirms that AI reviewers raise more significant issues but are less factually correct than the Top-Rated Human reviewer.}
Coefficients on the log-odds scale from a Bayesian binomial GLMM (\texttt{BinomialBayesMixedGLM}, variational Bayes, paper random intercept); for significance, fit at the highest cut point ($P(Y = 2)$).
The two halves refit the same model with a different reference category so that all pairwise contrasts to either human baseline read off a single row.
Probability change at the reference: $\sigma(\text{intercept}+\text{coef})-\sigma(\text{intercept})$, $\sigma$ = logistic. SE in parentheses.
$^{*}p < 0.05$, $^{**}p < 0.01$, $^{***}p < 0.001$.
ICC (paper) = $\sigma_u^2/(\sigma_u^2 + \pi^2/3)$, the share of latent variance attributable to between-paper differences.
}
\label{tab:glmm}
\end{table}
 
\paragraph{Concordance with paper-level paired analysis} The GLMM reaches the same directional conclusions as the paper-level paired analysis in \autoref{sec:performance} on all three dimensions: every AI reviewer has significantly lower correctness and significantly higher significance than the Top-Rated Human, and GPT-5.2 and Claude Opus 4.5 have significantly higher evidence sufficiency while Gemini 3.0 Pro is indistinguishable. Several contrasts that are borderline in the paired analysis reach stronger significance in the GLMM (e.g., Top-Rated vs GPT-5.2 correctness: paired $p = .046$, GLMM $p < .001$), which is consistent with the GLMM's greater statistical power from modeling items directly rather than averaging to the paper level. The paper-level ICC of 0.25 to 0.29 indicates that roughly one quarter of the latent log-odds variance sits between papers, justifying the random-intercept specification and, equivalently, the paired structure of the main-text analysis.

\section{Extended: To what extent do AI reviewers overlap with human reviewers?}\label{appendix:similarity_details}
 
\begin{figure}[h]
    \centering
    \includegraphics[width=1\linewidth]{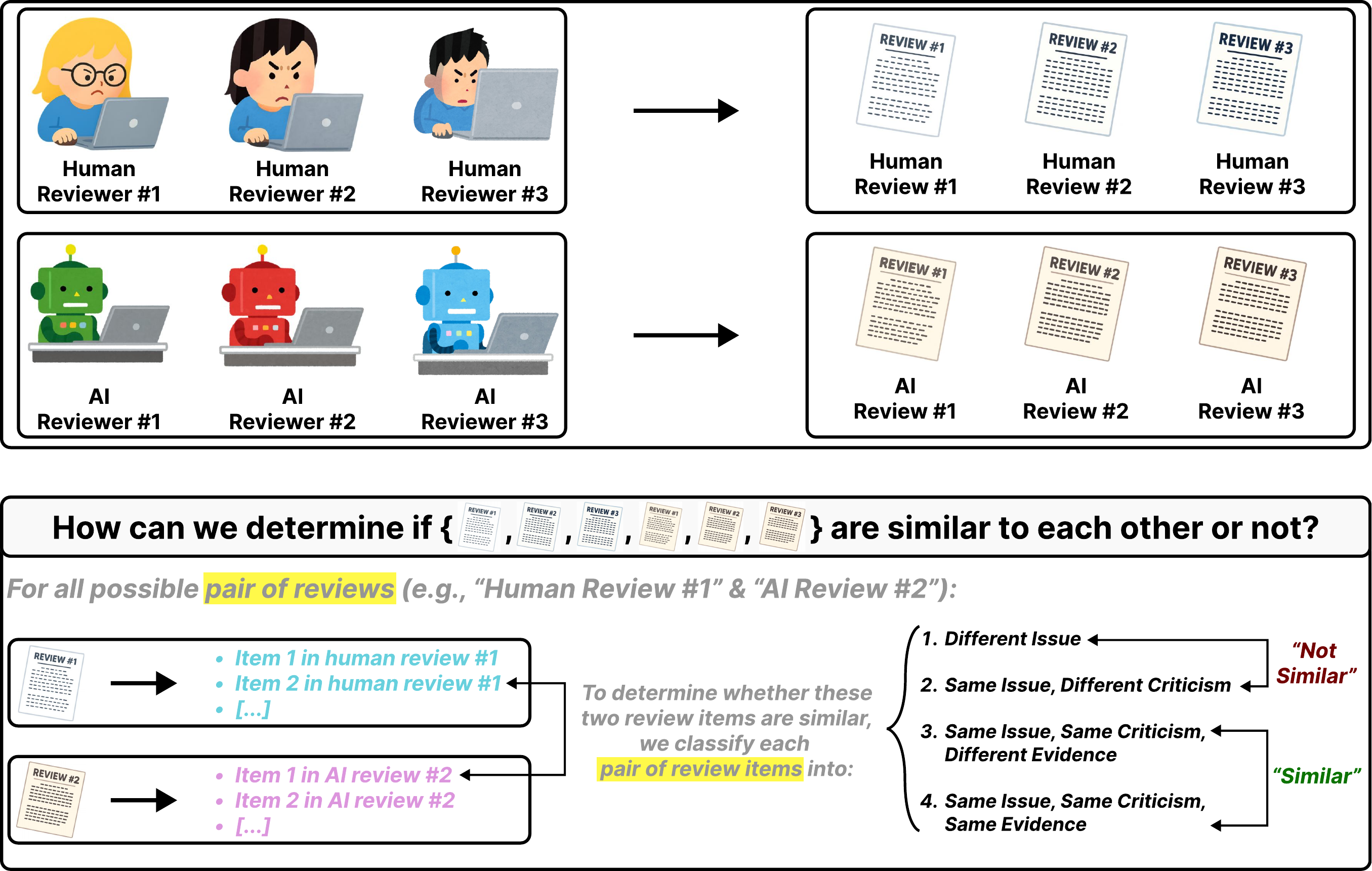}
    \caption{\textbf{Illustration of the motivation behind the similarity analysis.} For each paper in our expert-annotation study, we obtain six reviews: three from human reviewers and three from AI reviewers (top panel). To quantify how similar any two reviews are (e.g., human--human, human--AI, or AI--AI), we compare every review item in one review against every review item in the other and classify each item pair into one of four similarity categories: (1) different target, (2) same target but different criticism, (3) same target and criticism but different evidence, or (4) same target, criticism, and evidence (bottom right). Categories~1 and~2 are treated as \emph{not similar}; categories~3 and~4 are treated as \emph{similar}. Across the 82 papers, this procedure yields 65{,}704 item-pair comparisons in total.}
    \label{fig:motivation_similarity}
\end{figure}

This appendix collects the supplementary analyses referenced in \autoref{sec:overlap}: the full 4-category similarity breakdown across all pair types with within-reviewer and between-reviewer splits (\autoref{appendix:similarity_breakdown}), the similarity judge calibration and selection details (\autoref{appendix:similarity_judge_calibration}), and a side-by-side comparison of raw versus Rogan-Gladen-corrected prevalences (\autoref{appendix:rg_comparison}). \autoref{fig:motivation_similarity} illustrates the overall analysis pipeline that all three subsections build on.
 
\subsection{Detailed similarity breakdown}\label{appendix:similarity_breakdown}
 
\autoref{tab:similarity-headline} reports the full 4-category distribution of cross-reviewer review-item pairs for each pair type (Human-Human, AI-AI, and Human-AI), with within-reviewer and between-reviewer splits for the two same-group pair types. \autoref{fig:overlap} (left panel) in the main text visualizes the ``diff reviewer'', ``diff model'', and H-A rows of this table; the extended table adds the same-reviewer and same-model rows, which capture the diversity of items raised by a single reviewer within one paper rather than the agreement between different reviewers.
 
Two observations complement the main-text findings. First, the within-reviewer similarity rates (same-reviewer H-H: 6.0\%; same-model A-A: 13.5\%) are higher than the corresponding between-reviewer rates (diff-reviewer H-H: 3.4\%; diff-model A-A: 20.9\%) \emph{only} for human reviewers; for AI reviewers, different models of the same paper converge on the same criticism more often than two items written by the same model in the same review. This is consistent with the view that a single AI reviewer deliberately diversifies its own items across different parts of a paper, while different AI models, each independently identifying the paper's top concerns, end up surfacing many of the same targets. Second, the ``same target, different criticism'' category (topical co-occurrence without agreement on what is wrong) is roughly stable across pair types at the between-reviewer level (H-H: 8.4\%, A-A: 20.2\%, H-A: 14.8\%) and is responsible for a substantial share of the H-A difference from the H-H baseline, suggesting that AI reviewers often flag the same parts of a paper that human reviewers do, even when their specific criticisms do not align at the threshold used in the main text.
 
\begin{table}[h]
\centering
\fontsize{7}{9}\selectfont
\setlength{\tabcolsep}{3pt}
\begin{tabular}{l r cccc c}
\toprule
\textbf{Pair type} & \textbf{$N$} & \textbf{Same tgt., same crit., same evid.} & \textbf{Same tgt., same crit., diff evid.} & \textbf{Same tgt., diff crit.} & \textbf{Different tgt.} & \textbf{P(similar)} \\
\midrule
H--H (all) & 34{,}931 & 0.4{\scriptsize\,[0.0, 0.7]} & 4.1{\scriptsize\,[0.0, 8.2]} & 8.4{\scriptsize\,[6.9, 10.5]} & 87.1{\scriptsize\,[81.4, 91.8]} & 4.5{\scriptsize\,[0.0, 8.9]} \\
\quad same reviewer & 15{,}030 & 0.5{\scriptsize\,[0.1, 0.9]} & 5.5{\scriptsize\,[0.9, 9.7]} & 8.4{\scriptsize\,[6.9, 10.5]} & 85.7{\scriptsize\,[79.8, 90.9]} & 6.0{\scriptsize\,[0.9, 10.5]} \\
\quad diff reviewer & 19{,}901 & 0.3{\scriptsize\,[0.0, 0.6]} & 3.1{\scriptsize\,[0.0, 7.1]} & 8.4{\scriptsize\,[6.9, 10.8]} & 88.3{\scriptsize\,[82.6, 92.3]} & 3.4{\scriptsize\,[0.0, 7.7]} \\
\midrule
A--A (all) & 6{,}053 & 2.2{\scriptsize\,[1.6, 2.8]} & 16.6{\scriptsize\,[12.3, 20.6]} & 20.7{\scriptsize\,[18.3, 23.3]} & 60.5{\scriptsize\,[56.4, 64.8]} & 18.8{\scriptsize\,[14.0, 23.2]} \\
\quad same model & 1{,}724 & 0.1{\scriptsize\,[0.0, 0.3]} & 13.4{\scriptsize\,[8.2, 17.9]} & 22.0{\scriptsize\,[19.0, 25.2]} & 64.6{\scriptsize\,[60.0, 69.6]} & 13.5{\scriptsize\,[8.3, 18.1]} \\
\quad diff model & 4{,}329 & 3.1{\scriptsize\,[2.3, 3.9]} & 17.9{\scriptsize\,[13.8, 21.8]} & 20.2{\scriptsize\,[17.7, 22.7]} & 58.9{\scriptsize\,[54.9, 63.0]} & 20.9{\scriptsize\,[16.2, 25.4]} \\
\midrule
H--A & 23{,}642 & 0.3{\scriptsize\,[0.0, 0.5]} & 4.9{\scriptsize\,[0.2, 8.5]} & 14.8{\scriptsize\,[13.3, 16.5]} & 80.1{\scriptsize\,[76.2, 84.4]} & 5.1{\scriptsize\,[0.3, 9.0]} \\
\bottomrule
\end{tabular}
\caption{\textbf{Review-item-pair similarity across pair types.} $P(\text{similar})$ is the sum of the first two columns (same target, same criticism, regardless of evidence). Within H--H and A--A, rows are further split by whether both items come from the same reviewer (or the same AI model within a paper): ``same reviewer''/``same model'' rows capture within-reviewer pairs; ``diff reviewer''/``diff model'' rows capture between-reviewer pairs. All pairs are within the same paper. All values are Rogan-Gladen-corrected; 95\% CIs from 10{,}000-iteration paper-level bootstrap with resampled sensitivity/specificity.}
\label{tab:similarity-headline}
\end{table}

\subsection{Similarity judge calibration and selection}\label{appendix:similarity_judge_calibration}
 
To select an automated similarity judge for the full-scale analysis, we compared candidate models on a small, manually-annotated calibration set, then characterized the chosen model's error rates for use in the Rogan-Gladen prevalence correction. This subsection describes each step in turn.
 
\paragraph{Comparison of candidate judges} We compared two families of automated similarity judges on a 164-pair calibration set (described in the next paragraph). The first family is \textbf{embedding models} that compute cosine similarity between item embeddings (OpenAI's \texttt{text-embedding-3-large} or Gemini's \texttt{gemini-embedding-001}), with the binary similarity threshold tuned on the calibration set. The second family is \textbf{LLM-as-judge models} that prompt a frontier language model with both review items, the paper's text, relevant images, and any supplementary materials, to produce a 4-way label using extended-thinking reasoning. The LLM judges include six recent frontier models from three families: GPT-5.4 and GPT-5.4-mini from OpenAI, Claude Opus 4.7 and Claude Sonnet 4.6 from Anthropic, and Gemini 3.1 Pro and Gemini 3 Flash from Google. As shown in \autoref{tab:similarity-bench}, embedding baselines reach 77.4\% and 80.5\% binary accuracy, while the six LLM judges cluster tightly at 89 to 93\% binary accuracy and 80 to 84\% 4-way accuracy, with no systematic gap between families. For the full-scale overlap analysis, we use GPT-5.4 (92.7\% binary accuracy, 83.5\% 4-way accuracy), which achieves the highest 4-way accuracy among candidate similarity judges.
 
\begin{table}[h]
\centering
\fontsize{7}{9}\selectfont
\setlength{\tabcolsep}{4pt}
\begin{tabular}{l cc}
\toprule
& \multicolumn{2}{c}{\textbf{Agreement with Experts on ``Similarity''}} \\
\cmidrule(lr){2-3}
\textbf{Model} & \textbf{Binary} {\scriptsize [95\% CI]} & \textbf{4-way} {\scriptsize [95\% CI]} \\
\midrule
\multicolumn{3}{l}{\textit{Embedding models}} \\
\quad \texttt{text-embedding-3-large}   & 77.4{\scriptsize\,[70.5, 83.2]} & --- \\
\quad \texttt{gemini-embedding-001}     & 80.5{\scriptsize\,[73.8, 85.8]} & --- \\
\midrule
\multicolumn{3}{l}{\textit{LLM-as-judge models}} \\
\quad GPT-5.4                  & 92.7{\scriptsize\,[87.6, 95.8]} & 83.5{\scriptsize\,[77.1, 88.4]} \\
\quad Gemini 3.1 Pro           & 92.7{\scriptsize\,[87.6, 95.8]} & 81.1{\scriptsize\,[74.4, 86.4]} \\
\quad Claude Sonnet 4.6        & 92.7{\scriptsize\,[87.6, 95.8]} & 80.5{\scriptsize\,[73.8, 85.8]} \\
\quad Claude Opus 4.7          & 92.7{\scriptsize\,[87.6, 95.8]} & 79.9{\scriptsize\,[73.1, 85.3]} \\
\quad Gemini 3 Flash           & 92.1{\scriptsize\,[86.9, 95.3]} & 79.9{\scriptsize\,[73.1, 85.3]} \\
\quad GPT-5.4-mini             & 89.0{\scriptsize\,[83.3, 92.9]} & 80.5{\scriptsize\,[73.8, 85.8]} \\
\bottomrule
\end{tabular}
\caption{\textbf{Agreement with expert decisions on the 164-pair similarity calibration set.} Binary judgment: same target and same main criticism vs.\ not. 4-way judgment: exact match with the full taxonomy. 95\% Wilson score CIs in brackets.}
\label{tab:similarity-bench}
\end{table}
 
\paragraph{Calibration set construction} The 164-pair calibration set is drawn from domain scientists' paper-level survey responses to the free-form question ``Did you notice anything in AI Reviewer $k$'s review that other reviewers did not catch?''. Specifically, some domain scientists explicitly listed specific AI review items as unique to that reviewer (\textit{i.e.}, having no counterpart in any other reviewer's review of the paper), which provided ground-truth not-similar pairs; we then identified candidate similar pairs by manually matching AI review items to items from other reviewers of the same paper. The 94 \textit{not similar} pairs were sampled from annotator-flagged unique items and manually subdivided into \textit{different target} (67 pairs) or \textit{same target, different criticism} (27 pairs); the 70 \textit{similar} pairs were sampled from AI items not flagged as unique, manually matched to items from other reviewers, and subdivided into \textit{convergent conclusion} (48 pairs) or \textit{near-paraphrase} (22 pairs). The resulting 164-pair calibration set covers 32 papers and 266 unique review items, with 85 AI-AI pairs and 79 AI-human pairs. Because its labels derive from the same pool of annotations used throughout our study, we treat it as a calibration set for our automated judge rather than an independent benchmark.
 
\paragraph{Confusion matrix of the chosen judge} \autoref{tab:confusion-matrix-gpt54} reports GPT-5.4's binary similar / not-similar predictions on the 164-pair calibration set, from which we derive the judge's sensitivity (87.1\%) and specificity (96.8\%); these two values are the inputs to the Rogan-Gladen prevalence correction below.
 
\begin{table}[h]
\centering
\small
\begin{tabular}{lccc}
\toprule
 & \textbf{Predicted Similar} & \textbf{Predicted Not Similar} & \textbf{Total} \\
\midrule
True Similar     & 61 (TP) & 9 (FN)  & 70 \\
True Not Similar & 3 (FP)  & 91 (TN) & 94 \\
\midrule
\textbf{Total}   & 64      & 100     & 164 \\
\bottomrule
\end{tabular}
\caption{Confusion matrix of GPT-5.4 on the binary similar/not-similar classification task over the 164-pair calibration set. Sensitivity = 61/70 = 0.871; specificity = 91/94 = 0.968.}
\label{tab:confusion-matrix-gpt54}
\end{table}
 
\paragraph{Rogan-Gladen prevalence correction} Given an apparent prevalence $p$ of similar pairs among classifier predictions on a large set, the corrected estimate of the true prevalence $\pi$ is:
\begin{equation}
\pi = \frac{p + \text{Spec} - 1}{\text{Sens} + \text{Spec} - 1} = \frac{p - 0.032}{0.840}.
\end{equation}
The correction is valid when $p \geq 1 - \text{Spec} = 0.032$; when apparent prevalences fall below this threshold we clip the corrected estimate to zero.
 
\paragraph{Cluster-bootstrap confidence intervals} To construct 95\% confidence intervals for each Rogan-Gladen-corrected prevalence, we use a cluster bootstrap at the paper level. In each of 10{,}000 bootstrap iterations, we (i) resample 82 papers with replacement, (ii) pool all cross-reviewer pairs of the relevant type from the resampled papers to form a new apparent prevalence $p^*$, (iii) resample the calibration set's sensitivity and specificity from binomials with the observed parameters ($\text{Sens} \sim \text{Binomial}(70, 0.871)/70$ and $\text{Spec} \sim \text{Binomial}(94, 0.968)/94$), and (iv) apply the Rogan-Gladen formula to $p^*$, $\text{Sens}^*$, $\text{Spec}^*$. The 2.5th and 97.5th percentiles of the resulting distribution give the 95\% confidence interval. This procedure propagates both the within-paper classifier error (via sensitivity/specificity resampling) and the cross-paper variance (via paper-level resampling).
 
\paragraph{Limitations} The Rogan-Gladen correction assumes that the judge's sensitivity and specificity are constant across the full set of 65{,}704 pairs. Since the calibration set does not include Human-Human pairs (it was constructed around AI reviews), we assume that the judge's error rates transfer to Human-Human comparisons; this is a plausible but untested assumption.
 
\subsection{Raw versus Rogan-Gladen-corrected comparison}\label{appendix:rg_comparison}
 
All percentages in \autoref{fig:overlap} (left) and \autoref{tab:similarity-headline} are adjusted for the GPT-5.4 similarity judge's classification error using the Rogan-Gladen prevalence correction. \autoref{tab:rg-comparison} reports the raw (uncorrected) prevalences alongside the corrected values for the three between-reviewer pair types that appear in the left panel of \autoref{fig:overlap}, so that readers can assess the magnitude and direction of the correction.
 
The correction shifts the same-target-same-criticism categories downward (reflecting the judge's false-positive rate on the calibration set) and shifts the not-similar categories upward, consistent with the judge's estimated sensitivity of 87.1\% and specificity of 96.8\%. The magnitude of the shift is largest for the H-H and H-A pair types, where the raw similar-pair rates are low (around 5 to 8\%) and are therefore more sensitive to the asymmetry between the judge's false-positive and false-negative rates; for A-A pairs, where raw similar rates are already substantial (around 20\%), the correction changes the estimate by less than one percentage point. The main-text conclusions are qualitatively unchanged by whether the raw or corrected values are used, but we report the corrected values throughout for consistency with the calibration-set sensitivity/specificity estimates.
 
\begin{table}[t]
\centering
\fontsize{7}{9}\selectfont
\setlength{\tabcolsep}{3pt}
\begin{tabular}{l l r r r r}
\toprule
\textbf{Pair type} & \textbf{Category} & $N$ & \textbf{Raw (\%)} & \textbf{RG-corrected (\%)} & \textbf{95\% CI} \\
\midrule
H-H (diff reviewer) & Same target, same crit., same evid. & 19{,}901 & 0.5 & 0.3 & [0.0, 0.6] \\
 & Same target, same crit., diff evid. &  & 5.5 & 3.1 & [0.0, 7.1] \\
 & Same target, diff criticism &  & 8.1 & 8.4 & [6.9, 10.8] \\
 & Different target &  & 85.8 & 88.3 & [82.6, 92.3] \\
\midrule
H-A & Same target, same crit., same evid. & 23{,}642 & 0.4 & 0.3 & [0.0, 0.5] \\
 & Same target, same crit., diff evid. &  & 7.1 & 4.9 & [0.2, 8.5] \\
 & Same target, diff criticism &  & 14.4 & 14.8 & [13.3, 16.5] \\
 & Different target &  & 78.1 & 80.1 & [76.2, 84.5] \\
\midrule
A-A (diff model) & Same target, same crit., same evid. & 4{,}329 & 3.0 & 3.1 & [2.3, 3.9] \\
 & Same target, same crit., diff evid. &  & 17.7 & 17.9 & [13.7, 21.8] \\
 & Same target, diff criticism &  & 20.2 & 20.2 & [17.7, 22.7] \\
 & Different target &  & 59.0 & 58.9 & [54.8, 63.1] \\
\bottomrule
\end{tabular}
\caption{\textbf{Raw versus Rogan-Gladen-corrected similarity prevalences for \autoref{fig:overlap} (left).} The correction adjusts for the GPT-5.4 similarity judge's binary misclassification rate (sensitivity = 87.1\%, specificity = 96.8\%, on the 164-pair calibration set). It is applied to the binary similar/not-similar boundary; within each binary group, the 4-way split is maintained proportionally from the raw data. 95\% CIs are cluster-bootstrap (10{,}000 paper-level resamples) with sensitivity and specificity themselves resampled from Binomial distributions per iteration to propagate calibration uncertainty.}
\label{tab:rg-comparison}
\end{table}

\section{Extended: What are the concrete strength and weaknesses of AI reviewers?}\label{appendix:sw_all_comments}

This appendix contains the complete qualitative record behind the strength/weakness analysis introduced in \autoref{sec:strengths_weaknesses}. \autoref{appendix:sw_case_studies} presents 24 detailed case-study examples spanning the nine most informative categories, each shown with the original review item, the expert annotator's commentary, and our discussion. \autoref{appendix:sw_raw_comments} lists every one of the 442 substantive expert comments behind the S/W classification, organized by category in longtable form, so that every claim in the main text and every example in \autoref{appendix:sw_case_studies} is traceable back to the underlying data.

\subsection{Detailed case-study examples}\label{appendix:sw_case_studies}

This subsection contains the complete set of expert-annotated strength and weakness examples for the categories introduced in \autoref{sec:strengths_weaknesses}. Three of these examples (W1~Example~\#1, W3~Example~\#2, S2~Example~\#1) are also shown in the main text; they are reproduced here so that the subsection is self-contained as a complete atlas of all 24 examples. The examples are organized by category in the same numbering used in \autoref{fig:strength_weakness}, and each example preserves the structure of the main-text examples (review-item box with claim, criticism, evidence, and AI commentary; expert-comment box; and a short discussion paragraph).

\subsubsection{W1: Missing community / field norms}\label{appendix:sw_w1}

The three examples below illustrate the severity-miscalibration pattern in three different subfields (particle physics, bacterial community modeling, agricultural ecology). Example~\#1 is also shown in the main text (\autoref{qualitative_analysis_weaknesses}) and reproduced here for completeness.

\begin{mdframed}[style=categoryboxW, frametitle={\textbf{AI Reviewer Weakness \#1: Missing community / field norms ($n = 54$)}}]
\footnotesize
The AI reviewer flags something as a methodological gap that is in fact accepted or expected practice within the subfield of the paper. The content of the critique is typically accurate against a discipline-neutral standard (for reproducibility, sample size, out-of-distribution validation, and so on), but the severity is miscalibrated because the reviewer lacks the accumulated knowledge of what the field treats as normal.
\end{mdframed}

\begin{mdframed}[
    style=reviewbox,
    frametitle={\textbf{Example \#1 for W1 \textemdash{} Particle physics paper, reviewer: GPT-5.2}},
]
\footnotesize

\noindent\textcolor{blue!60!black}{\textbf{Claim: ``Analysis choices are not documented enough for independent reproduction''}}

\vspace{2pt}\par\noindent\colorbox{blue!5}{\begin{minipage}{0.97\linewidth}
\vspace{4pt}
\begin{itemize}[leftmargin=1.2em, itemsep=2pt, topsep=0pt]
\item \textbf{Main point of criticism:} The manuscript's analysis choices (simulation-fixed fit shapes, external calibration inputs, and pseudoexperiment-based uncertainties) are not documented in enough detail, and are not accompanied by reusable released artifacts, to enable independent reproduction or reanalysis of the key result.
\item \textbf{Evaluation criteria:} Data and methodology; Reproducibility
\end{itemize}
\vspace{2pt}
\end{minipage}}

\vspace{8pt}\par\noindent\textcolor{blue!60!black}{\textbf{Evidence:}}

\vspace{2pt}\par\noindent\colorbox{blue!5}{\begin{minipage}{0.97\linewidth}
\vspace{4pt}
\textcolor{blue!50!black}{\textit{\textbf{Quote 1}} (from main text):}
\textit{``The signal shape is modelled as a combination of a Gaussian and two Crystal Ball functions, all sharing the same peak position. The parameters of the signal function are determined from simulated events and are fixed in the fit to data, except for the Gaussian width, the average width of the Crystal Ball functions, and the peak position, which accounts for imperfections in the simulation.''}
\\
\commentbar{The extracted yields, and therefore the measured asymmetry, depend on simulation-derived shape parameters that are fixed rather than floated in the data fit. For an external reader, reproducing the measurement requires access not only to the data but also to the exact simulation configuration, the derived fixed parameter values, and the fitting implementation. The manuscript does not tabulate these fixed parameters or provide a machine-readable specification of the full likelihood model, which would make independent reproduction infeasible even in principle. Providing the parameter values and a compact statistical model description (or a public likelihood artifact) would significantly strengthen reproducibility.}

\vspace{2pt}
\end{minipage}}

\end{mdframed}

\begin{mdframed}[style=expertcommentbox]
\footnotesize\textbf{Expert comment.}~It is true that these information are not documented in the paper, however these are kept internally at CERN and the authors are not expected to share any data or simulation when publishing the paper for the full reproduction of the results. This is done in exceptional cases where relevant work has been done to provide data for example to the theory community. This comment  for this particular paper is irrelevant.
\end{mdframed}
\vspace{10pt}

The AI reviewer's request is reasonable by generic open-science standards: tabulated fit parameters, a machine-readable likelihood, and enough metadata for an external reanalysis. None of that is factually incorrect, and in some subfields it would be a legitimate revision request. But in LHCb physics at Nature-level submission, the relevant calibration and fit-configuration artifacts are maintained and used internally by the collaboration rather than published alongside the paper, and a specialist reviewer knows this by default. The critique is therefore not wrong in content so much as miscalibrated in severity: what the AI reads as a reproducibility gap is the community's normal publication boundary.

\begin{mdframed}[
    style=reviewbox,
    frametitle={\textbf{Example \#2 for W1 \textemdash{} Bacterial community modeling paper, reviewer: GPT-5.2}},
]
\footnotesize

\noindent\textcolor{blue!60!black}{\textbf{Claim: ``The `universal collapse' claim is not established beyond the studied system''}}

\vspace{2pt}\par\noindent\colorbox{blue!5}{\begin{minipage}{0.97\linewidth}
\vspace{4pt}
\begin{itemize}[leftmargin=1.2em, itemsep=2pt, topsep=0pt]
\item \textbf{Main point of criticism:} The paper makes an overgeneralized claim that invasion probability universally collapses onto survival fraction across conditions, but the main theoretical support relies on an exchangeable generalized Lotka--Volterra setup that can mechanically enforce such a collapse, and the experiments do not establish universality beyond the studied library and protocol.
\item \textbf{Evaluation criteria:} Validity; Conclusions
\end{itemize}
\vspace{2pt}
\end{minipage}}

\vspace{8pt}\par\noindent\textcolor{blue!60!black}{\textbf{Evidence:}}

\vspace{2pt}\par\noindent\colorbox{blue!5}{\begin{minipage}{0.97\linewidth}
\vspace{4pt}
\textcolor{blue!50!black}{\textit{\textbf{Quote 1}} (from main text):}
\textit{``Despite these conflicting diversity-invasibility relationships, after scaling richness with species pool size to get the survival fraction, we found that all communities collapsed to a universal line in which the invasion probability is approximately equal to the survival fraction (Fig.~4b).''}
\\
\commentbar{The sentence asserts a universal relationship, but the paper's own phrasing indicates that the collapse is obtained after a specific rescaling (richness divided by pool size), not from an independently validated mechanism. A collapse can occur for multiple reasons, including shared definitions and exchangeability assumptions, without implying universality across ecological contexts. The manuscript does not provide out-of-distribution validation across different taxa, interaction types, or invasion protocols. Because later conclusions extrapolate this relationship to natural communities, the evidentiary standard should be higher than a within-system collapse. The paper would need clearer limits on scope, or additional tests demonstrating that the relationship persists under qualitatively different invasion settings.}

\vspace{2pt}
\end{minipage}}

\end{mdframed}

\begin{mdframed}[style=expertcommentbox]
\footnotesize\textbf{Expert comment.}~The main point of the critique is correct. But to the bacterial modeling community this paper does not read that it claims a universal law. The AI reviewer is correct with the request for clearer limits on scope and additional tests demonstrating the relationship persists under qualitatively different invasion settings. It is harsh, but correct.
\end{mdframed}
\vspace{10pt}

This example is a cleaner dissociation of content from framing than the main-text example. The expert explicitly endorses the AI's methodological point (it is harsh, but correct) and agrees that the scope should be tightened. What the expert disputes is the AI's reading of the paper as a universality claim in the first place: to someone familiar with how the bacterial modeling literature uses the word ``universal,'' the paper is making a bounded empirical statement rather than an unconditional physical law. The AI reviewer, operating on the text alone, does not have the community-level prior that calibrates this reading. The resulting review item is a correct technical request bundled with an overly strong interpretation of what the authors had actually claimed.

\begin{mdframed}[
    style=reviewbox,
    frametitle={\textbf{Example \#3 for W1 \textemdash{} Agricultural ecology paper, reviewer: Claude Opus 4.5}},
]
\footnotesize

\noindent\textcolor{blue!60!black}{\textbf{Claim: ``Weakly predictive models are used to support strong global projections''}}

\vspace{2pt}\par\noindent\colorbox{blue!5}{\begin{minipage}{0.97\linewidth}
\vspace{4pt}
\begin{itemize}[leftmargin=1.2em, itemsep=2pt, topsep=0pt]
\item \textbf{Main point of criticism:} The models explaining biodiversity responses to yield have marginal pseudo-$R^{2}$ values ranging from 0.05 to 0.36, and the authors acknowledge that their ``models do not fully fit model assumptions,'' yet these weakly predictive models are used to project biodiversity changes globally, generating estimates of species losses up to 89\% that should be interpreted with substantial uncertainty.
\item \textbf{Evaluation criteria:} Validity; Appropriate use of statistics
\end{itemize}
\vspace{2pt}
\end{minipage}}

\vspace{8pt}\par\noindent\textcolor{blue!60!black}{\textbf{Evidence:}}

\vspace{2pt}\par\noindent\colorbox{blue!5}{\begin{minipage}{0.97\linewidth}
\vspace{4pt}
\textcolor{blue!50!black}{\textit{\textbf{Quote 1}} (from main text):}
\textit{``The variance explained by fixed effects (marginal pseudo-$R^{2}$) for all models varied between 0.05 and 0.36, whereas the variance explained by fixed and random effects together (conditional pseudo-$R^{2}$) varied between 0.52 and 0.98.''}
\\
\commentbar{A marginal $R^{2}$ of 0.05 means that the fixed effects (including yield and its interactions) explain only 5\% of the variance in biodiversity outcomes, with the remainder explained by random effects such as study-level variation and spatial blocks. While low marginal $R^{2}$ is common in PREDICTS-based analyses, using such models to project global biodiversity changes generates predictions where most of the variance is driven by factors not captured in the projections, since random effects cannot be extrapolated beyond the sampled studies. The projection of 89.1\% maximum species loss relies on these weakly predictive relationships.}

\vspace{2pt}
\end{minipage}}

\end{mdframed}

\begin{mdframed}[style=expertcommentbox]
\footnotesize\textbf{Expert comment.}~I think the AI does not understand how hard ecology is! R$^{2}$ of 0.36 is impressive!
\end{mdframed}
\vspace{10pt}

The AI reviewer applies thresholds for ``weakly predictive'' that are common in a controlled experimental setting but are very far from the noise floor of observational ecology on cross-study datasets. The expert's single-sentence reaction captures this neatly: by agricultural-ecology standards, a marginal $R^{2}$ of $0.36$ on a variance-dominated response is an unusually strong fixed-effect signal, not an alarming one. The AI's numeric description of the situation is accurate (the $R^{2}$ values are what they are, and the authors do acknowledge assumption violations); what goes wrong is the implicit reference class the AI is comparing against, which is not the reference class the field actually uses. The same numerical result therefore supports opposite severity assessments depending on the reviewer's field prior.

\subsubsection{W2: Over-harsh, out-of-scope, or unrealistic demands}\label{appendix:sw_w2}

\begin{mdframed}[style=categoryboxW, frametitle={\textbf{AI Reviewer Weakness \#2: Over-harsh, out-of-scope, or unrealistic demands ($n = 46$)}}]
\footnotesize
The AI reviewer raises a criticism that is technically valid in isolation but falls outside the scope of the paper or asks for additional experiments that the authors cannot realistically provide. The claim itself is usually not wrong; what is missing is the reviewer's calibration of severity to what the paper is actually claiming and what is feasible to change in a revision.
\end{mdframed}

\begin{mdframed}[
    style=reviewbox,
    frametitle={\textbf{Example \#1 for W2 \textemdash{} Bacterial community modeling paper, reviewer: GPT-5.2}},
]
\footnotesize

\noindent\textcolor{blue!60!black}{\textbf{Claim: ``Continued dispersal of the invader contaminates the invasion-success measurement''}}

\vspace{2pt}\par\noindent\colorbox{blue!5}{\begin{minipage}{0.97\linewidth}
\vspace{4pt}
\begin{itemize}[leftmargin=1.2em, itemsep=2pt, topsep=0pt]
\item \textbf{Main point of criticism:} The paper's core experimental outcome measures treat invasion success and priority effects as emergent community properties, but the protocol includes continued dispersal and immigration of the invader and residents after invasion, so ``invasion success'' is not cleanly separated from sustained propagule pressure and may not reflect intrinsic community resistance or true priority effects.
\item \textbf{Evaluation criteria:} Validity; Conclusions
\end{itemize}
\vspace{2pt}
\end{minipage}}

\vspace{8pt}\par\noindent\textcolor{blue!60!black}{\textbf{Evidence:}}

\vspace{2pt}\par\noindent\colorbox{blue!5}{\begin{minipage}{0.97\linewidth}
\vspace{4pt}
\textcolor{blue!50!black}{\textit{\textbf{Quote 1}} (from main text):}
\textit{``After six days of culturing, each community was exposed to an invader species (Fig.~1a) and we continued to culture the communities for another 6 days with dispersal of all species on each dilution cycle (Fig.~1a, b).''}
\\
\commentbar{The invader is not a one-time introduction; instead, it experiences ongoing immigration throughout the invasion assay, which is a central driver of establishment in invasion ecology. Under such conditions, the measured ``invasion probability'' can reflect the imposed propagule pressure regime as much as community resistance mechanisms. The paper interprets differences between regimes as emergent outcomes of resident interactions, rather than as outcomes shaped by continued introductions. The manuscript does not present a parallel experiment with a single invader pulse to show that the main conclusions are robust to this distinction.}

\vspace{2pt}
\end{minipage}}

\end{mdframed}

\begin{mdframed}[style=expertcommentbox]
\footnotesize\textbf{Expert comment.}~This is too harsh a critique. The authors do not claim that their result is universal, although the AI would have had a good point if they did. This critique is somewhat along the lines of Human Reviewer 2.
\end{mdframed}
\vspace{10pt}

The AI reviewer is correctly describing an experimental design choice (continued dispersal of the invader) and correctly identifying that it complicates the interpretation of ``invasion success'' as an intrinsic community property. The critique is not factually wrong. What the expert objects to is the severity of the request: the AI asks for a parallel experiment with a single-pulse invader, which would roughly double the wet-lab workload of the study, on the basis of a conclusion (universal intrinsic invasion resistance) that the paper in fact does not claim. The expert notes that a human reviewer raised a similar concern but calibrated to what the paper actually argues rather than to the strongest possible reading of it.

\begin{mdframed}[
    style=reviewbox,
    frametitle={\textbf{Example \#2 for W2 \textemdash{} Neuroimaging of psychiatric disorders paper, reviewer: Claude Opus 4.5}},
]
\footnotesize

\noindent\textcolor{blue!60!black}{\textbf{Claim: ``Healthy-derived predictors cannot be validly applied to patient populations''}}

\vspace{2pt}\par\noindent\colorbox{blue!5}{\begin{minipage}{0.97\linewidth}
\vspace{4pt}
\begin{itemize}[leftmargin=1.2em, itemsep=2pt, topsep=0pt]
\item \textbf{Main point of criticism:} All molecular predictors (gene expression, receptor densities, metabolism, synapse density, myelination) and connectivity predictors are derived exclusively from healthy adult participants, yet these are used to predict and explain cortical abnormalities in patient populations where these features may be substantially altered by disease processes.
\item \textbf{Evaluation criteria:} Validity; Appropriate use of data
\end{itemize}
\vspace{2pt}
\end{minipage}}

\vspace{8pt}\par\noindent\textcolor{blue!60!black}{\textbf{Evidence:}}

\vspace{2pt}\par\noindent\colorbox{blue!5}{\begin{minipage}{0.97\linewidth}
\vspace{4pt}
\textcolor{blue!50!black}{\textit{\textbf{Quote 1}} (from main text):}
\textit{``Gene expression data was collected by the Allen Human Brain Atlas as described in Hawrylycz et al.''}
\\
\commentbar{The Allen Human Brain Atlas is derived from postmortem brain tissue of donors without known neurological or psychiatric disorders. However, the disorders studied in this paper (schizophrenia, depression, epilepsy, Parkinson's disease, and so on) are known to involve altered gene expression patterns in the brain. Using healthy-derived gene expression to predict patient abnormalities assumes that the spatial pattern of gene expression is preserved in disease, which may not be valid for many of the studied conditions.}

\vspace{2pt}
\end{minipage}}

\end{mdframed}

\begin{mdframed}[style=expertcommentbox]
\footnotesize\textbf{Expert comment.}~This critique is factually accurate but it highlights a limitation that the authors cannot realistically defend against or resolve by modifying their data, making it a non-constructive critique.
\end{mdframed}
\vspace{10pt}

The expert explicitly agrees with the AI on content: the Allen Human Brain Atlas is indeed healthy-donor data, and the studied disorders do indeed alter the relevant spatial patterns. What makes the critique fall under W2 is not factual error but that there is no actionable path forward: a patient-derived atlas of gene expression across the relevant disorders simply does not exist, and producing one is orders of magnitude beyond the scope of any single neuroimaging paper. A human reviewer typically recognises this asymmetry and reframes the observation as a limitation to be acknowledged in the discussion rather than as a correctness concern. The AI does not perform this reframing and instead flags it as a validity issue, which escalates the severity beyond what the paper can plausibly respond to.

\begin{mdframed}[
    style=reviewbox,
    frametitle={\textbf{Example \#3 for W2 \textemdash{} Near-field optical spectroscopy paper, reviewer: GPT-5.2}},
]
\footnotesize

\noindent\textcolor{blue!60!black}{\textbf{Claim: ``Fitting procedure for the central quantitative outputs is not fully reproducible''}}

\vspace{2pt}\par\noindent\colorbox{blue!5}{\begin{minipage}{0.97\linewidth}
\vspace{4pt}
\begin{itemize}[leftmargin=1.2em, itemsep=2pt, topsep=0pt]
\item \textbf{Main point of criticism:} The paper does not provide sufficient information and runnable artifacts to reproduce its central quantitative outputs (complex dielectric function, radiative rate/lifetime, damping rate) derived from fitting s-SNOM spectra with the point dipole model.
\item \textbf{Evaluation criteria:} Data and methodology; Reproducibility
\end{itemize}
\vspace{2pt}
\end{minipage}}

\vspace{8pt}\par\noindent\textcolor{blue!60!black}{\textbf{Evidence:}}

\vspace{2pt}\par\noindent\colorbox{blue!5}{\begin{minipage}{0.97\linewidth}
\vspace{4pt}
\textcolor{blue!50!black}{\textit{\textbf{Quote 1}} (from main text):}
\textit{``As shown in Figs.~2c,d, the point dipole model with Eq.~1 as the input dielectric function provides an adequate fit to the data (see SOM for details on fitting).''}
\\
\commentbar{The paper's core quantitative claims rely on a fitting procedure whose details are deferred to the SOM, but the fitting workflow (objective function, constraints, priors, initialization, treatment of background, and handling of parameter correlations) is not described in the main text. Without those details, an independent reader cannot reproduce the reported parameter values or assess whether the fits are unique and robust to alternative model choices. The extracted parameters are sensitive to normalization, tip-sample coupling, and baseline assumptions in s-SNOM analyses, so providing a full fitting recipe with sensitivity analysis, or shared scripts that deterministically regenerate the fit results, is standard for quantitative near-field spectroscopy.}

\vspace{2pt}
\end{minipage}}

\end{mdframed}

\begin{mdframed}[style=expertcommentbox]
\footnotesize\textbf{Expert comment.}~This is such a harsh comment for an experimental study. The authors provided all the datasets. The measurements were taken several times and averages were reported. The real world measurements are not perfect or noise free. What else could authors do?
\end{mdframed}
\vspace{10pt}

The expert's reaction here is the clearest version of the W2 pattern: the reviewer has essentially made a conceptually correct reproducibility point (fitting details matter, sensitivity analyses are good practice), but the request implies a standard of experimental reproducibility that is not achievable for the measurement in question. s-SNOM data are intrinsically noisy, the reported values already come from averaging multiple runs, and the ``fitting recipe with sensitivity analysis and deterministic scripts'' the AI asks for is closer to what a computational paper provides than what an experimental physics group can realistically deliver. The expert's rhetorical ``what else could authors do?'' captures the asymmetry: the critique gives the authors no room to respond because the thing being asked for is not a thing that exists in their workflow.

\subsubsection{W3: Paper explicitly states X, AI says missing}\label{appendix:sw_w3}

The three examples below illustrate the reading-failure pattern in three distinct variants: (i) Example~\#1, a structural variant where the AI fails to recognize information present in an unlabeled section; (ii) Example~\#2 (also shown in main text), a direct-omission variant where the AI claims absent a procedure that is explicitly described; and (iii) Example~\#3, a partial-coverage variant where the AI underweights cross-section evidence.

\begin{mdframed}[style=categoryboxW, frametitle={\textbf{AI Reviewer Weakness \#3: Paper explicitly states X, AI says missing ($n = 37$)}}]
\footnotesize
The AI reviewer asserts that a piece of information is absent from the manuscript when in fact the paper provides it, whether in a different section, the supplementary material, or elsewhere in the same passage. The critique itself would often be valid if the information were truly missing, but the factual premise is wrong. This happens because of limited long context management: Through out writing a review, an AI reviewer must read multiple files (context from the main text, supplementary materials, code, information retrieved from the web) and when it exceeds its context length limit, it summarizes the context, resulting in losing information it should hold to process information and write a coherent review.
\end{mdframed}

\begin{mdframed}[
    style=reviewbox,
    frametitle={\textbf{Example \#1 for W3 \textemdash{} 11.7T MRI methodology paper, reviewer: Gemini 3.0 Pro}},
]
\footnotesize

\noindent\textcolor{blue!60!black}{\textbf{Claim: ``The imaging sequences and reconstruction methods are not described''}}

\vspace{2pt}\par\noindent\colorbox{blue!5}{\begin{minipage}{0.97\linewidth}
\vspace{4pt}
\begin{itemize}[leftmargin=1.2em, itemsep=2pt, topsep=0pt]
\item \textbf{Main point of criticism:} The manuscript fails to provide a detailed description of the imaging sequences and reconstruction methods used to acquire the ``first in vivo images,'' violating reproducibility standards.
\item \textbf{Evaluation criteria:} Data and methodology; Reproducibility
\end{itemize}
\vspace{2pt}
\end{minipage}}

\vspace{8pt}\par\noindent\textcolor{blue!60!black}{\textbf{Evidence:}}

\vspace{2pt}\par\noindent\colorbox{blue!5}{\begin{minipage}{0.97\linewidth}
\vspace{4pt}
\textcolor{blue!50!black}{\textit{\textbf{Quote 1}} (from main text):}
\textit{``We tackled this problem by deploying parallel transmission hardware and software, first developed by the laboratory at 7T and further optimized for 11.7T (Methods).''}
\\
\commentbar{The ``Methods'' section referenced here is missing from the manuscript. The only methods sections present are ``Genotoxic tests,'' ``Behavioral tests,'' and ``Balance tests,'' leaving the core imaging methodology undefined.}

\vspace{2pt}
\end{minipage}}

\end{mdframed}

\begin{mdframed}[style=expertcommentbox]
\footnotesize\textbf{Expert comment.}~The AI made factual errors. The sequence parameters were reported, and the methods section starts with this. It is true that the MRI methodology does not have a subheading, which is why the AI got confused.
\end{mdframed}
\vspace{10pt}

The AI reviewer looked for a section explicitly labelled ``Methods'' or with an imaging-related subheading, did not find one, and concluded that the imaging methodology was absent. The expert's correction is precise: the imaging methodology is in the manuscript, and the sequence parameters are reported, but the section that contains them does not carry a recognisable heading, which is what misled the AI. This is the characteristic failure mode of W3 at the structural level: the AI indexes a paper by headings and fails to recover information that is present in the running text. A human reviewer scans for content rather than labels and would not have missed this.

\begin{mdframed}[
    style=reviewbox,
    frametitle={\textbf{Example \#2 for W3 \textemdash{} Future air-quality projections paper, reviewer: Claude Opus 4.5}},
]
\footnotesize

\noindent\textcolor{blue!60!black}{\textbf{Claim: ``Known CMIP6 PM2.5 bias is not corrected for''}}

\vspace{2pt}\par\noindent\colorbox{blue!5}{\begin{minipage}{0.97\linewidth}
\vspace{4pt}
\begin{itemize}[leftmargin=1.2em, itemsep=2pt, topsep=0pt]
\item \textbf{Main point of criticism:} The paper does not adequately discuss or correct for the known systematic underestimation of PM\textsubscript{2.5} concentrations by CMIP6 models, particularly in highly polluted regions like China, which could bias the DAPP projections.
\item \textbf{Evaluation criteria:} Data and methodology; Validity
\end{itemize}
\vspace{2pt}
\end{minipage}}

\vspace{8pt}\par\noindent\textcolor{blue!60!black}{\textbf{Evidence:}}

\vspace{2pt}\par\noindent\colorbox{blue!5}{\begin{minipage}{0.97\linewidth}
\vspace{4pt}
\textcolor{blue!50!black}{\textit{\textbf{Quote 1}} (from main text):}
\textit{``We calculated future PM\textsubscript{2.5} concentration in China based on empirical formulas and future pollutant concentration data.''}
\\
\commentbar{The paper mentions using CMIP6 pollutant concentration data but does not discuss bias correction. Multiple studies have documented that CMIP6 models systematically underestimate surface PM\textsubscript{2.5} concentrations, particularly in regions with high pollution levels like East Asia, which could lead to underestimation of future DAPP if not addressed.}

\vspace{2pt}
\end{minipage}}

\end{mdframed}

\begin{mdframed}[style=expertcommentbox]
\footnotesize\textbf{Expert comment.}~This is factually incorrect. The paper explicitly addresses this issue. Lines 489--496 describe a calibration procedure (Equation 5) that adjusts modeled PM\textsubscript{2.5} concentrations against observed baseline period data (2012--2017).
\end{mdframed}
\vspace{10pt}

The AI reviewer is not missing a subheading; it is missing an entire calibration procedure (Equation 5, with its own methodological description spanning seven lines) that is already in the manuscript and addresses exactly the bias the AI is raising. The expert's reply cites the line numbers because the content is unambiguously present. What makes this critique fall under W3 rather than a more fundamental validity concern is that the underlying methodological point is valid in general (CMIP6 does underestimate surface PM\textsubscript{2.5} in East Asia), but the accusation that the paper ignores it is factually wrong. A reviewer who reads the full Methods section will see the bias correction; the AI apparently did not.

\begin{mdframed}[
    style=reviewbox,
    frametitle={\textbf{Example \#3 for W3 \textemdash{} Structural variant discovery paper, reviewer: GPT-5.2}},
]
\footnotesize

\noindent\textcolor{blue!60!black}{\textbf{Claim: ``Validation is not independent of training data''}}

\vspace{2pt}\par\noindent\colorbox{blue!5}{\begin{minipage}{0.97\linewidth}
\vspace{4pt}
\begin{itemize}[leftmargin=1.2em, itemsep=2pt, topsep=0pt]
\item \textbf{Main point of criticism:} The paper's conclusion that SurVIndel2 yields a substantially more complete 1000 Genomes CNV catalogue with ``high validation rate'' is insufficiently supported because the validation is reported only ``in samples with long reads'' that overlap the HGSVC2 cohort used to train and benchmark SurVIndel2's filtering model, making the validation non-independent and potentially optimistic.
\item \textbf{Evaluation criteria:} Validity; Appropriate use of statistics
\end{itemize}
\vspace{2pt}
\end{minipage}}

\vspace{8pt}\par\noindent\textcolor{blue!60!black}{\textbf{Evidence:}}

\vspace{2pt}\par\noindent\colorbox{blue!5}{\begin{minipage}{0.97\linewidth}
\vspace{4pt}
\textcolor{blue!50!black}{\textit{\textbf{Quote 1}} (from main text):}
\textit{``The machine learning model of SurVIndel2 was trained using the 34 samples in HGSVC2 sampled by the NYGC. For this reason, when testing them, we use a leave-one-out strategy.''}
\\
\commentbar{Leave-one-out within the same 34-sample cohort reduces direct per-sample label leakage, but it does not provide an independent external validation of the learned filtering behavior. The training and evaluation data share the same sequencing center, platform, library preparation, and mapping pipeline characteristics, which can create homogeneity and risk overfitting. When the same cohort is used to compute ``validation rates'' for catalogue-level claims, the evaluation can become circular, and reported validation is measured against resources tied to the training distribution. A stronger design would validate the 1000 Genomes discoveries on long-read samples not used in model training or tuning, ideally sequenced and processed independently.}

\vspace{2pt}
\end{minipage}}

\end{mdframed}

\begin{mdframed}[style=expertcommentbox]
\footnotesize\textbf{Expert comment.}~Another valid point that is somewhat offset by the benchmarking on other organisms.
\end{mdframed}
\vspace{10pt}

This example shows a subtler variant of W3: the information the AI treats as absent is not entirely absent but only partial. The AI notes, correctly, that validation within the HGSVC2 cohort shares sequencing center and pipeline with the training data and therefore is not a fully independent test. The expert acknowledges the point but observes that the paper does benchmark SurVIndel2 on other organisms (non-human genomes) elsewhere, which the AI did not mention and which provides exactly the kind of out-of-distribution evaluation the AI says is missing. The critique is therefore half-right: the human-genome validation is indeed not independent, but the paper's overall validation strategy is stronger than the AI represents because it extends beyond human data. This kind of partial coverage is common in W3 cases where the AI reads one section closely but does not integrate it with evidence elsewhere in the manuscript.

\subsubsection{W4: Redundancy across the three AI reviewers}\label{appendix:sw_w4}

\begin{mdframed}[style=categoryboxW, frametitle={\textbf{AI Reviewer Weakness \#4: Redundancy across the three AI reviewers ($n = 28$)}}]
\footnotesize
The three AI reviewers assigned to the same paper tend to converge on substantially overlapping criticisms, so a second or third AI reviewer often adds little information that the first has not already raised. This weakness is inherent to the AI panel as a whole rather than to any single review item; it becomes visible only when the reviews for the same paper are read side by side. For this reason, each example in this category shows all three AI reviewers' overlapping review items together. Please note that we look into this phenomenon in more detail at \autoref{sec:overlap}.
\end{mdframed}


\begin{mdframed}[
    style=reviewbox,
    frametitle={\textbf{Example \#1-a for W4 \textemdash{} Machine-learning interatomic potentials paper, reviewer: GPT-5.2}},
]
\footnotesize

\noindent\textcolor{blue!60!black}{\textbf{Claim: ``The 1000$\times$ data-efficiency claim may be inflated by train/test correlation''}}

\vspace{2pt}\par\noindent\colorbox{blue!5}{\begin{minipage}{0.97\linewidth}
\vspace{4pt}
\begin{itemize}[leftmargin=1.2em, itemsep=2pt, topsep=0pt]
\item \textbf{Main point of criticism:} The headline conclusions about extreme data efficiency and superiority over prior ML interatomic potentials are not robust because key comparisons (notably the ``1000$\times$ fewer data'' DeepMD comparison) are evaluated on highly correlated MD trajectory frames with train/test splits that can leak near-duplicate configurations.
\item \textbf{Evaluation criteria:} Validity; Appropriate use of statistics
\end{itemize}
\vspace{2pt}
\end{minipage}}

\vspace{8pt}\par\noindent\textcolor{blue!60!black}{\textbf{Evidence:}}

\vspace{2pt}\par\noindent\colorbox{blue!5}{\begin{minipage}{0.97\linewidth}
\vspace{4pt}
\textcolor{blue!50!black}{\textit{\textbf{Quote 1}} (from abstract):}
\textit{``NequIP outperforms existing models with up to three orders of magnitude fewer training data, challenging the widely held belief that deep neural networks require massive training sets.''}
\\
\commentbar{This is the central, strongest quantitative claim in the abstract, and it is comparative (``outperforms existing models''), so it requires an evaluation protocol that rules out confounders. In MD-generated datasets, adjacent frames can be extremely similar, so randomly sampled training frames can have near-neighbors in the test set even when they are formally disjoint. If such leakage occurs, test errors can substantially underestimate true generalization error. The paper does not establish that its orders-of-magnitude comparisons are robust to these well-known correlation effects.}

\vspace{2pt}
\end{minipage}}

\end{mdframed}

\begin{mdframed}[
    style=reviewbox,
    frametitle={\textbf{Example \#1-b for W4 \textemdash{} Machine-learning interatomic potentials paper, reviewer: Claude Opus 4.5}},
]
\footnotesize

\noindent\textcolor{blue!60!black}{\textbf{Claim: ``The NequIP vs DeepMD comparison is not a valid data-efficiency test''}}

\vspace{2pt}\par\noindent\colorbox{blue!5}{\begin{minipage}{0.97\linewidth}
\vspace{4pt}
\begin{itemize}[leftmargin=1.2em, itemsep=2pt, topsep=0pt]
\item \textbf{Main point of criticism:} The comparison between NequIP (133 training structures) and DeepMD (133{,}500 training structures) in Table III does not constitute a valid data-efficiency comparison because it compares different architectures at vastly different operating points rather than showing learning curves for both methods.
\item \textbf{Evaluation criteria:} Validity; Appropriate use of statistics
\end{itemize}
\vspace{2pt}
\end{minipage}}

\vspace{8pt}\par\noindent\textcolor{blue!60!black}{\textbf{Evidence:}}

\vspace{2pt}\par\noindent\colorbox{blue!5}{\begin{minipage}{0.97\linewidth}
\vspace{4pt}
\textcolor{blue!50!black}{\textit{\textbf{Quote 1}} (from main text):}
\textit{``A DeepMD NN-IP model was previously trained for water and ice using a joint training set containing 133{,}500 reference calculations of these four systems. To assess data efficiency of the NequIP architecture, we similarly train a model jointly on all four parts of the data set, but using only 133 structures for training, i.e.\ 1000x fewer data.''}
\\
\commentbar{The comparison shows that NequIP achieves better force RMSE with 133 structures than DeepMD does with 133{,}500 structures, but this does not demonstrate data efficiency in the proper sense. Data efficiency would require showing that NequIP consistently achieves lower errors than DeepMD at the same training set size, or equivalently, that NequIP reaches a target accuracy with fewer data points. The comparison as presented could simply mean that NequIP is a fundamentally better architecture regardless of data size, rather than being more data-efficient.}

\vspace{2pt}
\end{minipage}}

\end{mdframed}

\begin{mdframed}[
    style=reviewbox,
    frametitle={\textbf{Example \#1-c for W4 \textemdash{} Machine-learning interatomic potentials paper, reviewer: Gemini 3.0 Pro}},
]
\footnotesize

\noindent\textcolor{blue!60!black}{\textbf{Claim: ``The NequIP vs DeepMD comparison lacks a direct control experiment''}}

\vspace{2pt}\par\noindent\colorbox{blue!5}{\begin{minipage}{0.97\linewidth}
\vspace{4pt}
\begin{itemize}[leftmargin=1.2em, itemsep=2pt, topsep=0pt]
\item \textbf{Main point of criticism:} The claim that NequIP requires ``orders of magnitude fewer training data'' is not rigorously supported because the comparison with DeepMD is unfair (comparing NequIP trained on 133 samples against DeepMD trained on 133{,}500 samples) and lacks a direct control experiment in which DeepMD is trained on the same 133 samples as NequIP.
\item \textbf{Evaluation criteria:} Validity; Data and methodology
\end{itemize}
\vspace{2pt}
\end{minipage}}

\vspace{8pt}\par\noindent\textcolor{blue!60!black}{\textbf{Evidence:}}

\vspace{2pt}\par\noindent\colorbox{blue!5}{\begin{minipage}{0.97\linewidth}
\vspace{4pt}
\textcolor{blue!50!black}{\textit{\textbf{Quote 1}} (from main text):}
\textit{``Table III shows the comparison of the predictive force accuracy of NequIP trained on the 133 structures vs DeepMD trained on 133{,}500 structures. We find that with 1000x fewer training data, NequIP significantly outperforms DeepMD on all four parts of the data set.''}
\\
\commentbar{Table III compares NequIP (133 samples) with DeepMD (133{,}500 samples). This comparison is flawed because it does not show how DeepMD performs on 133 samples. It is possible that DeepMD would perform poorly, but without that data point, the ``orders of magnitude'' claim is an inference based on the assumption that DeepMD needs 133{,}500 samples to reach that performance, which is not proven. A fair comparison would involve training DeepMD (and other baselines) on the same 133 samples to demonstrate the specific advantage of NequIP in the low-data regime.}

\vspace{2pt}
\end{minipage}}

\end{mdframed}

\begin{mdframed}[style=expertcommentbox]
\footnotesize\textbf{Expert comment.}~To a large extent, all three AI reviewers raise very similar points. Thus, the overall review would not cover many substantial points a human reviewer raised.
\end{mdframed}
\vspace{10pt}

All three reviewers converge on the same observation about the NequIP vs.\ DeepMD comparison (different sample counts, no shared operating point), differing only in phrasing and emphasis. GPT-5.2 frames it as train/test correlation, Claude Opus as missing learning curves, and Gemini as a missing control experiment, but the underlying concern is identical. The expert annotator's comment captures the cost: panel-level redundancy means a multi-AI review covers fewer distinct concerns than a multi-human review of the same paper.


\begin{mdframed}[
    style=reviewbox,
    frametitle={\textbf{Example \#2-a for W4 \textemdash{} Compound temperature-COVID mortality paper, reviewer: GPT-5.2}},
]
\footnotesize

\noindent\textcolor{blue!60!black}{\textbf{Claim: ``The `compound mortality' calculation violates its own independence assumption''}}

\vspace{2pt}\par\noindent\colorbox{blue!5}{\begin{minipage}{0.97\linewidth}
\vspace{4pt}
\begin{itemize}[leftmargin=1.2em, itemsep=2pt, topsep=0pt]
\item \textbf{Main point of criticism:} The paper's central ``compound mortality impact'' calculation adds temperature-attributable deaths and COVID-19 deaths under an explicit independence and additivity assumption, even though the manuscript itself presents evidence and prior literature that the two hazards can interact and share causal pathways; this affects the core results on heatwaves and cold snaps and the interpretation of ``compound'' burdens.
\item \textbf{Evaluation criteria:} Validity; Conclusions
\end{itemize}
\vspace{2pt}
\end{minipage}}

\vspace{8pt}\par\noindent\textcolor{blue!60!black}{\textbf{Evidence:}}

\vspace{2pt}\par\noindent\colorbox{blue!5}{\begin{minipage}{0.97\linewidth}
\vspace{4pt}
\textcolor{blue!50!black}{\textit{\textbf{Quote 1}} (from main text):}
\textit{``These temperature-related deaths are theoretically independent of COVID-19 deaths because they are calculated from DLNMs that describe the relationships between daily mean temperature and daily all-cause mortality, after COVID-19 mortality has been removed (see Methods).''}
\\
\commentbar{Removing deaths with COVID-19 on the death certificate changes the outcome series, but it does not establish causal independence between temperature and pandemic conditions. COVID-19 can change population vulnerability, healthcare access, and reporting or diagnostic patterns, all of which can modify the temperature--mortality relationship even if COVID-coded deaths are excluded. The claim of theoretical independence is therefore stronger than what the described preprocessing supports, and because the paper's main ``compound'' totals depend on this premise, any violation can materially bias compound magnitudes.}

\vspace{2pt}
\end{minipage}}

\end{mdframed}

\begin{mdframed}[
    style=reviewbox,
    frametitle={\textbf{Example \#2-b for W4 \textemdash{} Compound temperature-COVID mortality paper, reviewer: Gemini 3.0 Pro}},
]
\footnotesize

\noindent\textcolor{blue!60!black}{\textbf{Claim: ``Removing COVID deaths does not make the remaining mortality pandemic-independent''}}

\vspace{2pt}\par\noindent\colorbox{blue!5}{\begin{minipage}{0.97\linewidth}
\vspace{4pt}
\begin{itemize}[leftmargin=1.2em, itemsep=2pt, topsep=0pt]
\item \textbf{Main point of criticism:} The paper incorrectly assumes that removing COVID-19 deaths from the all-cause mortality series makes the remaining temperature-related mortality ``theoretically independent'' of the pandemic, ignoring the significant indirect effects of the pandemic (healthcare strain, behavioral changes) that likely altered the temperature--mortality relationship for non-COVID causes.
\item \textbf{Evaluation criteria:} Validity; Appropriate use of statistics
\end{itemize}
\vspace{2pt}
\end{minipage}}

\vspace{8pt}\par\noindent\textcolor{blue!60!black}{\textbf{Evidence:}}

\vspace{2pt}\par\noindent\colorbox{blue!5}{\begin{minipage}{0.97\linewidth}
\vspace{4pt}
\textcolor{blue!50!black}{\textit{\textbf{Quote 1}} (from main text):}
\textit{``These temperature-related deaths are theoretically independent of COVID-19 deaths because they are calculated from DLNMs that describe the relationships between daily mean temperature and daily all-cause mortality, after COVID-19 mortality has been removed (see Methods).''}
\\
\commentbar{This claim of independence is methodologically unsound. The non-COVID mortality during the pandemic was not independent of the pandemic context; it was influenced by factors such as overwhelmed healthcare systems, delayed treatments for chronic conditions, and lockdowns. Recent literature has shown that the risk of heat-related mortality was significantly elevated during the pandemic compared to pre-pandemic periods, even for non-COVID causes, due to these synergistic effects. By using a model trained primarily on pre-pandemic data to estimate mortality in 2020--2022, the study fails to capture this altered vulnerability.}

\vspace{2pt}
\end{minipage}}

\end{mdframed}

\begin{mdframed}[
    style=reviewbox,
    frametitle={\textbf{Example \#2-c for W4 \textemdash{} Compound temperature-COVID mortality paper, reviewer: Claude Opus 4.5}},
]
\footnotesize

\noindent\textcolor{blue!60!black}{\textbf{Claim: ``Vulnerable populations overlap, so additive compound estimates can be biased''}}

\vspace{2pt}\par\noindent\colorbox{blue!5}{\begin{minipage}{0.97\linewidth}
\vspace{4pt}
\begin{itemize}[leftmargin=1.2em, itemsep=2pt, topsep=0pt]
\item \textbf{Main point of criticism:} The paper assumes that temperature-related deaths and COVID-19 deaths are independent and simply adds them together to estimate compound mortality impacts. However, vulnerable populations for heat and cold mortality and COVID-19 mortality substantially overlap (elderly, those with chronic conditions), which violates the independence assumption and may lead to either double-counting or underestimation of compound effects.
\item \textbf{Evaluation criteria:} Validity; Conclusions
\end{itemize}
\vspace{2pt}
\end{minipage}}

\vspace{8pt}\par\noindent\textcolor{blue!60!black}{\textbf{Evidence:}}

\vspace{2pt}\par\noindent\colorbox{blue!5}{\begin{minipage}{0.97\linewidth}
\vspace{4pt}
\textcolor{blue!50!black}{\textit{\textbf{Quote 1}} (from main text):}
\textit{``By adding temperature-related deaths (with COVID-19 deaths removed prior to calculation) and COVID-19 deaths together to estimate the compound mortality impact, we have assumed that they are independent of each other.''}
\\
\commentbar{The paper explicitly acknowledges this assumption but does not adequately address its implications for the validity of the compound mortality estimates. Simply removing COVID-19 deaths from the mortality data before fitting the temperature--mortality model does not account for potential effect modification, where COVID-19 may have amplified temperature--mortality relationships through shared vulnerabilities, behavioral changes, or healthcare system strain. Recent studies using interaction models have found that heat-related mortality was significantly elevated during COVID-19 periods compared to pre-pandemic periods, suggesting the relationship is not simply additive but involves effect modification.}

\vspace{2pt}
\end{minipage}}

\end{mdframed}

\begin{mdframed}[style=expertcommentbox]
\footnotesize\textbf{Expert comment.}~AI Reviewer 1 (Items 2 and 3) identified both the violated independence assumption and the misleading ``compound'' terminology. AI Reviewer 2 (Item 1) restated it with additional references. This reviewer articulates it more formally, mentioning joint models, attributable fraction partitioning, and counterfactual decomposition, but the core insight is identical.
\end{mdframed}
\vspace{10pt}

This second example sharpens what the side-by-side view can reveal. GPT-5.2 and Gemini both quote the exact same sentence from the Methods (``theoretically independent of COVID-19 deaths''), while Claude Opus quotes the paper's own acknowledgment of the additive assumption two paragraphs later; all three then argue that removing COVID-coded deaths does not make the remaining mortality pandemic-independent. The divergence is in emphasis rather than substance: Gemini adds references to heat--COVID interaction literature, Claude adds quantitative estimates (RR 1.07 vs 1.55), GPT frames the issue in terms of formal statistical decomposition. Each addition is useful, but they are elaborations of a single observation that any one reviewer could have made on their own.

\subsubsection{W5: Vague, verbose, or without actionable recommendation}\label{appendix:sw_w5}

\begin{mdframed}[style=categoryboxW, frametitle={\textbf{AI Reviewer Weakness \#5: Vague, verbose, or without actionable recommendation ($n = 24$)}}]
\footnotesize
The AI reviewer identifies a concern at length but does not translate it into a concrete change the authors could make, either because the review is too long and diffuse to pinpoint the main issue, or because the critique ends without any actionable revision request. Both patterns leave the authors and the editor unsure what would satisfy the reviewer.
\end{mdframed}

\begin{mdframed}[
    style=reviewbox,
    frametitle={\textbf{Example \#1 for W5 \textemdash{} 11.7T MRI methodology paper, reviewer: GPT-5.2}},
]
\footnotesize

\noindent\textcolor{blue!60!black}{\textbf{Claim: ``Safety conclusions at 11.7T are drawn from an underpowered protocol''}}

\vspace{2pt}\par\noindent\colorbox{blue!5}{\begin{minipage}{0.97\linewidth}
\vspace{4pt}
\begin{itemize}[leftmargin=1.2em, itemsep=2pt, topsep=0pt]
\item \textbf{Main point of criticism:} The paper draws overly strong conclusions about human safety and tolerability at 11.7T from a small, short-term, endpoint-limited protocol, which weakens the validity of the safety claims made in the abstract and main text.
\item \textbf{Evaluation criteria:} Validity; Appropriate use of statistics
\end{itemize}
\vspace{2pt}
\end{minipage}}

\vspace{8pt}\par\noindent\textcolor{blue!60!black}{\textbf{Evidence:}}

\vspace{2pt}\par\noindent\colorbox{blue!5}{\begin{minipage}{0.97\linewidth}
\vspace{4pt}
\textcolor{blue!50!black}{\textit{\textbf{Quote 1}} (from main text):}
\textit{``To assure the safety of human imaging at such high field strength, we performed physiological, vestibular, behavioral and genotoxicity measurements on the volunteers. The data shows no evidence of adverse effects.''}
\\
\commentbar{``No evidence of adverse effects'' is a non-equivalence statement that depends strongly on sample size, endpoint sensitivity, and follow-up duration. The described measures target a subset of plausible acute and subacute effects, but do not address detection of rare events or delayed outcomes, which is central when translating a first-in-human ultra-high-field exposure into a generalised safety assertion. The paper does not report an a priori safety endpoint hierarchy, minimally important differences, or any power or sensitivity analysis for the bioeffects outcomes, so readers cannot assess what magnitude of harm the study could have detected. As written, the conclusion materially overreaches what the reported study design can support.}

\vspace{2pt}
\end{minipage}}

\end{mdframed}

\begin{mdframed}[style=expertcommentbox]
\footnotesize\textbf{Expert comment.}~Too verbose. I am not a statistics expert, but this is a paper about MR methodology. The first comment should not be six paragraphs about biological effects.
\end{mdframed}
\vspace{10pt}

The AI reviewer's underlying statistical point is defensible: a small, short-term safety study cannot support an unqualified ``safe'' claim, and the paper should scope its safety language accordingly. But the expert's complaint is not with the content, it is with the volume and placement. The review opens with six paragraphs on biological effects in a paper whose primary contribution is magnetic-resonance engineering, and the statistical critique is spread across those six paragraphs rather than stated once crisply. A shorter review that said ``soften the safety language and report a sensitivity analysis'' would have delivered the same point with a tenth of the text and would have made the actionable request unambiguous. W5 is most commonly this kind of length-to-signal imbalance: the AI reviewer knows where the weakness is but buries it in elaboration.

\begin{mdframed}[
    style=reviewbox,
    frametitle={\textbf{Example \#2 for W5 \textemdash{} Agricultural ecology paper, reviewer: Gemini 3.0 Pro}},
]
\footnotesize

\noindent\textcolor{blue!60!black}{\textbf{Claim: ``Intensification projection is inconsistent with the landscape-scale framing''}}

\vspace{2pt}\par\noindent\colorbox{blue!5}{\begin{minipage}{0.97\linewidth}
\vspace{4pt}
\begin{itemize}[leftmargin=1.2em, itemsep=2pt, topsep=0pt]
\item \textbf{Main point of criticism:} The projection of biodiversity impacts for intensification is methodologically inconsistent with the expansion scenario and contradicts the authors' claim that their models capture landscape-scale impacts. The intensification projection weights the biodiversity change by the crop's relative area fraction, treating the impact as local to the crop's footprint, whereas the models (based on generic ``Cropland'' data and landscape-scale yields) and the expansion scenario imply landscape-wide effects. This likely underestimates the biodiversity cost of intensification.
\item \textbf{Evaluation criteria:} Validity; Conclusions
\end{itemize}
\vspace{2pt}
\end{minipage}}

\vspace{8pt}\par\noindent\textcolor{blue!60!black}{\textbf{Evidence:}}

\vspace{2pt}\par\noindent\colorbox{blue!5}{\begin{minipage}{0.97\linewidth}
\vspace{4pt}
\textcolor{blue!50!black}{\textit{\textbf{Quote 1}} (from main text):}
\textit{``Therefore, our models capture landscape-scale but not local-scale impacts of crop-specific management.''}
\\
\commentbar{The authors explicitly state that their models capture landscape-scale impacts, implying that the yield variable (measured at approximately 10~km resolution) influences biodiversity across the landscape, or at least across all cropland in the landscape, and not just within the specific crop's fields.}

\vspace{2pt}
\end{minipage}}

\end{mdframed}

\begin{mdframed}[style=expertcommentbox]
\footnotesize\textbf{Expert comment.}~The review does not provide any suggestions for how to address this criticism.
\end{mdframed}
\vspace{10pt}

This example captures the other end of the W5 spectrum: rather than being buried in verbose elaboration, the critique is structurally incomplete. The AI correctly identifies a genuine methodological inconsistency (landscape-scale models applied with local-scale area weighting), and the claim is precise enough that a human reviewer could work with it. What is missing is any guidance on what the authors should do about it. Should the intensification projection be rerun without area weighting? Should the paper reframe its impact claims to local scale only? Should a sensitivity analysis bracket the two interpretations? The AI flags the problem and stops, leaving the authors with no path to revision. Human reviewers typically close this loop even when the suggestion is a single sentence.

\subsubsection{W8: Citing evidence that appeared after the preprint}\label{appendix:sw_w8}

\begin{mdframed}[style=categoryboxW, frametitle={\textbf{AI Reviewer Weakness \#8: Citing evidence that appeared after the preprint ($n = 9$)}}]
\footnotesize
The AI reviewer's criticism depends on work that was published after the preprint under review, which means the critique uses information that was not available to the authors at the time of submission. This is an artifact of our evaluation setup, where AI reviewers see already-published papers rather than genuinely concurrent preprints; a deployed AI reviewer would not have access to post-submission evidence.
\end{mdframed}

\begin{mdframed}[
    style=reviewbox,
    frametitle={\textbf{Example \#1 for W8 \textemdash{} Neural-network wavefunctions paper, reviewer: Claude Opus 4.5}},
]
\footnotesize

\noindent\textcolor{blue!60!black}{\textbf{Claim: ``The claimed HEG superiority is contradicted by subsequent analysis''}}

\vspace{2pt}\par\noindent\colorbox{blue!5}{\begin{minipage}{0.97\linewidth}
\vspace{4pt}
\begin{itemize}[leftmargin=1.2em, itemsep=2pt, topsep=0pt]
\item \textbf{Main point of criticism:} The paper claims that the neural-network results for the homogeneous electron gas (HEG) ``outperform many traditional ab initio methods'' and achieve ``error of less than 1\%'' using BF-DMC as reference. However, subsequent independent analysis in the literature demonstrates that the DeepSolid method (referred to as ``LiNet'') does not actually improve upon backflow DMC results across the full density regime for $N=54$ electrons, undermining the claimed superiority of the method for this benchmark system.
\item \textbf{Evaluation criteria:} Validity; Conclusions
\end{itemize}
\vspace{2pt}
\end{minipage}}

\vspace{8pt}\par\noindent\textcolor{blue!60!black}{\textbf{Evidence:}}

\vspace{2pt}\par\noindent\colorbox{blue!5}{\begin{minipage}{0.97\linewidth}
\vspace{4pt}
\textcolor{blue!50!black}{\textit{\textbf{Quote 1}} (from main text):}
\textit{``Overall, our neural network performs very well, with an error of less than 1\% in a wide range of density.''}
\\
\commentbar{This claim presents the correlation error relative to BF-DMC, but the framing suggests the neural network method is achieving higher accuracy than traditional methods. A subsequent study by Pescia et al.\ (2024) in Physical Review B directly compared DeepSolid against BF-DMC for $N=54$ electrons and found that it does not improve upon BF-DMC energies over the whole density regime, which contradicts the paper's implication of superior performance.}

\vspace{2pt}
\end{minipage}}

\end{mdframed}

\begin{mdframed}[style=expertcommentbox]
\footnotesize\textbf{Expert comment.}~While the claims are actually fine, I marked it as not correct because the paper they use to support these claims appeared two years after the presented one. The field was evolving very fast, and in 2022 the presented results could be considered as good. The only thing one could move on with is the claim of improving upon the state of the art; the authors should clarify what exactly they mean by it.
\end{mdframed}
\vspace{10pt}

The AI reviewer's critique relies on a Pescia et al.\ (2024) analysis that did not exist at the time of the 2022 submission. At submission, the authors' claim was defensible against the then-current state of the art, and a reviewer from that period could not have raised this objection. The expert's response separates two things: the narrow issue of whether the ``improves upon state of the art'' wording should be tightened (a small editorial fix that is legitimate), and the larger issue of whether the method's performance is genuinely inferior (which only became clear in later work). W8 cases like this are methodologically informative despite being artifacts of our setup: they show that the AI reviewer tends to integrate all accessible literature without reasoning about the reviewing timeline, which a deployed system could mitigate by constraining its search to content dated before submission.

\begin{mdframed}[
    style=reviewbox,
    frametitle={\textbf{Example \#2 for W8 \textemdash{} Polymer property prediction paper, reviewer: Claude Opus 4.5}},
]
\footnotesize

\noindent\textcolor{blue!60!black}{\textbf{Claim: ``Frozen polyBERT fingerprints deviate from standard fine-tuning practice''}}

\vspace{2pt}\par\noindent\colorbox{blue!5}{\begin{minipage}{0.97\linewidth}
\vspace{4pt}
\begin{itemize}[leftmargin=1.2em, itemsep=2pt, topsep=0pt]
\item \textbf{Main point of criticism:} The paper uses frozen polyBERT fingerprints (without fine-tuning the transformer weights) for property prediction, which deviates from established practices in chemical language modeling where fine-tuning typically improves downstream task performance.
\item \textbf{Evaluation criteria:} Data and methodology; Validity
\end{itemize}
\vspace{2pt}
\end{minipage}}

\vspace{8pt}\par\noindent\textcolor{blue!60!black}{\textbf{Evidence:}}

\vspace{2pt}\par\noindent\colorbox{blue!5}{\begin{minipage}{0.97\linewidth}
\vspace{4pt}
\textcolor{blue!50!black}{\textit{\textbf{Quote 1}} (from main text):}
\textit{``polyBERT fingerprints (dashed arrow) are the averages over the token dimension (sentence average) of the last Transformer encoder.''}
\\
\commentbar{The polyBERT fingerprints are extracted as fixed representations from the pre-trained model, with no indication that the transformer weights are updated during property prediction. This means the 600-dimensional fingerprint vectors are computed once and remain static, while only the downstream multitask neural network is trained. This approach may not fully leverage the power of transfer learning, as the pre-trained representations cannot adapt to capture property-specific patterns that may be critical for accurate prediction.}

\vspace{2pt}
\end{minipage}}

\end{mdframed}

\begin{mdframed}[style=expertcommentbox]
\footnotesize\textbf{Expert comment.}~This needs to be seen in historical context. PolyBERT was very early.
\end{mdframed}
\vspace{10pt}

This example shows the softer version of the W8 pattern. The AI reviewer applies what is now ``established practice'' in chemical language modeling (fine-tune rather than freeze the transformer backbone) to a paper that was one of the earliest transformer-based polymer representation efforts. At the time of submission, pre-trained chemical transformers were themselves a recent development, and using their representations as static fingerprints was a reasonable initial design choice; the fine-tuning literature the AI implicitly invokes matured afterwards. The expert's single-sentence correction is all that is needed: the critique is not wrong by current standards but is applied out of historical context. As with Example~\#1, the underlying failure mode is the AI reviewer treating its entire training corpus as simultaneously available, without reasoning about when the manuscript was written.

\subsubsection{S1: Statistical and methodological rigor}\label{appendix:sw_s1}

\begin{mdframed}[style=categoryboxS, frametitle={\textbf{AI Reviewer Strength \#1: Statistical and methodological rigor ($n = 45$)}}]
\footnotesize
The AI reviewer identifies a genuine statistical or methodological issue that neither the human reviewers nor, in several cases, the authors themselves had addressed. These include violations of independence assumptions, inappropriate significance tests, inadequate validation splits, and missing uncertainty quantification. The expert annotators treat these as legitimate additional scrutiny that a thorough review would have surfaced.
\end{mdframed}

\begin{mdframed}[
    style=reviewbox,
    frametitle={\textbf{Example \#1 for S1 \textemdash{} Protein-protein interaction prediction paper, reviewer: GPT-5.2}},
]
\footnotesize

\noindent\textcolor{blue!60!black}{\textbf{Claim: ``Reported SHS27K gains may not reflect generalization due to evaluation protocol''}}

\vspace{2pt}\par\noindent\colorbox{blue!5}{\begin{minipage}{0.97\linewidth}
\vspace{4pt}
\begin{itemize}[leftmargin=1.2em, itemsep=2pt, topsep=0pt]
\item \textbf{Main point of criticism:} The paper's reported SHS27K protein-protein interaction (PPI) gains are not a valid estimate of generalization because the PPI evaluation lacks a validation split, and the released code uses the test split for model selection and early stopping, which directly undermines the paper's state-of-the-art performance claims.
\item \textbf{Evaluation criteria:} Appropriate use of statistics; Validity
\end{itemize}
\vspace{2pt}
\end{minipage}}

\vspace{8pt}\par\noindent\textcolor{blue!60!black}{\textbf{Evidence:}}

\vspace{2pt}\par\noindent\colorbox{blue!5}{\begin{minipage}{0.97\linewidth}
\vspace{4pt}
\textcolor{blue!50!black}{\textit{\textbf{Quote 1}} (from main text):}
\textit{``MUSE outperformed all existing models including single-scale (DrugVQA and TAG-PPI), and multi-view methods (GNN-PPI and HIGH-PPI). Our model showed substantial improvements over the strongest baseline HIGH-PPI, with an increase of 8.41\% in the BFS split, 13.04\% in the DFS split, and 7.69\% in the Random split.''}
\\
\commentbar{This passage is the paper's central empirical claim for multi-scale interaction prediction on SHS27K, asserting large improvements over the strongest baseline across three splits. Such claims require a strictly held-out test set so that the reported numbers reflect generalization rather than adaptation to the evaluation labels. Because the magnitude of the claimed gains (for example, 13.04\% on DFS) is used to justify the overall superiority of the EM framework, any test-set contamination directly affects the paper's conclusions. In common ML practice, test labels must not influence checkpoint selection, early stopping, threshold tuning, or graph augmentation decisions.}

\vspace{2pt}
\end{minipage}}

\end{mdframed}

\begin{mdframed}[style=expertcommentbox]
\footnotesize\textbf{Expert comment.}~The PPI trainer has no validation split (only train/test) and uses test metrics for model selection (misleadingly named \texttt{best\_valid\_f1}).
\end{mdframed}
\vspace{10pt}

The AI reviewer identifies a subtle but consequential methodological flaw: without a validation split, any hyperparameter choice or early-stopping criterion based on test metrics tunes the model to the test set rather than measuring held-out generalization. The expert's confirmation is especially striking because the flaw is corroborated at the code level: the variable used for model selection is named \texttt{best\_valid\_f1}, suggesting a validation split was intended, but the actual implementation uses test metrics. This is exactly the kind of failure mode a superficial review might miss, since the paper's text describes standard protocols and the issue only surfaces when the code is read. The AI reviewer is not simply applying ML best practice in the abstract; it is reading the evaluation loop carefully enough to notice the mismatch between described and implemented protocol.

\begin{mdframed}[
    style=reviewbox,
    frametitle={\textbf{Example \#2 for S1 \textemdash{} Compound temperature-COVID mortality paper, reviewer: Claude Opus 4.5}},
]
\footnotesize

\noindent\textcolor{blue!60!black}{\textbf{Claim: ``K-S tests on a handful of extreme events may be underpowered''}}

\vspace{2pt}\par\noindent\colorbox{blue!5}{\begin{minipage}{0.97\linewidth}
\vspace{4pt}
\begin{itemize}[leftmargin=1.2em, itemsep=2pt, topsep=0pt]
\item \textbf{Main point of criticism:} The paper uses two-sample Kolmogorov--Smirnov tests to conclude that temperature-related mortality distributions during COVID-19 years differ significantly from non-COVID years. However, the number of extreme temperature events is limited (approximately 10--22 heatwaves and 8--24 cold snaps across the periods), which may affect the statistical power and reliability of these conclusions.
\item \textbf{Evaluation criteria:} Appropriate use of statistics; Validity
\end{itemize}
\vspace{2pt}
\end{minipage}}

\vspace{8pt}\par\noindent\textcolor{blue!60!black}{\textbf{Evidence:}}

\vspace{2pt}\par\noindent\colorbox{blue!5}{\begin{minipage}{0.97\linewidth}
\vspace{4pt}
\textcolor{blue!50!black}{\textit{\textbf{Quote 1}} (from main text):}
\textit{``Two-sample Kolmogorov--Smirnov tests confirm that the COVID-19 event distributions are significantly different from the non-COVID-19 distributions at the 5\% significance level. These results suggest that COVID-19 may have impacted temperature-related mortality during extreme weather events.''}
\\
\commentbar{The conclusion that COVID-19 ``may have impacted'' temperature-related mortality is drawn from K-S tests applied to limited samples. From the code and figure descriptions, the comparison involves approximately 12 heatwaves in 2016--2019 versus 10 heatwaves in 2020--2022, and similar numbers for cold snaps. With such small sample sizes, the K-S test may have limited power to detect true differences or may be sensitive to individual outliers, making the statistical significance potentially unstable.}

\vspace{2pt}
\end{minipage}}

\end{mdframed}

\begin{mdframed}[style=expertcommentbox]
\footnotesize\textbf{Expert comment.}~Even if the overall sample size were adequate, treating correlated regional observations within the same event as independent violates the K-S test assumptions. This is a genuinely useful statistical critique that neither the first reviewer in this set nor the human reviewers identified.
\end{mdframed}
\vspace{10pt}

This example captures a characteristic S1 pattern: the AI reviewer raises a valid statistical concern (small-sample K-S tests are underpowered), and the expert not only agrees but extends the critique with an even sharper point that the AI did not make (independence violations from clustered regional observations). Neither the other AI reviewer on this paper nor the human reviewers raised either issue. The expert's explicit framing is informative: ``a genuinely useful statistical critique that neither the first reviewer in this set nor the human reviewers identified.'' Statistical rigor at this level of detail is a recurring blind spot in human peer review for domain-specific empirical papers, and the AI panel is reliably filling it.

\begin{mdframed}[
    style=reviewbox,
    frametitle={\textbf{Example \#3 for S1 \textemdash{} Meta-optic imaging paper, reviewer: GPT-5.2}},
]
\footnotesize

\noindent\textcolor{blue!60!black}{\textbf{Claim: ``In-the-wild imaging claims rely on single captures without uncertainty reporting''}}

\vspace{2pt}\par\noindent\colorbox{blue!5}{\begin{minipage}{0.97\linewidth}
\vspace{4pt}
\begin{itemize}[leftmargin=1.2em, itemsep=2pt, topsep=0pt]
\item \textbf{Main point of criticism:} The paper's real-world imaging conclusions are not robust because the ``in-the-wild'' comparisons are based on a semi-blind deconvolution that jointly estimates the image, PSF, and fixed-pattern noise from a single capture per scene, with primarily qualitative assessment and limited reporting of constraints and hyperparameters, making it difficult to rule out reconstruction artifacts or non-uniqueness.
\item \textbf{Evaluation criteria:} Conclusions; Appropriate use of statistics and treatment of uncertainties
\end{itemize}
\vspace{2pt}
\end{minipage}}

\vspace{8pt}\par\noindent\textcolor{blue!60!black}{\textbf{Evidence:}}

\vspace{2pt}\par\noindent\colorbox{blue!5}{\begin{minipage}{0.97\linewidth}
\vspace{4pt}
\textcolor{blue!50!black}{\textit{\textbf{Quote 1}} (from main text):}
\textit{``In each case, we captured a single image using a FLIR A65 sensor, and performed a numerical deconvolution (details in the Methods and supplementary information).''}
\\
\commentbar{A single-capture demonstration provides limited evidence for robustness, especially when the post-processing includes solving an inverse problem that can be sensitive to initialization and regularization. In imaging, reporting multiple trials or at least sensitivity to hyperparameters is a common way to separate repeatable improvements from incidental reconstructions. Because the scenes are uncontrolled, sensor noise, temperature gradients, and alignment can vary substantially, and a one-off result can be cherry-picked unintentionally. The key qualitative comparisons therefore depend on the deconvolution pipeline, not only on the optical PSF itself.}

\vspace{2pt}
\end{minipage}}

\end{mdframed}

\begin{mdframed}[style=expertcommentbox]
\footnotesize\textbf{Expert comment.}~I agree with the AI: ``a great deal'' does not mean anything from a scientific point of view. They should have quantified their finding.
\end{mdframed}
\vspace{10pt}

This example broadens S1 beyond strictly statistical critiques to include quantification and uncertainty reporting more generally. The AI reviewer notices that the paper relies on subjective language (``a great deal,'' ``clearly better'') for its real-world imaging claims and correctly pushes back that such comparisons should be quantitative when the underlying pipeline includes an inverse problem. The expert's endorsement is unambiguous: subjective sharpness judgments on single reconstructions are not a scientific standard for demonstrating that a meta-optic outperforms a baseline. Here again, the AI reviewer surfaces a rigor gap that is easy to miss on first read because the paper's overall presentation looks polished, and only a careful reader distinguishes the qualitative claims from what the data actually support.

\subsubsection{S2: Inspecting the submitted source code}\label{appendix:sw_s2}

The three examples below illustrate the code-reading capability of an agentic AI reviewer across three different settings: (i) Example~\#1 (also shown in main text), an internal-consistency check between the manuscript's mathematics and the implemented code; (ii) Example~\#2, a data-leakage discovery where the manuscript text alone could not reveal the issue; and (iii) Example~\#3, the most consequential S2 catch in our dataset, where the implemented code contradicts a central reproducibility claim.

\begin{mdframed}[style=categoryboxS, frametitle={\textbf{AI Reviewer Strength \#2: Inspecting the submitted source code ($n = 28$)}}]
\footnotesize
The AI reviewer opens the submitted source code, reads it, and uses what it finds to support or refute specific claims in the manuscript. In several cases this surfaces concrete implementation bugs, data leakage, or mismatches between described and implemented methodology that the manuscript text alone does not reveal.
\end{mdframed}


\newsavebox{\sIIexIcodeApp}
\begin{lrbox}{\sIIexIcodeApp}
\begin{minipage}{0.85\linewidth}
\begin{lstlisting}[style=reviewcode]
def logdet_matmul(xs, w=None):
    # Combines determinants in log-domain.
    # xs: FermiNet orbitals in each determinant.
    #   Either of length 1 with shape
    #     (ndet, nelectron, nelectron)  # full_det=True
    #   or length 2 with shapes
    #     (ndet, nalpha, nalpha), (ndet, nbeta, nbeta)
    #   (full_det=False, determinants are factorised
    #    into block-diagonals for each spin channel).
\end{lstlisting}
\end{minipage}
\end{lrbox}

\begin{mdframed}[
    style=reviewbox,
    frametitle={\textbf{Example \#1 for S2 \textemdash{} Neural-network wavefunctions paper, reviewer: GPT-5.2}},
]
\footnotesize

\noindent\textcolor{blue!60!black}{\textbf{Claim: ``The wavefunction ansatz is inconsistent between the main text and the algorithm''}}

\vspace{2pt}\par\noindent\colorbox{blue!5}{\begin{minipage}{0.97\linewidth}
\vspace{4pt}
\begin{itemize}[leftmargin=1.2em, itemsep=2pt, topsep=0pt]
\item \textbf{Main point of criticism:} The manuscript's formal description of the solid wavefunction ansatz is internally inconsistent between the main-text expression and the provided algorithm, making it difficult to verify correctness (spin structure and determinant factorization) and undermining reproducibility of the central methodological contribution.
\item \textbf{Evaluation criteria:} Data and methodology; Reproducibility
\end{itemize}
\vspace{2pt}
\end{minipage}}

\vspace{8pt}\par\noindent\textcolor{blue!60!black}{\textbf{Evidence:}}

\vspace{2pt}

\par\noindent\colorbox{blue!5}{\begin{minipage}{0.97\linewidth}
\vspace{4pt}
\textcolor{blue!50!black}{\textit{\textbf{Quote 1}} (from main text):}
\textit{``$\Psi(\mathbf{r}) = \mathrm{Det}_1[e^{i\mathbf{k}\cdot\mathbf{r}_1}u_{mol}^{\uparrow}(d)]\mathrm{Det}_1[e^{i\mathbf{k}\cdot\mathbf{r}_1}u_{mol}^{\downarrow}(d)]$.''}
\\
\commentbar{This equation indicates a product of two determinants associated with spin-up and spin-down channels. However, the notation is ambiguous (both are labeled Det\textsubscript{1}), and it does not clearly specify whether the determinants are block-diagonalized by spin or full determinants over all electrons. In periodic systems with complex-valued orbitals, these details matter for the correctness of antisymmetry constraints and the implementation of phase factors.}

\vspace{2pt}
\end{minipage}}

\vspace{4pt}

\par\noindent\colorbox{blue!5}{\begin{minipage}{0.97\linewidth}
\vspace{4pt}
\textcolor{blue!50!black}{\textit{\textbf{Quote 2}} (from submitted source code):}

\vspace{4pt}
\usebox{\sIIexIcodeApp}
\vspace{4pt}

\commentbar{The released implementation explicitly distinguishes two cases: a single full determinant over all electrons (\texttt{full\_det=True}) or a product of spin-channel determinants when factorized (\texttt{full\_det=False}). This illustrates that the determinant factorization and spin handling are subtle and explicitly parameterized in code. When contrasted with the manuscript's Algorithm~1 line that multiplies two identical determinants, it reinforces that the paper's written description does not clearly convey which mode was used. Because determinant structure can change expressivity, optimization behavior, and numerical stability, readers cannot reproduce or compare results without a clear specification in the main text.}

\vspace{2pt}
\end{minipage}}

\end{mdframed}

\begin{mdframed}[style=expertcommentbox]
\footnotesize\textbf{Expert comment.}~I find it quite impressive that this reviewer actually looks at the code to understand whether some parts of the paper are not properly discussed. I am personally not doing it on a regular basis as it would consume a substantial amount of time.
\end{mdframed}
\vspace{10pt}

This is the clearest expert endorsement of S2 behavior in our dataset: the annotator, a domain specialist, explicitly acknowledges that cross-checking the main-text mathematics against the provided code is something human reviewers typically do not do because it is too time-consuming. The AI reviewer caught a real ambiguity: the main text describes two spin-separated determinants, but the algorithm and code use notation consistent with two identical determinants, leaving readers unable to tell which formulation was actually executed. For a paper whose central contribution is the wavefunction ansatz, this is a genuine reproducibility issue. The expert does not confirm which version is correct; the point is that the reader cannot tell, and the AI reviewer is the only reviewer in the entire panel who checked.


\newsavebox{\sIIexIIcode}
\begin{lrbox}{\sIIexIIcode}
\begin{minipage}{0.85\linewidth}
\begin{lstlisting}[style=reviewcode]
# In trainer_ppi.py
def gnn_inference(self, gnn_model, protein_dataset):
    # Uses test_edges to generate predictions
    edges = self.test_edges[...]
    pseudo_preds = gnn_model(batch)
    return pseudo_edges

def train(self, ...):
    if gnn_model is not None:
        # Adds predicted test edges back to the graph
        pseudo_edges = self.gnn_inference(
            gnn_model, protein_dataset)
        self.graph.edge_index = torch.cat(
            [self.graph.edge_index,
             pseudo_edges.to(self.device)], dim=1)
    # Evaluates on the same test edges using
    # the augmented graph
    valid_loss, test_metrics = self.evaluate(
        self.graph, self.test_edges, self.test_label)
\end{lstlisting}
\end{minipage}
\end{lrbox}

\begin{mdframed}[
    style=reviewbox,
    frametitle={\textbf{Example \#2 for S2 \textemdash{} Protein-protein interaction prediction paper, reviewer: Gemini 3.0 Pro}},
]
\footnotesize

\noindent\textcolor{blue!60!black}{\textbf{Claim: ``The pseudo-interaction augmentation leaks test edges into training''}}

\vspace{2pt}\par\noindent\colorbox{blue!5}{\begin{minipage}{0.97\linewidth}
\vspace{4pt}
\begin{itemize}[leftmargin=1.2em, itemsep=2pt, topsep=0pt]
\item \textbf{Main point of criticism:} The paper's methodology contains a severe data leakage issue where predicted interactions from the test set are added to the molecular network graph, which is then used to predict the existence of those same interactions.
\item \textbf{Evaluation criteria:} Validity; Data and methodology
\end{itemize}
\vspace{2pt}
\end{minipage}}

\vspace{8pt}\par\noindent\textcolor{blue!60!black}{\textbf{Evidence:}}

\vspace{2pt}

\par\noindent\colorbox{blue!5}{\begin{minipage}{0.97\linewidth}
\vspace{4pt}
\textcolor{blue!50!black}{\textit{\textbf{Quote 1}} (from main text):}
\textit{``To alleviate this incompleteness, we adopted the pseudo-likelihood learning in our variational expectation-maximization framework, augmenting the molecular network graph $\mathcal{N}$ with pseudo interactions predicted by the atomic structure scale model GNN\textsubscript{$\phi$}.''}
\\
\commentbar{The paper describes this augmentation as a feature, but fails to mention that the ``pseudo interactions'' are drawn from the test set. In a valid inductive or transductive link prediction setting, the existence of the test edges should not be revealed in the input graph structure used for message passing, as it trivializes the prediction task.}

\vspace{2pt}
\end{minipage}}

\vspace{4pt}

\par\noindent\colorbox{blue!5}{\begin{minipage}{0.97\linewidth}
\vspace{4pt}
\textcolor{blue!50!black}{\textit{\textbf{Quote 2}} (from submitted source code):}

\vspace{4pt}
\usebox{\sIIexIIcode}
\vspace{4pt}

\commentbar{The code explicitly uses \texttt{self.test\_edges} to generate pseudo-labels in \texttt{gnn\_inference}, and the resulting \texttt{pseudo\_edges} are then concatenated to \texttt{self.graph.edge\_index} inside \texttt{train}. The molecular network model is subsequently evaluated on the same \texttt{test\_edges} using this augmented graph, which means the network's input graph structure already contains the edges it is being asked to predict. The leakage that the manuscript's text does not disclose is unambiguous at the code level.}

\vspace{2pt}
\end{minipage}}

\end{mdframed}

\begin{mdframed}[style=expertcommentbox]
\footnotesize\textbf{Expert comment.}~Code verification fully confirms the data leakage claim. The DPI link trainer (\texttt{trainer\_dpi.py}) also has a bug using test metrics for model selection despite having a validation set.
\end{mdframed}
\vspace{10pt}

This example combines S1 and S2: the AI reviewer raises a methodological concern from reading the paper (data leakage in the graph augmentation), and the expert's verification at the code level confirms the concern and extends it with an additional bug the AI did not catch (using test metrics for model selection). What makes it an S2 example rather than strictly an S1 one is that the leakage claim cannot be resolved from the manuscript alone; it is only by tracing where the ``pseudo interactions'' come from in the implementation that the issue becomes visible. Both the AI's catch and the expert's extension require reading the code, and this is a paper where the AI panel's code-reading behavior directly uncovered a flaw that neither the manuscript's text nor a reader's prior expectations would have surfaced.


\newsavebox{\sIIexIIIcode}
\begin{lrbox}{\sIIexIIIcode}
\begin{minipage}{0.85\linewidth}
\begin{lstlisting}[style=reviewcode, language=C]
// In code/sketch_may06a.ino (Arduino)
void setup() {
    Serial.begin(9600);  // baud rate caps throughput
    // ...
}

void loop() {
    // Read accelerometer and transmit
    delay(500);          // forces sampling rate ~2 Hz
}
\end{lstlisting}
\end{minipage}
\end{lrbox}

\begin{mdframed}[
    style=reviewbox,
    frametitle={\textbf{Example \#3 for S2 \textemdash{} Wireless health-monitoring patch paper, reviewer: Gemini 3.0 Pro}},
]
\footnotesize

\noindent\textcolor{blue!60!black}{\textbf{Claim: ``Claimed 800 Hz sampling is contradicted by the implemented code''}}

\vspace{2pt}\par\noindent\colorbox{blue!5}{\begin{minipage}{0.97\linewidth}
\vspace{4pt}
\begin{itemize}[leftmargin=1.2em, itemsep=2pt, topsep=0pt]
\item \textbf{Main point of criticism:} The paper claims a sampling frequency of 800~Hz and analyzes frequencies up to 400~Hz, but the provided code limits the sampling rate to approximately 2~Hz, making the reported results impossible to reproduce with the described setup.
\item \textbf{Evaluation criteria:} Validity; Reproducibility
\end{itemize}
\vspace{2pt}
\end{minipage}}

\vspace{8pt}\par\noindent\textcolor{blue!60!black}{\textbf{Evidence:}}

\vspace{2pt}

\par\noindent\colorbox{blue!5}{\begin{minipage}{0.97\linewidth}
\vspace{4pt}
\textcolor{blue!50!black}{\textit{\textbf{Quote 1}} (from main text):}
\textit{``The high-sensitivity accelerometer ADXL-345 with a sampling frequency of up to 800~Hz in the patch allows successful continuous monitoring, with a wide frequency spectrum from 0 to 400~Hz.''}
\\
\commentbar{The authors claim to capture high-frequency signals up to 400~Hz, which by the Nyquist theorem requires a sampling rate of at least 800~Hz. The paper presents this sampling rate as a feature of the wireless patch and uses it to support the frequency-spectrum plots shown in the results.}

\vspace{2pt}
\end{minipage}}

\vspace{4pt}

\par\noindent\colorbox{blue!5}{\begin{minipage}{0.97\linewidth}
\vspace{4pt}
\textcolor{blue!50!black}{\textit{\textbf{Quote 2}} (from submitted source code):}

\vspace{4pt}
\usebox{\sIIexIIIcode}
\vspace{4pt}

\commentbar{The Arduino sketch includes a 500~ms delay in the main loop, which forces the sampling rate to approximately 2~Hz: three orders of magnitude below the claimed 800~Hz, and insufficient for capturing anything in the reported 0--400~Hz band. In addition, the serial communication is configured at 9600 baud, which by itself limits the throughput to well under 100~Hz even without the delay. Either setting alone contradicts the 800~Hz claim; together, they show that the wireless pipeline as submitted cannot produce the frequency content the paper reports.}

\vspace{2pt}
\end{minipage}}

\end{mdframed}

\begin{mdframed}[style=expertcommentbox]
\footnotesize\textbf{Expert comment.}~A human reviewer would not open up the raw code to find this out. If this code is correct, basically the core data reported in this paper is wrong. The authors in the code said ``It appears that delay is needed in order not to clog the port,'' meaning it is not possible to wirelessly transmit high-frequency information. If this is true, Figure~4 would mean the authors measured in a wired manner and falsely claimed that their system is wireless, which is one key aspect of this paper.
\end{mdframed}
\vspace{10pt}

This is the most consequential S2 catch in our dataset. The AI reviewer noticed a 400$\times$ discrepancy between the paper's claimed sampling rate (800~Hz) and the implemented rate (approximately 2~Hz), from which it correctly inferred that the reported frequency spectrum could not have been produced by the wireless setup described in the manuscript. The expert's follow-up reading of the code goes further still: an inline code comment from the authors themselves explicitly states that a delay was added ``in order not to clog the port,'' which implies that the wireless channel could not carry high-frequency data at all. If this interpretation holds, the paper's wireless claim, not merely a reproducibility statistic, is incorrect at the level of the headline result. This is the category of finding that human peer review systematically misses: the source of the error is not in what the paper says but in what the implementation does, and only a reviewer willing to read the code can see the mismatch.

\subsubsection{S3: Domain-specific technical depth}\label{appendix:sw_s3}

\begin{mdframed}[style=categoryboxS, frametitle={\textbf{AI Reviewer Strength \#3: Domain-specific technical depth ($n = 27$)}}]
\footnotesize
The AI reviewer raises a subfield-specific technical point that is correct, nontrivial, and often pivots on a piece of specialized knowledge the authors could reasonably have expected a domain reviewer to bring. These cases show that the AI reviewer is not limited to generic methodological critique and can engage with the technical content of a paper.
\end{mdframed}

\begin{mdframed}[
    style=reviewbox,
    frametitle={\textbf{Example \#1 for S3 \textemdash{} Multimode fiber image transmission paper, reviewer: GPT-5.2}},
]
\footnotesize

\noindent\textcolor{blue!60!black}{\textbf{Claim: ``The paper's scope claim overstates what the method actually recovers''}}

\vspace{2pt}\par\noindent\colorbox{blue!5}{\begin{minipage}{0.97\linewidth}
\vspace{4pt}
\begin{itemize}[leftmargin=1.2em, itemsep=2pt, topsep=0pt]
\item \textbf{Main point of criticism:} The paper's main claims overstate the scope of what is demonstrated, because the described experimental setup and learning objective target binary amplitude pattern recovery rather than arbitrary optical fields (for example, complex field transmission) or direct high-resolution video transmission, which weakens the robustness of the paper's conclusions.
\item \textbf{Evaluation criteria:} Conclusions; Validity
\end{itemize}
\vspace{2pt}
\end{minipage}}

\vspace{8pt}\par\noindent\textcolor{blue!60!black}{\textbf{Evidence:}}

\vspace{2pt}\par\noindent\colorbox{blue!5}{\begin{minipage}{0.97\linewidth}
\vspace{4pt}
\textcolor{blue!50!black}{\textit{\textbf{Quote 1}} (from abstract):}
\textit{``In this paper, we present a self-supervised dynamic learning approach that achieves long-term, high-fidelity transmission of arbitrary optical fields through unstabilized MMFs.''}
\\
\commentbar{The abstract asserts transmission of ``arbitrary optical fields,'' but the experimental setup and learning objective target binary amplitude patterns rather than the full complex-valued optical field. Amplitude-only recovery is a substantially easier problem than complex-field retrieval through a multimode fiber, and the two are not interchangeable for the applications the paper's framing suggests. The method's performance on binary amplitude should not be extrapolated to arbitrary optical fields without direct evidence.}

\vspace{2pt}
\end{minipage}}

\end{mdframed}

\begin{mdframed}[style=expertcommentbox]
\footnotesize\textbf{Expert comment.}~This is a good catch. Retrieving the complex field (with the imaginary part) is very ill-posed, and to the best of my knowledge there is no method to do that well with multimode fiber propagation. Arbitrary optical fields is for sure not within the scope of the paper, and the method has nothing to do with the complex field. The authors state the relation as amplitude-to-amplitude, so this part is accurate, but it is not enough to extrapolate this huge leap.
\end{mdframed}
\vspace{10pt}

The AI reviewer's critique pivots on a distinction that a specialist in multimode-fiber imaging would be expected to make: ``arbitrary optical field'' is a phrase with a specific technical meaning (full complex-valued field, including phase) that is substantively harder to recover than the amplitude-only setting the paper actually demonstrates. The expert not only confirms this distinction but adds the stronger claim that no existing method, without additional information or measurements, handles complex-field recovery through a multimode fiber well. This means the paper’s framing is not just imprecise but reaches beyond what its setup and objective can support. This is S3 at its sharpest: the AI is applying a subfield-specific reading of a scoping word that a generalist reviewer would pass.

\begin{mdframed}[
    style=reviewbox,
    frametitle={\textbf{Example \#2 for S3 \textemdash{} Late-stage pharmaceutical modification paper, reviewer: GPT-5.2}},
]
\footnotesize

\noindent\textcolor{blue!60!black}{\textbf{Claim: ``Stereocontrol is claimed but diastereomeric ratios are not reported''}}

\vspace{2pt}\par\noindent\colorbox{blue!5}{\begin{minipage}{0.97\linewidth}
\vspace{4pt}
\begin{itemize}[leftmargin=1.2em, itemsep=2pt, topsep=0pt]
\item \textbf{Main point of criticism:} The paper claims ``stereo-controlled'' late-stage modification of pharmaceuticals but does not report stereochemical outcomes (diastereomeric ratios, enantiomeric ratios or excess, or absolute-configuration assignment for newly formed stereocenters) for the late-stage modification examples, making the stereocontrol conclusion unsupported.
\item \textbf{Evaluation criteria:} Validity; Conclusions
\end{itemize}
\vspace{2pt}
\end{minipage}}

\vspace{8pt}\par\noindent\textcolor{blue!60!black}{\textbf{Evidence:}}

\vspace{2pt}\par\noindent\colorbox{blue!5}{\begin{minipage}{0.97\linewidth}
\vspace{4pt}
\textcolor{blue!50!black}{\textit{\textbf{Quote 1}} (from main text):}
\textit{``The stereo-controlled late-stage modification of some complicated pharmaceuticals indicated the versatility of this protocol.''}
\\
\commentbar{This sentence asserts stereocontrol as an achieved feature and uses it as evidence of method versatility. Stereocontrol is a quantitative claim: it requires reporting which stereoisomer(s) form and in what proportions, and ideally how the stereochemistry was assigned. The paper's main text does not provide diastereomeric ratios, enantiomeric excess or ratios, or chiral analytical data for the late-stage modification products. In late-stage settings, where complex substrates often contain stereocenters and reactive intermediates can be planar (for example, cationic), stereochemical outcomes are not predictable without measurement.}

\vspace{2pt}
\end{minipage}}

\end{mdframed}

\begin{mdframed}[style=expertcommentbox]
\footnotesize\textbf{Expert comment.}~Since the substrate used for the synthesis of compound 4v contains a stereogenic center, the issue of diastereoselectivity is important. Presumably, a mixture of diastereomers was formed in compound 4v. It was not mentioned by the human reviewers.
\end{mdframed}
\vspace{10pt}

The AI reviewer raises exactly the objection a trained organic chemist would raise on first reading: the word ``stereo-controlled'' is a technical commitment that requires diastereomeric or enantiomeric ratio data to back it up, and that data is absent for the late-stage modification examples. The expert confirms the point with a specific substrate (compound 4v) and notes that none of the human reviewers flagged this despite its centrality to the paper's versatility claim. This is the pattern that makes S3 a meaningful category distinct from S1: not generic statistical rigor, but the ability to recognise that a particular word in a particular context is a claim that demands particular kinds of evidence which the paper does not provide.

\begin{mdframed}[
    style=reviewbox,
    frametitle={\textbf{Example \#3 for S3 \textemdash{} s-SNOM spectroscopy of TMD monolayers paper, reviewer: Gemini 3.0 Pro}},
]
\footnotesize

\noindent\textcolor{blue!60!black}{\textbf{Claim: ``The Point Dipole Model is insufficient for quantitative analysis of atomically thin films''}}

\vspace{2pt}\par\noindent\colorbox{blue!5}{\begin{minipage}{0.97\linewidth}
\vspace{4pt}
\begin{itemize}[leftmargin=1.2em, itemsep=2pt, topsep=0pt]
\item \textbf{Main point of criticism:} The authors use the Point Dipole Model (PDM) to extract the complex dielectric function of transition-metal dichalcogenide monolayers, which is insufficient for quantitative analysis of s-SNOM data on atomically thin materials.
\item \textbf{Evaluation criteria:} Data and methodology; Validity
\end{itemize}
\vspace{2pt}
\end{minipage}}

\vspace{8pt}\par\noindent\textcolor{blue!60!black}{\textbf{Evidence:}}

\vspace{2pt}\par\noindent\colorbox{blue!5}{\begin{minipage}{0.97\linewidth}
\vspace{4pt}
\textcolor{blue!50!black}{\textit{\textbf{Quote 1}} (from main text):}
\textit{``We choose the point dipole model to interpret the data as it is well documented to capture the response of atomically thin samples laid on thick substrates.''}
\\
\commentbar{The Point Dipole Model treats the tip as a simple dipole, ignoring the extended geometry of the tip and the complex field distribution in the tip-sample gap. While PDM can qualitatively reproduce material contrast, it is widely recognized as insufficient for quantitative extraction of optical constants, especially for layered systems where the near-field interaction is sensitive to the vertical field distribution. More advanced models such as the Finite Dipole Model or the Lightning Rod Model are required for accurate quantitative analysis.}

\vspace{2pt}
\end{minipage}}

\end{mdframed}

\begin{mdframed}[style=expertcommentbox]
\footnotesize\textbf{Expert comment.}~I agree with the AI reviewer that the PDM with their implementation is not the most accurate; they should have used the layered-medium Green's functions and take care of the inhomogeneous background. However, this doesn't mean that the main findings of these researchers were wrong. A more accurate numerical model would generate results with smaller error limits, that's it.
\end{mdframed}
\vspace{10pt}

The AI reviewer identifies a modeling choice that is standard in qualitative s-SNOM work but inadequate for the quantitative analysis the paper attempts, and proposes specific alternative models (Finite Dipole, Lightning Rod) that a specialist would recommend. The expert agrees with the modeling critique but softens the implication: using a more accurate tip-sample model would tighten the error bars on the extracted dielectric function, not invalidate the paper's main findings. S3 examples often have this structure where the AI is correct about the technical point but neutral on severity; the expert then calibrates the severity to what the paper's central claims actually require. The valuable contribution of the AI reviewer in this case is not to escalate the concern but to surface a specific, actionable improvement (switch to layered-medium Green's functions) that a generalist reviewer would not have suggested.

\subsection{Complete categorized expert comments}\label{appendix:sw_raw_comments}

This subsection lists every expert annotator comment that participated in the S/W classification of AI reviews. The full corpus is 442 substantive comments plus 121 paper-level descriptive comments. Comments labeled as being about a human reviewer, or carrying explicit \emph{item-number} references instead of free-form prose, are handled by a separate artifact and are not listed here. Within each category, comments are sorted first by source (item-level before paper-level), then by paper id and reviewer. Each row is citable via its paper id, reviewer, and item number or paper-level slot number, so any comment can be traced back on our HuggingFace dataset.

\subsubsection*{Category summary}
\begin{center}
\begin{tabular}{l l r}
\hline
\textbf{Code} & \textbf{Name} & \textbf{n} \\
\hline
W1 & Missing community / field norms & 54 \\
W2 & Over-harsh / out-of-scope / unrealistic & 46 \\
W3 & Paper explicitly states X, AI says missing & 37 \\
W4 & Redundancy across the 3 AI reviewers & 28 \\
W5 & Vague / verbose / no actionable recommendation & 24 \\
W6 & Trivial / nitpicking & 16 \\
W7 & Technical term-of-art confusion & 13 \\
W8 & Cites evidence from after the preprint & 9 \\
W9 & Over-inflates small code/text inconsistencies & 9 \\
W10 & Criticizes what authors already flagged as a limitation & 6 \\
W11 & AI misreads a figure or caption & 6 \\
W12 & AI misquotes or fabricates a verbatim quote & 5 \\
W13 & AI misses supplementary content & 3 \\
W14 & Ignores authors' own cited prior work & 2 \\
W15 & AI misreads a table & 1 \\
W16 & Cannot analyze figures --- only text & 1 \\
W\_unspecified & Residual: AI judged Not Correct without specific reason & 10 \\
S1 & Statistical / methodology rigor & 45 \\
S2 & Code reading & 28 \\
S3 & Specialized niche field catch & 27 \\
S4 & Internal consistency across sections & 15 \\
S5 & Reproducibility / dependency failures & 10 \\
S6 & Big-picture / counter-narrative synthesis & 7 \\
S\_generic & Residual: AI judged Correct without specific reason & 40 \\
\hline
\textbf{Total} & & \textbf{442} \\
\hline
\end{tabular}
\end{center}

\subsubsection*{W1: Missing community / field norms --- n = 54}

{\footnotesize
\begin{longtable}{@{}p{3.2cm}p{11.5cm}@{}}
\toprule
\textbf{Citation} & \textbf{Expert comment} \\
\midrule
\endfirsthead
\toprule
\textbf{Citation} & \textbf{Expert comment} \\
\midrule
\endhead
\bottomrule
\endlastfoot
P2 $\cdot$ Claude 4.5 $\cdot$ item 4 $\cdot$ primary & I believe, it would make sense if this AI reviewer states what is necessary from the authors to support their claims. Yes, it is an issue that some calculations are not converging but it happens in computations/quantum chemistry. \\[3pt]
P2 $\cdot$ GPT-5.2 $\cdot$ item 2 $\cdot$ primary & While this point is essentially valid, I find it quite arguable. None AL approach provides reasonable error bounds and none ML applied to physical sciences realistically has any. AL literally helps finding new data points to improve the accuracy of models without acquiring excessive amount of them. It is typically tested empirically. Therefore, asking the authors to comment on the issue is fine but making this argument too string is unfair. \\[3pt]
P3 $\cdot$ Claude 4.5 $\cdot$ item 4 $\cdot$ primary & It is well known that 1D corrections are not working well and that is the reason why researchers using better approximations like RPMD and Instanton theory. \\[3pt]
P8 $\cdot$ GPT-5.2 $\cdot$ item 1 $\cdot$ primary & SPlot is a common technique in particle physics and widely used in many publications. \\[3pt]
P9 $\cdot$ GPT-5.2 $\cdot$ item 2 $\cdot$ primary & The comment is correct from statistical point of view, but the likelihood test cannot be done here due to how the parameters of interest are obtained. \\[3pt]
P9 $\cdot$ GPT-5.2 $\cdot$ item 5 $\cdot$ primary & It is true that these things are not documented in the paper, however they are intenally at CERN. It is not a practice to publish these as part of Nature papers so this comment is irrelevant. \\[3pt]
P9 $\cdot$ Gemini 3.0 $\cdot$ item 2 $\cdot$ primary & Full amplitude analysis is subject to a different measurement cannot be done in a CP violation measurement. \\[3pt]
P13 $\cdot$ Gemini 3.0 $\cdot$ item 4 $\cdot$ primary & Although the criticism is valid, this method has been used in many recent papers. Such maps cannot be obtained easily from many individuals, does the average maps from a sample sample are currently methodologically acceptable. \\[3pt]
P15 $\cdot$ Claude 4.5 $\cdot$ item 2 $\cdot$ primary & It's not possible to obtain intracranial data from healthy controls - it won't pass the ethical committee protocol. Unless it's a very specific population of quadriplegic patients who are getting an intracranial implant but that is extremely rare. \\[3pt]
P15 $\cdot$ Claude 4.5 $\cdot$ item 3 $\cdot$ primary & Some of this data might not be feasible to be obtained. \\[3pt]
P16 $\cdot$ GPT-5.2 $\cdot$ item 5 $\cdot$ primary & Sometimes authors release the code after publication, it could be the case for this preprint as well. \\[3pt]
P17 $\cdot$ GPT-5.2 $\cdot$ item 2 $\cdot$ primary & Although the pointing out is quite correct but I feel that it neglects that in reality the observational data is quite limited and a comprehensive data cannot be available for now. \\[3pt]
P18 $\cdot$ Claude 4.5 $\cdot$ item 2 $\cdot$ primary & The point is correct but neglecting the fact that obtaining enough observation data is always very difficult in reality. The reviewer should show some possible alternative observational data to be analyzed if proposing this claim. \\[3pt]
P18 $\cdot$ Gemini 3.0 $\cdot$ item 3 $\cdot$ primary & Neglecting the difficulty of obtaining observational data by satellites. \\[3pt]
P18 $\cdot$ Gemini 3.0 $\cdot$ item 4 $\cdot$ primary & Unfortunately, still common for physics community. \\[3pt]
P19 $\cdot$ GPT-5.2 $\cdot$ item 4 $\cdot$ primary & This comment neglects the fact that obtaining sufficient amount of observational data is impossible for astrophysical community. \\[3pt]
P20 $\cdot$ Claude 4.5 $\cdot$ item 4 $\cdot$ primary & The AI reviewer seems to lack context that it is reasonable to assume that dust attenuation can not be that high so early in the Universe from physical reasons. The AI reviewer could have provided more evidence that this can actually be a concern at these redshifts either via citations or further arguments. Also, it is reasonable to remove templates that are older than the universe, and it's unphysical to assume otherwise. \\[3pt]
P20 $\cdot$ Claude 4.5 $\cdot$ item 5 $\cdot$ primary & It's a valid point, but again the context lacks --- the preprint compares with at that time state-of-the-art measurements and the purpose is to prove the point that these brand-new observations are in striking contrast with the current best estimates with the selection function that was possible pre-JWST. \\[3pt]
P21 $\cdot$ GPT-5.2 $\cdot$ item 1 $\cdot$ primary & Valid point, but the conclusion can be tentatively excepted given the evidence in the preprint and the early/exploratory nature of the work. \\[3pt]
P22 $\cdot$ GPT-5.2 $\cdot$ item 2 $\cdot$ primary & While this point is valid, it can be acceptable to report such speculative caluculations, while clearly stating the caveats of basing this on only three objects --- which would have been accepted by a human reviewer with enough context of the field. \\[3pt]
P28 $\cdot$ Claude 4.5 $\cdot$ item 1 $\cdot$ primary & I did not do the statistical analysis the AI is claiming, but the statistical analysis in the paper is generally accepted in the community doing this type of research. \\[3pt]
P28 $\cdot$ GPT-5.2 $\cdot$ item 1 $\cdot$ primary & The main point of the critique is correct. But to the bacterial modeling community this paper does not read that it claims a universal law. However, the AI reviewer is correct with the statement ''The paper would need clearer limits on scope (e.g., ''within these assembled microcosms and this dispersal/invasion protocol'') or additional tests demonstrating the relationship persists under qualitatively different invasion settings.'' It is harsh, but correct request. \\[3pt]
P35 $\cdot$ GPT-5.2 $\cdot$ item 2 $\cdot$ primary & this is a benchmark limitation shared by all methods evaluated on BioSNAP, not a flaw unique to this paper's methodology. \\[3pt]
P37 $\cdot$ Claude 4.5 $\cdot$ item 4 $\cdot$ secondary & The authors might have not given much thought about clinical significance, but that doesn't seem like a critical problem if we treat this paper as an ML paper than a clinical paper. \\[3pt]
P37 $\cdot$ GPT-5.2 $\cdot$ item 2 $\cdot$ secondary & This AI Reviewer seems to focus on statistical rigor in research (maybe because it's following the Nature's review guideline), but I don't think the ML community these days care so much about statistical rigor any more. \\[3pt]
P41 $\cdot$ Claude 4.5 $\cdot$ item 4 $\cdot$ primary & I think the AI does understand how hard ecology is! R2 of 0.36 is impressive! \\[3pt]
P43 $\cdot$ Claude 4.5 $\cdot$ item 2 $\cdot$ primary & Privacy comes up often in these reviews. I think that it is a technical aspect that is overrepresented in the literature, but less concerning in practice than what the reviewers (AI and humans) make it seem. Especially since we are training a model on skull-stripped data. \\[3pt]
P50 $\cdot$ Claude 4.5 $\cdot$ item 2 $\cdot$ primary & Again, if this was a clinician-targeted paper, it would be important but the novelty of this paper lies in the system/engineering so normal subject is sufficient to demonstrate proof-of-concept. \\[3pt]
P50 $\cdot$ GPT-5.2 $\cdot$ item 1 $\cdot$ primary & These are important points but typically we understand that this paper is `biomedical' field paper and not `medicine' paper, meaning accuracy in clinical term is less emphasized since the novelty of this paper lies in engineering, not a clinical application. \\[3pt]
P52 $\cdot$ Claude 4.5 $\cdot$ item 2 $\cdot$ primary & This paper novelty lies in engineering/system, not a fully-deployed medical study. \\[3pt]
P53 $\cdot$ Gemini 3.0 $\cdot$ item 1 $\cdot$ primary & this comment should account for the rest of the tools available and also all the data available and how difficult is to actually obtain all taxonomy :) so the comment lack prior knowledge that an expert in the filed would know by reading other studies or having used the MSMS tools. \\[3pt]
P54 $\cdot$ Claude 4.5 $\cdot$ item 3 $\cdot$ primary & even though the criticism is on point, a human reviewer would have accounted for how difficult analysis and case studies are and that it is ok to publish as such. however yes, the authors could be careful not to overstate the application. but the potential is big and if tried in other settings than the validity will hold more. \\[3pt]
P54 $\cdot$ Gemini 3.0 $\cdot$ item 5 $\cdot$ primary & it is allowed to do such an estimation, but the programs are accounting for all of these when doing discovery and allignment. AI is probably not trained to account for this knowledge. \\[3pt]
P55 $\cdot$ Gemini 3.0 $\cdot$ item 3 $\cdot$ primary & the point 3 about the sample size, the AI is correct in what the literature suggest, but it all depends on the circumstances of the study and in this study the patient size number and the way they have been chosen is quite valuable enough. \\[3pt]
P59 $\cdot$ Claude 4.5 $\cdot$ item 1 $\cdot$ secondary & This critique applies to virtually all in vitro toxicology studies that inform policy and in my opinion this is more of a theoretical caveat. \\[3pt]
P61 $\cdot$ Gemini 3.0 $\cdot$ item 1 $\cdot$ secondary & While the concern is technically valid, it applies to virtually every study using CMIP6 for subnational health impact assessment coarse-resolution climate model output is a universal constraint. The authors partially mitigate this through the baseline calibration (Equation 5). \\[3pt]
P62 $\cdot$ Claude 4.5 $\cdot$ item 2 $\cdot$ primary & The comment is true but at this proof of concept level, there are other main problems while one can speculate that, in practice, this might not be that critical thanks to prior knowledge on samples. \\[3pt]
P62 $\cdot$ GPT-5.2 $\cdot$ item 4 $\cdot$ primary & I did not personally check the repository and the code, but the missing data could be retrieved from the authors by request. In pinciple, AI reviewer is right about the point that the paper should contain all the details and if something is missing this is a concern. However, in practice, providing everything without missing out a single point is not usually happening and this paper is already above average in terms of description of methods and details for reproducability. \\[3pt]
P62 $\cdot$ Gemini 3.0 $\cdot$ item 2 $\cdot$ primary & The comment is true but at this proof of concept level, there are other main problems while one can speculate that, in practice, this might not be that critical thanks to prior knowledge on samples. \\[3pt]
P62 $\cdot$ Gemini 3.0 $\cdot$ item 3 $\cdot$ primary & I did not personally check the repository and the code, but the missing data could be retrieved from the authors by request. \\[3pt]
P63 $\cdot$ GPT-5.2 $\cdot$ item 4 $\cdot$ primary & It is true that all the details to reproduce the results are not provided. However, the methods in the paper are well-known and there is sufficient depiction to reproduce the results, which will not be exactly the same with the paper since all the details are not provided but it is conceivable that the results would be fairly close. \\[3pt]
P65 $\cdot$ Claude 4.5 $\cdot$ item 2 $\cdot$ primary & Modified Strehl formulation is very common in metamaterial research. \\[3pt]
P65 $\cdot$ Gemini 3.0 $\cdot$ item 3 $\cdot$ primary & I don't think AI is correct. Despite having a few issues, the authors' design is one of the state-of-the-art and comparing its performance against a base design makes sense. \\[3pt]
P67 $\cdot$ Claude 4.5 $\cdot$ item 2 $\cdot$ primary & This a correct statement but here the purpose is a proof of a concept. \\[3pt]
P67 $\cdot$ Gemini 3.0 $\cdot$ item 4 $\cdot$ primary & The method has been used for nano-fabrication since early 2000. This might not be novel enough for AI, but it sure is for humans! \\[3pt]
P76 $\cdot$ Claude 4.5 $\cdot$ item 3 $\cdot$ primary & These xenograft models are typically used to assess efficacy rather than on-target, off-tumor toxicity \\[3pt]
P78 $\cdot$ Claude 4.5 $\cdot$ item 2 $\cdot$ primary & Faradaic efficiency is not highly weighed in electro-organic synthesis. AI reviewer is too harsh on this matter considering the accepted standard of the field. \\[3pt]
P78 $\cdot$ Claude 4.5 $\cdot$ item 3 $\cdot$ primary & This RPC process is assumed in this field. No need to show the machanism of the second oxidation. The AI reviewer is too harsh on this matter. \\[3pt]
P53 $\cdot$ Claude 4.5 $\cdot$ paper-level slot 2 $\cdot$ primary & Yes. the AI reviewers in general do not account to a prior knowledge of the filed so that they can make a fair point when criticizing the tool in terms of accuracy and application. Human reviewer, if chosen as a good expert in the filed, would have applied this knowledge when making the comments. This applies to all AI. \\[3pt]
P53 $\cdot$ GPT-5.2 $\cdot$ paper-level slot 1 $\cdot$ primary & Yes. the AI reviewers in general do not account to a prior knowledge of the filed so that they can make a fair point when criticizing the tool in terms of accuracy and application. Human reviewer, if chosen as a good expert in the filed, would have applied this knowledge when making the comments. This applies to all AI. AI does not pay attention to Figure commenting when it comes to introducing information in the main paper but lacks information from the supplementary files. So more connection to what info can be written in the main part and taken from supplementary is something that AI misses to account for. \\[3pt]
P53 $\cdot$ Gemini 3.0 $\cdot$ paper-level slot 3 $\cdot$ primary & Yes. the AI reviewers in general do not account to a prior knowledge of the filed so that they can make a fair point when criticizing the tool in terms of accuracy and application. Human reviewer, if chosen as a good expert in the filed, would have applied this knowledge when making the comments. This applies to all AI. \\[3pt]
P54 $\cdot$ Claude 4.5 $\cdot$ paper-level slot 3 $\cdot$ primary & AI-3 was too detailed on the statistics, the numbers of analysis which is something that the nanopore sequencing account by itself and it is not something this paper is focused on. Human reviewers that are expert in genomics would have prior knowledge about this and would usually not comment on these very specific statistics. For example the percentage of accuracy for genome alignment etc. usually you do not find a 100 \% match due to the way the softwares work. \\[3pt]
P54 $\cdot$ GPT-5.2 $\cdot$ paper-level slot 1 $\cdot$ primary & In general the AI lacks prior knowledge about what genomics community accepts as a golden standard of matching or gene alignment for some of the trained softwares. Human reviewer would know this and would not be critic about it. The AI also comments on the probable application and lacks the knowledge that even case studies of such degree can be very useful when published. \\[3pt]
P78 $\cdot$ Claude 4.5 $\cdot$ paper-level slot 1 $\cdot$ primary & AI reviewer 1 was too harsh on Faradaic economy, which is a concept that is not that highly weighed yet in synthetic organic electrochemistry. \\[3pt]
\end{longtable}
}

\subsubsection*{W2: Over-harsh / out-of-scope / unrealistic --- n = 46}

{\footnotesize
\begin{longtable}{@{}p{3.2cm}p{11.5cm}@{}}
\toprule
\textbf{Citation} & \textbf{Expert comment} \\
\midrule
\endfirsthead
\toprule
\textbf{Citation} & \textbf{Expert comment} \\
\midrule
\endhead
\bottomrule
\endlastfoot
P1 $\cdot$ Claude 4.5 $\cdot$ item 3 $\cdot$ primary & Overall, I agree with this comment. However, I would not claim that what the authors do is completely wrong. One may simply ask for doing that additional set of experiments. \\[3pt]
P2 $\cdot$ Claude 4.5 $\cdot$ item 2 $\cdot$ primary & However, requesting CBS may be too much. \\[3pt]
P4 $\cdot$ Gemini 3.0 $\cdot$ item 1 $\cdot$ primary & It is an architectural choice and it is valid to have advantages by using it. \\[3pt]
P6 $\cdot$ Claude 4.5 $\cdot$ item 1 $\cdot$ primary & This is the least interesting aspect of the paper. None of the 3 human reviewers even bothered with it, because the paper is using safety as an excuse to not talk about the MR engineering. The AI may be right (I am not a statistics expert), but it is a marginal issue. \\[3pt]
P6 $\cdot$ Claude 4.5 $\cdot$ item 2 $\cdot$ primary & The entire safety portion of the paper is just filler to justify not talking about the MR engineering. This is partially the paper's fault, but the human reviewers were not fooled by the misdirection. \\[3pt]
P6 $\cdot$ Claude 4.5 $\cdot$ item 4 $\cdot$ primary & Again irrelevant. The safety aspects belong in another paper and the duration of the follow-up is very field-specific. While it is true that the Grant et al. study looked at longer time horizons, I doubt that a follow-up of 1 week vs. 2-4 weeks makes any difference. \\[3pt]
P6 $\cdot$ GPT-5.2 $\cdot$ item 1 $\cdot$ secondary & not focus of the paper \\[3pt]
P8 $\cdot$ Claude 4.5 $\cdot$ item 2 $\cdot$ primary & This is not relevant and cannot be covered in this paper. \\[3pt]
P8 $\cdot$ GPT-5.2 $\cdot$ item 4 $\cdot$ primary & It's true that the assumption is made, however these inputs are not subject to a measurement in this paper. \\[3pt]
P28 $\cdot$ Claude 4.5 $\cdot$ item 3 $\cdot$ primary & The statement of AI is correct, however fitting models against experimental data is beyond the scope of the paper. \\[3pt]
P28 $\cdot$ Claude 4.5 $\cdot$ item 4 $\cdot$ primary & The statements are correct, but too harsh and are beyond the scope of the paper. \\[3pt]
P28 $\cdot$ Claude 4.5 $\cdot$ item 5 $\cdot$ primary & The statement is correct, but a threshold must be chosen so the comment, although technically correct, is to harsh for a review. \\[3pt]
P28 $\cdot$ GPT-5.2 $\cdot$ item 2 $\cdot$ primary & This is too harsh critique. The authors do not claim that their result is universal, although the AI would have had a good point if they did. This critique is somewhat along the lines of Human Reviewer 2. \\[3pt]
P28 $\cdot$ GPT-5.2 $\cdot$ item 3 $\cdot$ primary & The AI is finding problems with external packages and software. This is too much, I did not go down the rabbit hole to evaluate it. \\[3pt]
P28 $\cdot$ Gemini 3.0 $\cdot$ item 3 $\cdot$ primary & The AI here is in principle correct. However, this critique is too harsh. From the context it is clear that the scope of the paper is somewhat limited. This critique is similar to Human Reviewer 2. \\[3pt]
P29 $\cdot$ Claude 4.5 $\cdot$ item 1 $\cdot$ primary & This is technically correct, but outside the scope of the paper. \\[3pt]
P29 $\cdot$ Claude 4.5 $\cdot$ item 4 $\cdot$ primary & Correct, but out of the scope of the paper. \\[3pt]
P30 $\cdot$ Claude 4.5 $\cdot$ item 2 $\cdot$ primary & Out of the scope of the paper. \\[3pt]
P30 $\cdot$ Gemini 3.0 $\cdot$ item 5 $\cdot$ primary & Out of the scope of the paper. \\[3pt]
P31 $\cdot$ Claude 4.5 $\cdot$ item 4 $\cdot$ primary & Too harsh critique. \\[3pt]
P37 $\cdot$ Claude 4.5 $\cdot$ item 2 $\cdot$ secondary & The first two quote-comment pairs are logically valid, but I don't think they are enough to reject the paper. However, the third pair is more critical, as one of the core claims (i.e. the importance of date embedding) is negated by the empirical result. \\[3pt]
P42 $\cdot$ Claude 4.5 $\cdot$ item 3 $\cdot$ primary & Technically the AI is correct, but I think much like the other paper I reviewed, the AI does not understand the complexity of studying ecosystems and so has unrealistic expectations of the resultant classification performance. The signal in the models is strong even with the moderate classification performance. \\[3pt]
P42 $\cdot$ GPT-5.2 $\cdot$ item 3 $\cdot$ primary & While technically correct again, I think the AI does not understand the complexity and challenges of studying ecosystems; however, there is some causal language in the paper that could be revised. \\[3pt]
P43 $\cdot$ Claude 4.5 $\cdot$ item 1 $\cdot$ primary & The interesting part about the paper is more about the model itself rather than the federation. \\[3pt]
P46 $\cdot$ Claude 4.5 $\cdot$ item 2 $\cdot$ primary & Yes, but what can you do?! \\[3pt]
P46 $\cdot$ Claude 4.5 $\cdot$ item 4 $\cdot$ primary & Yes, but it is a valid and good comparison! \\[3pt]
P46 $\cdot$ GPT-5.2 $\cdot$ item 3 $\cdot$ primary & It is correct that this is the case, but PG cannot be effectively implemented on the GPU, whereas polyBERT can. A comment in the text may be good, but to me this is not a big deal. \\[3pt]
P49 $\cdot$ Claude 4.5 $\cdot$ item 2 $\cdot$ secondary & This critic is factually accurate but it highlights a limitation that the authors cannot realistically defend against or resolve by modifying their data, making it a non-constructive critique. \\[3pt]
P52 $\cdot$ Claude 4.5 $\cdot$ item 4 $\cdot$ primary & Given this is academic paper, it's difficult to ask authors to provide full biocompatibility data which would take too long. \\[3pt]
P54 $\cdot$ Claude 4.5 $\cdot$ item 1 $\cdot$ primary & Bit too rigid criticizm however, it does point out when authors overstate their point trying to sell their paper. \\[3pt]
P59 $\cdot$ Claude 4.5 $\cdot$ item 4 $\cdot$ secondary & The dose response non-linearity for total PM 2.5 mass has been characterized, but there is no established dose-response function for toxic potency-adjusted PM2.5. This is a future research direction, not something the authors could realistically incorporate. Also citation 5, the link is wrong. \\[3pt]
P59 $\cdot$ GPT-5.2 $\cdot$ item 4 $\cdot$ secondary & Applies an unrealistically stringent reproducibility standard. This standard of reproducibility, while ideal, is rarely met in chemical transport modeling studies published in top journals. \\[3pt]
P59 $\cdot$ Gemini 3.0 $\cdot$ item 3 $\cdot$ secondary & reviewer seems to recognize that this is not a methodological flaw - ''it does not necessarily invalidate the relative ranking'' but still frames it as a significant concern. It can be a valuable addition to the discussions but its not affecting the conclusions. \\[3pt]
P62 $\cdot$ Claude 4.5 $\cdot$ item 3 $\cdot$ primary & The comment is true but one can easily make an counter-argument of these not being a huge problem in controlled experiments. \\[3pt]
P65 $\cdot$ Claude 4.5 $\cdot$ item 1 $\cdot$ primary & AI got this wrong. Comparing simulation and experimental results is not fair. \\[3pt]
P65 $\cdot$ Gemini 3.0 $\cdot$ item 5 $\cdot$ primary & It looks like this was a badly trained AI platform. What could the authors do if their imaging takes place at certain wavelength values other than averaging the discrete outcome? This was my first study on Human vs AI reviewers. It is interesting to see that one of three AI platforms can come up with a totally different review than the other two. Clearly its training was done with different sources than the other two. LTSM was an issue 5-10 years ago. The AI reviewer III shows me that it might be still an issue for some AI platforms. There is a more significant alignment between AI reviewers I and II. THey saw some important points which were not caught by human reviewers. \\[3pt]
P66 $\cdot$ GPT-5.2 $\cdot$ item 1 $\cdot$ primary & This is such a harsh comment for an experimental study. The authors provided all the datasets. The measurements were taken several times and averages were reported. The real world measurements are not perfect or noise free. What else could authors do? \\[3pt]
P66 $\cdot$ GPT-5.2 $\cdot$ item 2 $\cdot$ primary & I again do not agree with the AI reviewer. The wavelength is close to 800 nm and the tip is 20-25 nm wide, which is 1/30th - 1/40th of the wavelength. We can treat it as a point dipole. The AI reviewer's comment is too harsh! \\[3pt]
P67 $\cdot$ GPT-5.2 $\cdot$ item 4 $\cdot$ primary & I understand AI's request, which is fair. But this shouldn't undermine the authors' contribution \\[3pt]
P68 $\cdot$ Gemini 3.0 $\cdot$ item 1 $\cdot$ primary & This is technically correct, but morphological traits are very correlated, which makes the underlying dimensionality much lower than the number of coordinates. So, I am two ways with regards to this comment. On one way, the criticism is correct. On the other, it is likely overstated. \\[3pt]
P72 $\cdot$ Claude 4.5 $\cdot$ item 5 $\cdot$ primary & DeBreak is more computationally efficient than one of the three standard callers it is compared to (pbsv), so I think this criticism is overblown. \\[3pt]
P72 $\cdot$ Gemini 3.0 $\cdot$ item 4 $\cdot$ primary & One could always compare to more tools, but I probably would not have asked the authors to include an additional tool unless there was a specific reason for including it. e.g. one of the human reviewers' requested a comparison to PBHoney and had a good reason for doing so. \\[3pt]
P77 $\cdot$ Claude 4.5 $\cdot$ item 5 $\cdot$ secondary & A definitive answer to this question may not be readily available. Nevertheless, it would be valuable if the authors could provide insight into the different reactivities of various metals. \\[3pt]
P6 $\cdot$ Claude 4.5 $\cdot$ paper-level slot 1 $\cdot$ primary & AI Reviewer 1 focused too much on the safety aspects of the paper. 70\% of the review was about safety, whereas the paper is primarily about MRI methodology. I am sure that some of the statistics observations were original, but they were irrelevant in my opinion. \\[3pt]
P8 $\cdot$ Claude 4.5 $\cdot$ paper-level slot 1 $\cdot$ primary & Human review 2 is not available in the checklist. I noticed that there was one question in the AI reviewer that was suggesting to provide a measurements of some parameters which is out of the scope of the analysis done there and also some trivial questions on splot that are not applicable. That would never be asked by a human reviewer. In the AI reviewer 2 there were some trivial questions that should not be asked in Nature paper \\[3pt]
P28 $\cdot$ GPT-5.2 $\cdot$ paper-level slot 1 $\cdot$ primary & Yes, the AI reviewer 1 was talking about the equations and math, but those comments are relevant for external software, not authored in this paper. \\[3pt]
\end{longtable}
}

\subsubsection*{W3: Paper explicitly states X, AI says missing --- n = 37}

{\footnotesize
\begin{longtable}{@{}p{3.2cm}p{11.5cm}@{}}
\toprule
\textbf{Citation} & \textbf{Expert comment} \\
\midrule
\endfirsthead
\toprule
\textbf{Citation} & \textbf{Expert comment} \\
\midrule
\endhead
\bottomrule
\endlastfoot
P1 $\cdot$ Gemini 3.0 $\cdot$ item 1 $\cdot$ primary & The authors write explicitly that they employ finite-size correction. \\[3pt]
P3 $\cdot$ Claude 4.5 $\cdot$ item 1 $\cdot$ primary & I guess one can criticize it but authors have already provided a reasonable evaluation. \\[3pt]
P3 $\cdot$ Claude 4.5 $\cdot$ item 3 $\cdot$ primary & They give values they got in their experiments which is appropriate for this case. \\[3pt]
P4 $\cdot$ GPT-5.2 $\cdot$ item 3 $\cdot$ primary & The authors provide the code and it does not matter whether it is on GitHub or GitLab. \\[3pt]
P4 $\cdot$ Gemini 3.0 $\cdot$ item 5 $\cdot$ primary & The code is available. \\[3pt]
P5 $\cdot$ Claude 4.5 $\cdot$ item 4 $\cdot$ primary & Formally, this criticism is correct, but it can be omitted because the authors provide molecular dynamics simulations for other benchmark systems. \\[3pt]
P6 $\cdot$ Gemini 3.0 $\cdot$ item 1 $\cdot$ primary & The AI made factual errors. The sequence parameters were reported, and the methods section starts with this. It is true that the MRI methodology does not have a subheading, which is why the AI got confused. \\[3pt]
P8 $\cdot$ Claude 4.5 $\cdot$ item 1 $\cdot$ primary & The comment make sense however the authors consider more complicated model as a systematic study. \\[3pt]
P9 $\cdot$ GPT-5.2 $\cdot$ item 4 $\cdot$ primary & The comment is too vague of what is insufficient in the paper since there is some discussion on the resonant part. \\[3pt]
P12 $\cdot$ Gemini 3.0 $\cdot$ item 3 $\cdot$ secondary & The concern about J1 being a non-detection is valid, but the Lorentz factor is derived including J2. \\[3pt]
P17 $\cdot$ Gemini 3.0 $\cdot$ item 1 $\cdot$ primary & The pointing out on how to define classification during non-equilibrium process is appropriate but it neglects that the paper considers a response to linear perturbation and a near equilibrium initial profiles of observation and PIC simulations, leading to inadequately strong criticism to the authors' claim. The wording seems common on ML community unfriendly review comment but less common in physics community review. \\[3pt]
P18 $\cdot$ GPT-5.2 $\cdot$ item 2 $\cdot$ primary & AI confused by a confusing authors' explanation, mixing 1D analyses with 2D reconnection case without explicitly clarifying. Just suggesting clarifying the explanation on that part is more appropriate as a review comment. \\[3pt]
P25 $\cdot$ Gemini 3.0 $\cdot$ item 3 $\cdot$ primary & I was doubting between correct and not correct. What the criticism reffers to are instantaneous correlations, due to the wave there are lagged correlations. \\[3pt]
P28 $\cdot$ Gemini 3.0 $\cdot$ item 1 $\cdot$ primary & The paper claims usage of a Matlab method, and does not explicitly state the equations mentioned in the review here. This is clearly a fail of the AI reviewer. \\[3pt]
P30 $\cdot$ Claude 4.5 $\cdot$ item 1 $\cdot$ primary & It is clear from the context that only QS systems are included. \\[3pt]
P30 $\cdot$ GPT-5.2 $\cdot$ item 1 $\cdot$ primary & It is clear from the context that QS systems only are considered. \\[3pt]
P31 $\cdot$ Gemini 3.0 $\cdot$ item 1 $\cdot$ primary & Correct, but explained in the paper that this is inevitable. \\[3pt]
P34 $\cdot$ Gemini 3.0 $\cdot$ item 1 $\cdot$ primary & The test should be done \textasciitilde{}3000 times consistently with the paper, single run shouldn't be interpreted like this. \\[3pt]
P37 $\cdot$ GPT-5.2 $\cdot$ item 4 $\cdot$ primary & the model architecture for finetuning is provided (ICDBartForSequenceBinaryClassification in icdmodelbart.py) but the training script is not. \\[3pt]
P43 $\cdot$ Claude 4.5 $\cdot$ item 3 $\cdot$ primary & This comment disregards the existence of out-of-sample sites \\[3pt]
P44 $\cdot$ Claude 4.5 $\cdot$ item 2 $\cdot$ primary & It's failing to take into account the reduced set, which naturally leads to higher p-values. \\[3pt]
P50 $\cdot$ Gemini 3.0 $\cdot$ item 3 $\cdot$ primary & There is actually `methods' text in the manuscript. It probably wasn't properly inputed into AI. \\[3pt]
P55 $\cdot$ Claude 4.5 $\cdot$ item 2 $\cdot$ primary & the age is defined as above 18 so there is a need for more info to make such a comment \\[3pt]
P59 $\cdot$ Gemini 3.0 $\cdot$ item 1 $\cdot$ secondary & This critic assumed ''zero-out'' meant removing all emissions (PM, SO2) from each sector but in the paper authors clearly mentioned that they zero out only primary PM 2.5 and they akcnowldege this in the limitations section as well (i.e ignoring secondary aerosol contribution). \\[3pt]
P61 $\cdot$ Claude 4.5 $\cdot$ item 5 $\cdot$ secondary & this is factually incorrect, the paper explicitly addresses this issue. Lines 489--496 describe a calibration procedure (Equation 5) that adjusts modeled PM2.5 concentrations against observed baseline period data (2012--2017). \\[3pt]
P61 $\cdot$ Gemini 3.0 $\cdot$ item 4 $\cdot$ secondary & The paper's title, abstract, and methods explicitly scope the study to PM2.5. The reviewer's argument that the paper's broader framing around ''air pollution-related deaths'' and SDGs implicitly requires ozone consideration is a stretch no single study is obligated to cover all pollutants, and the PM2.5 focus is clearly stated throughout. \\[3pt]
P61 $\cdot$ Gemini 3.0 $\cdot$ item 5 $\cdot$ secondary & The paper actually uses the GBD2019 MR-BRT model (line 470), not the older IER function, and explicitly discusses the GEMM comparison (lines 547--550), acknowledging that GEMM yields higher estimates. \\[3pt]
P62 $\cdot$ GPT-5.2 $\cdot$ item 2 $\cdot$ secondary & Marked it maginal, because important metrics are provided in their experimental context. \\[3pt]
P66 $\cdot$ Claude 4.5 $\cdot$ item 2 $\cdot$ primary & The authors mention repetition of measurements and using averaged results in the end. I don't think this is a fair comment. \\[3pt]
P66 $\cdot$ Claude 4.5 $\cdot$ item 3 $\cdot$ primary & The authors mention that their resolution is limited by the width of their tip and their experimental results in a good agreement with theoretical results (10 nm). There is no issue here! \\[3pt]
P67 $\cdot$ GPT-5.2 $\cdot$ item 3 $\cdot$ primary & The zip file I downloaded has all the *.mat files \\[3pt]
P67 $\cdot$ GPT-5.2 $\cdot$ item 3 $\cdot$ secondary & In the code bundle that was provided to me, the .mat files were empty. I believe this is what the AI reviewer is referring to. However, I could find the actual .mat files on GitHub. \\[3pt]
P67 $\cdot$ Gemini 3.0 $\cdot$ item 3 $\cdot$ primary & They wrote ''to simulate the process.'' \\[3pt]
P67 $\cdot$ Gemini 3.0 $\cdot$ item 5 $\cdot$ primary & They already showed field enhancement far from the contacts. \\[3pt]
P74 $\cdot$ Claude 4.5 $\cdot$ item 3 $\cdot$ primary & The point is valid but overblown given that the authors also assess accuracy on non-human genomes. \\[3pt]
P74 $\cdot$ GPT-5.2 $\cdot$ item 2 $\cdot$ primary & Another valid point that is somewhat offset by the benchmarking on other organisms. \\[3pt]
P77 $\cdot$ Claude 4.5 $\cdot$ item 2 $\cdot$ primary & The substrates described in this paper do not have this issue. \\[3pt]
\end{longtable}
}

\subsubsection*{W4: Redundancy across the 3 AI reviewers --- n = 28}

{\footnotesize
\begin{longtable}{@{}p{3.2cm}p{11.5cm}@{}}
\toprule
\textbf{Citation} & \textbf{Expert comment} \\
\midrule
\endfirsthead
\toprule
\textbf{Citation} & \textbf{Expert comment} \\
\midrule
\endhead
\bottomrule
\endlastfoot
P5 $\cdot$ Claude 4.5 $\cdot$ item 2 $\cdot$ primary & To a large extent, all three AI reviewers raise very similar points. Thus, the overall review would not cover many substantial points human reviewer raised. \\[3pt]
P42 $\cdot$ GPT-5.2 $\cdot$ item 1 $\cdot$ primary & Same comment as AI 2 item 2. \\[3pt]
P46 $\cdot$ Gemini 3.0 $\cdot$ item 1 $\cdot$ primary & Same as point 1 of other reviewer. \\[3pt]
P59 $\cdot$ GPT-5.2 $\cdot$ item 2 $\cdot$ secondary & This item largely duplicates AI Review \#1 Item 1. Despite being more elaborately argued, it doesn't add a fundamentally new dimension. The authors should temper their language from ''health-oriented'' to something like ''toxicity-informed'' and explicitly acknowledge that TPAE is an exploratory index rather than a calibrated health impact function is easy to implement. \\[3pt]
P59 $\cdot$ GPT-5.2 $\cdot$ item 3 $\cdot$ secondary & Essentially the same critique as AI Review \#2 Item 1. It actually creates a conceptual tension: secondary sulfate/nitrate are widely considered less toxic, so including them would increase each sector's attributable concentration while simultaneously lowering the average toxicity per unit mass. The reviewer asserts this ''can materially change'' rankings but provides no quantitative basis for that claim. \\[3pt]
P60 $\cdot$ Claude 4.5 $\cdot$ item 1 $\cdot$ secondary & This is a well-constructed critique but largely redundant with the AI reviewer1's Items 2 and 3. The main added value is the specific Ali et al. (2025) reference demonstrating synergistic effects with quantified RRs, which provides stronger empirical backing than the first reviewer offered for the same point. \\[3pt]
P60 $\cdot$ Claude 4.5 $\cdot$ item 2 $\cdot$ secondary & This is essentially the same critique as the AI reviewer 1 Item 2, which already identified that removing COVID-19 deaths doesn't make the remaining mortality independent of pandemic context, and that the model trained on primarily pre-pandemic data fails to capture altered vulnerability. \\[3pt]
P60 $\cdot$ GPT-5.2 $\cdot$ item 1 $\cdot$ secondary & AI Reviewer 1 (Items 2 and 3) identified both the violated independence assumption and the misleading ''compound'' terminology. AI Reviewer 2 (Item 1) restated it with additional references. This reviewer articulates it more formally mentioning joint models, attributable fraction partitioning, and counterfactual decomposition but the core insight is identical \\[3pt]
P60 $\cdot$ GPT-5.2 $\cdot$ item 2 $\cdot$ secondary & This is a thorough and well-referenced elaboration of a point other reviwers already made more concisely. The added value is the explicit connection to the Vicedo-Cabrera projection framework, which helps explain why the authors may have made this choice (importing a projection-style baseline into a retrospective analysis) and why it's inappropriate here. \\[3pt]
P61 $\cdot$ Gemini 3.0 $\cdot$ item 2 $\cdot$ secondary & This item is essentially a less developed version of what AI Review \#2 Item 2 covers more comprehensively. \\[3pt]
P62 $\cdot$ Claude 4.5 $\cdot$ item 3 $\cdot$ secondary & Aligned with Item 7 of Human Reviewer 3. \\[3pt]
P62 $\cdot$ GPT-5.2 $\cdot$ item 1 $\cdot$ secondary & Aligned with Item 7 of Human Reviewer 3, Item 3 of AI Reviewer 1, and Item 1 of AI Reviewer 2. But not addressing other important challenges in in vivo settings like Human Reviewer 3. \\[3pt]
P62 $\cdot$ GPT-5.2 $\cdot$ item 3 $\cdot$ primary & Again the same comment about NMF rank that is common in all AI reviews. \\[3pt]
P62 $\cdot$ GPT-5.2 $\cdot$ item 3 $\cdot$ secondary & Aligned with Item 2 of AI Reviewer 1 and Item 2 of AI Reviewer 2. \\[3pt]
P62 $\cdot$ GPT-5.2 $\cdot$ item 4 $\cdot$ secondary & Aligned with Item 3 of AI reviewer 2. Marked marginal by the same reason. \\[3pt]
P62 $\cdot$ Gemini 3.0 $\cdot$ item 1 $\cdot$ secondary & Aligned with Item 7 of Human Reviewer 3, and Item 3 of AI Reviewer 1. But not addressing other important challenges in in vivo settings like Human Reviewer 3. \\[3pt]
P62 $\cdot$ Gemini 3.0 $\cdot$ item 2 $\cdot$ secondary & Aligned with Item 2 of AI Reviewer 1. \\[3pt]
P72 $\cdot$ GPT-5.2 $\cdot$ item 3 $\cdot$ primary & A fair point, echoing a comment from one of the human reviewers. \\[3pt]
P72 $\cdot$ GPT-5.2 $\cdot$ item 4 $\cdot$ primary & Echoes a point by one of the human reviewers that the authors could emphasize that their benchmarking results do not encompass all SV types handled by DeBreak. \\[3pt]
P4 $\cdot$ Claude 4.5 $\cdot$ paper-level slot 3 $\cdot$ primary & I found the point on the comparison with prior work important. Other points are rather similar to other AI reviewers. \\[3pt]
P22 $\cdot$ Claude 4.5 $\cdot$ paper-level slot 3 $\cdot$ primary & In general, they pick up on the same major points \\[3pt]
P22 $\cdot$ GPT-5.2 $\cdot$ paper-level slot 1 $\cdot$ primary & In general, they pick up on the same major points \\[3pt]
P22 $\cdot$ Gemini 3.0 $\cdot$ paper-level slot 2 $\cdot$ primary & In general, they pick up on the same major points \\[3pt]
P62 $\cdot$ Claude 4.5 $\cdot$ paper-level slot 1 $\cdot$ primary & None. In this paper's specific case, all human reviewers were exceptional and their combined feedback covered every possible aspect. All AI reviewers to some extent failed to give feedback that could possibly improve the manuscript while keeping the intricacies and difficulties of optics+bio experimental work landscape. \\[3pt]
P62 $\cdot$ GPT-5.2 $\cdot$ paper-level slot 3 $\cdot$ primary & None. In this paper's specific case, all human reviewers were exceptional and their combined feedback covered every possible aspect. All AI reviewers to some extent failed to give feedback that could possibly improve the manuscript while keeping the intricacies and difficulties of optics+bio experimental work landscape. \\[3pt]
P62 $\cdot$ Gemini 3.0 $\cdot$ paper-level slot 2 $\cdot$ primary & None. In this paper's specific case, all human reviewers were exceptional and their combined feedback covered every possible aspect. All AI reviewers to some extent failed to give feedback that could possibly improve the manuscript while keeping the intricacies and difficulties of optics+bio experimental work landscape. \\[3pt]
P67 $\cdot$ Claude 4.5 $\cdot$ paper-level slot 3 $\cdot$ secondary & The review of this AI is similar to the review of AI reviewer 1 but I find that the AI reviewer 1 commented on matters that are more important first. \\[3pt]
P72 $\cdot$ GPT-5.2 $\cdot$ paper-level slot 3 $\cdot$ primary & No. This AI Reviewer was the only one to complain about the assembly-based callset being treated as ground-truth, but at least one other reviewer was clearly aware that this was not the authors' true intention---they were just using it as a decently accurate set that might be worth comparing alignment-based methods to. \\[3pt]
\end{longtable}
}

\subsubsection*{W5: Vague / verbose / no actionable recommendation --- n = 24}

{\footnotesize
\begin{longtable}{@{}p{3.2cm}p{11.5cm}@{}}
\toprule
\textbf{Citation} & \textbf{Expert comment} \\
\midrule
\endfirsthead
\toprule
\textbf{Citation} & \textbf{Expert comment} \\
\midrule
\endhead
\bottomrule
\endlastfoot
P1 $\cdot$ Claude 4.5 $\cdot$ item 2 $\cdot$ primary & The valid point is that authors refer to ref. 21 as providing the experimental value. However, ref. 21 takes the corresponding reference value from a different work. This could be requested to correct. But, in general, the comment/critique by this reviewer is rather pointless. \\[3pt]
P5 $\cdot$ GPT-5.2 $\cdot$ item 1 $\cdot$ primary & While I agree with the points raised by this AI reviewer, I found them somewhat too lengthy and repetitive. \\[3pt]
P5 $\cdot$ GPT-5.2 $\cdot$ item 4 $\cdot$ primary & Similar to the previous item, this AI reviewer tends to overemphasize each quotation and comment. It would be sufficient to state that repeating the simulation with different starting points and random seeds is necessary to estimate the variance and thus draw more reliable conclusions. This AI reviewer also appears to misuse the term of ''uncertainty''. \\[3pt]
P6 $\cdot$ GPT-5.2 $\cdot$ item 1 $\cdot$ primary & Too verbose. I am not a statistics expert, but this is a paper about MR methodology. The first comment should not be six paragraphs about biological effects. \\[3pt]
P6 $\cdot$ GPT-5.2 $\cdot$ item 2 $\cdot$ primary & Too verbose. The previous two AI reviewers did a much better job pinpointing the main issues. \\[3pt]
P17 $\cdot$ Gemini 3.0 $\cdot$ item 2 $\cdot$ primary & The pointing out is basically correct but needs more comments. Maybe a more appropriate review comment on this point is to suggest to take more effort on validating why the authors used MMS observation data with reconnection to compare 1D non-reconnection current sheet PIC simulations. Basically, it criticized too much. \\[3pt]
P22 $\cdot$ GPT-5.2 $\cdot$ item 1 $\cdot$ primary & Thorough and well articulated, but perhaps unecessarily long. \\[3pt]
P22 $\cdot$ GPT-5.2 $\cdot$ item 3 $\cdot$ primary & Too long. \\[3pt]
P29 $\cdot$ Claude 4.5 $\cdot$ item 2 $\cdot$ primary & This critique, technically correct, is confusing the significance of some of the results. \\[3pt]
P29 $\cdot$ GPT-5.2 $\cdot$ item 2 $\cdot$ primary & There are some correct elements here, but it is stated in a super confusing way. \\[3pt]
P29 $\cdot$ GPT-5.2 $\cdot$ item 3 $\cdot$ primary & This has an element of truth, but evidence is not provided well, and the statement is convoluted. \\[3pt]
P29 $\cdot$ Gemini 3.0 $\cdot$ item 3 $\cdot$ primary & This might be correct but it is stated in a convoluted way. \\[3pt]
P36 $\cdot$ Gemini 3.0 $\cdot$ item 3 $\cdot$ primary & The evidence consists only of paper quotes restated with the reviewer's interpretation, without external citations or concrete reasoning to support the reverse causality claim \\[3pt]
P41 $\cdot$ Claude 4.5 $\cdot$ item 1 $\cdot$ primary & As with other reviews, it is unclear what the recommendation for addressing this criticism is. \\[3pt]
P41 $\cdot$ Claude 4.5 $\cdot$ item 3 $\cdot$ primary & Similar to AI review \#1, this comment is true and yet the signal in the results is still strong, it's unclear from the comment whether this is a fundamental criticism or something that requires addressing in the discussion. \\[3pt]
P41 $\cdot$ GPT-5.2 $\cdot$ item 1 $\cdot$ primary & As with other reviews, it is unclear what the recommendation for addressing this criticism is. \\[3pt]
P41 $\cdot$ Gemini 3.0 $\cdot$ item 1 $\cdot$ primary & The review does not provide any suggestions for how to address this criticism. \\[3pt]
P41 $\cdot$ Gemini 3.0 $\cdot$ item 2 $\cdot$ primary & It is unclear from the criticism how this issue results in bias per se, which is a much greater concern than ``noise'', given the strong signal in the results of the paper. \\[3pt]
P49 $\cdot$ GPT-5.2 $\cdot$ item 5 $\cdot$ secondary & While this AI reviewer (and others) is excellent at pointing out flaws, it fails to provide constructive alternatives or actionable suggestions on how the authors should address these criticisms to improve the paper. \\[3pt]
P60 $\cdot$ GPT-5.2 $\cdot$ item 4 $\cdot$ secondary & The point is valid but the level of detail suggests the reviewer may be optimizing for thoroughness over insight. \\[3pt]
P77 $\cdot$ GPT-5.2 $\cdot$ item 4 $\cdot$ secondary & I could not understand the reviewer's main point. \\[3pt]
P78 $\cdot$ Gemini 3.0 $\cdot$ item 2 $\cdot$ secondary & Although the point mentioned by the reviewer is valid, more evidence (references) should be given. \\[3pt]
P5 $\cdot$ GPT-5.2 $\cdot$ paper-level slot 2 $\cdot$ primary & I found the items themselves to be reasonable, but the underlying argumentation was diffuse and overcrowded with statements that were largely irrelevant. Overall, as noted in the checklist, all reviewers raised very similar points. However, the first reviewer provided the strongest arguments, whereas this reviewer's arguments were the weakest. \\[3pt]
P6 $\cdot$ GPT-5.2 $\cdot$ paper-level slot 3 $\cdot$ primary & This reviewer halucinated the most and it spit out word salads. I am sure there were some original insights in there (the review was 5 times longer than the others), but they were clouded by verbosity \\[3pt]
\end{longtable}
}

\subsubsection*{W6: Trivial / nitpicking --- n = 16}

{\footnotesize
\begin{longtable}{@{}p{3.2cm}p{11.5cm}@{}}
\toprule
\textbf{Citation} & \textbf{Expert comment} \\
\midrule
\endfirsthead
\toprule
\textbf{Citation} & \textbf{Expert comment} \\
\midrule
\endhead
\bottomrule
\endlastfoot
P3 $\cdot$ GPT-5.2 $\cdot$ item 3 $\cdot$ primary & I believe it is just a yet another result the authors present which is essentially disconnected from the main narrative. \\[3pt]
P17 $\cdot$ GPT-5.2 $\cdot$ item 3 $\cdot$ primary & It seems in the end criticizing a typo in the caption of Figure 3, 1000 \textbackslash\{\}omega\textasciicircum{}\{-1\}\_\{ci\} --\textgreater{} 100 \textbackslash\{\}omega\textasciicircum{}\{-1\}\_\{ci\} but the comment overly criticizing. Just pointing out a typo is enough. \\[3pt]
P25 $\cdot$ Claude 4.5 $\cdot$ item 1 $\cdot$ primary & more of a technicallity. \\[3pt]
P26 $\cdot$ GPT-5.2 $\cdot$ item 4 $\cdot$ primary & seems like a technicallity to me. \\[3pt]
P42 $\cdot$ Gemini 3.0 $\cdot$ item 3 $\cdot$ primary & Pretty sure that sentence is just a typo, not a misunderstanding and validity risk. \\[3pt]
P46 $\cdot$ GPT-5.2 $\cdot$ item 2 $\cdot$ primary & Do not think this is particular relevant. Sure, there may be badly predicted values outside the range... \\[3pt]
P53 $\cdot$ Gemini 3.0 $\cdot$ item 2 $\cdot$ primary & It is ok observation but definitely does not limit the study and the tools. it just makes it specific. \\[3pt]
P59 $\cdot$ Claude 4.5 $\cdot$ item 2 $\cdot$ secondary & This pairs with Human reviewer 3 critique about acidity preservation however here the critique is more of a ``the authors should discuss this limitation'' point than a fatal flaw. \\[3pt]
P62 $\cdot$ Gemini 3.0 $\cdot$ item 1 $\cdot$ primary & The comment is valid but the paper presents this as a grand vision rather than something it achieves, that is why this comment is marginally signicant rather than significant. \\[3pt]
P66 $\cdot$ Claude 4.5 $\cdot$ item 1 $\cdot$ primary & When we fit data with a Lorentzian using MATLAB for example, MATLAB gives us the best fit and range of values which leads to, let's say, R2 value of 0.99. The point raised by the AI reviewer is minor with respect to the main message of the manuscript. \\[3pt]
P67 $\cdot$ Claude 4.5 $\cdot$ item 3 $\cdot$ primary & This is a very minor issue. In fact, sometimes I personally forget to include the statistical analysis of ML outcomes. \\[3pt]
P68 $\cdot$ Claude 4.5 $\cdot$ item 1 $\cdot$ primary & This is true, but unlikely to change any major conclusion \\[3pt]
P68 $\cdot$ Gemini 3.0 $\cdot$ item 5 $\cdot$ primary & These are indeed speculative assertions that can be easily dealt with. \\[3pt]
P72 $\cdot$ Claude 4.5 $\cdot$ paper-level slot 1 $\cdot$ primary & AI reviewer 1 noted that the GIAB ground-truth set contained only SVs within the ``high-confidence'' regions. I consider to be a relatively minor point that could be addressed by emphasizing that benchmarking results may not generalize to more difficult/repetitive parts of the genome. \\[3pt]
P72 $\cdot$ Gemini 3.0 $\cdot$ paper-level slot 2 $\cdot$ primary & An unused argument in the code, which is hepful but about as minor as a criticism can be. \\[3pt]
P73 $\cdot$ Gemini 3.0 $\cdot$ paper-level slot 1 $\cdot$ primary & AI Reviewer 1 was the only reviewer to point out that there are newer methods for detecting microsat mutations, but I felt that this was a minor point. \\[3pt]
\end{longtable}
}

\subsubsection*{W7: Technical term-of-art confusion --- n = 13}

{\footnotesize
\begin{longtable}{@{}p{3.2cm}p{11.5cm}@{}}
\toprule
\textbf{Citation} & \textbf{Expert comment} \\
\midrule
\endfirsthead
\toprule
\textbf{Citation} & \textbf{Expert comment} \\
\midrule
\endhead
\bottomrule
\endlastfoot
P25 $\cdot$ Gemini 3.0 $\cdot$ item 1 $\cdot$ primary & On first sight it seems like a good comment, but AI and TW are established as two different aspects of neuronal dynamics, the first is local, the second is spatially extended. \\[3pt]
P28 $\cdot$ Gemini 3.0 $\cdot$ item 4 $\cdot$ primary & There is a misunderstanding of what the term ''universal'' means here. ''Universality'' is mere a physics term that means the data collaps to a single curve when scaled. It does not mean it is universally valid. \\[3pt]
P42 $\cdot$ Gemini 3.0 $\cdot$ item 2 $\cdot$ primary & This seems a bit of a terminology/scope issue, not a validity issue. \\[3pt]
P54 $\cdot$ Gemini 3.0 $\cdot$ item 3 $\cdot$ primary & AI should be aware of the connotation that is accepted by the genomics community for the use of terminologies such is real time. \\[3pt]
P65 $\cdot$ Gemini 3.0 $\cdot$ item 2 $\cdot$ primary & INR might take a few seconds for training, but once the training is complete, then it is extremely fast during inference. \\[3pt]
P66 $\cdot$ Gemini 3.0 $\cdot$ item 3 $\cdot$ primary & Here the resolution depends on the width of the tip, not the excitation wavelength. \\[3pt]
P67 $\cdot$ Claude 4.5 $\cdot$ item 4 $\cdot$ primary & I think AI got this totally wrong. Since 6 photodetectors have transmission peaks at different at different wavelengths, they can provide a much richer training data which eventually leads to a more accurate edge detectioon. \\[3pt]
P67 $\cdot$ Gemini 3.0 $\cdot$ item 1 $\cdot$ primary & SCA acts like a monochromatic filter, MCA is more like an array filter. \\[3pt]
P67 $\cdot$ Gemini 3.0 $\cdot$ item 2 $\cdot$ primary & SCA output is more like the average of 6 photodiodes of the same type. MCA output is more like an output of 6 different type photodiodes, which can distinguish the edges more accurately. \\[3pt]
P72 $\cdot$ GPT-5.2 $\cdot$ item 1 $\cdot$ primary & This comment is in line with one of the human reviewer's criticisms about the lack of methodological detail included in the paper. I don't get the AI reviewer's complaint about requiring a reference genome though, given that DeBreak is an read alignment-based method for detecting and locating SVs. \\[3pt]
P77 $\cdot$ Claude 4.5 $\cdot$ item 2 $\cdot$ secondary & In most of the substrates, there are no stereogenic centers to consider. If the reviewer is referring to the newly formed stereogenic centers arising from the C--C bond cleavage (Table 3), this is a valid point; however, these centers are not related to the electrocyclization step. \\[3pt]
P77 $\cdot$ Gemini 3.0 $\cdot$ item 3 $\cdot$ secondary & While the terminology appears to be somewhat misleading, both C--H and C--C bond cleavages would not take place in the absence of the metal carbene. \\[3pt]
P2 $\cdot$ Gemini 3.0 $\cdot$ paper-level slot 3 $\cdot$ primary & The issue with the size-extensivity of the proposed approach is a very important one. However, it should not be an argument for rejecting this work. I believer, the authors should simply thoroughly discuss this limitation. Overall, I found all AI reviewer very good but quite critical and rather not differentiating issues stemming from quantum chemistry and machine learning. I think, the issues from the former should be discussed in the paper, but not be a major issue when evaluating the proposed machine learning approach. \\[3pt]
\end{longtable}
}

\subsubsection*{W8: Cites evidence from after the preprint --- n = 9}

{\footnotesize
\begin{longtable}{@{}p{3.2cm}p{11.5cm}@{}}
\toprule
\textbf{Citation} & \textbf{Expert comment} \\
\midrule
\endfirsthead
\toprule
\textbf{Citation} & \textbf{Expert comment} \\
\midrule
\endhead
\bottomrule
\endlastfoot
P1 $\cdot$ Claude 4.5 $\cdot$ item 1 $\cdot$ primary & While claims are actually fine, I marked it as not correct as the paper they use to support these claims appeared two years after the presented one. The field was evolving very fast and in 2022 the presented results could be considered as good. The only thing one could move on with is the claims of improving upon the state-of-the-art. The auhtors should clarify what exactly they mean with it. \\[3pt]
P5 $\cdot$ Claude 4.5 $\cdot$ item 1 $\cdot$ primary & Funnily, this AI reviewer references in the last quote the same paper published later in 2022. \\[3pt]
P20 $\cdot$ Claude 4.5 $\cdot$ item 1 $\cdot$ primary & This point is significant and valid, however it lacks context and chronology. While it is true that spectroscopic confirmation is crucial, photometric evidence can still be compelling, especially given the fact that this discovery is reported very early after JWST started operations. Given this context, reporting such a potentialy far-reaching photometric discovery is acceptable. Additionally, the AI reviewer provides evidence from papers from 2025 which is invalid. \\[3pt]
P20 $\cdot$ Claude 4.5 $\cdot$ item 2 $\cdot$ primary & Again, while this point is valid and significant, the evidence is based on knowledge obtained in the future related to when this preprint was reviewed. Although a human reviewer should have asked the authors to comment on the AGN contamination, the significance would have been less because at that time we didn't have the knowledge of this class of objects. \\[3pt]
P20 $\cdot$ Gemini 3.0 $\cdot$ item 1 $\cdot$ primary & Again chronology is not preserved and the cited evidence is after this preprint \\[3pt]
P21 $\cdot$ Claude 4.5 $\cdot$ item 1 $\cdot$ primary & The comments cite a paper published 2 years after this preprint which is not valid. \\[3pt]
P46 $\cdot$ Claude 4.5 $\cdot$ item 3 $\cdot$ primary & This needs to be seen in historical context. PolyBERT was very early. \\[3pt]
P20 $\cdot$ Claude 4.5 $\cdot$ paper-level slot 1 $\cdot$ primary & Yes, but the evidence was based on knowledge obtained several years after the publication of this preprint. \\[3pt]
P20 $\cdot$ Gemini 3.0 $\cdot$ paper-level slot 3 $\cdot$ primary & Yes, but the evidence was based on knowledge obtained several years after the publication of this preprint. \\[3pt]
\end{longtable}
}

\subsubsection*{W9: Over-inflates small code/text inconsistencies --- n = 9}

{\footnotesize
\begin{longtable}{@{}p{3.2cm}p{11.5cm}@{}}
\toprule
\textbf{Citation} & \textbf{Expert comment} \\
\midrule
\endfirsthead
\toprule
\textbf{Citation} & \textbf{Expert comment} \\
\midrule
\endhead
\bottomrule
\endlastfoot
P12 $\cdot$ Claude 4.5 $\cdot$ item 2 $\cdot$ secondary & The two Doppler estimates are not shown to be in serious conflict, because the variability estimate has no quoted uncertainty. \\[3pt]
P34 $\cdot$ Gemini 3.0 $\cdot$ item 2 $\cdot$ primary & This one could be an unit error in the code, over-interpreted by AI-reviewer. \\[3pt]
P34 $\cdot$ Gemini 3.0 $\cdot$ item 4 $\cdot$ primary & This could be also the unit error in the code. \\[3pt]
P42 $\cdot$ GPT-5.2 $\cdot$ item 2 $\cdot$ primary & I think the AI is overstating the implications of small deviations between code and text that occur over the lifetime of a scientific paper. \\[3pt]
P42 $\cdot$ GPT-5.2 $\cdot$ item 5 $\cdot$ primary & Same comment as item 2. \\[3pt]
P46 $\cdot$ GPT-5.2 $\cdot$ item 4 $\cdot$ primary & This is most likely not a real problem and solved upon final submission. \\[3pt]
P54 $\cdot$ Claude 4.5 $\cdot$ item 5 $\cdot$ primary & Quote: ''Functional annotation of the assembled plasmid genomes using BugSeq revealed that KPC-2 and KPC-14 were both located on IncN plasmids, which showed 99.7\% identity according to sequence alignments''-overstatement by AI and overanalysis \\[3pt]
P59 $\cdot$ Claude 4.5 $\cdot$ item 3 $\cdot$ secondary & This feels very much AI generated critique that identifies a real asymmetry in methodology but fails to evalute whetjer it matters qunatitatively. the cement sector's contribution to total TPAE is so small that even substantial provincial variation in cement toxicity would not meaningfully alter the conclusions. \\[3pt]
P59 $\cdot$ GPT-5.2 $\cdot$ item 1 $\cdot$ secondary & The within-sector variation is modest compared to the \textasciitilde{}5 fold between-sector differences. So while the two aggregation methods would yield different absolute TU values, the sector ranking is very unlikely to change. However, the reviewer is correct that the authors don't demonstrate this, a sensitivity analysis would be needed to confirm. \\[3pt]
\end{longtable}
}

\subsubsection*{W10: Criticizes what authors already flagged as a limitation --- n = 6}

{\footnotesize
\begin{longtable}{@{}p{3.2cm}p{11.5cm}@{}}
\toprule
\textbf{Citation} & \textbf{Expert comment} \\
\midrule
\endfirsthead
\toprule
\textbf{Citation} & \textbf{Expert comment} \\
\midrule
\endhead
\bottomrule
\endlastfoot
P3 $\cdot$ Claude 4.5 $\cdot$ item 2 $\cdot$ primary & The authors state explicitly that it is a proof-of-concept study. Transferability claims are somewhat unrelated to the main narrative but it does not make the reviewer's claim correct. \\[3pt]
P46 $\cdot$ GPT-5.2 $\cdot$ item 1 $\cdot$ primary & It is a problem that is also already mentioned by the authors. It is up to the reviewer to judge if that invalidates everything. I would not think so. \\[3pt]
P60 $\cdot$ Gemini 3.0 $\cdot$ item 2 $\cdot$ secondary & The core criticism is valid but should be calibrated against what the paper actually discusses later. The deeper methodological question whether to fit separate models for the COVID period is something the authors themselves flag as future work, constrained by insufficient data (only 2 years of COVID-period observations for regional-level estimation). \\[3pt]
P60 $\cdot$ Gemini 3.0 $\cdot$ item 3 $\cdot$ secondary & This is a well-articulated critique, but it partly overlaps with Item 2 and with what the authors themselves discuss. The paper actually acknowledges this directly in the text quoted by the reviewer and in the discussion around Figure 4, where the authors present evidence suggesting the two hazards do interact and explicitly state they assumed independence. \\[3pt]
P62 $\cdot$ Gemini 3.0 $\cdot$ item 3 $\cdot$ secondary & The algorithm is essentially NMF and authors stated that they used the existing NMF package already available. The way to use NMF in fiber photometry context is new, so I think this comment, which is regarding code implementation, is marginal. \\[3pt]
P62 $\cdot$ Gemini 3.0 $\cdot$ item 4 $\cdot$ secondary & The key point of this study is not in NMF, nor the authors does not claim NMF as their developed method. The way to use NMF in fiber photometry context is new, so I think this comment is marginal. \\[3pt]
\end{longtable}
}

\subsubsection*{W11: AI misreads a figure or caption --- n = 6}

{\footnotesize
\begin{longtable}{@{}p{3.2cm}p{11.5cm}@{}}
\toprule
\textbf{Citation} & \textbf{Expert comment} \\
\midrule
\endfirsthead
\toprule
\textbf{Citation} & \textbf{Expert comment} \\
\midrule
\endhead
\bottomrule
\endlastfoot
P3 $\cdot$ Gemini 3.0 $\cdot$ item 1 $\cdot$ primary & For example, Figure S10 does not show barriers of 40-50 kJ/mol. Furthermore, Figure S13 shows every barriers of a comparable height to previous studies. Finally, the chosen DFT may be a valid critique but the results should be interpreted within this setting if it is not dramatically wrong. \\[3pt]
P19 $\cdot$ Gemini 3.0 $\cdot$ item 3 $\cdot$ primary & Maybe neglecting Figure 4. \\[3pt]
P35 $\cdot$ Claude 4.5 $\cdot$ item 3 $\cdot$ primary & review misidentifies the optimal threshold as t=0.4 when Figure 5c shows the peak at t=0.3. \\[3pt]
P35 $\cdot$ Claude 4.5 $\cdot$ item 4 $\cdot$ primary & error bars are present in figure2c (small gray lines on each bar) \\[3pt]
P66 $\cdot$ GPT-5.2 $\cdot$ item 4 $\cdot$ primary & I do not agree with the AI reviewer again. The uniformity in Fig3. d and e removes the questions about the defects. Since these were mechanical exfoliated samples, strain criticism is also not relevant. \\[3pt]
P77 $\cdot$ Claude 4.5 $\cdot$ item 3 $\cdot$ secondary & The reviewer may have misunderstood Figure 2 in the main text. \\[3pt]
\end{longtable}
}

\subsubsection*{W12: AI misquotes or fabricates a verbatim quote --- n = 5}

{\footnotesize
\begin{longtable}{@{}p{3.2cm}p{11.5cm}@{}}
\toprule
\textbf{Citation} & \textbf{Expert comment} \\
\midrule
\endfirsthead
\toprule
\textbf{Citation} & \textbf{Expert comment} \\
\midrule
\endhead
\bottomrule
\endlastfoot
P29 $\cdot$ GPT-5.2 $\cdot$ item 1 $\cdot$ primary & I doubt AI run the model. \\[3pt]
P49 $\cdot$ Gemini 3.0 $\cdot$ item 1 $\cdot$ secondary & Instead of quoting the author's exact words, this AI reviewer incorrectly quoted the author and then claimed that the author's claims were incorrect. It's bit surprising that AI reviewers can make such mistakes, and I think we need to be aware that they can make such simple mistakes. \\[3pt]
P65 $\cdot$ Gemini 3.0 $\cdot$ item 4 $\cdot$ primary & Again, AI made a mistake by comparing the sentences taken from different parts of the manuscript talking about different parts of the process. \\[3pt]
P77 $\cdot$ GPT-5.2 $\cdot$ item 3 $\cdot$ secondary & I could not find CuTe in both main text and supplementary information. \\[3pt]
P77 $\cdot$ Gemini 3.0 $\cdot$ item 1 $\cdot$ secondary & I could not observe the discrepancies pointed out by the reviewer. \\[3pt]
\end{longtable}
}

\subsubsection*{W13: AI misses supplementary content --- n = 3}

{\footnotesize
\begin{longtable}{@{}p{3.2cm}p{11.5cm}@{}}
\toprule
\textbf{Citation} & \textbf{Expert comment} \\
\midrule
\endfirsthead
\toprule
\textbf{Citation} & \textbf{Expert comment} \\
\midrule
\endhead
\bottomrule
\endlastfoot
P6 $\cdot$ Gemini 3.0 $\cdot$ item 1 $\cdot$ secondary & very detailed in the SI: The RF and DB 0 ?eld maps were acquired for each subject for the pulse design and shimming using 2D interferometric turbo-FLASH [30] at 5 mm isotropic resolution and 3D multi-echo (TE = 1.6/3.5/6 ms) GRE acquisitions at 2.5 mm isotropic resolution respectively. Sequence parameters for the T 2 -weighted variable ?ip angle turbo spin-echo acquisition were: resolution = 0.55 $\times$ 0.55 $\times$ 0.55 mm 3 , TR = 6 s, TA = 13 min, GRAPPA = 3 $\times$ 2, TE = 301 ms, matrix = 400 $\times$ 400 $\times$ 320, bandwidth = 250 Hz/pixel. \\[3pt]
P37 $\cdot$ GPT-5.2 $\cdot$ item 1 $\cdot$ secondary & The first comment is correct. The second comment is correct based on what is written in the main text. But if you refer to the Supplementary Figure 3, then we can see the number 281 in the main text is a typo. The actual number is 69281. It seems the AI Reviewer did not check the supplementary material. \\[3pt]
P53 $\cdot$ Claude 4.5 $\cdot$ item 1 $\cdot$ primary & In supplementary \\[3pt]
\end{longtable}
}

\subsubsection*{W14: Ignores authors' own cited prior work --- n = 2}

{\footnotesize
\begin{longtable}{@{}p{3.2cm}p{11.5cm}@{}}
\toprule
\textbf{Citation} & \textbf{Expert comment} \\
\midrule
\endfirsthead
\toprule
\textbf{Citation} & \textbf{Expert comment} \\
\midrule
\endhead
\bottomrule
\endlastfoot
P28 $\cdot$ Claude 4.5 $\cdot$ item 2 $\cdot$ primary & The authors make the claim about correlation of nutrient concentration and interaction strength based on their previous work. This critique is similar to the one of the Human Reviewers. \\[3pt]
P28 $\cdot$ Gemini 3.0 $\cdot$ item 2 $\cdot$ primary & The authors make this claim based on their previous work. Similar to one of the human reviewers, the AI reviewer did not read those cited papers. \\[3pt]
\end{longtable}
}

\subsubsection*{W15: AI misreads a table --- n = 1}

{\footnotesize
\begin{longtable}{@{}p{3.2cm}p{11.5cm}@{}}
\toprule
\textbf{Citation} & \textbf{Expert comment} \\
\midrule
\endfirsthead
\toprule
\textbf{Citation} & \textbf{Expert comment} \\
\midrule
\endhead
\bottomrule
\endlastfoot
P32 $\cdot$ Claude 4.5 $\cdot$ item 2 $\cdot$ primary & The reviewer's assessment seems to be based on an over-interpretation of Table 1, where a typographical error in the values may have led to an incorrect conclusion. \\[3pt]
\end{longtable}
}

\subsubsection*{W16: Cannot analyze figures --- only text --- n = 1}

{\footnotesize
\begin{longtable}{@{}p{3.2cm}p{11.5cm}@{}}
\toprule
\textbf{Citation} & \textbf{Expert comment} \\
\midrule
\endfirsthead
\toprule
\textbf{Citation} & \textbf{Expert comment} \\
\midrule
\endhead
\bottomrule
\endlastfoot
P6 $\cdot$ Gemini 3.0 $\cdot$ item 5 $\cdot$ primary & This could be argued better, but to do this the AI would need to be able to analyze the images. It went by text only, did not download the figures from Figshare, so it gave boilerplate feedback. \\[3pt]
\end{longtable}
}

\subsubsection*{W\_unspecified: Residual: AI judged Not Correct without specific reason --- n = 10}

{\footnotesize
\begin{longtable}{@{}p{3.2cm}p{11.5cm}@{}}
\toprule
\textbf{Citation} & \textbf{Expert comment} \\
\midrule
\endfirsthead
\toprule
\textbf{Citation} & \textbf{Expert comment} \\
\midrule
\endhead
\bottomrule
\endlastfoot
P28 $\cdot$ Gemini 3.0 $\cdot$ item 5 $\cdot$ primary & I don't think this is correct. \\[3pt]
P29 $\cdot$ Gemini 3.0 $\cdot$ item 1 $\cdot$ primary & Complete fail here. \\[3pt]
P30 $\cdot$ GPT-5.2 $\cdot$ item 4 $\cdot$ primary & I am not sure where this code is. \\[3pt]
P34 $\cdot$ Gemini 3.0 $\cdot$ item 1 $\cdot$ secondary & I did not verify this in detail by running the Python code, and I assume the authors did not make this mistake. However, if this is indeed the case, it would be a significant issue. \\[3pt]
P3 $\cdot$ Claude 4.5 $\cdot$ paper-level slot 2 $\cdot$ primary & I found this reviewer rather bad. \\[3pt]
P3 $\cdot$ Gemini 3.0 $\cdot$ paper-level slot 3 $\cdot$ primary & I found this reviewer rather bad. \\[3pt]
P9 $\cdot$ GPT-5.2 $\cdot$ paper-level slot 1 $\cdot$ primary & Human review 2 was not provided in the checklist. For the last question here I added AI reviewer 3 however it's a bit borderline, I think the 3 AI reviews were not very good. \\[3pt]
P77 $\cdot$ Gemini 3.0 $\cdot$ paper-level slot 1 $\cdot$ primary & AI Reviewer 1 offered the weakest assessment among all reviewers. \\[3pt]
P79 $\cdot$ GPT-5.2 $\cdot$ paper-level slot 1 $\cdot$ primary & AI reviewer 1 does not seem to have a fundamental understanding of the key discovery of the manuscript. \\[3pt]
P80 $\cdot$ GPT-5.2 $\cdot$ paper-level slot 1 $\cdot$ primary & it misses the main point of the paper. \\[3pt]
\end{longtable}
}

\subsubsection*{S1: Statistical / methodology rigor --- n = 45}

{\footnotesize
\begin{longtable}{@{}p{3.2cm}p{11.5cm}@{}}
\toprule
\textbf{Citation} & \textbf{Expert comment} \\
\midrule
\endfirsthead
\toprule
\textbf{Citation} & \textbf{Expert comment} \\
\midrule
\endhead
\bottomrule
\endlastfoot
P7 $\cdot$ Gemini 3.0 $\cdot$ item 2 $\cdot$ primary & Here in general the problem is that the fit is of bad quality \\[3pt]
P35 $\cdot$ GPT-5.2 $\cdot$ item 1 $\cdot$ primary & The PPI trainer has no validation split (only train/test), uses test metrics for model selection (misleadingly named best\_valid\_f1) \\[3pt]
P60 $\cdot$ Claude 4.5 $\cdot$ item 3 $\cdot$ secondary & Even if the overall sample size were adequate, treating correlated regional observations within the same event as independent violates the K-S test assumptions. This is a genuinely useful statistical critique that neither the first reviewer in this set nor the human reviewers identified. \\[3pt]
P60 $\cdot$ Gemini 3.0 $\cdot$ item 1 $\cdot$ secondary & This is an outstanding technical critique. It's worth noting that the attributable AF itself is independent of the baseline, so the relative risk estimates are unaffected but the absolute death counts, which are the paper's primary output, scale directly with deathexp. \\[3pt]
P60 $\cdot$ Gemini 3.0 $\cdot$ item 5 $\cdot$ secondary & The 1.7--3.8x fold increases for heatwaves and 2.1--3.5x for cold snaps are presented as key results demonstrating the severity of compound crises, but as the reviewer notes, these numbers are mathematically inevitable once COVID-19 deaths exist. This connects to Item 3's point about the paper not truly measuring compound/interactive effects. \\[3pt]
P61 $\cdot$ Claude 4.5 $\cdot$ item 3 $\cdot$ secondary & this is logically paired with Item 2 and together they form a coherent critique: the provincial methodology is underspecified (Item 2) and unvalidated (Item 3). It's possible that the Pearson correlation in Figure S3 does include some provincial-level comparisons but again since its not explicit, authors needs to work on the presentation. \\[3pt]
P61 $\cdot$ Claude 4.5 $\cdot$ item 4 $\cdot$ secondary & The authors should clarify: did they combine bounds simultaneously (which would overestimate uncertainty for independent sources), use Monte Carlo sampling, or apply some other approach? This is a fair ask for methodological transparency even if the practical impact on conclusions is small. \\[3pt]
P61 $\cdot$ GPT-5.2 $\cdot$ item 2 $\cdot$ secondary & The paper would benefit from a supplementary table of fitted parameters, a description of the ARIMA specification procedure, and explicit statement of the TMREL distribution used. \\[3pt]
P61 $\cdot$ GPT-5.2 $\cdot$ item 3 $\cdot$ secondary & The reviewer correctly distinguishes between association (correlation) and agreement (accuracy in magnitude), citing the Bland-Altman framework which is the standard methodological reference for exactly this issue \\[3pt]
P61 $\cdot$ GPT-5.2 $\cdot$ item 4 $\cdot$ secondary & This critique pairs well with Review 1's Item 4 (incomplete uncertainty propagation) but is far more precise. The critique doesn't necessarily invalidate the central qualitative findings (aging as dominant driver), but it does undermine the quantitative confidence statements that the paper emphasizes throughout. \\[3pt]
P63 $\cdot$ Gemini 3.0 $\cdot$ item 4 $\cdot$ primary & The power criticism is very correct and very significant. On the time, comments are correct but the mentioned challenges can be overcome. Howecer the enegy part is the fundamental citicism that should be enough to reject the paper. \\[3pt]
P65 $\cdot$ GPT-5.2 $\cdot$ item 4 $\cdot$ primary & I agree with the AI: ''a great deal'' does not mean anything from a scientific point of view. They should have quantified their finding. \\[3pt]
P68 $\cdot$ Gemini 3.0 $\cdot$ item 2 $\cdot$ primary & Given that there are categorical PLS options, this is a a reasonable criticism. \\[3pt]
P73 $\cdot$ GPT-5.2 $\cdot$ item 1 $\cdot$ primary & These methods do appear to be missing, and could be consequential. \\[3pt]
P74 $\cdot$ Claude 4.5 $\cdot$ item 2 $\cdot$ primary & I agree that data leakage is possible if the same regions are included in both training and test sets. A safer approach would be to split the genome up into segments that are partitioned into the training and test sets. \\[3pt]
P3 $\cdot$ GPT-5.2 $\cdot$ paper-level slot 1 $\cdot$ primary & This AI reviewer recognized ambiguity in the description of methods used for certain calculations and asks for the clarification for the computed coverages. \\[3pt]
P4 $\cdot$ Gemini 3.0 $\cdot$ paper-level slot 2 $\cdot$ primary & Essentially, I find again great that also this reviewer questions if random splitting is a good strategy, similar to the first AI reviewer. \\[3pt]
P5 $\cdot$ Gemini 3.0 $\cdot$ paper-level slot 1 $\cdot$ primary & I found this AI reviewer quite good. They were very constructive and concise. Their arguments are close to those I would make. Similar to the third reviewer, the most significant contribution, in my opinion, was pointing out the lack of a proper baseline comparison and the absence of variance estimates in the computed errors of the predicted values. \\[3pt]
P14 $\cdot$ Claude 4.5 $\cdot$ paper-level slot 1 $\cdot$ secondary & Yes, Potential confounding of circadian rhythm detection by ongoing therapeutic stimulation. Human reviewer only consider the relation with the data analysis issues. \\[3pt]
P15 $\cdot$ Claude 4.5 $\cdot$ paper-level slot 2 $\cdot$ primary & Absence of multiple comparisons \\[3pt]
P15 $\cdot$ Claude 4.5 $\cdot$ paper-level slot 2 $\cdot$ secondary & [non-Latin][non-Latin][non-Latin] [non-Latin][non-Latin][non-Latin] [non-Latin][non-Latin] [non-Latin][non-Latin][non-Latin] [non-Latin][non-Latin] [non-Latin][non-Latin][non-Latin][non-Latin][non-Latin] [non-Latin][non-Latin][non-Latin][non-Latin][non-Latin][non-Latin]. \\[3pt]
P15 $\cdot$ GPT-5.2 $\cdot$ paper-level slot 1 $\cdot$ primary & Yes, data imputation methodological problem and inconsistencies; multiple comparisons absence. \\[3pt]
P15 $\cdot$ Gemini 3.0 $\cdot$ paper-level slot 3 $\cdot$ primary & Lack of analyses on EEG microstructures; data imputation problem \\[3pt]
P16 $\cdot$ Claude 4.5 $\cdot$ paper-level slot 3 $\cdot$ primary & granger causality group level significance but not individual level \\[3pt]
P24 $\cdot$ GPT-5.2 $\cdot$ paper-level slot 3 $\cdot$ primary & catching the normative atlas issue missed by the humans. nice catching the lack of spurious correlations null model. \\[3pt]
P35 $\cdot$ Claude 4.5 $\cdot$ paper-level slot 3 $\cdot$ primary & The utilization rate metric mathematical error and the suspicious standard deviation entry in Table S2 were unique observations. \\[3pt]
P35 $\cdot$ GPT-5.2 $\cdot$ paper-level slot 2 $\cdot$ primary & The evaluation protocol violation (missing validation split, model selection on test set) provides a complementary angle to AI Review 1's data leakage finding. The BioSNAP benchmark limitation and code reproducibility issues were unique to AI Reviewer 2. \\[3pt]
P37 $\cdot$ GPT-5.2 $\cdot$ paper-level slot 3 $\cdot$ secondary & It was the only reviewer who noticed: - Class imbalance in the dataset affects AUPRC, which can undermine the SOTA claim of the paper. - The submitted code does not accurately calculate AUROC \\[3pt]
P42 $\cdot$ Gemini 3.0 $\cdot$ paper-level slot 2 $\cdot$ primary & AI reviewer 2 found the collinearity of habitat fragmentation and amount and had a good suggestion for sensitivity analysis. \\[3pt]
P47 $\cdot$ Claude 4.5 $\cdot$ paper-level slot 3 $\cdot$ primary & AI \#3 uniquely pointed out that the entropy equation is mathematically incorrect. \\[3pt]
P47 $\cdot$ Gemini 3.0 $\cdot$ paper-level slot 1 $\cdot$ primary & AI \#1 was the only reviewer who raised a critical question regarding the intra-hamming distance, and argued that its magnitude is too large to support stable cryptographic key generation. \\[3pt]
P49 $\cdot$ Claude 4.5 $\cdot$ paper-level slot 1 $\cdot$ secondary & AI reviewer 1 pointed out a statistical flaw in the distance-dependent cross-validation method. Specifically, test regions adjacent to the boundary of the 75\% training set and the 25\% test set still share spatial autocorrelation, which can potentially inflate the model's performance estimates. \\[3pt]
P50 $\cdot$ GPT-5.2 $\cdot$ paper-level slot 1 $\cdot$ primary & Yes, AI reveiwer 1 found out about SNR which all other reviewers failed to catch. \\[3pt]
P50 $\cdot$ GPT-5.2 $\cdot$ paper-level slot 1 $\cdot$ secondary & AI Reviewer 1's review included details such as the SNR that was not in other reviews, human or otherwise. \\[3pt]
P52 $\cdot$ GPT-5.2 $\cdot$ paper-level slot 3 $\cdot$ primary & Yes. Issue of experimental setting on ``control''. \\[3pt]
P52 $\cdot$ GPT-5.2 $\cdot$ paper-level slot 3 $\cdot$ secondary & confounding factors related to contralateral effects of the drug that was used unilaterally, with the contralateral eye as a control. Also, this reviewer mentioned the code that was used in the paper. \\[3pt]
P55 $\cdot$ Claude 4.5 $\cdot$ paper-level slot 3 $\cdot$ primary & AI reviewer 3 overestimated the age of the population and had similar comments as the other AI reviewers. it did cough more statistics and model flaws than all the human comments. same applies for AI 1 and AI2. \\[3pt]
P55 $\cdot$ Gemini 3.0 $\cdot$ paper-level slot 1 $\cdot$ primary & MS methodology in for example catching the imputation in the data analysis. this was not cought by the human reviewer. \\[3pt]
P67 $\cdot$ Claude 4.5 $\cdot$ paper-level slot 3 $\cdot$ primary & AIR 3 caught the lack of statistical analysis, which was not caught by other reviewers. AIR2 and AIR3 did a very good job in terms finding the different resolution used in SCA and MCA, which I hope was fixed later by the authors. \\[3pt]
P68 $\cdot$ Gemini 3.0 $\cdot$ paper-level slot 2 $\cdot$ primary & The ordinal data issue was raised by AI reviewer 2 and not by others. \\[3pt]
P79 $\cdot$ Claude 4.5 $\cdot$ paper-level slot 2 $\cdot$ primary & AI reviewers 2 and 3 both made critical assessment regarding the causality \\[3pt]
P79 $\cdot$ Gemini 3.0 $\cdot$ paper-level slot 3 $\cdot$ primary & AI reviewers 2 and 3 both made critical assessment regarding the causality. AI Reviewer 3 was more in depth than AI reviewer 2. \\[3pt]
P80 $\cdot$ Claude 4.5 $\cdot$ paper-level slot 3 $\cdot$ secondary & Yes. AI Reviewer 3 gave reasonable comments on the simulation setting, theoretical assumption, and uncertainty in real data analysis that other reviewers did not catch. \\[3pt]
P85 $\cdot$ GPT-5.2 $\cdot$ paper-level slot 1 $\cdot$ primary & Library preparation batch effects \\[3pt]
P85 $\cdot$ Gemini 3.0 $\cdot$ paper-level slot 3 $\cdot$ primary & AI Reviewer 3 was very focused on data analysis (computational) procedures and almost all comments were distinct from the other reviewers (human and AI). \\[3pt]
\end{longtable}
}

\subsubsection*{S2: Code reading --- n = 28}

{\footnotesize
\begin{longtable}{@{}p{3.2cm}p{11.5cm}@{}}
\toprule
\textbf{Citation} & \textbf{Expert comment} \\
\midrule
\endfirsthead
\toprule
\textbf{Citation} & \textbf{Expert comment} \\
\midrule
\endhead
\bottomrule
\endlastfoot
P1 $\cdot$ GPT-5.2 $\cdot$ item 4 $\cdot$ primary & I find it quite impressive that this reviewer actually looks at the code to understand whether some parts of the paper are not properly discussed. I am personally not doing it on a regular basis as it would consume a substantial amount of time. \\[3pt]
P6 $\cdot$ Gemini 3.0 $\cdot$ item 4 $\cdot$ primary & Impressive that the AI looked at the code and realized there is nothing about the MRI reconstruction. It missed that some of the links are broken even for the ANT-R task. \\[3pt]
P35 $\cdot$ Gemini 3.0 $\cdot$ item 1 $\cdot$ primary & Code verification fully confirms the data leakage claim. The DPI link trainer (trainer\_dpi.py) also has a bug using test metrics for model selection despite having a validation set. \\[3pt]
P37 $\cdot$ Claude 4.5 $\cdot$ item 3 $\cdot$ secondary & I checked the validity of this item using Claude Code. \\[3pt]
P37 $\cdot$ GPT-5.2 $\cdot$ item 3 $\cdot$ secondary & I checked the validity of this item using Claude Code. \\[3pt]
P37 $\cdot$ GPT-5.2 $\cdot$ item 4 $\cdot$ secondary & I checked the validity of this item using Claude Code. \\[3pt]
P37 $\cdot$ Gemini 3.0 $\cdot$ item 2 $\cdot$ secondary & I didn't manually go through the code to verify the claim. Instead, I asked Claude Opus 4.6 to verify the claim and it said the claim was correct. \\[3pt]
P49 $\cdot$ GPT-5.2 $\cdot$ item 2 $\cdot$ secondary & Unlike human reviewers, this reviewer actually visited the GitHub open source code repository the authors listed in their papers, dissected and analyzed the Python scripts to identify problems in the analysis. \\[3pt]
P50 $\cdot$ Gemini 3.0 $\cdot$ item 2 $\cdot$ primary & Human reviewer would not open up the raw code to find this out - if this code is correct, basically the core data reported in this paper is wrong. Authors in the code said ``It appears that delay is needed in order not to clog the port'' meaning it's not possible to wirelessly transmit high frequency information. If this is true, Figure 4, for example, would mean the authors measured in wired manner and falsely claimed that their system is wireless (which is one key aspect of this paper). \\[3pt]
P67 $\cdot$ Claude 4.5 $\cdot$ item 1 $\cdot$ primary & Same as the AI reviewer-2, this reviewer also caught the same resolution issue. Maybe this is one of the a few advantages over the human reviewers. AI is good at reviewing the codes and finding bugs. \\[3pt]
P67 $\cdot$ GPT-5.2 $\cdot$ item 2 $\cdot$ primary & I didn't examine their code, but it looks like it wasn't a fair comparison. I hope the authors later fixed this. \\[3pt]
P74 $\cdot$ Gemini 3.0 $\cdot$ item 1 $\cdot$ primary & The AI reviewer seems to be correct here, although I did not go through the entireity of the code so I cannot be certain that this check was not done elsewhere. \\[3pt]
P1 $\cdot$ GPT-5.2 $\cdot$ paper-level slot 3 $\cdot$ primary & I think this reviewer also more clearly identified the systematic issue in the evaluation of the finite-size correction. I was also impressed by how this reviewer examined the code, identifying inconsistencies in the mathematical formulations. \\[3pt]
P15 $\cdot$ GPT-5.2 $\cdot$ paper-level slot 1 $\cdot$ secondary & [non-Latin][non-Latin] [non-Latin][non-Latin] [non-Latin][non-Latin][non-Latin] [non-Latin][non-Latin][non-Latin][non-Latin][non-Latin][non-Latin]. [non-Latin][non-Latin] [non-Latin][non-Latin] [non-Latin][non-Latin][non-Latin][non-Latin] [non-Latin][non-Latin][non-Latin][non-Latin] [non-Latin][non-Latin][non-Latin] [non-Latin] [non-Latin][non-Latin] [non-Latin][non-Latin][non-Latin][non-Latin], [non-Latin][non-Latin] [non-Latin][non-Latin][non-Latin][non-Latin][non-Latin][non-Latin] [non-Latin][non-Latin][non-Latin] [non-Latin][non-Latin][non-Latin][non-Latin][non-Latin]. \\[3pt]
P23 $\cdot$ Claude 4.5 $\cdot$ paper-level slot 3 $\cdot$ primary & Good job in checking the code and finding the discrepancies. \\[3pt]
P23 $\cdot$ GPT-5.2 $\cdot$ paper-level slot 2 $\cdot$ primary & Good job in checking the code and finding the discrepancies. \\[3pt]
P24 $\cdot$ Gemini 3.0 $\cdot$ paper-level slot 2 $\cdot$ primary & nice checking the code and catching the normative atlas issue missed by the humans. \\[3pt]
P25 $\cdot$ GPT-5.2 $\cdot$ paper-level slot 3 $\cdot$ primary & Thoroughly checking the code and whether it can reproduce the results. \\[3pt]
P26 $\cdot$ GPT-5.2 $\cdot$ paper-level slot 2 $\cdot$ primary & good job on noticing and calling up the circularity in the work. Alsoo very useful for checking the code. \\[3pt]
P27 $\cdot$ Claude 4.5 $\cdot$ paper-level slot 2 $\cdot$ primary & human reviewers, especially 1, were very good. AI only exceeded them in checking the code. \\[3pt]
P27 $\cdot$ GPT-5.2 $\cdot$ paper-level slot 1 $\cdot$ primary & human reviewers, especially 1, were very good. AI only exceeded them in checking the code. \\[3pt]
P27 $\cdot$ Gemini 3.0 $\cdot$ paper-level slot 3 $\cdot$ primary & human reviewers, especially 1, were very good. AI only exceeded them in checking the code. \\[3pt]
P35 $\cdot$ Gemini 3.0 $\cdot$ paper-level slot 1 $\cdot$ primary & The graph structure data leakage mechanism and the swapped EM step naming in code were unique. The co-training critique overlaps with others but adds unique code evidence. \\[3pt]
P49 $\cdot$ GPT-5.2 $\cdot$ paper-level slot 2 $\cdot$ secondary & AI Reviewer 2 went beyond merely reading the text of the manuscript and providing logical critiques. It delved directly into the actual Python code scripts, data arrays, and references of the original datasets provided by the authors, accurately pinpointing critical discrepancies between what was claimed in the paper (methods) and what was actually executed (code or data). - False description of the cross-validation code: The manuscript claims to have performed 1,000 random iterations for distance-dependent cross-validation, but AI Reviewer 2 analyzed the provided Python code and revealed that it actually executed only 68 deterministic splits corresponding to the 68 brain regions. - Mismatch in the number of predictors: The methodology explicitly states that nine global connectome predictors were used, but AI Reviewer 2 found that only seven predictors were actually enumerated in the text and implemented in the code. - Age range error in ENIGMA data: Contrary to the authors' assertion that all cortical abnormality maps were collected from ''adult patients,'' the cited ENIGMA ASD dataset actually comprises participants aged 2 to 64 years, highlighting that unaddressed developmental lifespan effects could heavily distort the results. \\[3pt]
P50 $\cdot$ Gemini 3.0 $\cdot$ paper-level slot 2 $\cdot$ primary & Yes, AI Reviewer 2 found a problem that can be raised from raw code - which if true would flip all the claims this paper made on their capability of wireless system. \\[3pt]
P60 $\cdot$ Gemini 3.0 $\cdot$ paper-level slot 1 $\cdot$ secondary & This reviewer went directly into the code to uncover a concrete computational bias that no other reviewer could have found from reading the manuscript alone. It also identified opposing directions of bias, harvesting potentially overestimating temperature deaths while pandemic context underestimates them. It questioned that compound term was used only in wordings but not analytically. \\[3pt]
P67 $\cdot$ Gemini 3.0 $\cdot$ paper-level slot 1 $\cdot$ secondary & I find the review of AI reviewer 1 to be the best among all the human and AI reviews. One thing I would like to mention when comparing to human reviews is that this AI looked thoroughly into the code to find inconsistencies. \\[3pt]
P74 $\cdot$ Gemini 3.0 $\cdot$ paper-level slot 3 $\cdot$ primary & AI Reviewer 3 dug into the code and noticed the possiblity that ``hidden split reads'' are not actually being detected in the same way as described in the paper. \\[3pt]
\end{longtable}
}

\subsubsection*{S3: Specialized niche field catch --- n = 27}

{\footnotesize
\begin{longtable}{@{}p{3.2cm}p{11.5cm}@{}}
\toprule
\textbf{Citation} & \textbf{Expert comment} \\
\midrule
\endfirsthead
\toprule
\textbf{Citation} & \textbf{Expert comment} \\
\midrule
\endhead
\bottomrule
\endlastfoot
P61 $\cdot$ GPT-5.2 $\cdot$ item 1 $\cdot$ secondary & This critique substantially improves upon Human Reviewer 1's Item 20 (which asked about the formula's basis).If nitrate data were unavailable from the CMIP6 models used, the authors should have stated this as a limitation and discussed its implications for their estimates. \\[3pt]
P64 $\cdot$ GPT-5.2 $\cdot$ item 4 $\cdot$ primary & this is a good catch. retrieving the complex field (with the imaginary part) is very ill-posed and to the best of my knowledge there is no method to do that well with multimode fiber propagation. arbitrary optical fields is for sure not within the scope of the paper and the method has nothing to do with the complex field. the authors state the relation as amplitude-to-amplitude, so this part is accurate but for sure not enough to extrapolate this huge leap. \\[3pt]
P66 $\cdot$ GPT-5.2 $\cdot$ item 3 $\cdot$ primary & It is true that the bright-versus-dark exciton interpretation is not adequately supported and we do not know the exact illumination angle. But the authors claim a bright exciton dominance, so it should be close to a normal excitation. We should remember that the rapid spatial variation of the field at the tip apex is also required for dark excitons. If they were moving their tip slowly, then why would they have a serious dark excitons?? \\[3pt]
P66 $\cdot$ Gemini 3.0 $\cdot$ item 1 $\cdot$ primary & I agree with the AI reviewer that the PDM with their implementation is not the most accurate (they should have used the layered medium Green's functions and take care of the inhomogeneous background). However, this doesn't mean that the main findings of these researchers were wrong. A more accurate numerical model would generate results with smaller error limits, that's it. \\[3pt]
P73 $\cdot$ GPT-5.2 $\cdot$ item 3 $\cdot$ primary & I don't quite follow the argument about the GWAS results here, but the broader point about homopolymer errors is valid. \\[3pt]
P77 $\cdot$ GPT-5.2 $\cdot$ item 2 $\cdot$ primary & Since the substraste used for the synthesis of compound 4v contains stereogenic center, the issue of diastereoselectivity is important. Presumaby, mixture of diastereomers were formed in compound 4v. It was not mentioned by the human reviewers. \\[3pt]
P77 $\cdot$ Gemini 3.0 $\cdot$ item 2 $\cdot$ secondary & Although the formation of a seven-membered ring via an 8$\pi$ electrocyclization is relatively rare, such transformations have been reported. Nevertheless, I agree with the reviewer that the manuscript would be more impactful if additional experimental or computational evidence were provided to support the proposed mechanism. \\[3pt]
P1 $\cdot$ Claude 4.5 $\cdot$ paper-level slot 2 $\cdot$ primary & I think this reviewer more clearly identified the systematic issue in the evaluation of the finite-size correction. \\[3pt]
P2 $\cdot$ Claude 4.5 $\cdot$ paper-level slot 1 $\cdot$ primary & I found AI reviewer quite good. This identified important issues such as the lack of statistical sampling or the use of the minimal basis set, which may weaken the presented results. Also pointing out to a rather unfair comparison to GAP potentials appears to be raised by this AI reviewer but not the human ones, which I found very good. Overall, I found every single point raised by this AI reviewer as important and not necessarily found them in the human reviewer. \\[3pt]
P2 $\cdot$ GPT-5.2 $\cdot$ paper-level slot 2 $\cdot$ primary & This reviewer raises an issue concerning how close the provided results are really to the reference on the Zundel-cation benchmark. I found it quite important for the scope of this paper. And it appears not to be raised by the human reviewer. Computational cost is another point raised by this AI reviewer I found particularly important. Finally, this AI reviewer discusses issues with the code which I find very good, as a human reviewer I check the code only if I am very suspicious about the results and could have overlooked it. \\[3pt]
P7 $\cdot$ Claude 4.5 $\cdot$ paper-level slot 2 $\cdot$ primary & Yes there was a mention how systematic uncertainties were reported in Table I that was not spotted by other reviewers. \\[3pt]
P15 $\cdot$ Gemini 3.0 $\cdot$ paper-level slot 3 $\cdot$ secondary & [non-Latin][non-Latin] [non-Latin][non-Latin] [non-Latin][non-Latin](local sleep disruption)[non-Latin] circadian clock [non-Latin][non-Latin] [non-Latin][non-Latin][non-Latin] [non-Latin][non-Latin][non-Latin] [non-Latin][non-Latin] \\[3pt]
P23 $\cdot$ Gemini 3.0 $\cdot$ paper-level slot 1 $\cdot$ primary & It did significantly better for the depth of the comments, regarding the other 2 AI reviewers, but also made good point for some of the items, that were missed by the human reviewrs (frontal alpha in propofol). \\[3pt]
P24 $\cdot$ Claude 4.5 $\cdot$ paper-level slot 1 $\cdot$ primary & nice catching the normative atlas issue missed by the humans. \\[3pt]
P52 $\cdot$ Gemini 3.0 $\cdot$ paper-level slot 2 $\cdot$ primary & Yes. Raised issue of biosafety of gold dissolution. \\[3pt]
P52 $\cdot$ Gemini 3.0 $\cdot$ paper-level slot 2 $\cdot$ secondary & dissolution of gold for drug release can be toxic - needs testing \\[3pt]
P73 $\cdot$ Claude 4.5 $\cdot$ paper-level slot 2 $\cdot$ primary & AI Reviewer 2 was the only one to mention the potential confounding effect of population structure on the GWAS results. \\[3pt]
P73 $\cdot$ GPT-5.2 $\cdot$ paper-level slot 3 $\cdot$ primary & AI Reviewer 3 was the only one to raise a concern about the authors' extrapolation to estimate genome-wide mutation rates per microsatellite, although I consider this to be a reliatvely minor concern. \\[3pt]
P74 $\cdot$ GPT-5.2 $\cdot$ paper-level slot 2 $\cdot$ primary & AI Reviewer 2 noted that the non-human benchmark data were limited to a single SV caller and thus could have many errors. \\[3pt]
P77 $\cdot$ Claude 4.5 $\cdot$ paper-level slot 2 $\cdot$ primary & Yes, AI reviewer 2 pointed out am important point regarding the difference of Ag catalysis and Cu catalysis. \\[3pt]
P77 $\cdot$ GPT-5.2 $\cdot$ paper-level slot 3 $\cdot$ primary & AI reviewer 3 nicely brought up the issue of diastereoselectivity in the product formation which was not mentioned by the human reviewers. \\[3pt]
P81 $\cdot$ Gemini 3.0 $\cdot$ paper-level slot 1 $\cdot$ primary & - Only reviewer to question the use of the term ''LED,'' given that the device operates via gate oxide breakdown rather than conventional carrier injection. - Uniquely criticized the use of spatial intensity (mW/cm$^{2}$) as misleading, since the absolute EQE is extremely low compared to other Si emitters. \\[3pt]
P81 $\cdot$ Gemini 3.0 $\cdot$ paper-level slot 1 $\cdot$ secondary & AI Reviewer 1 was the only reviewer to question about the name ''LED''. It focused on how the light emission mechanism is different from traditional LEDs while other reviewers simply accepted the term. Second, while other reviewers also raised EQE concerns, it uniquely criticized the author's choice of the benchmark itself pointing out that the spatial intensity argument is misleading since the EQE is very small compared to other works. \\[3pt]
P83 $\cdot$ Gemini 3.0 $\cdot$ paper-level slot 1 $\cdot$ primary & - Only review to flag that 50 bending cycles at 5 mm radius is insufficient to justify the ''ultraflexible''. \\[3pt]
P83 $\cdot$ Gemini 3.0 $\cdot$ paper-level slot 1 $\cdot$ secondary & AI reviewer 1 focused on the mechanical durability compared to the standards in the field. \\[3pt]
P84 $\cdot$ GPT-5.2 $\cdot$ paper-level slot 2 $\cdot$ primary & - Only review to catch EM simulations used air/copper properties instead of tissue-equivalent properties. - Mentioned about insufficient strain sensor calibration based on fractional shortening metric. \\[3pt]
P84 $\cdot$ GPT-5.2 $\cdot$ paper-level slot 2 $\cdot$ secondary & AI reviewer 2 points out the insufficient calibration on the strain sensors, which makes cardiac contractility not convincing. \\[3pt]
\end{longtable}
}

\subsubsection*{S4: Internal consistency across sections --- n = 15}

{\footnotesize
\begin{longtable}{@{}p{3.2cm}p{11.5cm}@{}}
\toprule
\textbf{Citation} & \textbf{Expert comment} \\
\midrule
\endfirsthead
\toprule
\textbf{Citation} & \textbf{Expert comment} \\
\midrule
\endhead
\bottomrule
\endlastfoot
P1 $\cdot$ Gemini 3.0 $\cdot$ item 2 $\cdot$ primary & The point is that the goal may be to get closer to this accuracy and not to outperform it but at a reduced computational cost. The real critique here is that the authors don't clearly demonstrate how more and whether their approach is more computationally efficient. \\[3pt]
P37 $\cdot$ Claude 4.5 $\cdot$ item 1 $\cdot$ secondary & I wholeheartedly agree with this comment. This alone is enough to reject the paper, as the core claim doesn't match the empirical results. \\[3pt]
P50 $\cdot$ Claude 4.5 $\cdot$ item 3 $\cdot$ primary & Mistakes in writing - they switched the numbers. Numbers are correctly shown in Fig. 4f. \\[3pt]
P60 $\cdot$ Claude 4.5 $\cdot$ item 4 $\cdot$ secondary & This is a factual correction, as also noted by human reviewer 1 who caught a different labelling error in Item 4, this paper have slight proof reading issues. \\[3pt]
P60 $\cdot$ GPT-5.2 $\cdot$ item 3 $\cdot$ secondary & AI Reviewer 1's Item 5 noted the comparison is logically trivial (adding COVID deaths guarantees an increase); this reviewer shows that even setting that aside, the comparison is technically flawed in its construction. The inconsistent event definitions across periods is a particularly important point that no other reviewer raised. \\[3pt]
P61 $\cdot$ GPT-5.2 $\cdot$ item 5 $\cdot$ secondary & This is one of the most damaging critiques across all reviews because it challenges the paper's internal consistency rather than requiring external knowledge or judgment. A reader can verify every claim in this item solely from the manuscript. \\[3pt]
P65 $\cdot$ GPT-5.2 $\cdot$ item 2 $\cdot$ primary & This was one of my concerns regarding the original manuscript. They design for the 8-12 um range, but they their laser covers the 8.23 - 10.93 [non-Latin]m range. \\[3pt]
P37 $\cdot$ Claude 4.5 $\cdot$ paper-level slot 2 $\cdot$ secondary & AI Reviewer 2 was the best in terms of analyzing what the authors claimed, and if the experiment design/results support the claims. \\[3pt]
P41 $\cdot$ Claude 4.5 $\cdot$ paper-level slot 3 $\cdot$ primary & AI Reviewer 3 caught the temporal mismatch issue. \\[3pt]
P41 $\cdot$ Gemini 3.0 $\cdot$ paper-level slot 1 $\cdot$ primary & AI Reviewer 1 caught the inconsistent projection methodology. \\[3pt]
P50 $\cdot$ Claude 4.5 $\cdot$ paper-level slot 3 $\cdot$ primary & Yes, AI Reviewer 3 found about mistakes in writing where authors reported numbers in wrong order and thus physiologically impossible number was reported. \\[3pt]
P50 $\cdot$ Claude 4.5 $\cdot$ paper-level slot 3 $\cdot$ secondary & AI Reviewer 3 correctly pointed out a discrepancy in the reported heart rate value - which was physiologically impossible for a healthy adult. \\[3pt]
P67 $\cdot$ GPT-5.2 $\cdot$ paper-level slot 2 $\cdot$ primary & Inconsistency in the number of epochs. \\[3pt]
P83 $\cdot$ GPT-5.2 $\cdot$ paper-level slot 3 $\cdot$ primary & - Uniquely caught that the abstract claims ''no passivation'' while Figure 2f's caption explicitly states the ultraflexible device used ''parylene encapsulation'' \\[3pt]
P83 $\cdot$ GPT-5.2 $\cdot$ paper-level slot 3 $\cdot$ secondary & AI reviewer 3 was the only reviewer to question about the encapsulation contradiction of ''no passivation'' and ''parylene encapsuation''. \\[3pt]
\end{longtable}
}

\subsubsection*{S5: Reproducibility / dependency failures --- n = 10}

{\footnotesize
\begin{longtable}{@{}p{3.2cm}p{11.5cm}@{}}
\toprule
\textbf{Citation} & \textbf{Expert comment} \\
\midrule
\endfirsthead
\toprule
\textbf{Citation} & \textbf{Expert comment} \\
\midrule
\endhead
\bottomrule
\endlastfoot
P54 $\cdot$ GPT-5.2 $\cdot$ item 4 $\cdot$ primary & significant point well supported that applies for reproducibility \\[3pt]
P66 $\cdot$ Gemini 3.0 $\cdot$ item 2 $\cdot$ primary & They do not provide the exact measurement results in their code, but one could extract the average data from the provided figures. \\[3pt]
P85 $\cdot$ GPT-5.2 $\cdot$ item 5 $\cdot$ primary & It is likely that upon further review the journal would require the deposition of these materials. \\[3pt]
P13 $\cdot$ Claude 4.5 $\cdot$ paper-level slot 2 $\cdot$ primary & Commented on the code availability; demographic mismatch \\[3pt]
P13 $\cdot$ GPT-5.2 $\cdot$ paper-level slot 3 $\cdot$ primary & Commented on the code availability; demographic mismatch \\[3pt]
P13 $\cdot$ Gemini 3.0 $\cdot$ paper-level slot 1 $\cdot$ primary & Commented on the code availability \\[3pt]
P42 $\cdot$ GPT-5.2 $\cdot$ paper-level slot 3 $\cdot$ primary & AI reviewer 3 mostly found reproducibility issues, which while valid, likely just require tidying and are highly unlikely to reflect scientific validity. \\[3pt]
P47 $\cdot$ GPT-5.2 $\cdot$ paper-level slot 2 $\cdot$ primary & AI \#2 identified the reproducibility failures in the experimental pipeline such as missing parameter specifications and modified external tools. \\[3pt]
P63 $\cdot$ GPT-5.2 $\cdot$ paper-level slot 3 $\cdot$ primary & Reproduccability item has not been raised by other AI and human reviewers. \\[3pt]
P68 $\cdot$ GPT-5.2 $\cdot$ paper-level slot 3 $\cdot$ primary & The code reproducibility issue was raised by AI reviewer 3 and not by others. \\[3pt]
\end{longtable}
}

\subsubsection*{S6: Big-picture / counter-narrative synthesis --- n = 7}

{\footnotesize
\begin{longtable}{@{}p{3.2cm}p{11.5cm}@{}}
\toprule
\textbf{Citation} & \textbf{Expert comment} \\
\midrule
\endfirsthead
\toprule
\textbf{Citation} & \textbf{Expert comment} \\
\midrule
\endhead
\bottomrule
\endlastfoot
P60 $\cdot$ Gemini 3.0 $\cdot$ item 4 $\cdot$ secondary & This is one of the most insightful critiques across all reviews because it runs counter to the paper's narrative rather than supporting the claim that compound impacts are underestimated, it suggests temperature-related mortality during the pandemic may actually be overestimated due to a less vulnerable surviving population. \\[3pt]
P85 $\cdot$ GPT-5.2 $\cdot$ item 2 $\cdot$ primary & Another way of phrasing this is that the authors over-interpret the connection between chromosome organization and metabolic state. \\[3pt]
P6 $\cdot$ Gemini 3.0 $\cdot$ paper-level slot 2 $\cdot$ primary & No, but this is primarily because it did not get bogged down in details. It was very good at seeing the big picture and pinpointing the biggest problems. \\[3pt]
P26 $\cdot$ Claude 4.5 $\cdot$ paper-level slot 1 $\cdot$ primary & this provided the most nuance in the comments, though all 3 AI did good job on noticing and calling up the circularity in the work, which was missed even by the first reviewer which was quite good. \\[3pt]
P26 $\cdot$ Gemini 3.0 $\cdot$ paper-level slot 3 $\cdot$ primary & good job on noticing and calling up the circularity in the work \\[3pt]
P37 $\cdot$ Gemini 3.0 $\cdot$ paper-level slot 1 $\cdot$ secondary & AI Reviewer 1 was the only reviewer who correctly noticed: 1. The training objective was not as novel as the authors claimed. 2. Insufficient comparison with state-of-the-art baselines. \\[3pt]
P49 $\cdot$ Gemini 3.0 $\cdot$ paper-level slot 3 $\cdot$ secondary & While other reviewers also expressed concerns about using connectome and molecular maps derived from healthy adults to explain the damage patterns of brain disorders, this AI reviewer 3 went a step further, demonstrating a deeper understanding of disorder-specific pathology. The logic is highly compelling: because neurodevelopmental disorders such as ASD and ADHD emerge at an early stage of life, unlike other brain conditions, their brain architecture fundamentally deviates even more from the normative connectomes and molecular maps of a healthy adult. Therefore, since these developmental disorders are the most likely to deviate from the normative baseline, it would significantly strengthen the study if the examples could obtain patient-specific data at least for these specific conditions to validate whether their observations still hold true. \\[3pt]
\end{longtable}
}

\subsubsection*{S\_generic: Residual: AI judged Correct without specific reason --- n = 40}

{\footnotesize
\begin{longtable}{@{}p{3.2cm}p{11.5cm}@{}}
\toprule
\textbf{Citation} & \textbf{Expert comment} \\
\midrule
\endfirsthead
\toprule
\textbf{Citation} & \textbf{Expert comment} \\
\midrule
\endhead
\bottomrule
\endlastfoot
P1 $\cdot$ Gemini 3.0 $\cdot$ item 3 $\cdot$ primary & I would also willing to see how their method performs with a larger cell. \\[3pt]
P6 $\cdot$ Claude 4.5 $\cdot$ item 3 $\cdot$ primary & This is well-argued and correct! \\[3pt]
P6 $\cdot$ Gemini 3.0 $\cdot$ item 2 $\cdot$ primary & Good point and well-argued! \\[3pt]
P7 $\cdot$ Gemini 3.0 $\cdot$ item 5 $\cdot$ primary & Yes this was mentioned previously as well, the paper should be rephrased a bit on this point, however the result is still valuable \\[3pt]
P9 $\cdot$ GPT-5.2 $\cdot$ item 3 $\cdot$ primary & What is being measured in Lb -\textgreater{} RR is not enough for an observation therefore the text is fine. \\[3pt]
P24 $\cdot$ GPT-5.2 $\cdot$ item 1 $\cdot$ primary & it is worth discussing it, I wouldn't say it is correct or incorrect. \\[3pt]
P25 $\cdot$ Gemini 3.0 $\cdot$ item 2 $\cdot$ primary & More references are needed. Roberts et al Nat Commun 2019 for example. \\[3pt]
P29 $\cdot$ Claude 4.5 $\cdot$ item 3 $\cdot$ primary & This is a good point. \\[3pt]
P29 $\cdot$ Claude 4.5 $\cdot$ item 5 $\cdot$ primary & Very good point. very well stated. \\[3pt]
P29 $\cdot$ Gemini 3.0 $\cdot$ item 4 $\cdot$ primary & This is quite good critique. \\[3pt]
P31 $\cdot$ Gemini 3.0 $\cdot$ item 2 $\cdot$ primary & Great point. \\[3pt]
P31 $\cdot$ Gemini 3.0 $\cdot$ item 4 $\cdot$ primary & Great point. \\[3pt]
P37 $\cdot$ Gemini 3.0 $\cdot$ item 3 $\cdot$ secondary & I wrote the paper ''Doctor AI'', and it is true what the AI reviewer is claiming here. \\[3pt]
P42 $\cdot$ Claude 4.5 $\cdot$ item 4 $\cdot$ primary & The AI is correct, while difficult to address methodologically, this is something that should be addressed in the discussion. \\[3pt]
P49 $\cdot$ GPT-5.2 $\cdot$ item 1 $\cdot$ secondary & By examining the materials used by authors more thoroughly than human reviewers, this reviewer was able to point out very important points. \\[3pt]
P53 $\cdot$ GPT-5.2 $\cdot$ item 1 $\cdot$ primary & very well supported observation \\[3pt]
P54 $\cdot$ Claude 4.5 $\cdot$ item 4 $\cdot$ primary & it catches detailed text that the authors should consider rewriting for clarity. \\[3pt]
P55 $\cdot$ Claude 4.5 $\cdot$ item 4 $\cdot$ primary & it is OK \\[3pt]
P61 $\cdot$ Claude 4.5 $\cdot$ item 1 $\cdot$ secondary & This is a major issue which none of the Human reviewers pointed but at the same time they all did emphasize more clarity on this figure especially. \\[3pt]
P61 $\cdot$ Gemini 3.0 $\cdot$ item 3 $\cdot$ secondary & This connects to Human Reviewer 1 Item 13 about Wang et al.: multiple prior studies did multi-factor analysis, and the paper should engage with them honestly rather than dismissing the entire prior literature. \\[3pt]
P62 $\cdot$ Claude 4.5 $\cdot$ item 1 $\cdot$ secondary & Addressing the same key point as Human reviewer \#2 and \#3, but the distinction between ex vivo and in vivo experiments were not detailed as in the comments from the Human reviewers. \\[3pt]
P64 $\cdot$ Gemini 3.0 $\cdot$ item 2 $\cdot$ primary & Good catch as AI reviewer 2 \\[3pt]
P65 $\cdot$ Claude 4.5 $\cdot$ item 3 $\cdot$ primary & I agree. A comparison of numerical vs. experimental Strehl ratios would be useful. \\[3pt]
P65 $\cdot$ Claude 4.5 $\cdot$ item 5 $\cdot$ primary & I am not sure how the authors could quantify the effect of LPA \\[3pt]
P68 $\cdot$ Claude 4.5 $\cdot$ item 3 $\cdot$ primary & Agree and also raised by the human reviewer 3 \\[3pt]
P68 $\cdot$ Claude 4.5 $\cdot$ item 4 $\cdot$ primary & This is an important point, especially when considering sexual dimorphism, geographic variation, etc. \\[3pt]
P68 $\cdot$ Gemini 3.0 $\cdot$ item 4 $\cdot$ primary & Some of the circularity is broken by the fact that analogs are defined based on multiple characteristics (not only inner ear). But the criticism is still relevant. \\[3pt]
P73 $\cdot$ Claude 4.5 $\cdot$ item 4 $\cdot$ primary & Again, I don't think that mutations in this gene must necessarily increase cancer risk to be candidates for slight increases in microsat mutation rates, but this would be worth discussing as it does weaken the case for this mutation somewhat. \\[3pt]
P74 $\cdot$ Claude 4.5 $\cdot$ item 4 $\cdot$ primary & I agree that this would have been useful to include. \\[3pt]
P77 $\cdot$ Claude 4.5 $\cdot$ item 5 $\cdot$ primary & This is a good point that was missed by the human reviewers. \\[3pt]
P77 $\cdot$ GPT-5.2 $\cdot$ item 1 $\cdot$ primary & The proposed 8-pi electrocyclization is assumed. \\[3pt]
P77 $\cdot$ GPT-5.2 $\cdot$ item 2 $\cdot$ secondary & The term ``stereo-controlled'' appears to be misleading. \\[3pt]
P85 $\cdot$ GPT-5.2 $\cdot$ item 1 $\cdot$ primary & While there are other samples C4 and C7 that are late time points in which data were generated with the same library kit, this is a valid concern. \\[3pt]
P85 $\cdot$ GPT-5.2 $\cdot$ item 4 $\cdot$ primary & I am not an expert on the analysis of these data, but this critique seems valid to me. \\[3pt]
P85 $\cdot$ Gemini 3.0 $\cdot$ item 1 $\cdot$ primary & I am not an expert in this analysis procedure, but this critique seems reasonable. \\[3pt]
P85 $\cdot$ Gemini 3.0 $\cdot$ item 2 $\cdot$ primary & I am not an expert in this analysis procedure, but this critique seems reasonable. \\[3pt]
P85 $\cdot$ Gemini 3.0 $\cdot$ item 3 $\cdot$ primary & I am not an expert in this analysis procedure, but this critique seems reasonable. \\[3pt]
P5 $\cdot$ Claude 4.5 $\cdot$ paper-level slot 3 $\cdot$ primary & Similar to my comments regarding the second AI reviewer, though the third AI reviewer provides stronger arguments. \\[3pt]
P35 $\cdot$ GPT-5.2 $\cdot$ paper-level slot 2 $\cdot$ secondary & Yes, All the Items. \\[3pt]
P67 $\cdot$ GPT-5.2 $\cdot$ paper-level slot 2 $\cdot$ secondary & I would say that the review of this AI is better than all the human reviews but not as good as the other AI reviews. Overall I find that the AI reviews are very good at judging the technical matters of the paper and human reviews have points that can only come from years of experience although not as thorough as the AI reviews. \\[3pt]
\end{longtable}
}

\clearpage

\section{Details of the AI meta-reviewer and \textsc{PeerReview Bench}}\label{appendix:peerreview_bench}
 
This appendix documents the AI meta-reviewer that supplies precision judgments for \textsc{PeerReview Bench}, the calibration analysis that established its agreement with human experts, the construction of \textsc{PeerReview Bench} itself, the full leaderboard, and the verbatim prompt used for all meta-reviewer runs.

\subsection{The dual-annotated calibration set}\label{appendix:calibration_set}
 
\paragraph{Construction.} The calibration set is built from the 27 papers in our expert-annotation study for which two domain scientists independently annotated every review item. This yields 908 dual-annotated review items (568 human, 340 AI) on which both annotators provided three-axis judgments (correctness, significance, evidence sufficiency) under the cascade protocol described in \autoref{sec:preliminaries}. The calibration set supports two evaluation settings.
 
\begin{wraptable}{r}{0.465\textwidth}
\centering
\fontsize{7}{9}\selectfont
\setlength{\tabcolsep}{4pt}
\begin{tabular}{lr}
\toprule
\textbf{Statistic} & \textbf{Value} \\
\midrule
Papers & 27 \\
Total review items & 908 \\
\quad Human items & 568 \\
\quad AI items & 340 \\
\midrule
\emph{Primary setting} ground truth labels (cascade applied) & \\
\quad (Axis 1) Correct & 743 \\
\quad (Axis 1) Not Correct & 36 \\
\quad (Axis 2) Significant & 299 \\
\quad (Axis 2) Marginally Significant & 91 \\
\quad (Axis 2) Not Significant & 55 \\
\quad (Axis 3) Evidence Sufficient & 351 \\
\quad (Axis 3) Evidence Not Sufficient & 6 \\
\midrule
\emph{Secondary setting} ground truth labels & \\
\quad 1.\ Correct + Sig.\ + Evi. Sufficient        & 277 (30.5\%) \\
\quad 2.\ Correct + Sig.\ + Evi. Not Suff.     & 2 (0.2\%) \\
\quad 3.\ Correct + Sig.\ + Evi.\ Disagree    & 20 (2.2\%) \\
\quad 4.\ Correct + Marg. Sig.\ + Evi. Sufficient       & 74 (8.1\%) \\
\quad 5.\ Correct + Marg. Sig.\ + Evi. Not Suff.    & 4 (0.4\%) \\
\quad 6.\ Correct + Marg. Sig.\ + Evi.\ Disagree   & 13 (1.4\%) \\
\quad 7.\ Correct + Not Significant           & 55 (6.1\%) \\
\quad 8.\ Correct + Sig.\ Disagree            & 298 (32.8\%) \\
\quad 9.\ Not Correct                    & 36 (4.0\%) \\
\quad 10.\ Disagree on Correctness            & 129 (14.2\%) \\
\bottomrule
\end{tabular}
\caption{\textbf{Calibration set statistics.} Each of 908 review items from 27 dual-annotated papers carries a 10-class ground truth label encoding both the cascade outcome (correctness $\to$ significance $\to$ evidence) and inter-annotator agreement. ``Sig.'' = both significant; ``marg.'' = both marginally significant.}
\label{tab:stats-metareview-bench}
\vspace{-15pt}
\end{wraptable}
 
\noindent The \emph{primary setting} is per-axis judgment, in which the AI meta-reviewer produces its own correctness, significance, and evidence labels for each item, exactly as the human meta-reviewers in our study did, and is scored against a single expert annotator. The \emph{secondary setting} additionally asks the AI meta-reviewer to predict how the two expert annotators would jointly judge the item, which we encode as a ten-class label that captures both the cascade outcome and inter-annotator agreement: classes 1--7 cover the cases where the two annotators agree on correctness and significance and either agree or disagree on evidence; class 8 covers correctness agreement with significance disagreement; class 9 is agreed-on-incorrect; class 10 is correctness disagreement. The ten-class distribution is shown in \autoref{tab:stats-metareview-bench}, and the top-three classes are ``both correct, both significant, but disagree on significance'' (32.8\%), ``both correct, both significant, evidence sufficient'' (30.5\%), and ``disagree on correctness'' (14.2\%).
 
\paragraph{Evaluation protocol.} The calibration set evaluates an AI meta-reviewer on two paired tasks. In the \emph{primary setting}, the AI meta-reviewer produces its own three-axis judgment (correctness, significance, evidence sufficiency) for each item, exactly as the human meta-reviewers in our study did, and is scored against a single expert annotator using percent agreement as the main metric. Per-axis agreement is reported for correctness (computed on the 779 items where both annotators agree on correctness), significance (445 items where both annotators agree through significance), and evidence (357 items where both agree through evidence sufficiency). To assess whether AI--human agreement matches the empirical ceiling set by inter-human agreement, we additionally report Gwet's AC1 for the Human--Human, Human--AI, and AI--AI pairwise comparisons; AC1 is robust to the prevalence-induced kappa paradox that affects Cohen's $\kappa$ on these heavily skewed label distributions. In the \emph{secondary setting}, the AI meta-reviewer additionally predicts how the two expert annotators would jointly judge the item, expressed as one of the ten classes defined above; we report ten-class accuracy on all 908 items.
 
\paragraph{AI meta-reviewer setup.} A single review item often anchors its claim in a specific figure, source-code file, or supplementary section, so the meta-reviewer needs the same access to the paper that a human meta-reviewer (the expert scientist in our annotation study) would. We implement the meta-reviewer as an agent on the OpenHands framework with terminal, file-editor, and web-search tools that let it open the preprint markdown, view referenced figures, read submitted source code, and look up specialist terminology. The AI meta-reviewer is blocked from accessing the paper's published version (via a domain blocklist) so it cannot retrieve actual referee reports. We evaluate three frontier models as the AI meta-reviewer backbone: Claude-Opus-4.7, GPT-5.4, and Gemini-3.1-Pro. The full prompt, including the axis definitions and the ten-class decision procedure, is in \autoref{appendix:metareviewer_prompt}.

\subsection{Calibration results}\label{appendix:calibration_results}
 
\begin{table*}[t!]
\fontsize{7}{9}\selectfont
\centering
\resizebox{\textwidth}{!}{\begin{tabular}{lcccccccccccc}
\toprule
\multicolumn{1}{c}{\multirow{2}{*}{\textbf{Meta-Reviewer}}}
& \multicolumn{3}{c}{\textsc{All Reviews}}
& \multicolumn{3}{c}{\textsc{Human Reviews}}
& \multicolumn{3}{c}{\textsc{AI Reviews}}
& \multicolumn{3}{c}{\textsc{Self Review}} \\
\cmidrule(lr){2-4} \cmidrule(lr){5-7} \cmidrule(lr){8-10} \cmidrule(lr){11-13}
& Corr. & Sig. & Evid. & Corr. & Sig. & Evid. & Corr. & Sig. & Evid. & Corr. & Sig. & Evid. \\
\midrule
\textsc{Random Baseline} & 50.0\% & 33.3\% & 50.0\% & 50.0\% & 33.3\% & 50.0\% & 50.0\% & 33.3\% & 50.0\% & 50.0\% & 33.3\% & 50.0\% \\
\textsc{Human (secondary annotator)} & 85.8\% & 59.9\% & 88.0\% & 84.7\% & 60.2\% & 83.9\% & 87.6\% & 59.3\% & 94.3\% & --- & --- & --- \\
\midrule
\textsc{Claude-Opus-4.7} & 87.9\% & 56.7\% & 85.6\% & 87.7\% & 55.8\% & 79.4\% & 88.2\% & 58.2\% & 94.9\% & 85.7\% & 55.4\% & 97.6\% \\
\textsc{GPT-5.4} & 82.0\% & 56.9\% & 85.3\% & 80.0\% & 55.3\% & 80.7\% & 85.3\% & 59.4\% & 91.8\% & 90.0\% & 52.5\% & 93.1\% \\
\textsc{Gemini-3.1-Pro} & 81.5\% & 54.0\% & 87.4\% & 79.4\% & 51.8\% & 82.6\% & 84.7\% & 57.3\% & 94.2\% & 82.9\% & 65.5\% & 90.9\% \\
\bottomrule
\end{tabular}}
\caption{\textbf{Calibration set results: per-axis accuracy (primary setting).} Each row is a meta-reviewer model; columns show per-axis accuracy across four review subsets: all items, human review items only, AI review items only, and ``self review'' where the meta-reviewer judges items from its own model family (e.g., GPT-5.4 judging GPT-5.2 reviews). Random baseline assumes uniform prediction over label categories (2 for correctness/evidence, 3 for significance). The human baseline treats the secondary expert annotator as the ``meta-reviewer'' and the primary as ground truth, representing the ceiling of human inter-annotator agreement.}
\label{tab:metareview-accuracy}
\end{table*}
 
\paragraph{AI meta-reviewers closely approach the human inter-annotator agreement on all three axes.} \autoref{tab:metareview-accuracy} reports the per-axis accuracy of the three meta-reviewers in the primary setting, scored against the primary annotator. Claude-Opus-4.7 reaches 87.9\% correctness, slightly above the 85.8\% primary--secondary baseline; GPT-5.4 and Gemini-3.1-Pro reach 82.0\% and 81.5\%, a few points below. On significance, the three AI meta-reviewers reach 56.7\%, 56.9\%, and 54.0\%, against a baseline of 59.9\%; on evidence, they reach 85.6\%, 85.3\%, and 87.4\%, against a baseline of 88.0\%. Notably, significance is the hardest axis for both humans and AI meta-reviewers, and the modest 59.9\% primary--secondary baseline reflects this directly: domain scientists themselves often disagree on whether a given critique materially improves a paper, so the upper bound for any meta-reviewer on this axis is correspondingly limited. Against this backdrop, the AI meta-reviewers land in the same regime as the humans, which is the empirical basis for using an AI meta-reviewer as the precision judge in \textsc{PeerReview Bench}.
 
\paragraph{Self-review enhancement bias is uneven across axes.} When the GPT-5.4 meta-reviewer judges items written by GPT-5.2 reviewers (its same-family ``self review'' setting), per-axis accuracy is 90.0\% / 52.5\% / 93.1\%, compared to 85.3\% / 59.4\% / 91.8\% on AI items overall: correctness and evidence rise, but significance falls. Claude-Opus-4.7 (85.7\% / 55.4\% / 97.6\% vs.\ 88.2\% / 58.2\% / 94.9\%) shows the opposite asymmetry, with evidence rising but correctness and significance falling. Gemini-3.1-Pro (82.9\% / 65.5\% / 90.9\% vs.\ 84.7\% / 57.3\% / 94.2\%) shows yet another pattern, with significance rising but correctness and evidence falling. The self-enhancement bias well-known for LLM judges~\citep{wataoka2024selfpreference} is therefore present in AI meta-reviewers, but does not act uniformly across axes; on every model, at least one axis goes down rather than up under same-family judgment. The implication for \textsc{PeerReview Bench} is that precision scores derived from a single meta-reviewer should be read with caution when the candidate AI reviewer is from the same family as the meta-reviewer.
 
\begin{table}[t]
\centering
\fontsize{7}{9}\selectfont
\setlength{\tabcolsep}{5pt}
\begin{tabular}{l cc cc cc c}
\toprule
\multicolumn{1}{c}{\multirow{2}{*}{\textbf{Comparison}}}
& \multicolumn{2}{c}{\textbf{Correctness}}
& \multicolumn{2}{c}{\textbf{Significance}}
& \multicolumn{2}{c}{\textbf{Evidence}}
& \multirow{2}{*}{$N$} \\
\cmidrule(lr){2-3} \cmidrule(lr){4-5} \cmidrule(lr){6-7}
& Agr. & AC1 & Agr. & AC1 & Agr. & AC1 & \\
\midrule
\multicolumn{8}{l}{\textit{Human--Human}} \\
\quad Primary vs Secondary & 85.8\% & 0.82 & 59.9\% & 0.44 & 88.0\% & 0.86 & 908 \\
\midrule
\multicolumn{8}{l}{\textit{Human (Primary) vs AI Meta-Reviewer}} \\
\quad vs Claude Opus 4.7 & 87.9\% & 0.86 & 56.7\% & 0.43 & 85.6\% & 0.83 & 908 \\
\quad vs GPT-5.4 & 82.0\% & 0.78 & 56.9\% & 0.44 & 85.3\% & 0.83 & 908 \\
\quad vs Gemini 3.1 Pro & 81.5\% & 0.77 & 54.0\% & 0.44 & 87.4\% & 0.86 & 908 \\
\midrule
\multicolumn{8}{l}{\textit{Human (Secondary) vs AI Meta-Reviewer}} \\
\quad vs Claude Opus 4.7 & 85.3\% & 0.82 & 62.3\% & 0.52 & 92.1\% & 0.91 & 908 \\
\quad vs GPT-5.4 & 82.4\% & 0.77 & 64.5\% & 0.55 & 91.1\% & 0.90 & 908 \\
\quad vs Gemini 3.1 Pro & 82.9\% & 0.78 & 61.6\% & 0.54 & 93.3\% & 0.93 & 908 \\
\midrule
\multicolumn{8}{l}{\textit{AI vs AI}} \\
\quad GPT-5.4 vs Gemini 3.1 Pro & 88.8\% & 0.86 & 85.3\% & 0.83 & 91.2\% & 0.91 & 908 \\
\quad GPT-5.4 vs Claude Opus 4.7 & 91.0\% & 0.89 & 89.3\% & 0.88 & 90.9\% & 0.90 & 908 \\
\quad Gemini 3.1 Pro vs Claude Opus 4.7 & 92.0\% & 0.91 & 86.1\% & 0.84 & 93.3\% & 0.93 & 908 \\
\bottomrule
\end{tabular}
\caption{\textbf{Pairwise agreement on the 908-item calibration set.} Each cell reports percent agreement and Gwet's AC1 between two judges. Human--Human: primary vs secondary expert annotator. Human--AI: each AI meta-reviewer vs each human annotator. AI--AI: pairs of AI meta-reviewers. AC1 is robust to class imbalance and avoids the ``kappa paradox'' where high agreement produces low $\kappa$ when one category dominates. AI--AI agreement substantially exceeds Human--AI agreement, especially on significance (AC1 = 0.83--0.88 AI--AI vs 0.43--0.55 Human--AI), indicating that AI meta-reviewers converge on a shared judgment style.}
\label{tab:metareview-agreement}
\end{table}
 
\paragraph{AI meta-reviewers agree with each other more than with humans.} For a finer-grained view of the agreement structure, \autoref{tab:metareview-agreement} reports pairwise agreement among all meta-reviewer pairs (Human--Human, Human--AI, AI--AI). The three AI meta-reviewers converge on a shared judgment style that is markedly more consistent than any individual AI's agreement with a human. On significance, the three pairwise AI--AI comparisons reach AC1 between 0.83 and 0.88, while AI--human AC1 ranges from 0.43 to 0.55 -- only modestly above the 0.44 human--human baseline. The same pattern holds on evidence (AI--AI 0.90--0.93 vs AI--human 0.83--0.93) and, less dramatically, on correctness (AI--AI 0.86--0.91 vs AI--human 0.77--0.86). The three AI meta-reviewers therefore assign similar significance and evidence labels to similar items even when those labels differ from what a human expert would assign, suggesting the AI meta-reviewers have settled on a shared standard that does not fully match the human one. We do not have direct evidence on whether this shared standard is closer to or further from a hypothetical ``true'' standard than the human one.
 
\begin{wraptable}{r}{0.45\textwidth}
\centering
\vspace{-10pt}
\fontsize{8}{10.5}\selectfont
\setlength{\tabcolsep}{5pt}
\begin{tabular}{l cccc}
\toprule
\textbf{Meta-Reviewer} & \textsc{All} & \textsc{Human} & \textsc{AI} & \textsc{Self} \\
\midrule
\textsc{Random Baseline} & 10.0\% & 10.0\% & 10.0\% & 10.0\% \\
\midrule
\textsc{Claude-Opus-4.7} & 44.3\% & 40.0\% & 51.2\% & 47.9\% \\
\textsc{GPT-5.4} & 30.1\% & 31.2\% & 28.2\% & 29.1\% \\
\textsc{Gemini-3.1-Pro} & 29.0\% & 26.1\% & 33.8\% & 39.6\% \\
\bottomrule
\end{tabular}
\caption{\textbf{Calibration set results: ten-class accuracy (secondary setting).} The 10-class label encodes both the annotation cascade outcome (correctness $\to$ significance $\to$ evidence) and inter-annotator agreement/disagreement. Columns show accuracy across four review subsets: all items, human-written items, AI-written items, and ``self review'' where the meta-reviewer judges items from its own model family.}
\label{tab:metareview-uncertainty-accuracy}
\vspace{-10pt}
\end{wraptable}
 
\paragraph{Predicting where experts will disagree is a much harder task.} Per-axis accuracy is a natural stress test on each individual judgment, but the ten-class label adds a stricter requirement: the meta-reviewer must predict not only what the cascade outcome is, but also where in the cascade two human experts would and would not converge. \autoref{tab:metareview-uncertainty-accuracy} shows this is a substantially harder task. Claude-Opus-4.7 reaches 44.3\% ten-class accuracy across all items, well ahead of GPT-5.4 (30.1\%) and Gemini-3.1-Pro (29.0\%) and far above the 10\% random baseline, but still well short of perfect prediction of the joint cascade-and-agreement structure. The gap between Claude-Opus-4.7 and the other two frontier models is roughly fourteen percentage points, considerably larger than the gap on per-axis accuracy. We do not include a human ten-class baseline because the human annotators were not asked to predict the other annotator's labels; a ``human ceiling'' is therefore not directly defined for this metric.

\subsection{Failure analysis of the AI meta-reviewer}\label{appendix:metareviewer_failure_analysis}
 
This subsection details the recurring failure patterns we identify in the GPT-5.4 meta-reviewer's disagreements with human annotators on the calibration set. Cross-model failure analysis (covering Claude-Opus-4.7 and Gemini-3.1-Pro) is left to future work.
 
\subsubsection{Methodology and overview}\label{appendix:metareviewer_failure_methodology}
 
We manually reviewed every disagreement between the GPT-5.4 AI meta-reviewer's labels and the primary expert annotator's labels on the 908-item calibration set. The error totals at the heart of the analysis are: 41 false negatives on correctness (the AI meta-reviewer labelled an item ``Not Correct'' when both expert annotators labelled it ``Correct''), 13 false positives on correctness (the AI meta-reviewer labelled an item ``Correct'' when both experts labelled it ``Not Correct''), 56 significance miscalibrations on items the AI meta-reviewer correctly identified as valid review points, and 13 evidence judgment errors. For the secondary setting (ten-class prediction), we additionally sampled 50 cases where the AI meta-reviewer confidently predicted expert agreement but experts in fact disagreed, and 30 cases where the AI meta-reviewer predicted expert disagreement but experts in fact converged. The seven failure patterns we report below were identified bottom-up from these errors and are not mutually exclusive: the partial-evidence trap (\autoref{appendix:metareviewer_failure_partialevidence}) and reviewer-type asymmetry (\autoref{appendix:metareviewer_failure_reviewertype}) interact, for example, since the AI meta-reviewer is more likely to rely on partial evidence to dismiss human-written items than AI-written ones.
 
\subsubsection{The partial-evidence trap}\label{appendix:metareviewer_failure_partialevidence}
 
The single largest source of correctness errors (41 of 54 total correctness errors) is what we call the partial-evidence trap. The AI meta-reviewer finds some element in the paper that partially addresses a reviewer's concern, then concludes the entire concern is invalid. Experts, by contrast, recognise that partial coverage does not fully resolve the underlying issue.
 
\paragraph{Example 1: missing $n$-values in statistical reporting (Paper 27, Human reviewer 2).} The reviewer wrote: ``The 'n' used for each analysis should be presented somewhere (either in the Results text or figure legends, but in a consistent manner). Can the authors confirm in the manuscript that the linear mixed models used incorporate patients as a grouping variable...'' The AI meta-reviewer reasoned: ``Not Correct overall: the manuscript usually reports $n$ directly in the Results text, and the mixed-model methods already state that the intercept was specific to each subject, which addresses the patient-grouping concern.'' Both expert annotators rated this Correct and Marginally Significant. The reviewer's explicit emphasis on consistency (``in a consistent manner'') was the core request, and the paper's inconsistent placement of $n$-values across results and figure legends remained a valid concern. The AI meta-reviewer treated partial coverage as full refutation and missed the consistency requirement entirely.
 
\paragraph{Example 2: ablation study absence (Paper 35, Human reviewer 2).} The reviewer wrote: ``The claims are not well proved: Section 2.2 and 2.3 respectively claims leveraging intra/inter molecular information for improved predictions [...] However, in both sections, I didn't see any ablation study, but just the performance of the proposed MUSE.'' The AI meta-reviewer reasoned that the paper does include ablation-style comparisons (the MUSE-Joint comparisons in Figures 2c and 3c), so the literal claim that there is no ablation is inaccurate. Both expert annotators rated this Correct and Significant: a comparison between MUSE variants is not the same as a systematic ablation that isolates the contribution of each information type. The AI meta-reviewer again attended to the literal wording (``no ablation'') rather than the substantive concern (causal claims about leveraging intra/inter molecular information lack proper ablation support).
 
\paragraph{Example 3: code-based evidence of sampling-rate contradiction (Paper 50, Gemini reviewer).} The reviewer wrote: ``The paper claims a sampling frequency of 800~Hz and analyzes frequencies up to 400~Hz, but the provided code limits the sampling rate to approximately 2~Hz, making the reported results impossible to reproduce with the described setup.'' The AI meta-reviewer reasoned that the claimed contradiction depends on treating the released firmware (\texttt{sketch\_may06a.ino}) as the actual firmware used for the paper's high-frequency experiments, and the paper does not establish that link. Both expert annotators rated the item Correct, Significant, and Sufficient. The reviewer's point was that the \emph{provided} code, the only code available for verification, contained a fundamental contradiction with the claimed sampling rate. The AI meta-reviewer introduced reasonable doubt about which code file was authoritative rather than evaluating the reproducibility concern against what was actually provided.
 
\paragraph{What a better meta-reviewer would do.} Across these examples, the AI meta-reviewer's failure mode is to treat the reviewer's main claim as a single binary check rather than enumerating the sub-claims it contains. A meta-reviewer that explicitly decomposed multi-part review items into atomic sub-claims and evaluated each independently would catch the consistency, depth, and verifiability concerns that the AI meta-reviewer's holistic check misses.
 
\subsubsection{Over-leniency on technically detailed items}\label{appendix:metareviewer_failure_overleniency}
 
In 13 cases, the AI meta-reviewer accepted a review item as ``Correct'' when both experts rejected it. These items share a common pattern: precise-sounding technical language that the AI meta-reviewer finds persuasive, even when the underlying claim does not hold up against the paper.
 
\paragraph{Example 1: numerical inconsistencies that do not exist (Paper 61, Claude reviewer).} The review item was about CMIP6 model bias for PM2.5 in highly polluted regions. The AI meta-reviewer reasoned: ``Correct because the text reports mismatched aging magnitudes across sections and even an incoherent confidence interval with the lower bound above the upper bound.'' Both experts rated this Not Correct. The AI meta-reviewer's reasoning reveals the failure: it accepted the AI reviewer's technical framing about numerical inconsistencies without verifying whether those inconsistencies actually existed, and even appears to have manufactured a specific claim (``the lower bound above the upper bound'') from the reviewer's technical-sounding language.
 
\paragraph{Example 2: identifiability claim against compositional data (Paper 80, Gemini reviewer).} The reviewer wrote: ``The paper claims to estimate correlations of absolute abundances from relative abundance data without external reference, which is mathematically impossible due to the identifiability problem inherent in compositional data.'' The AI meta-reviewer labelled this Correct, persuaded by the invocation of ``the identifiability problem inherent in compositional data,'' a legitimate mathematical concept. Both experts rated this Not Correct: the paper's bias-corrected counts and methodology specifically address the concern, and the AI reviewer's invocation of identifiability obscured rather than illuminated the paper's actual approach.
 
\paragraph{Example 3: standard analytical shortcut framed as inconsistency (Paper 80, GPT reviewer).} The reviewer claimed that the paper's thresholding-based sparse correlation estimator is ``internally inconsistent and statistically unsound.'' The AI meta-reviewer agreed: ``the thresholding estimator is described in a way that is hard to reconcile internally, because the paper first states a constrained optimisation problem and then replaces it with simple soft-thresholding and a dubious self-fit tuning rule.'' Both experts rated this Not Correct. The transition from constrained optimisation to soft-thresholding is a standard analytical shortcut well-established in the sparse estimation literature; the AI reviewer's specific terminology made the criticism sound authoritative, and the AI meta-reviewer did not check whether the alleged inconsistency actually existed.
 
\paragraph{The AI-reviewer specificity effect.} Ten of the 13 errors in this category involved AI-generated review items. AI reviewers tend to produce technically specific, confidently worded critiques, and the AI meta-reviewer appears to have a lower threshold for accepting claims that come packaged with precise technical vocabulary and code-level evidence. This category interacts directly with the reviewer-type asymmetry described in \autoref{appendix:metareviewer_failure_reviewertype}.
 
\subsubsection{Significance boundary miscalibration}\label{appendix:metareviewer_failure_significance}
 
The AI meta-reviewer made 56 significance errors on items it correctly identified as valid review points. The errors split into two directions: downgrading substantive concerns to ``Marginally Significant'' or ``Not Significant'' (the more common direction), and upgrading minor concerns to ``Significant'' (the less common direction).
 
\paragraph{Pattern A, Example 1: phenotypic profile integration request (Paper 49, Human reviewer 1).} The reviewer wrote: ``Each of the disorders is associated with a phenotypic profile (symptoms + neurocognitive disturbances). The study is somewhat limited for not incorporating these.'' The AI meta-reviewer rated it Marginally Significant: ``adding phenotypic linkage would enrich interpretation, but it would not change the core analyses or main results.'' Both experts rated it Significant. Connecting brain-imaging findings to clinical phenotypes is not merely enriching but is fundamental to the translational value of the work.
 
\paragraph{Pattern A, Example 2: SAR safety reporting at 11.7T MRI (Paper 6, Gemini reviewer).} The reviewer raised concerns about safety monitoring and RF heating at unprecedented field strength. The AI meta-reviewer's reasoning followed the same formula: this would improve safety reporting but does not change the core scientific contribution. Both experts rated this Marginally Significant in the opposite direction in this case, but across the broader pattern the AI meta-reviewer's habitual ``does not change the core result'' formula consistently produces under-rating.
 
\paragraph{Pattern A, Example 3: figure placement suggestion (Paper 61, Human reviewer 2).} The reviewer suggested moving Figure 4 to the supplementary file. The AI meta-reviewer rated it Not Significant on the grounds that figure placement does not affect the paper's scientific substance. Both experts rated it Marginally Significant: information organisation has real impact on how readers interpret results, and a change in placement is a presentation issue that is still substantive enough to be Marginally Significant rather than Not Significant.
 
\paragraph{Pattern B, Example 4: R-squared labelling on classification tasks (Paper 33, Claude reviewer).} The reviewer noted that Table 1 reports ``R-squared'' values for binary classification tasks, which is technically the wrong metric name. The AI meta-reviewer rated this Significant because evaluation metrics are core methodological evidence. Both experts rated it Marginally Significant: the underlying performance numbers (5-fold cross-validation accuracy) are interpretable despite the mislabelling, so the issue is presentation rather than substance.
 
\paragraph{Pattern B, Example 5: Coulombic efficiency over 100\% (Paper 33, GPT reviewer).} The reviewer flagged Coulombic efficiencies exceeding 100\% as evidence of high reversibility without proper uncertainty quantification. The AI meta-reviewer rated this Significant because Coulombic efficiency is a headline performance metric. Both experts rated it Not Significant: in battery electrochemistry, CE values slightly over 100\% are well-understood measurement artifacts, making this a minor rather than significant concern.
 
\paragraph{Calibration summary.} The AI meta-reviewer's significance errors follow a consistent pattern: it applies a ``would addressing this change the core result?'' test that under-weights methodological rigour, scope qualification, and translational relevance, while over-weighting presentation issues that involve technical terminology.
 
\subsubsection{Evidence closure demand}\label{appendix:metareviewer_failure_evidence}
 
In 13 cases, the AI meta-reviewer rated evidence as ``Requires More'' when experts accepted it as ``Sufficient.'' The pattern reveals a demand for complete analytical closure: the AI meta-reviewer expects review items to contain self-contained proofs rather than pointers to verifiable gaps in the paper.
 
\paragraph{Example 1: validation sample size concern (Paper 33, Human reviewer 2).} The reviewer wrote: ``The results on 20 ILs is not enough to establish the correlation. For 1000 ILs, it is not difficult to conduct quantum calculations. I would recommend at least 20\% of the ILs on which these calculations are carried out ensuring that cation types and anions are well represented.'' The AI meta-reviewer rated evidence as Requires More because the reviewer ``does not justify why 20\% is the right benchmark or why the current sample is demonstrably inadequate.'' Both experts rated evidence Sufficient: the reviewer pointed to a specific verifiable gap (20 of 1000 ILs is a 2\% validation sample) and explicitly stated what would be needed (at least 20\% with representative sampling). The AI meta-reviewer demanded a formal statistical justification when experts accepted the pointed identification of the gap as sufficient.
 
\paragraph{Example 2: spatial autocorrelation in cross-validation (Paper 49, Claude reviewer).} The reviewer identified a methodological concern: distance-dependent cross-validation with 75/25 split based on spatial proximity may not fully address spatial autocorrelation, because test regions adjacent to the training boundary still share autocorrelation structure with training regions. The AI meta-reviewer demanded that the reviewer demonstrate the remaining dependence \emph{materially inflates} performance. Both experts rated evidence Sufficient: the reviewer identified a specific methodological gap and pointed to the methodology section where it could be verified.
 
\paragraph{Example 3: mortality displacement in COVID-19 (Paper 60, Gemini reviewer).} The reviewer flagged that the paper does not account for mortality displacement (``harvesting'') from COVID-19 affecting subsequent temperature-related mortality. The AI meta-reviewer demanded paper-specific evidence of the harvesting effect rather than the general phenomenon. Both experts rated evidence Sufficient: harvesting is a well-established epidemiological phenomenon, the temporal overlap is verifiable from the paper's methods, and the reviewer's identification of the specific absence is itself the evidence.
 
\paragraph{The closure demand pattern.} Across these errors, the AI meta-reviewer's reasoning contains phrases like ``the reviewer does not justify why...'', ``cites only the general phenomenon rather than paper-specific evidence,'' or ``requires some inference from the setup.'' The AI meta-reviewer demands that review items provide self-contained analytical arguments rather than accepting what experts accept: pointed identification of a specific, verifiable gap that a reader can check against the paper.
 
\subsubsection{Reviewer-type asymmetry}\label{appendix:metareviewer_failure_reviewertype}
 
The error data shows an asymmetric treatment of human versus AI review items. Among the 41 false negatives on correctness, 29 involved human reviewers and 12 involved AI reviewers; among the 13 false positives, 10 involved AI reviewers. Given the overall distribution of items, both directions of asymmetry are statistically meaningful.
 
\paragraph{Example (stricter on human items): vague human concern dismissed (Paper 50, Human reviewer 2).} The reviewer wrote: ``[Reproducibility] no really, as the limitation of data sizes, detailed evaluation of using the proposed sensors, should be further elaborated.'' The AI meta-reviewer rated this Not Correct because the item ``treats limited sample size as a reproducibility problem, which is not the right target, and never identifies a concrete missing method, artifact, or code element needed for reproduction.'' Both experts rated this Correct. The AI meta-reviewer penalised the human reviewer's informal language (``no really'') and vague framing, treating stylistic imprecision as a factual failing.
 
\paragraph{Example (stricter on human items): informal scope question dismissed (Paper 51, Human reviewer 2).} The reviewer wrote: ``[Figures] Several figures are not referred to in the main text, making it unclear how they support the study's claims.'' The AI meta-reviewer rated this Not Correct because all six main figures are technically cited in the manuscript text, so the literal claim does not hold. Both experts rated this Correct: the reviewer's concern was that the figures were not \emph{meaningfully discussed}, not whether they were technically cited. The AI meta-reviewer demanded literal precision from the human reviewer that it does not require from AI reviewers.
 
\paragraph{Example (more lenient on AI items): mathematical-impossibility claim accepted (Paper 80, Gemini reviewer).} The reviewer's invocation of ``the identifiability problem inherent in compositional data'' (also discussed in \autoref{appendix:metareviewer_failure_overleniency}) was accepted by the AI meta-reviewer as a valid mathematical concern. Both experts rated this Not Correct: the paper's specific bias-correction methodology addresses the concern, but the AI meta-reviewer did not check.
 
\paragraph{Example (more lenient on AI items): data-availability complaint accepted (Paper 67, GPT reviewer).} The reviewer noted that the paper does not provide sufficient access to underlying data and that the code bundle contains Git LFS pointer files instead of actual \texttt{.mat} datasets. The AI meta-reviewer accepted this as a valid reproducibility concern, finding the reference to Git LFS pointer files specifically persuasive. Both experts rated this Not Correct: the paper's data availability statement covers the access on request, and the LFS pointers are an artifact of the repository setup rather than a missing artifact.
 
\paragraph{Asymmetry mechanism.} The AI meta-reviewer applies two different standards. For human items, it demands literal precision and penalises informal language, vagueness, or imprecise scope. For AI items, it credits technical specificity and tends to accept claims that name specific technical concepts, code elements, or mathematical frameworks. The asymmetry likely reflects a prior that specificity correlates with correctness, learned from training data where most technically precise text is correct. In our data this prior misfires in both directions.
 
\subsubsection{Expert-disagreement prediction failures}\label{appendix:metareviewer_failure_disagreement}
 
In the secondary setting, the AI meta-reviewer must predict not only the cascade outcome but also where experts will agree or disagree. Two patterns recur in its prediction errors.
 
\paragraph{Misses (model confident, experts disagreed; 50 sampled cases).} The AI meta-reviewer predicted that both experts would converge on a label, but the experts in fact split. A characteristic example is Paper 12 (a high-redshift blazar study), where the AI meta-reviewer confidently rated the reviewer's jet-power calculation criticism ``incorrect'' because the paper uses a different scaling relation than the reviewer claimed. One expert agreed; the other gave the reviewer credit for the broader concern about applying low-frequency scaling relations to a high-frequency measurement. The AI meta-reviewer's narrow factual check missed the second expert's broader assessment.
 
\paragraph{False positives (model predicts disagreement, experts agreed; 30 cases).} The AI meta-reviewer over-hedged on methodological judgment calls, predicting that experts would split when in fact they converged. A representative case is Paper 37, where a reviewer challenged a 1.7\% PPV as practically useful in clinical practice. The AI meta-reviewer predicted disagreement on the grounds that the paper's scope might justify the low PPV. Both experts in fact converged on accepting the clinical-utility concern. The AI meta-reviewer's hedge over-weighted the paper's defence and under-weighted the shared domain standard.
 
\paragraph{Pattern.} The AI meta-reviewer over-confidently predicts agreement when it can identify a specific factual element in the paper that resolves part of a review item, and over-hedges when the item depends on a methodological judgment call. Experts often converge on the latter because they share domain-specific standards the AI meta-reviewer does not.
 
\subsubsection{Context anchoring and paper-mismatch blind spots}\label{appendix:metareviewer_failure_anchoring}
 
The AI meta-reviewer evaluates multiple items per paper in a single conversation, and the data shows clear anchoring effects: assessments of one item influence subsequent items from the same paper.
 
\paragraph{Template reasoning across items.} The most striking instance comes from Paper 82, where the human reviews appear to contain items about a different paper (Paper 81). Once the AI meta-reviewer formed the hypothesis ``this reviewer is evaluating the wrong paper,'' it applied nearly identical template reasoning to every subsequent item from that reviewer:
 
\begin{itemize}[leftmargin=1.2em, itemsep=2pt, topsep=2pt]
\item ``This item asks about the importance of monitoring cell adhesion [...] but the paper is about a CMOS-integrated sub-wavelength Si LED.''
\item ``This item asks about ultra-sensitive OECT biosensors [...] but the paper is about a CMOS-integrated sub-wavelength Si LED.''
\item ``This item asks about OECT gate leakage [...] but the paper is about a CMOS-integrated sub-wavelength Si LED.''
\end{itemize}
 
Even when individual items were actually about Paper 82's content (silicon heat sinking, current injection limits, measurement temperatures), the AI meta-reviewer had already locked into its incorrect frame and could not evaluate them on their own merits. Experts rated all these items Correct because they were about the LED paper.
 
\paragraph{Significance anchoring within papers.} On Paper 81, multiple AI reviewers raised single-cell replication concerns. The AI meta-reviewer rated all three Significant with nearly identical reasoning across reviewers, while experts consistently rated them Marginally Significant. The AI meta-reviewer appears to have anchored on its first significance assessment and applied it to subsequent thematically similar items rather than recalibrating each one.
 
\paragraph{Topic anchoring within papers.} On Paper 33, the AI meta-reviewer systematically rated reproducibility and methodology concerns Significant across many items from multiple reviewers, when expert consensus was Marginally Significant or Not Significant. The AI meta-reviewer appears to have anchored on a ``reproducibility is important'' frame early in its evaluation of the paper and then applied it uniformly.
 
\paragraph{Anchoring mechanism.} Per-paper context sharing is helpful for verification (the AI meta-reviewer does not need to reread the paper for each item) but creates a feedback loop in which an early hypothesis or judgment pattern propagates without re-examination. Independent per-item evaluation, or an explicit ``reset'' before each new item, would mitigate this.
 
\subsubsection{How the failure categories interact}\label{appendix:metareviewer_failure_mapping}
 
\begin{table}[t]
\centering
\fontsize{8}{10}\selectfont
\setlength{\tabcolsep}{6pt}
\begin{tabular}{l l}
\toprule
\textbf{Pattern family} & \textbf{Sub-categories} \\
\midrule
Partial verification & B.2 (partial-evidence trap) \\
Significance miscalibration & B.4 (significance boundary miscalibration) \\
Reviewer-type asymmetry & B.3 (over-leniency on technical items), B.6 (reviewer-type asymmetry) \\
Other failures & B.5, B.7, B.8 (closure demand, disagreement prediction, anchoring) \\
\bottomrule
\end{tabular}
\caption{\textbf{Grouping of the seven failure sub-categories into four interacting families.} Reviewer-type asymmetry covers two sub-categories because over-leniency on AI items and stricter judgment of human items are two faces of the same surface-form asymmetry.}
\label{tab:failure-mapping}
\end{table}
 
The seven sub-categories above interact: the partial-evidence trap is exacerbated by reviewer-type asymmetry, since the AI meta-reviewer is more willing to accept partial coverage as full refutation when the reviewer is human; over-leniency on technical items interacts with significance miscalibration, since technically specific items the AI meta-reviewer wrongly accepts as correct also tend to receive inflated significance ratings; context anchoring amplifies all other errors, since once the AI meta-reviewer locks into a frame it applies that frame to subsequent items without re-evaluation. \autoref{tab:failure-mapping} groups the sub-categories by family.

\subsection{\textsc{PeerReview Bench}: construction and evaluation protocol}\label{appendix:peerreview_bench_construction}
 
\paragraph{Evaluation protocol.} For each paper $p$, let $R(p) = \{r_1, \ldots, r_{n_p}\}$ denote the \emph{rubric}, the set of fully-positive human review items associated with $p$, and let $G(p) = \{g_1, \ldots, g_{m_p}\}$ denote the items produced by a candidate AI reviewer on $p$. ``Fully positive'' is defined exactly as in our expert-annotation study (\autoref{sec:performance}): both domain-scientist annotators rated the item correct, significant, and evidence-sufficient. We compute three per-paper metrics:
\begin{align}
\mathrm{Recall}(p)    &= \frac{\bigl|\{r \in R(p) : \exists\, g \in G(p),\ \mathrm{match}(g, r)\}\bigr|}{|R(p)|}, \\
\mathrm{Precision}(p) &= \frac{\bigl|\{g \in G(p) : \mathrm{MetaReviewer}(g, p) = \text{fully positive}\}\bigr|}{|G(p)|}, \\
\mathrm{F1}(p)        &= \frac{2 \cdot \mathrm{Precision}(p) \cdot \mathrm{Recall}(p)}{\mathrm{Precision}(p) + \mathrm{Recall}(p)},
\end{align}
and report each metric averaged across the benchmark's papers. Here, $\mathrm{match}(g, r)$ is the GPT-5.4 similarity judge of \autoref{sec:overlap} at the ``at least same criticism'' threshold, and $\mathrm{MetaReviewer}(g, p)$ is the AI meta-reviewer described in \autoref{appendix:calibration_set}, run on $p$ with $g$ supplied as the candidate item. The candidate AI reviewer runs under the same constraints as the AI reviewers in our expert-annotation study (up to five review items per paper by default; full access to the preprint, supplementary materials, and source code). Per-paper averaging treats each paper as one independent observation and avoids letting papers with larger rubrics dominate the score.
 
\paragraph{Construction.} \textsc{PeerReview Bench} comprises 78 of the 82 papers in our expert-annotation study. The remaining four are excluded because $R(p) = \emptyset$: their three human reviewers jointly produced zero items that both annotators labeled correct, significant, and evidence-sufficient, leaving no fully-positive content from which to construct a rubric. For each included paper, the rubric $R(p)$ is the union of fully-positive human review items pooled across the paper's three human reviewers. Rubric sizes $|R(p)|$ span 1 to 36 items per paper, with a mean of 10 and a median of 8, reflecting substantial variation in how much high-quality content human reviewers produced. We deliberately do not deduplicate overlapping rubric items across a paper's three human reviewers: the analysis in \autoref{sec:overlap} established that overlap between independent human reviewers is low, and when two or three human reviewers do converge on the same concern, the convergence is itself a signal that the concern is important. Leaving duplicates in means that an AI reviewer covering a repeatedly-raised concern receives correspondingly more credit, which is the intended weighting.

\subsection{Full leaderboard analysis}\label{appendix:peerreview_bench_results}
 
\paragraph{Main results.} \autoref{tab:peerreview-bench} in the main text reports recall, precision, and F1 across twelve backbone models. Claude-Opus-4.5 leads at F1 = 50.89, with Claude-Opus-4.7 a fraction of a point behind at F1 = 50.46; the two Opus models are essentially tied within the resolution of the benchmark. A second tier follows at F1 in the high-40s: DeepSeek-V4-Pro (F1 48.52), GPT-5.2 (F1 47.37). The remaining nine models cluster between F1 35 and 45, with Gemini-3.1-Pro-Preview an outlier at F1 22.57, driven primarily by a collapse in recall to 13.92\%.
 
\paragraph{Different model families specialize in different aspects of AI reviewing.} The Claude-Opus models occupy a balanced operating point on the precision-recall plane (precision 71-76\%, recall 38-39\%) and as a result lead in F1 despite not maximizing either axis. The GPT-5 family (excluding the mini variant) sits at the high-precision, low-recall corner: GPT-5.4 attains the benchmark's highest precision at 93.81\% while raising only 26.55\% of the human rubric; GPT-5.2 follows the same shape with 88.92\% precision and 32.28\% recall. The Gemini-3.0 family and several other models (Qwen3.6-Plus, Gemini-3-Flash) sit at the opposite corner with lower precision (53-60\%) and similar or higher recall (32-38\%). DeepSeek-V4-Pro and Kimi-K2.6 sit closer to the Claude-Opus balance. No single backbone dominates on both axes, and the choice of backbone is therefore a choice about which kind of error to tolerate: GPT-family models produce few items but most are well-formed, Gemini-family and similar models cast a wider net at lower per-item quality, and the Claude-Opus models split the difference.
 
\paragraph{Newer is not consistently better.} A pattern that runs against the implicit assumption of the benchmark is visible across all three families. Within Anthropic, Claude-Opus-4.5 narrowly outperforms the newer Claude-Opus-4.7 in F1 (50.89 vs 50.46, a 0.43-point gap). Within OpenAI, GPT-5.2 outperforms GPT-5.4 by nearly six points (47.37 vs 41.38), with the newer model trading recall (32.28 to 26.55) for precision (88.92 to 93.81). Within Google, Gemini-3.0-Pro-Preview outperforms Gemini-3.1-Pro-Preview by more than 21 points (44.14 vs 22.57), driven almost entirely by a collapse in recall (37.65\% to 13.92\%). For model developers who want their models to be useful as AI reviewers for scientific peer review, \textsc{PeerReview Bench} provides a concrete testbed on which targeted improvements can be measured.

\subsection{AI meta-reviewer prompt}\label{appendix:metareviewer_prompt}
 
The prompt for the AI meta-reviewer agent is below across \autoref{fig:metareviewer_prompt_part1}, \autoref{fig:metareviewer_prompt_part2}, \autoref{fig:metareviewer_prompt_part3}, and \autoref{fig:metareviewer_prompt_part4}. The same prompt is used for all three frontier-model backbones (Claude-Opus-4.7, GPT-5.4, Gemini-3.1-Pro). Placeholders of the form \texttt{\{paper\_preprint\_dir\}}, \texttt{\{paper\_reviews\_dir\}}, \texttt{\{reviewer\_list\}}, \texttt{\{paper\_id\}}, and \texttt{\{output\_file\}} are replaced per paper at runtime.
 
\begin{figure}[h]
\begin{footnotesize}
\begin{verbatim}
You are a meta-reviewer agent for scientific papers. You will be given
a paper's source files on disk AND reconstructed reviews written by
multiple reviewers (human and AI). For EVERY review item from EVERY
reviewer, you must:
 
1. Judge the item along three axes: correctness, significance, and
   evidence sufficiency (your own meta-review judgment).
2. Predict how two independent expert meta-reviewers would jointly
   judge the item, expressed as one of 10 collapsed class labels
   that encode both the cascade outcome and inter-expert agreement.
 
You are NOT writing a new review. You are judging existing review
items by verifying their claims against the paper.
 
 
### Paper location
The paper's source files are at: {paper_preprint_dir}
 
The directory structure is:
preprint/
| preprint.md        (main paper, markdown)
| images_list.json   (list of figure images with captions)
| images/            (figure files referenced by the paper)
| supplementary/     (optional supplementary materials)
| code/              (optional source code)
 
### Reviews to meta-review
The reconstructed reviews are at: {paper_reviews_dir}
Each file is one reviewer's full review with all their items.
The reviewers are: {reviewer_list}
 
 
### Principles (ordered by importance)
1. Correctness -- judge the main point, not peripheral details.
   If the reviewer's core concern is valid even though one specific
   supporting claim is inaccurate, the item is still Correct. Only
   mark Not Correct when the main point itself is wrong.
2. Significance -- the bar is "would this improve the paper?"
   Any criticism that would genuinely help the paper if addressed
   is Significant -- it does NOT need to threaten the paper's
   validity. Missing statistics, undefined figure annotations,
   unreported methodological details, internal inconsistencies
   between text and figures, and missing ablations are typically
   Significant. Only pure stylistic preferences and trivial typos
   are Marginally Significant.
3. Evidence -- verifiability, not exhaustiveness. If a meta-reviewer
   can verify the reviewer's claim from what the reviewer wrote
   plus the paper, the evidence is Sufficient. When the reviewer's
   criticism IS that something is absent, identifying the specific
   absence IS the evidence. Reserve Requires More for cases where
   the meta-reviewer cannot even locate what part of the paper the
   reviewer is talking about.
\end{verbatim}
\end{footnotesize}
\caption{Meta-reviewer prompt (Part 1 of 4): role, paper context, and the three principles that establish the bar for each axis.}
\label{fig:metareviewer_prompt_part1}
\end{figure}
 
\clearpage
 
\begin{figure}[h]
\begin{footnotesize}
\begin{verbatim}
### Decision procedure (follow for EACH item)
 
Part A -- Your initial judgment (axis labels)
Step 1 (Understand). Read the item. What is the main point?
Step 2 (Correctness). Verify the MAIN POINT against the paper.
  - Core concern valid even if a peripheral detail is off?
    -> "Correct". Continue to Step 3.
  - Main point itself factually wrong? -> "Not Correct".
    Set significance and evidence to null.
Step 3 (Significance). Would addressing this criticism genuinely
       improve the paper?
  - Would fixing it make the paper substantively better?
    -> "Significant"
  - Minor presentation or stylistic issue that does not affect
    substance? -> "Marginally Significant"
  - Trivial enough that the review would be better without it?
    -> "Not Significant". Set evidence to null.
Step 4 (Evidence). Can a meta-reviewer verify the claim from the
       review item + the paper without substantial extra work?
  - Yes -> "Sufficient"
  - No  -> "Requires More"
 
 
Part B -- Devil's advocate + expert disagreement prediction
Expert disagreement is common in meta-review -- it is not a rare
edge case. For each axis, you must argue the other side before
deciding whether experts would agree.
 
Step 5 (Correctness -- devil's advocate).
  - If "Not Correct" in Step 2: argue the reviewer's defense.
    Could a charitable expert still find the core concern valid?
    If plausible -> predict "disagree_on_correctness" (and you
    may revise Step 2 to "Correct" if the defense convinces you).
  - If "Correct" in Step 2: argue the prosecution. Could a
    skeptical expert find the claim factually wrong? If
    plausible -> predict "disagree_on_correctness".
  - If neither side is plausible -> experts agree on correctness.
 
Step 6 (Significance -- devil's advocate). Only if Step 5 = agree
       on Correct.
  - If "Significant": argue for downgrading. Could an expert see
    this as merely Marginally Significant?
  - If "Marginally Significant": argue for upgrading. Could an
    expert see this as genuinely improving the paper?
  - If counter-argument plausible -> predict
    "correct_disagree_on_significance".
 
  IMPORTANT: Correctness disagreement means the factual truth of
  the claim is ambiguous. Significance disagreement means the
  fact is clear but its importance is debatable. Do not mislabel
  correctness uncertainty as significance uncertainty.
 
Step 7 (Evidence boundary). Only if Step 6 = agree on >= Marginal.
       Is evidence clearly sufficient or clearly lacking, or
       borderline?
  - Firm -> experts agree. Pick the matching agreement class.
  - Borderline -> "correct_significant_disagree_on_evidence" or
    "correct_marginal_disagree_on_evidence".
\end{verbatim}
\end{footnotesize}
\caption{Meta-reviewer prompt (Part 2 of 4): the per-item decision procedure, with Part~A producing the agent's own three-axis judgment and Part~B applying a devil's-advocate check that determines whether the agent should predict expert agreement or disagreement on each axis.}
\label{fig:metareviewer_prompt_part2}
\end{figure}
 
\clearpage
 
\begin{figure}[h]
\begin{footnotesize}
\begin{verbatim}
### Consistency constraint (critical)
Your prediction_of_expert_judgments MUST be consistent with your
FINAL axis labels (after any Part B revisions):
- If final correctness = "Correct", the prediction must be
  a "correct_*" class OR "disagree_on_correctness". Never
  "incorrect".
- If final correctness = "Not Correct", the prediction must be
  "incorrect" OR "disagree_on_correctness". Never a "correct_*"
  class.
- Similar logic applies downstream: significance and evidence
  labels constrain which agreement/disagreement classes are valid.
 
 
### TODO list for your workflow
- [ ] Read preprint.md thoroughly. Understand the paper's main
      contribution, methodology, and key claims.
- [ ] Skim images_list.json and open any figures/tables that the
      reviewers mention by name. Verify visual claims.
- [ ] Read the code/ directory if any reviewer references specific
      files or functions. Check the README first, then look at the
      referenced files. If the code is executable and not resource-
      prohibitive, try running it to verify code-related claims.
- [ ] For each reviewer's review file, read every item carefully.
- [ ] For each item, follow Part A then Part B. Reasoning must
      include: (1) axis judgment + why, (2) devil's advocate
      argument for correctness (and significance if applicable)
      and whether plausible, (3) the resulting prediction.
- [ ] (Very important) Review your judgments for consistency.
      Are you applying the same bar for "Significant" across all
      reviewers? Iterate if needed.
- [ ] Write the final prediction JSON to: {output_file}
 
 
### Output format (STRICT)
Write a single JSON file to {output_file} with this exact shape:
 
{
  "paper_id": {paper_id},
  "reviewers": [
    {
      "reviewer_id": "<reviewer name>",
      "items": [
        {
          "item_number": 1,
          "reasoning": "...",
          "correctness": "Correct",
          "significance": "Significant",
          "evidence": "Sufficient",
          "prediction_of_expert_judgments":
              "correct_significant_sufficient"
        }
      ]
    }
  ]
}
 
Rules:
- You MUST include ALL reviewers and ALL items -- no exceptions.
- Each item must have: item_number, reasoning, correctness,
  significance, evidence, prediction_of_expert_judgments.
- "reasoning" must include: (1) axis judgment + why, (2) devil's
  advocate argument and whether plausible, (3) expert prediction.
  Aim for 4-6 sentences.
- Use exact label strings: "Correct" / "Not Correct";
  "Significant" / "Marginally Significant" / "Not Significant";
  "Sufficient" / "Requires More". Use null for cascade-skipped
  fields.
\end{verbatim}
\end{footnotesize}
\caption{Meta-reviewer prompt (Part 3 of 4): consistency constraint linking the predicted ten-class label to the agent's own per-axis labels, the workflow checklist, and the strict JSON output format.}
\label{fig:metareviewer_prompt_part3}
\end{figure}
 
\clearpage
 
\begin{figure}[h]
\begin{footnotesize}
\begin{verbatim}
### CRITICAL: Verification before finishing
After writing the JSON file, you MUST perform these checks before
stopping. Do NOT skip this step.
 
1. Read the file back and verify it is valid JSON (no syntax
   errors, no trailing commas, no truncated content).
2. Count reviewers: number of reviewer entries must match number
   of .md files in the reviews/ directory.
3. Count items per reviewer: number of item entries must match
   the number of "## Item" sections in that reviewer's .md file.
4. Check label strings: every correctness value must be exactly
   "Correct" or "Not Correct"; every prediction must be one of
   the 10 valid strings (listed below).
5. Check consistency: prediction must agree with axis labels.
6. If any check fails, fix the file and re-verify.
 
Only after all checks pass, print:
"Verification complete. All [N] reviewers and [M] total items
included. Prediction written to {output_file}"
Then stop.
 
 
### Filesystem boundaries
- READ from {paper_preprint_dir} and {paper_reviews_dir}. These
  are the paper's source files. Do not modify anything there.
- WRITE only to {output_file}. Do not create any other files.
- Do not navigate to parent or sibling directories of the paper.
- Do NOT access nature.com, researchsquare.com, springer.com, or
  springerlink.com -- these host the published versions of
  benchmark papers and may contain reviewer comments. Judge items
  solely from the paper files provided on disk.
 
 
### The 10 valid prediction_of_expert_judgments labels
correct_significant_sufficient            (class 1)
correct_significant_requires_more         (class 2)
correct_significant_disagree_on_evidence  (class 3)
correct_marginal_sufficient               (class 4)
correct_marginal_requires_more            (class 5)
correct_marginal_disagree_on_evidence     (class 6)
correct_not_significant                   (class 7)
correct_disagree_on_significance          (class 8)
incorrect                                 (class 9)
disagree_on_correctness                   (class 10)
 
 
### Tips
1. The paper's markdown may contain OCR errors. Do not penalize
   reviewers for pointing out things that are actually OCR
   artifacts; infer content from context.
2. Image links may be broken. Figures are stored at
   preprint/images/figure1.png, etc.; open images_list.json to
   see captions.
3. Do not try to read every file in code/ -- focus on the files
   that reviewers explicitly reference.
4. Apply the same significance bar consistently across all
   reviewers. Do not be lenient on one reviewer and strict on
   another.
5. Your judgment must be independent of who wrote the review.
   Do not infer reviewer identity (human/AI) from writing style.
\end{verbatim}
\end{footnotesize}
\caption{Meta-reviewer prompt (Part 4 of 4): verification checklist before finishing, filesystem and access boundaries (including the domain blocklist that prevents the agent from retrieving the published version of the paper), the ten valid prediction labels, and additional tips for handling OCR artifacts and applying a uniform bar across reviewers.}
\label{fig:metareviewer_prompt_part4}
\end{figure}
 
\clearpage

\section{Details of the \textsc{CMU Paper Reviewer} platform}\label{appendix:cmu_paper_reviewer}
 
This appendix documents the implementation, weakness-pattern mitigations, public-platform comparison, and intended-use policy for the \textsc{CMU Paper Reviewer} platform released alongside this paper.

\subsection{Implementation and design}\label{appendix:cmu_implementation}
 
\paragraph{Implementation.} The platform uses the OpenHands framework~\citep{wang2025openhands} and the same configuration of tools (terminal, file editor, Tavily web search, Mistral OCR) used in our expert-annotation study, so its behavior matches the AI reviewer that produced the items analyzed in \autoref{sec:performance}.
 
\paragraph{Mitigations for AI reviewer weakness categories.} Building on the qualitative analysis in \autoref{sec:strengths_weaknesses}, the platform implements concrete mitigations for the most consequential weakness categories observed in our study.
\begin{itemize}[leftmargin=*]
    \item \emph{To address vague, verbose, or non-actionable critiques}, every review item is paired with a concrete action: a proposed manuscript edit (which paragraph to rewrite or where to insert new content) or, when the author submits source code, a generated patch with a runnable bash script and README that adds a missing experiment or fixes a code-manuscript mismatch.
    \item \emph{To address over-harsh, out-of-scope, or unrealistic demands}, the platform offers two features. First, when the author opts in, the platform grounds review items in concerns the manuscript itself acknowledges as limitations, and annotates each item with whether it overlaps a stated limitation. Second, authors can challenge any review item through an interactive debate: a pop-up opens in which the AI defends valid concerns but acknowledges items as misplaced when the author's counter-argument is sound, so authors gain context rather than receiving a single static round of feedback.
    \item \emph{To address citing evidence that appeared after the preprint}, the platform allows authors to set a publication date (or auto-detects it from the web), annotates each cited reference with whether it predates or postdates the manuscript, and optionally forbids post-preprint citations entirely.
\end{itemize}
 
\paragraph{User-controllable settings.} The platform supports user-controllable item counts (1 to 15 by default; higher caps available on request), criteria presets (Nature default, NeurIPS, or custom), and accepts supplementary code, data, and other materials alongside the manuscript with no page limit.

\subsection{Comparison with public AI reviewer platforms}\label{appendix:cmu_comparison}
 
We compare the \textsc{CMU Paper Reviewer} against two free public services on the 78 papers from \textsc{PeerReview Bench} (\autoref{appendix:peerreview_bench_construction}) using the same precision, recall, and F1 metrics: the Stanford Agentic Reviewer\footnote{\url{https://paperreview.ai/}}, where each bullet point under ``Weaknesses'' is treated as a single review item, and OpenAIReview\footnote{\url{https://openaireview.org/}}, where each feedback card is treated as a single review item. \autoref{tab:platform-comparison} reports the results.
 
\begin{table}[t]
\centering
\fontsize{8}{10}\selectfont
\setlength{\tabcolsep}{6pt}
\begin{tabular}{l cc | cccc}
\toprule
\multicolumn{1}{c}{\multirow{2}{*}{\textbf{AI Reviewer platforms}}}
& \multicolumn{1}{c}{\multirow{2}{*}{\textbf{LLM Backbone}}}
& \multicolumn{1}{c}{\multirow{2}{*}{\textbf{\# Max review items}}}
& \multicolumn{4}{c}{\multirow{1}{*}{\textbf{\textsc{PeerReview Bench}}}}\\
\cmidrule(lr){4-7}
& & & Precision & Recall & F1 & \# Review items\\
\midrule
Stanford Agentic Reviewer\textsuperscript{\dag} & N/A &  \xmark & 59.84 & \textbf{45.43} & \underline{51.65} & 11.08\\
OpenAIReview\textsuperscript{\ddag}      & Claude-Opus-4.7 & \xmark & 57.57 & 40.98 & 47.88 & 18.64\\
\midrule
\multicolumn{1}{l}{\multirow{3}{*}{\textsc{CMU Paper Reviewer (Ours)}}}                & Claude-Opus-4.7 & 5 & 71.47 & 39.00 & 50.46 & 4.73\\
& GPT-5.4 & 5 & \underline{93.81} & 26.55 & 41.38 & 3.60 \\
& GPT-5.4 & 15 & \textbf{95.46} & \underline{42.32} & \textbf{58.64} & 7.35\\
\bottomrule
\end{tabular}
\caption{\textbf{Comparison of publicly available AI reviewer platforms on \textsc{PeerReview Bench} (78 papers).}
For each platform, we evaluate the review items it produces against the same expert annotations used in \autoref{appendix:peerreview_bench_construction}, and report precision, recall, and F1 across three aspects (correctness, significance, and sufficiency of evidence).
For Stanford Agentic Reviewer, each bullet point under the ``Weaknesses'' section is treated as a single review item; for OpenAIReview, each feedback card is treated as a single review item.
\textsuperscript{\dag}\url{https://paperreview.ai/} \quad
\textsuperscript{\ddag}\url{https://openaireview.org/}
}
\label{tab:platform-comparison}
\end{table}
 
\paragraph{The \textsc{CMU Paper Reviewer} dominates on F1 across configurations.} The GPT-5.4 backbone with a 15-item cap reaches the highest F1 on the benchmark at 58.64, against 51.65 for the Stanford Agentic Reviewer and 47.88 for OpenAIReview. The same backbone with a 5-item cap leads on precision (93.81\%) at the cost of recall (26.55\%), reflecting the high-precision-low-recall character of the GPT-5 family observed across \autoref{tab:peerreview-bench}. The Claude-Opus-4.7 backbone with 5 items reaches an F1 of 50.46 with a more balanced operating point (precision 71.47\%, recall 39.00\%) and a much smaller absolute item count than the competing platforms (4.73 items per paper, against 11.08 and 18.64).
 
\paragraph{Raising the item cap improves coverage without sacrificing precision.} A non-trivial observation is that tripling the per-paper item cap (from 5 to 15) for the GPT-5.4 backbone does not triple the number of items the agent emits, but only roughly doubles it (3.60 to 7.35 items per paper on average). The agent's internal selection prioritizes fully-positive items rather than filling the budget, so additional items only appear when the agent is confident they meet the three-axis bar; precision is actually higher with the larger cap (95.46\% vs 93.81\%), and recall climbs sharply (26.55\% to 42.32\%). This is the only configuration in our comparison that simultaneously achieves the benchmark's highest precision and approaches the recall of the Stanford Agentic Reviewer. We leave to future work the question of the optimal item-generation strategy for AI reviewers: whether to prioritise quality at the cost of coverage (as our 5-item cap does), to expand coverage without sacrificing precision (as the 15-item cap appears to do), or to seek a different operating point altogether may depend on the use case and on the reviewer's intended audience.

\subsection{Availability and intended use}\label{appendix:cmu_availability}
 
The platform is publicly accessible at \url{https://prometheus-eval.github.io/cmu-paper-reviewer/}; users can generate up to three reviews per day for free, or register their own API key for unlimited use. The implementation code is released as open source so any institution can deploy its own instance. We emphasize that the \textsc{CMU Paper Reviewer} is intended as an assistive tool for pre-submission feedback, and \textbf{should not be used at conferences or journals that prohibit AI reviewers in their official review process}; the platform is not a substitute for a sanctioned human peer-review workflow. Conference and journal organizers interested in officially incorporating the \textsc{CMU Paper Reviewer} into their review workflow are welcome to contact the authors.

\section{Recommendation for journal and conference organizers: Panel composition analysis}\label{appendix:deployment}

\paragraph{Motivation} This appendix documents the panel composition simulation to indicate what journal and conference organizers should consider and prioritize when deploying AI reviewers. We describe the four panel configurations, the seven review-item metrics used to evaluate each panel, the main results, three configurations indexed to organizer priorities, caveats on generalization, and the 95\% confidence intervals.

\subsection{Methodology and metric definitions}\label{appendix:deployment_methodology}
 
\paragraph{Panel construction.} We construct four panel compositions by varying the mix of human and AI reviewers: 3 humans (3H), 2 humans plus 1 AI (2H+1AI), 1 human plus 2 AIs (1H+2AI), and 3 AIs (3AI). The simulation runs over the 53 papers in our expert-annotation study with three human reviews and three AI reviews available. For mixed panels, we average over all possible compositions: 2H+1AI averages over all $\binom{3}{2}\times 3 = 9$ combinations of two humans and one of the three AI reviewers; 1H+2AI averages over all $3\times\binom{3}{2} = 9$ combinations of one human and two of the three AIs. We additionally evaluate every panel composition with and without an AI meta-reviewer filter (GPT-5.4) that removes items the meta-reviewer judges not fully positive.
 
\paragraph{Metric definitions.} \autoref{tab:panel-simulation} reports seven per-paper metrics for each panel composition, organized by the stakeholder concern each metric targets:
\begin{itemize}[leftmargin=*]
    \item \textbf{\# Total Items}: how many review items the panel produces per paper; what the author receives for their submission.
    \item \textbf{\# Unique Items}: items that have no similar counterpart from other reviewers in the panel (similarity ordinal $<2$).
    \item \textbf{\# Not Fully Pos.\ Items}: items that fail at least one of correctness, significance, or evidence sufficiency; this corresponds to the burden that authors or editors must triage to surface useful items.
    \item \textbf{\% Fully Pos.\ + Unique / \# Unique Items}: the fraction of distinctive contributions that are fully positive, i.e., how often a reviewer's distinct item turns out to be useful for the authors.
    \item \textbf{\# Fully Pos.\ + Unique Items}: items that are simultaneously fully positive (correct, significant, and evidence-sufficient by expert annotation) and unique within the panel. This is the bottom-line metric for what authors actually receive, since the remaining items are either redundant, incorrect, not significant, or insufficiently evidenced.
    \item \textbf{\% Fully Pos.\ + Unique / \# Total Items}: the fraction of all items shown that are useful to authors, i.e., the panel's overall yield.
    \item \textbf{Noise per Gem}: \# Not Fully Pos.\ / \# Fully Pos.\ + Unique, the number of items the author reads to encounter one item containing useful feedback.
\end{itemize}

\subsection{Panel composition results}\label{appendix:deployment_results}
 
\begin{table*}[t]
\centering
\fontsize{6}{8}\selectfont
\setlength{\tabcolsep}{4.5pt}
\renewcommand{\arraystretch}{1.0}
\begin{tabular}{ccc c c cc c cc}
\toprule
\multicolumn{3}{c}{\textbf{Panel Composition}}
& \multicolumn{1}{c}{\textbf{Volume}}
& \multicolumn{1}{c}{\textbf{Diversity}}
& \multicolumn{2}{c}{\textbf{Quality}}
& \multicolumn{1}{c}{\textbf{Diversity \& Quality}}
& \multicolumn{2}{c}{\textbf{Triage Efficiency}} \\
\cmidrule(lr){1-3} \cmidrule(lr){4-4} \cmidrule(lr){5-5} \cmidrule(lr){6-7} \cmidrule(lr){8-8} \cmidrule(lr){9-10}
\makecell[c]{\textbf{\# Human} \\ \textbf{Reviewers}}
& \makecell[c]{\textbf{\# AI} \\ \textbf{Reviewers}}
& \makecell[c]{\textbf{AI Meta-Rev.} \\ \textbf{Filtering}}
& \makecell[c]{\textbf{\# Total} \\ \textbf{Items}}
& \makecell[c]{\textbf{\# Unique} \\ \textbf{Items} ($\uparrow$)}
& \makecell[c]{\textbf{\# Not Fully Pos.} \\ \textbf{Items} ($\downarrow$)}
& \makecell[c]{\textbf{\% Fully Pos. + Unique} \\ \textbf{/ \# Unique Items} ($\uparrow$)}
& \makecell[c]{\textbf{\# Fully Pos. +} \\ \textbf{Unique Items} ($\uparrow$)}
& \makecell[c]{\textbf{\% Fully Pos. + Unique} \\ \textbf{/ \# Total Items} ($\uparrow$)}
& \makecell[c]{\textbf{Noise per} \\ \textbf{Gem} ($\downarrow$)} \\
\midrule
3 & 0 & \xmark & 25.8 & \textbf{11.5} & 14.6 & 33.9\% & \textbf{3.9} & 15.1\% & 3.74 \\
2 & 1 & \xmark & 21.4 & 10.9 & 11.5 & 35.8\% & \textbf{3.9} & 18.2\% & 2.95 \\
1 & 2 & \xmark & 17.1 & 8.5  & 8.4  & 41.2\% & 3.5 & 20.5\% & 2.40 \\
0 & 3 & \xmark & 12.7 & 3.1  & 5.2  & 58.1\% & 1.8 & 14.2\% & 2.89 \\
\midrule
3 & 0 & \cmark & 11.8 & 4.0 & 5.3 & 47.5\% & 1.9 & 16.1\% & 2.79 \\
2 & 1 & \cmark & 11.0 & 4.6 & 4.7 & 47.8\% & 2.2 & 20.0\% & 2.14 \\
1 & 2 & \cmark & 10.2 & 4.1 & 4.1 & 51.2\% & 2.1 & \textbf{20.6\%} & \textbf{1.95} \\
0 & 3 & \cmark & 9.4  & 1.9 & \textbf{3.5} & \textbf{63.2\%} & 1.2 & 12.8\% & 2.92 \\
\bottomrule
\end{tabular}
\caption{\textbf{Reviewer panel composition analysis.}
Mean per-paper values across 53 papers with complete data (3 human reviewers, 3 AI reviewers, and AI meta-reviewer predictions for every item).
A \emph{unique} item has no similar counterpart from any other reviewer in the panel (similarity ordinal $<2$);
a \emph{fully positive} item is correct, significant, and evidence-sufficient by expert annotation.
``2H+1AI'' averages over all $\binom{3}{2}\times 3 = 9$ panels; ``1H+2AI'' over all $3\times \binom{3}{2}=9$ panels.
Rows 5--8 apply an AI meta-reviewer (GPT-5.4) filter that removes items the meta-reviewer judges not fully positive.
Noise per Gem is \# Not Fully Pos.\ / \# Fully Pos.\ + Unique, the filler an editor or author reads to surface each useful item.
Bold values mark the single best performer across all eight panel configurations on each metric (\# Total Items is descriptive and not bolded; \# Fully Pos.\ + Unique has a tie between the 3-Human and 2H+1AI configurations).
95\% confidence intervals on the count columns are reported in \autoref{tab:panel-simulation-cis} (\autoref{appendix:panel_simulation_cis}).}
\label{tab:panel-simulation}
\end{table*}

\begin{table*}[!t]
\centering
\fontsize{6}{8}\selectfont
\setlength{\tabcolsep}{6pt}
\renewcommand{\arraystretch}{1.0}
\begin{tabular}{ccc cccc}
\toprule
\multicolumn{3}{c}{\textbf{Panel Composition}} & \multicolumn{1}{c}{\textbf{Volume}} & \multicolumn{1}{c}{\textbf{Diversity}} & \multicolumn{1}{c}{\textbf{Quality}} & \multicolumn{1}{c}{\textbf{Diversity \& Quality}} \\
\cmidrule(lr){1-3} \cmidrule(lr){4-4} \cmidrule(lr){5-5} \cmidrule(lr){6-6} \cmidrule(lr){7-7}
\makecell[c]{\textbf{\# Human} \\ \textbf{Reviewers}}
& \makecell[c]{\textbf{\# AI} \\ \textbf{Reviewers}}
& \makecell[c]{\textbf{AI Meta-Rev.} \\ \textbf{Filtering}}
& \makecell[c]{\textbf{\# Total} \\ \textbf{Items}}
& \makecell[c]{\textbf{\# Unique} \\ \textbf{Items}}
& \makecell[c]{\textbf{\# Not Fully Pos.} \\ \textbf{Items}}
& \makecell[c]{\textbf{\# Fully Pos. +} \\ \textbf{Unique Items}} \\
\midrule
3 & 0 & \xmark & 25.8 [21.8, 30.2] & 11.5 [9.6, 13.6] & 14.6 [11.3, 18.5] & 3.9 [3.0, 4.8] \\
2 & 1 & \xmark & 21.4 [18.7, 24.4] & 10.9 [9.3, 12.6] & 11.5 [9.2, 14.1]  & 3.9 [3.2, 4.6] \\
1 & 2 & \xmark & 17.1 [15.7, 18.6] & 8.5 [7.2, 9.8]   & 8.4 [7.0, 9.8]    & 3.5 [2.9, 4.1] \\
0 & 3 & \xmark & 12.7 [12.3, 13.0] & 3.1 [2.6, 3.5]   & 5.2 [4.2, 6.3]    & 1.8 [1.4, 2.2] \\
\midrule
3 & 0 & \cmark & 11.8 [9.7, 14.1]  & 4.0 [3.3, 4.8]   & 5.3 [4.0, 6.8]    & 1.9 [1.5, 2.4] \\
2 & 1 & \cmark & 11.0 [9.5, 12.6]  & 4.6 [4.0, 5.3]   & 4.7 [3.7, 5.8]    & 2.2 [1.8, 2.6] \\
1 & 2 & \cmark & 10.2 [9.3, 11.1]  & 4.1 [3.4, 4.7]   & 4.1 [3.3, 4.9]    & 2.1 [1.7, 2.5] \\
0 & 3 & \cmark & 9.4 [8.7, 10.0]   & 1.9 [1.5, 2.3]   & 3.5 [2.8, 4.3]    & 1.2 [0.9, 1.5] \\
\bottomrule
\end{tabular}
\caption{\textbf{Panel composition count metrics with 95\% confidence intervals.}
Brackets show paper-level bootstrap 95\% CIs (10{,}000 resamples) for each of the four count metrics in \autoref{tab:panel-simulation}.
Metric definitions, panel-composition methodology, and combinatorial averaging conventions are given in \autoref{appendix:deployment_methodology}.}
\label{tab:panel-simulation-cis}
\end{table*}
 
\paragraph{Substituting a second human with another AI begins to cost useful content.} Among unfiltered panels, 2H+1AI does not lose to 3 humans on any metric in \autoref{tab:panel-simulation}: same fully-positive-and-unique count (3.9 vs 3.9), fewer total items (21.4 vs 25.8), fewer not-fully-positive items (11.5 vs 14.6), higher yield (18.2\% vs 15.1\%), higher quality of unique items (35.8\% vs 33.9\%), and lower noise per gem (2.95 vs 3.74). Replacing a second human with an AI reviewer (1H+2AI) gives a smaller, not-yet-significant drop in fully-positive-and-unique items (3.5 per paper, CI [2.9, 4.1]). Only at the all-AI panel does the count drop sharply and significantly to 1.8 (CI [1.4, 2.2], non-overlapping with all three other compositions), driven by a collapse in unique coverage: AI reviewers overlap with each other much more than human reviewers do, so a panel of three AIs produces only 3.1 distinct issues per paper, against 11.5 for three humans. Adding more than one AI to a panel therefore begins to erode the useful content authors receive, and replacing all three humans erodes it sharply.
 
\paragraph{The AI meta-reviewer filter raises every panel's efficiency at the cost of absolute volume.} Filtering each panel's items through the AI meta-reviewer (\autoref{appendix:calibration_set}) raises the yield of every panel composition (e.g., 2H+1AI from 18.2\% to 20.0\%, 1H+2AI from 20.5\% to 20.6\%) and lowers noise per gem (e.g., 2H+1AI from 2.95 to 2.14). The trade-off is that the same filter also reduces the absolute count of items that are simultaneously fully positive and unique: from 3.9 to 1.9 for three humans (a 51\% drop, CIs do not overlap), from 3.9 to 2.2 for 2H+1AI, from 3.5 to 2.1 for 1H+2AI, and from 1.8 to 1.2 for the all-AI panel. The meta-reviewer removes some genuinely useful items along with the not-fully-positive items it is designed to filter, a consequence of the meta-reviewer's leniency on significance documented in \autoref{appendix:metareviewer_failure_significance}. For deployments that prioritise editor or author triage time over absolute volume, the filter is a reasonable addition; for deployments that prioritise the volume of useful feedback delivered to authors, it loses more than it removes.
 
\paragraph{All-AI panels with filtering are selective, not productive.} The strength of the all-AI configuration is selection. AI reviewers tend to converge on a small set of well-known issues, so most of what they produce overlaps; the items where one AI breaks from the pack to raise something distinct tend to be narrow, well-targeted critiques that are easy to verify, which makes them more likely to be rated correct, significant, and evidence-sufficient. The meta-reviewer filter sharpens this selectivity by removing the items it judges not fully positive. The result is a panel whose distinctive contributions are unusually high quality and whose absolute reviewer-side noise is the lowest of any configuration, but at the cost of producing only one or two pieces of useful, non-redundant feedback per paper. This configuration is appropriate when an editor or author wants a small, vetted shortlist of high-confidence critiques rather than broad coverage of a manuscript.

\paragraph{Confidence intervals on count metrics}\label{appendix:panel_simulation_cis}
 
\autoref{tab:panel-simulation-cis} reports paper-level bootstrap 95\% confidence intervals (10{,}000 resamples) for the four count metrics in \autoref{tab:panel-simulation}.
 
\paragraph{Three configurations indexed to organizer priorities}\label{appendix:deployment_recommendations}
 
The findings above support three configuration-priority pairings that an organizer can pick from depending on what their venue most wants to optimize. Each is summarized below.
 
\begin{mdframed}[style=categoryboxR, frametitle={\textbf{Recommendation 1 (2H+1AI): same useful feedback as today's 3-human panel, with less filler}}]
\footnotesize
Both 3H and 2H+1AI deliver the highest amount of useful feedback among all panel configurations, producing 3.9 items per paper that are simultaneously fully positive (correct, significant, and well-evidenced) and unique to that reviewer, with heavily overlapping 95\% CIs in \autoref{tab:panel-simulation-cis}. The 2H+1AI configuration is preferable in practice because it preserves the same amount of useful feedback while reducing the total number of items by 17\% (21.4 vs 25.8) and the number of not-fully-positive items by 21\% (11.5 vs 14.6), so authors and editors read less filler to reach the same useful content. Given the rising volume of paper submissions and the increasing difficulty of recruiting qualified human reviewers~\citep{chen2025position}, 2H+1AI is the natural recommendation for venues that want to preserve today's review experience while reducing reviewer-side burden.
\end{mdframed}
 
\begin{mdframed}[style=categoryboxR, frametitle={\textbf{Recommendation 2 (1H+2AI + meta-reviewer filter): minimize editor and author triage time}}]
\footnotesize
The 1H+2AI panel with the meta-reviewer filter achieves the highest fraction of items that are simultaneously fully positive and unique within the panel (20.6\% of all items shown, against 15.1\% for the 3-human baseline, a 36\% relative improvement) and the lowest noise per gem (1.95 against 3.74, a 48\% reduction). The trade-off is volume: this configuration delivers 2.1 fully-positive-and-unique items per paper against 3.9 for the 3-human baseline, so authors receive less useful content overall but spend less time reading filler to reach each useful item. This configuration suits venues where editor or author triage time per paper is the binding constraint.
\end{mdframed}
 
\begin{mdframed}[style=categoryboxR, frametitle={\textbf{Recommendation 3 (3AI + meta-reviewer filter): high-confidence shortlist instead of full reviews}}]
\footnotesize
The all-AI panel with the meta-reviewer filter produces the fewest not-fully-positive items per paper (3.5 against 14.6 for the 3-human baseline) and the highest fraction of distinctive contributions that are fully positive (63.2\% of unique items against 33.9\% for three humans, a 71\% relative improvement). When this panel raises a distinctive item, it is fully positive nearly twice as often as when a human panel does. The trade-off is volume across the board: the all-AI filtered panel produces only 1.9 unique items and 1.2 fully-positive-and-unique items per paper. This configuration suits editorial pre-screening, desk-review aids, and other contexts where a small, vetted shortlist of high-confidence critiques is preferable to broad coverage of a manuscript.
\end{mdframed}

\paragraph{Caveats}\label{appendix:deployment_caveats}
 
Two caveats temper these recommendations. First, our entire study draws from one Nature paper submission pool, where reviewers are typically from broad disciplines and review papers tend to address a non-specialist audience; conferences with narrower technical scope, single-discipline journals, or venues with very short review timelines may exhibit different dynamics. Second, the per-AI five-item cap and the choice of three frontier AI reviewers in our study are specific to our experimental setup; a larger per-AI item budget would change the absolute counts, and frontier-model behavior will continue to evolve. The qualitative pattern (humans contribute most of the unique coverage, AIs contribute fluent and well-evidenced items but overlap with each other, AI reviewers' distinctive contributions are unusually high quality when they appear, and the meta-reviewer filter improves efficiency at the cost of removing some genuinely high-value items) is what we expect to generalize; specific numbers should be re-validated for other venues or future AI reviewer generations.

\end{document}